\documentclass{article}

\PassOptionsToPackage{numbers}{natbib}

 \usepackage[preprint]{neurips_data_2021}

\usepackage[utf8]{inputenc} 
\usepackage[T1]{fontenc}    
\usepackage{hyperref}       
\usepackage{url}            
\usepackage{booktabs}       
\usepackage{amsfonts}       
\usepackage{amsmath}
\usepackage{nicefrac}       
\usepackage{microtype}      
\usepackage{xcolor}         
\usepackage[inline]{enumitem}

\newcommand{\psp}{\textsc{PeopleSansPeople}}

\let\oldfrac\frac
\renewcommand{\frac}[2]{%
  \mathchoice
    {\oldfrac{#1}{#2}}
    {#1/#2}
    {#1/#2}
    {#1/#2}
}

\usepackage{graphicx}
\usepackage{rotating}
\usepackage{subcaption} 
\usepackage{adjustbox}

\usepackage{float}

\usepackage{multirow}
\usepackage[compact]{titlesec}
\usepackage{algorithm}
\usepackage{algpseudocode}
\usepackage[group-separator={,}]{siunitx}
\titlespacing{\section}{0pt}{*0.3}{*0.3}
 \titlespacing{\subsection}{0pt}{*0.3}{*0.3}
\title{\psp: \\A Synthetic Data Generator for Human-Centric Computer Vision} 

\begin{document}

\maketitle

\vspace{-1.5cm}
\vbox{
    \centering
    \textbf{Salehe Erfanian Ebadi, You-Cyuan Jhang, Alex Zook, Saurav Dhakad,} 
    \vfill
    \textbf{Adam Crespi, Pete Parisi, Steven Borkman, 
    Jonathan Hogins, Sujoy Ganguly} 
    \vfill
    \text{Unity Technologies} 
    \vfill
    \texttt{\{salehe.erfanianebadi, youcyuan, alex.zook, saurav.dhakad, adamc, pete.parisi, steven.borkman, jonathanh, sujoy.ganguly\} @unity3d.com}
}
\vspace{0.3cm}
\begin{abstract}
In recent years, person detection and human pose estimation have made great strides, helped by large-scale labeled datasets. 
However, these datasets had no guarantees or analysis of human activities, poses, or context diversity. 
Additionally, privacy, legal, safety, and ethical concerns may limit the ability to collect more human data. 
An emerging alternative to real-world data that alleviates some of these issues is synthetic data. 
However, creation of synthetic data generators is incredibly challenging and prevents researchers from exploring their usefulness.
Therefore, we release a human-centric synthetic data generator \psp{} which contains simulation-ready 3D human assets, a parameterized lighting and camera system, and generates 2D and 3D bounding box, instance and semantic segmentation, and COCO pose labels. 
Using \psp{}, we performed benchmark synthetic data training using a Detectron2 Keypoint R-CNN variant~\citep{wu2019detectron2}. 
%
%
We found that pre-training a network using synthetic data and fine-tuning on various sizes of real-world data
resulted in a keypoint AP increase of $+38.03$ ($44.43 \pm 0.17$ vs.~$6.40$) for few-shot transfer (limited subsets of COCO-person train~\citep{lin2014microsoftcoco}), and an increase of $+1.47$ ($63.47 \pm 0.19$ vs.~$62.00$) for abundant real data regimes, outperforming models trained with the same real data alone. We also found that our models outperformed those pre-trained with ImageNet with a keypoint AP increase of $+22.53$ ($44.43 \pm 0.17$ vs.~$21.90$) for few-shot transfer and $+1.07$ ($63.47 \pm 0.19$ vs.~$62.40$) for abundant real data regimes.
This freely-available data generator\footnote{\psp{} template Unity environment, benchmark binaries, and source code is available at: \url{https://github.com/Unity-Technologies/PeopleSansPeople}} should enable a wide range of research into the emerging field of simulation to real transfer learning in the critical area of human-centric computer vision.
\end{abstract}
\section{Introduction}
\label{introduction}
Over the last decade, computer vision has relied on supervised machine learning to solve increasingly complex vision tasks. The challenge with supervised machine learning is the need for large labeled datasets, and as the tasks become increasingly complex, so too do the datasets. This need for increasingly complex labeled data is particularly acute for human-centric computer vision tasks. While straightforward tasks, such as person detection, can use simple bounding boxes, more complicated applications (e.g., activity recognition, motion analysis, augmented reality (AR)) require granular skeleton predictions.  

To fuel the development of human pose estimation models, researchers established a set of benchmark datasets~\citep{andriluka14cvpr, lin2014microsoftcoco, li2018crowdpose} using real-world data and human annotators. However, real-world images collected using consumer cameras are limited to a smaller collection of images under limited diversity of human activities. In addition to serious privacy and ethical concerns with human data, certain safety-critical human activities (e.g., humans falling, injury-prone activities in sports) are almost impossible to collect in the real world, and regulations~\citep{voigt_eu_2017, bukaty_california_2019} put restrictions on how real-world data can be collected and used. Moreover, granular labels like keypoints occluded by foreground objects and self-occlusion~\citep{lin2014microsoftcoco}, and other granular labeling tasks such as instance segmentation require human annotators to follow the guidelines carefully and can be open to interpretation and error-prone.





\begin{figure}[htb] 
    \centering
    \begin{subfigure}[t]{0.326\textwidth}
        \raisebox{-\height}{\includegraphics[width=\textwidth]{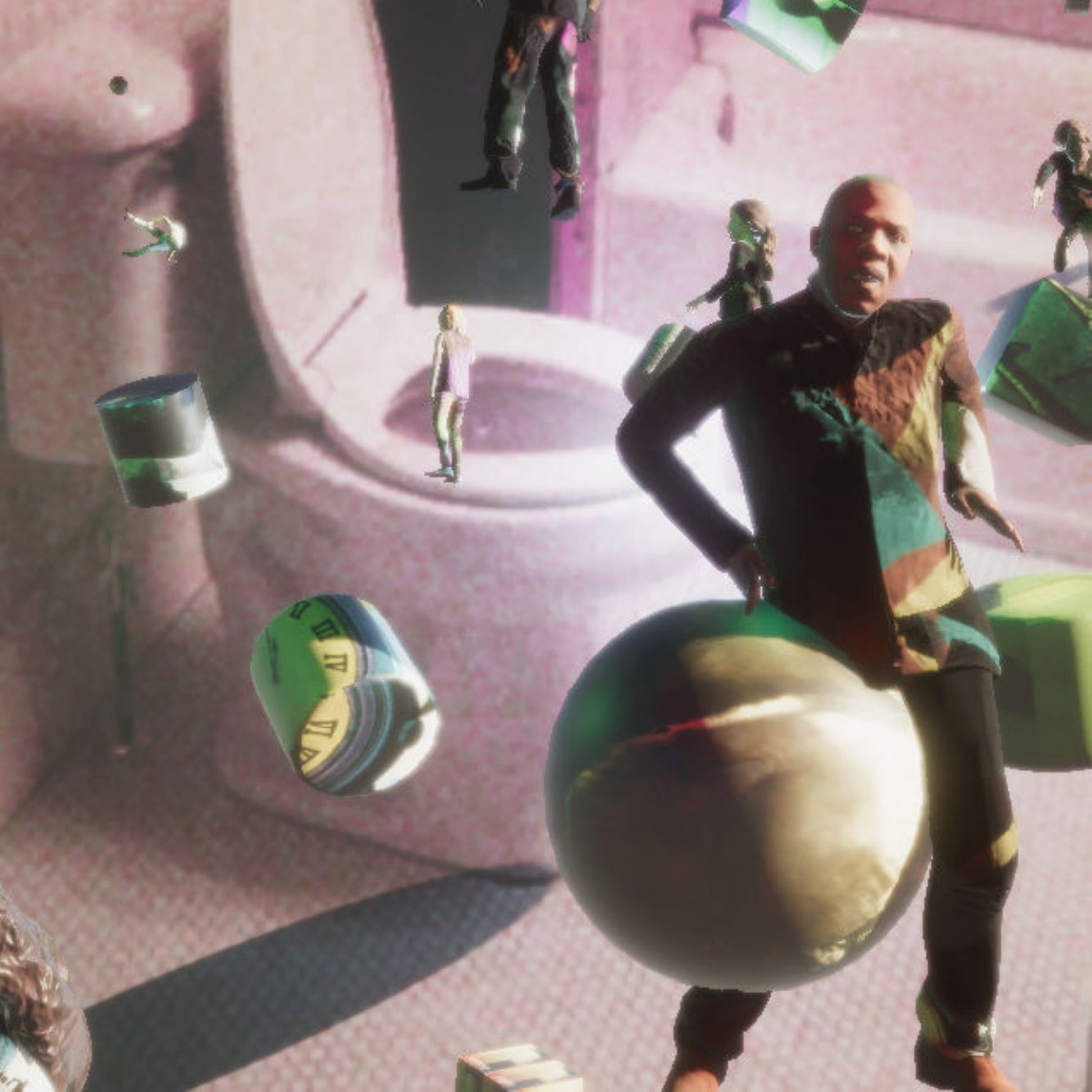}}
    \end{subfigure}
    \begin{subfigure}[t]{0.326\textwidth}
        \raisebox{-\height}{\includegraphics[width=\textwidth]{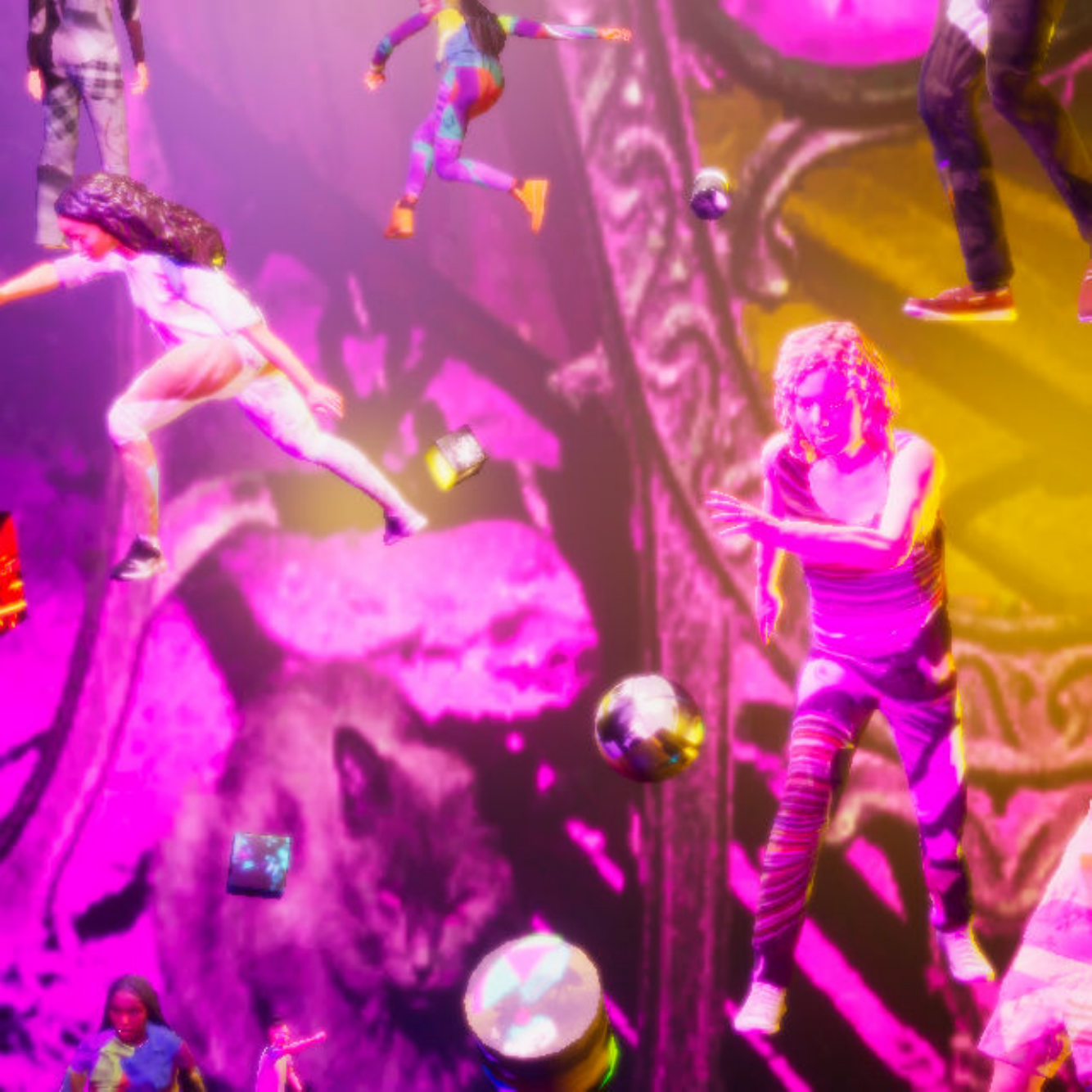}}
    \end{subfigure}
    \begin{subfigure}[t]{0.326\textwidth}
        \raisebox{-\height}{\includegraphics[width=\textwidth]{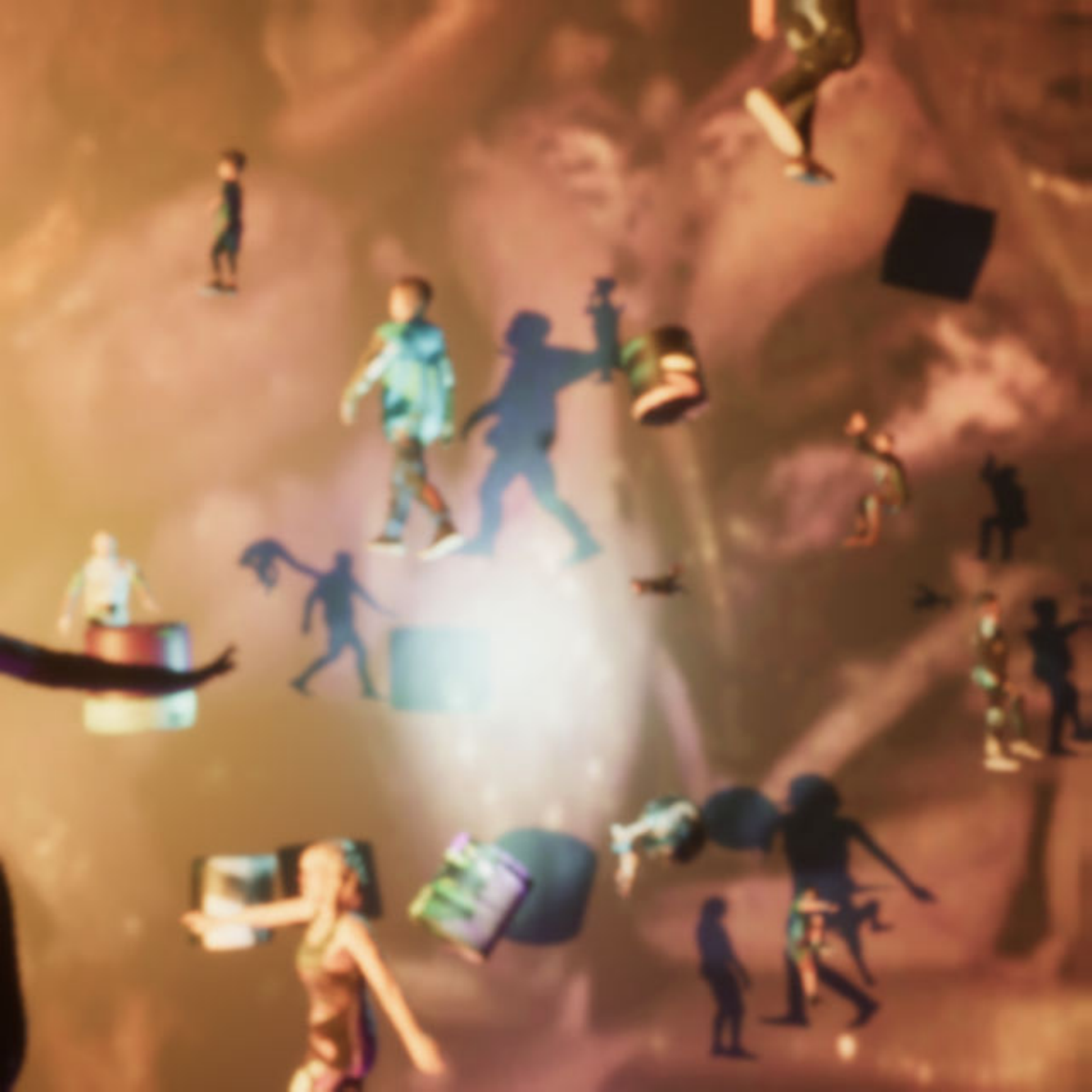}}
    \end{subfigure}
    \\
    \begin{subfigure}[t]{0.326\textwidth}
        \raisebox{-\height}{\includegraphics[width=\textwidth]{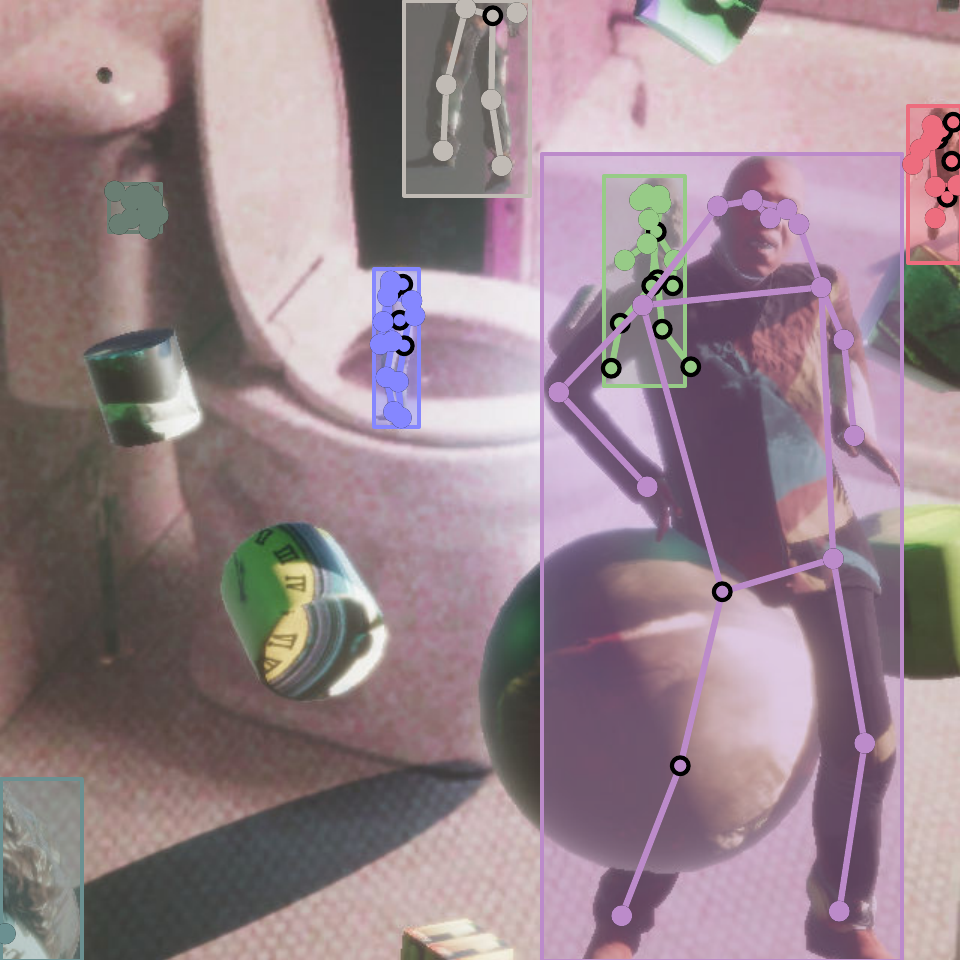}}
    \end{subfigure}
    \begin{subfigure}[t]{0.326\textwidth}
        \raisebox{-\height}{\includegraphics[width=\textwidth]{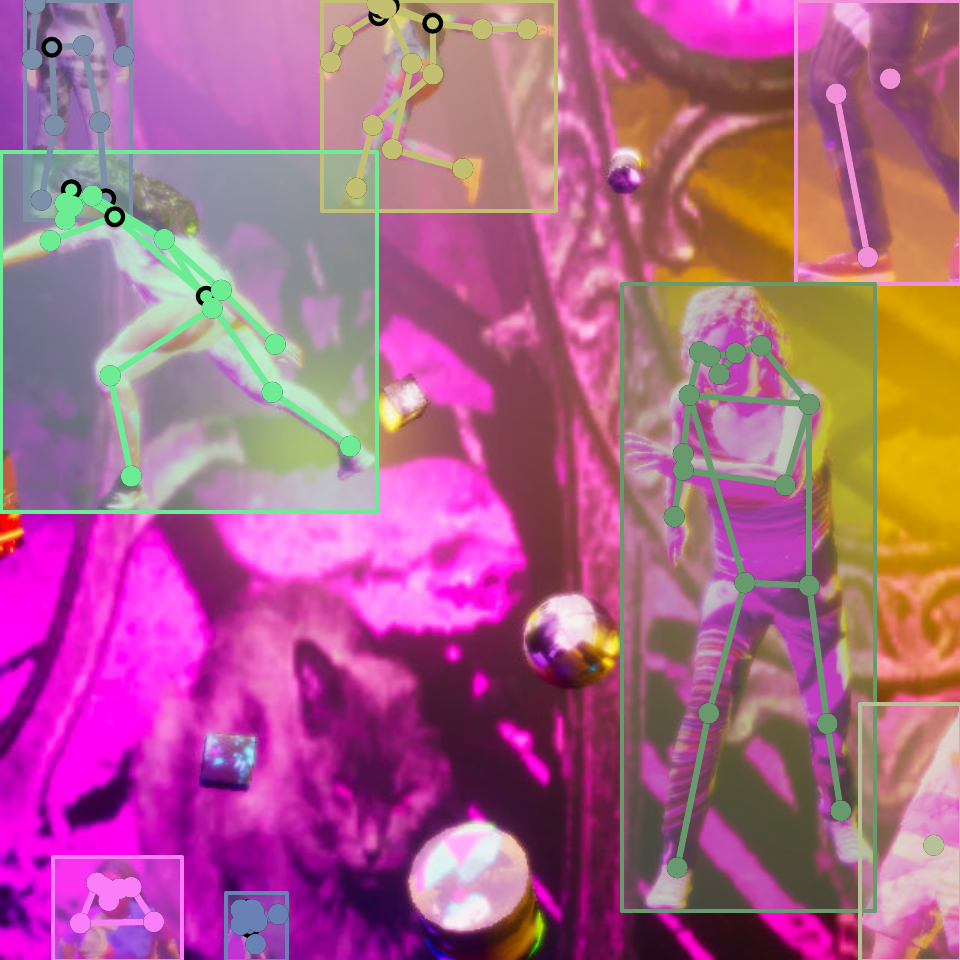}}
    \end{subfigure}
    \begin{subfigure}[t]{0.326\textwidth}
        \raisebox{-\height}{\includegraphics[width=\textwidth]{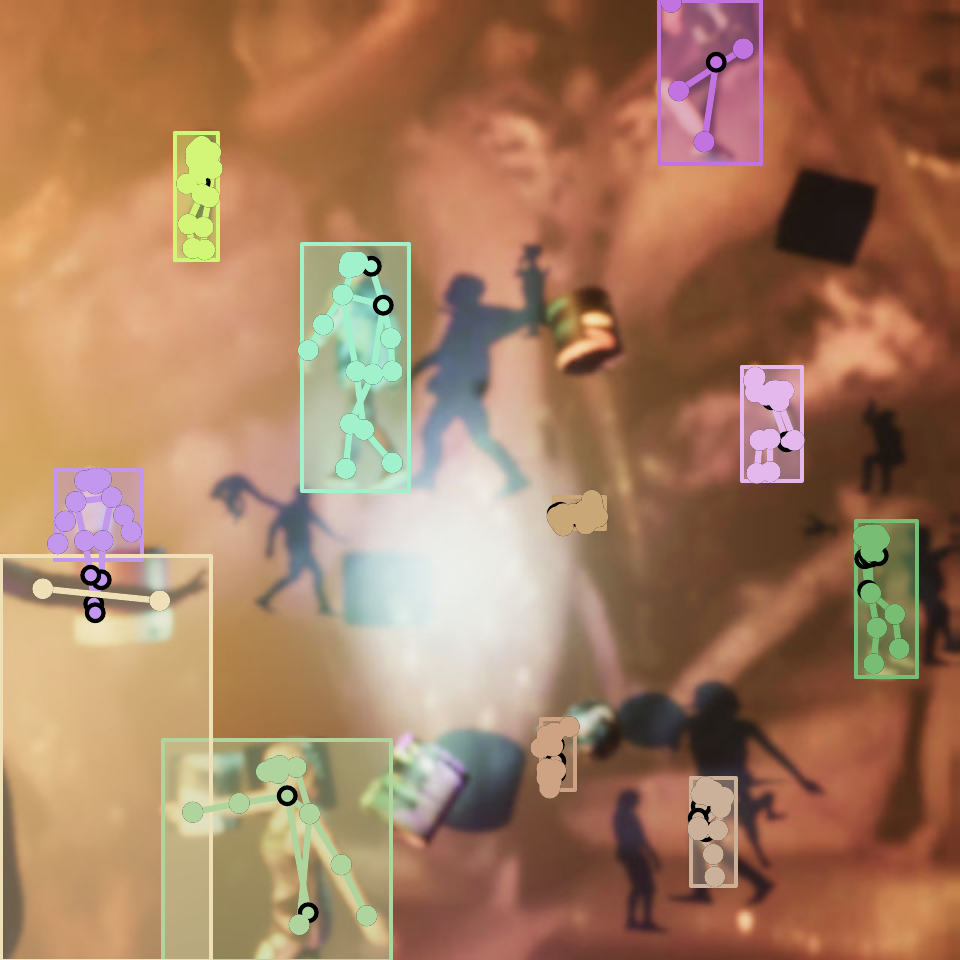}}
    \end{subfigure}
\caption{\textbf{Sample \psp{} Images and Labels.} Top row: three sample synthetic images generated using \psp{}. Bottom row: the same images with generated bounding box and COCO pose labels.}
\label{fig:psp_synth_data_examples}%
\end{figure}

Synthetic data offers an alternative to real-world data that bypasses data variability, privacy, and annotation concerns.
Although tools like the Unity Perception package~\citep{borkman2021unity} make it easy to adjust a scene and generate a dataset with perfect annotations, it is still challenging to source high-quality 3D assets with good diversity to produce valuable datasets. 
Therefore, we introduce a benchmark environment
built using Unity and the Perception package targeting human-centric computer vision (Fig.~\ref{fig:psp_synth_data_examples}).
We affectionately name our synthetic data generator \psp{}, combining \textit{People} + \textit{Sans (middle English for without)} + \textit{People}, which is a data generator aimed at human-centric computer vision without using human data. It includes:
\begin{itemize}[leftmargin=*]
    \item macOS and Linux binaries capable of generating large-scale 
    $1M+$
    datasets with JSON annotations; 
    \item 28 3D human models of varying age and ethnicity, with varying clothing, hair, and skin colors; 
    \item 39 animations clips, with fully randomized humanoid placement, size, and rotation, to generate diverse arrangements of people;
    \item fully-parameterized lighting (position, color, angle, and intensity) and camera settings; 
    \item a set of object primitives to act as distractors and occluders; and
    \item a set of natural images to act as backgrounds and textures for objects.
\end{itemize}
In addition to the binary files mentioned above, we release a Unity template project that helps lower the barrier of entry for the community, by helping them get started with creating their own version of a human-centric data generator. The users can bring their own sourced 3D assets into this environment and further its capabilities by modifying the already-existing domain randomizers or defining new ones. This environment comes with the full functionalities described for the binary files, except with:
\begin{itemize}[leftmargin=*]
    \item 4 example 3D human models with varying clothing colors;
    \item 8 example animations clips, with fully randomized humanoid placement, size, and rotation, to generate diverse arrangements of people; and
    \item a set of natural images of grocery items from Unity Perception package~\citep{borkman2021unity} to act as backgrounds and textures for objects.
\end{itemize}
\section{Related Work}

Traditionally computer vision models have been trained using large scale human-labeled datasets such as PASCAL VOC~\citep{everingham2010pascalvoc}, NYU-Depth V2~\citep{silberman2012nyudepthv2}, MS COCO~\citep{lin2014microsoftcoco}, and SUN RGB-D~\citep{song2015sunrgbd}.
While being powerful resources, producing these datasets is costly, and these static data sources do not allow researchers to create datasets appropriate to their task of interest.
In response, researchers have adopted simulators to control data generation for desired target tasks.
SYNTHIA~\citep{ros2016synthia}, virtual KITTI~\citep{gaidon2016virtual}, CARLA~\citep{dosovitskiy2017carla}, VIPER~\citep{richter2017benchmarks}, and Synscapes~\citep{wrenninge2018synscapes} provide synthetic datasets for computer vision tasks relevant to autonomous vehicle navigation in cities.
Hypersim~\citep{roberts2020hypersim} and OpenRooms~\citep{li2021openrooms} develop simulators for indoor object detection. 
Robotic simulators include AI-2THOR~\citep{kolve2017ai2thor}, Habitat~\citep{savva2019habitat, szot2021habitat2}, NVIDIA Isaac Sim~\citep{nvidia2019isaac}, and iGibson~\citep{shen2020igibson} focus largely on embodied AI tasks.
More generic tools for object detection dataset generation include BlenderProc~\citep{denninger2019blenderproc}, BlendTorch~\citep{heindl2020blendtorch}, NVISII~\citep{morrical2021nvisii}, and the Unity Perception package \citep{borkman2021unity}.
\psp{} contributes to this growing body of tools by addressing humans as a vital part of the dataset and enabling human-centric computer vision tasks.

Synthetic people datasets are challenging to build due to the complexity of human bodies and the significant variations seen in their poses and identities. Several efforts have used learned models to extract 3D posed humans from existing datasets and composited these into new scenes to produce larger synthetic datasets~\citep{pishchulin2011people, varol2017humans, kviatkovsky2020people, hassan2021populating}.
Model approximation quality and biases existing in the training data constrain the generalization capabilities of the learned models, in turn limiting the synthetic data.
Alternatively, we can use simulators to generate richly labeled datasets, derived from hand-crafted scenes~\citep{bak2018people}, existing games like GTA V~\citep{fabbri2018jta, hu2019sailvos, hu2021sailvos3d}, or game engines~\citep{roberto2017video}.
\psp{} builds on this line of work using the Unity Perception package~\citep{borkman2021unity} to produce labeled synthetic data.
We use high-quality human assets and rendering pipelines to generate labeled image data, enabling researchers to produce diverse human data.
\psp{} allows researchers to exchange the provided assets and scene components, allowing a level of customization absent from previous efforts. 
It also enables researchers to leverage existing sources of high-quality digital human assets~\citep{epic2021metahuman,renderpeople}.

Simulators are valuable for data generation as they provide control over the data generation itself and facilitate tuning of the dataset to enable simulation to real (sim2real) transfer.
Domain randomization \citep{tobin2017domain} is a technique that is used to introduce diversity into the generated data, by randomizing the parameters of the simulator.
Domain randomization has been applied to tasks including object detection~\citep{tremblay2018domain,hinterstoisser2019annotation,morrical2021nvisii,heindl2020blendtorch}, robotic manipulation~\citep{tobin2017domain,tremblay2018pose}, and autonomous vehicle navigation~\citep{kar2019metasim,devaranjan2020metasim2,prakash2019structured,prakash2020sim2sg}.
\psp{} enables researchers to use synthetic data with domain randomization in tasks involving people as part of the target class, expanding the space of simulator capabilities in existing and new domains, like autonomous vehicle driving and human pose estimation and tracking.



\section{\psp{}}
\label{sec:psp}



\psp{} is a parametric data generator with a 3D scene populated by 3D human assets in a variety of poses and distractor objects with natural textures. 
We package the data generator as a binary that exposes several parameters for variation via a simple JSON configuration file. 
\psp{} generates RGB images and corresponding labels for the human assets with 2D and 3D bounding boxes, semantic and instance segmentation masks, and the COCO keypoint labels in a JSON. 
Additionally, it emits scene meta-data for statistical comparison and analysis.
Fig.~\ref{fig:psp_synth_data_examples} shows a few examples for the generated data and corresponding labels.
In this section we will describe the components of the PeopleSansPeople synthetic data generator. 
Since our main topic of interest for this work revolves around a human-centric task, much of our 3D environment design went into creating fully-parametric models of humans. 
With such parameter sets, we are able to capture some fundamental intrinsic and extrinsic aspects of variations for our human models. 
Then we show how the human models are inserted in highly-randomized environments to capture data with high diversity.

\subsection{3D Assets}
\label{subsec:3dassets}
\begin{figure}[hbt]
    \centering
    \begin{subfigure}[t]{1\textwidth}
        \raisebox{-\height}{\includegraphics[width=\textwidth, trim={1cm 0.5cm 1cm 14.5cm},clip]{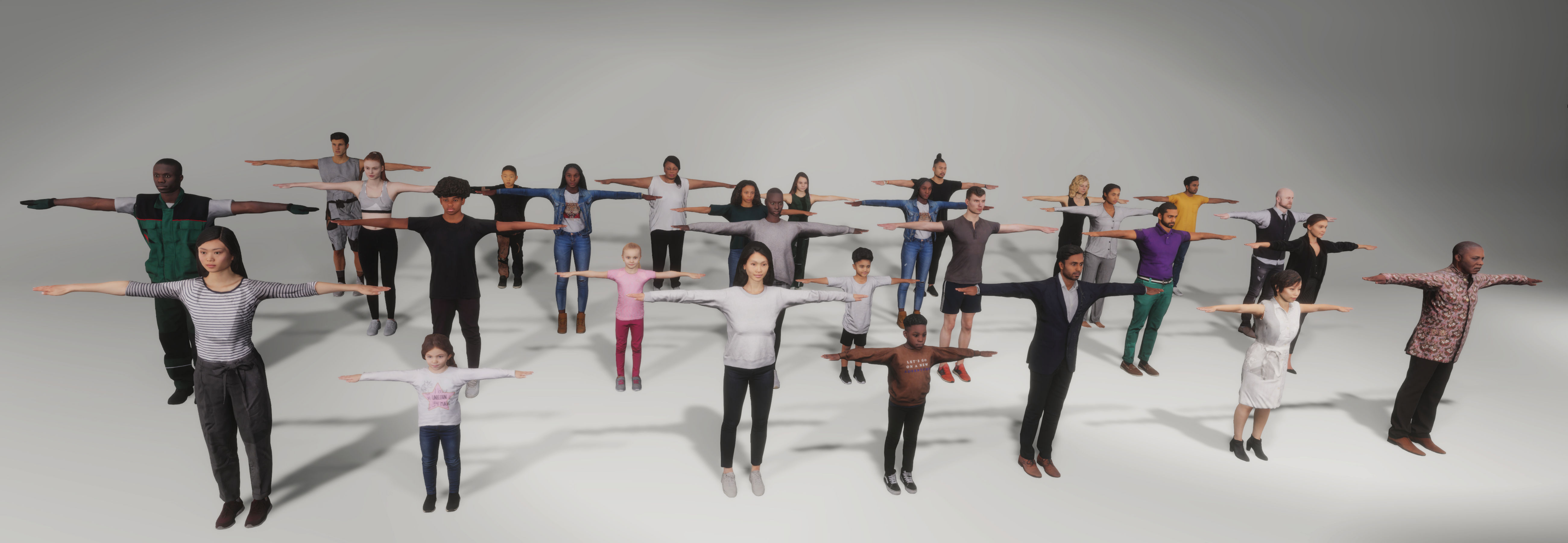}}
        \caption{}
    \end{subfigure}
    \hfill
    \\
    \vspace{0.1cm}
    \begin{subfigure}[t]{1\textwidth}
        \raisebox{-\height}{\includegraphics[width=\textwidth, trim={1cm 0.5cm 1cm 14.5cm},clip]{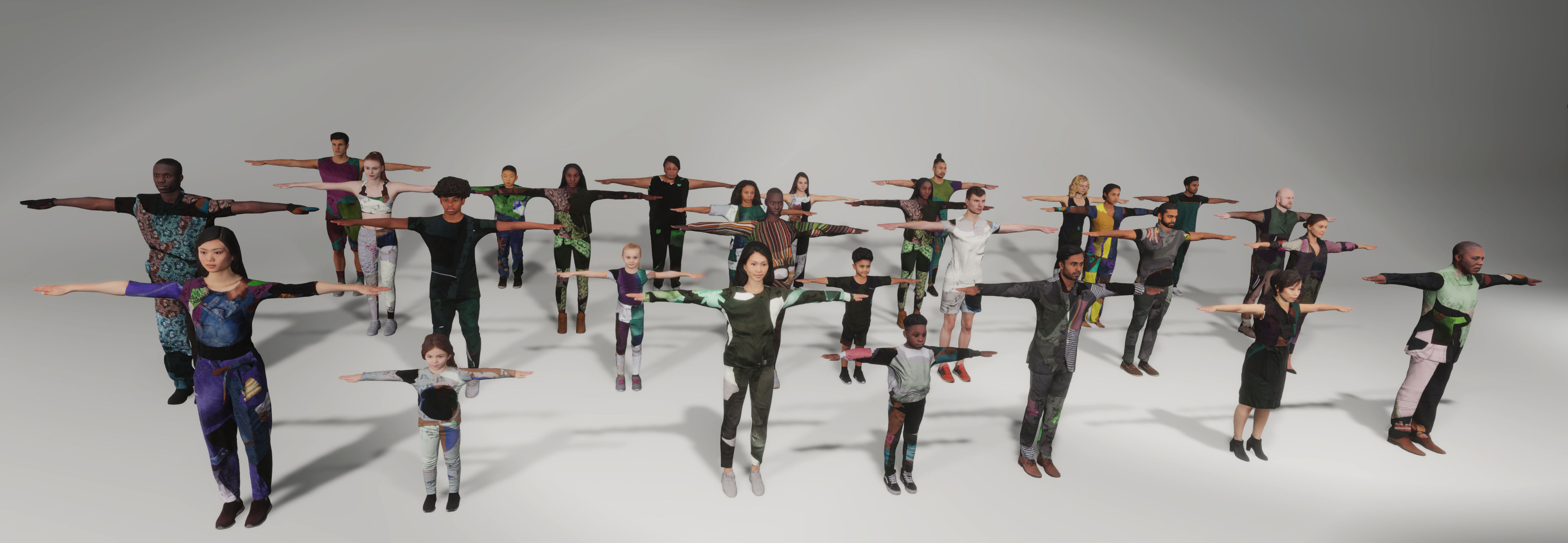}}
        \caption{}
    \end{subfigure}
    \caption{\textbf{\psp{} 3D Human Models.} a) 28 scanned 3D human models~\citep{renderpeople} used in the environment with default pose and clothing textures. b) One example of clothing texture variations enabled by the \psp \hspace{0.1em} Shader Graph.}
     \label{fig:renderpeople_assets}%
\end{figure}

\psp{} has a set of 28 scanned 3D human models from RenderPeople~\citep{renderpeople}.
These models are ethnically- and age-diverse, fully re-topologized, rigged, and skinned with high-quality textures (Fig~\ref{fig:renderpeople_assets}a). 
We added a character control rig to pose or animate them with motion capture data. 

We had to alter the assets to manipulate the material elements for clothing at run-time. 
Specifically, we redrew some red, green, blue, and alpha channels that make up the mask textures. 
Additionally, we created a Shader Graph\footnote{Unity Shader Graph: \url{https://unity.com/shader-graph}} in Unity that allows us to swap the material elements of our human assets and change the hue and texture of clothing (Fig.~\ref{fig:renderpeople_assets}b). 
These changes allowed us to import the human models into Unity, place them into a scene, animate them, and change their clothing texture and color.

To generate diverse poses for our human assets, we gathered a set of 39 animations from Mixamo~\citep{mixamo} which range from simple motions such as idling, walking, and running to more complex ones such as planking, performing break-dance, and fighting. 
We downloaded these animation clips as FBX for Unity at 24 fps and no keyframe reduction\footnote{Refer to the following URL for Adobe Mixamo's legal notices and redistribution policy \url{https://helpx.adobe.com/creative-cloud/faq/mixamo-faq.html}}.
Lastly, we ensured proper re-targeting of all the animation clips to our RenderPeople human assets.


\subsection{Unity Environment}
\begin{figure}[ht] 
    \centering
    \begin{subfigure}[t]{0.49\textwidth}
        \raisebox{-\height}{\includegraphics[width=\textwidth]{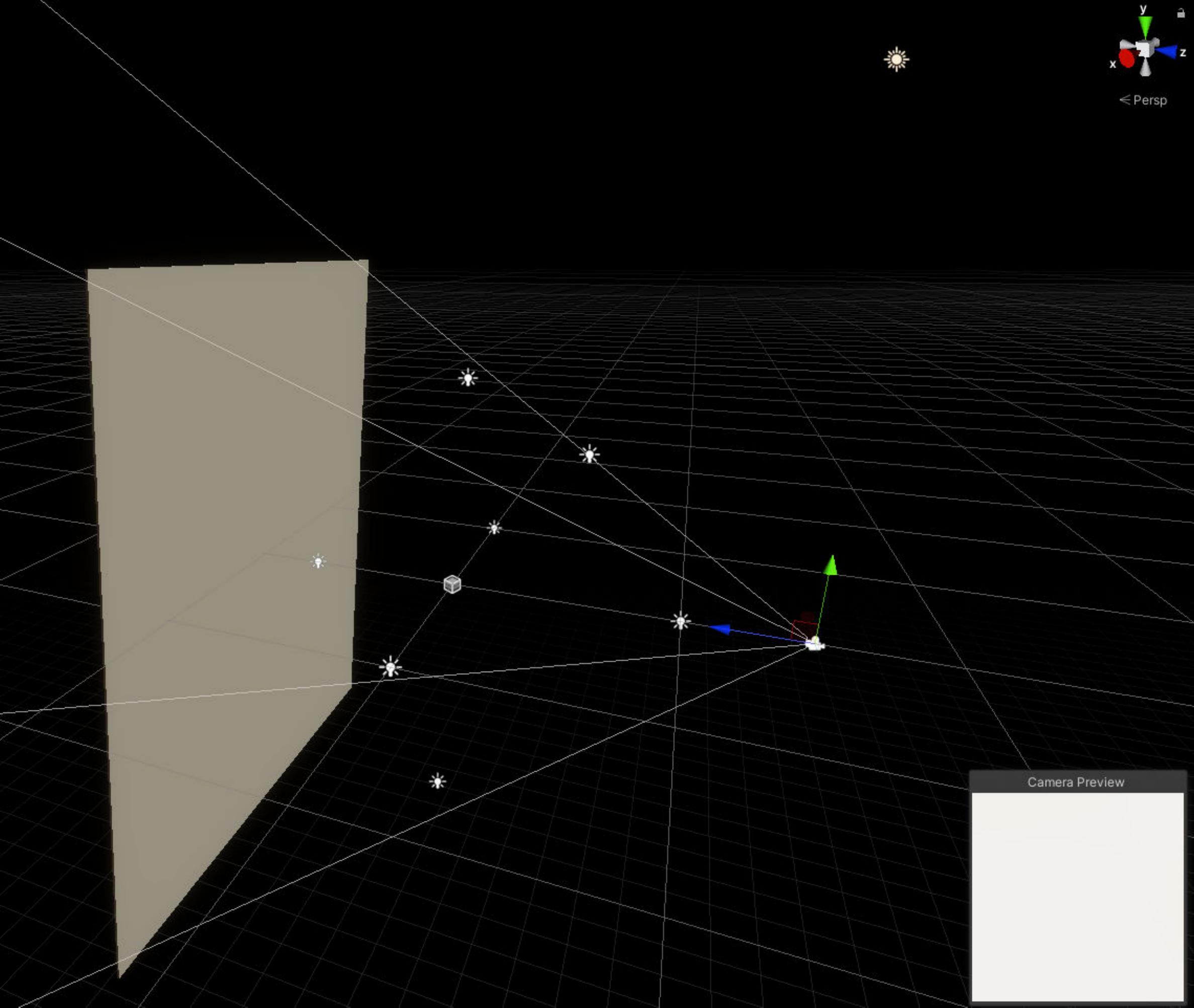}}
        \caption{ }
    \end{subfigure}
    \begin{subfigure}[t]{0.49\textwidth}
        \raisebox{-\height}{\includegraphics[width=\textwidth]{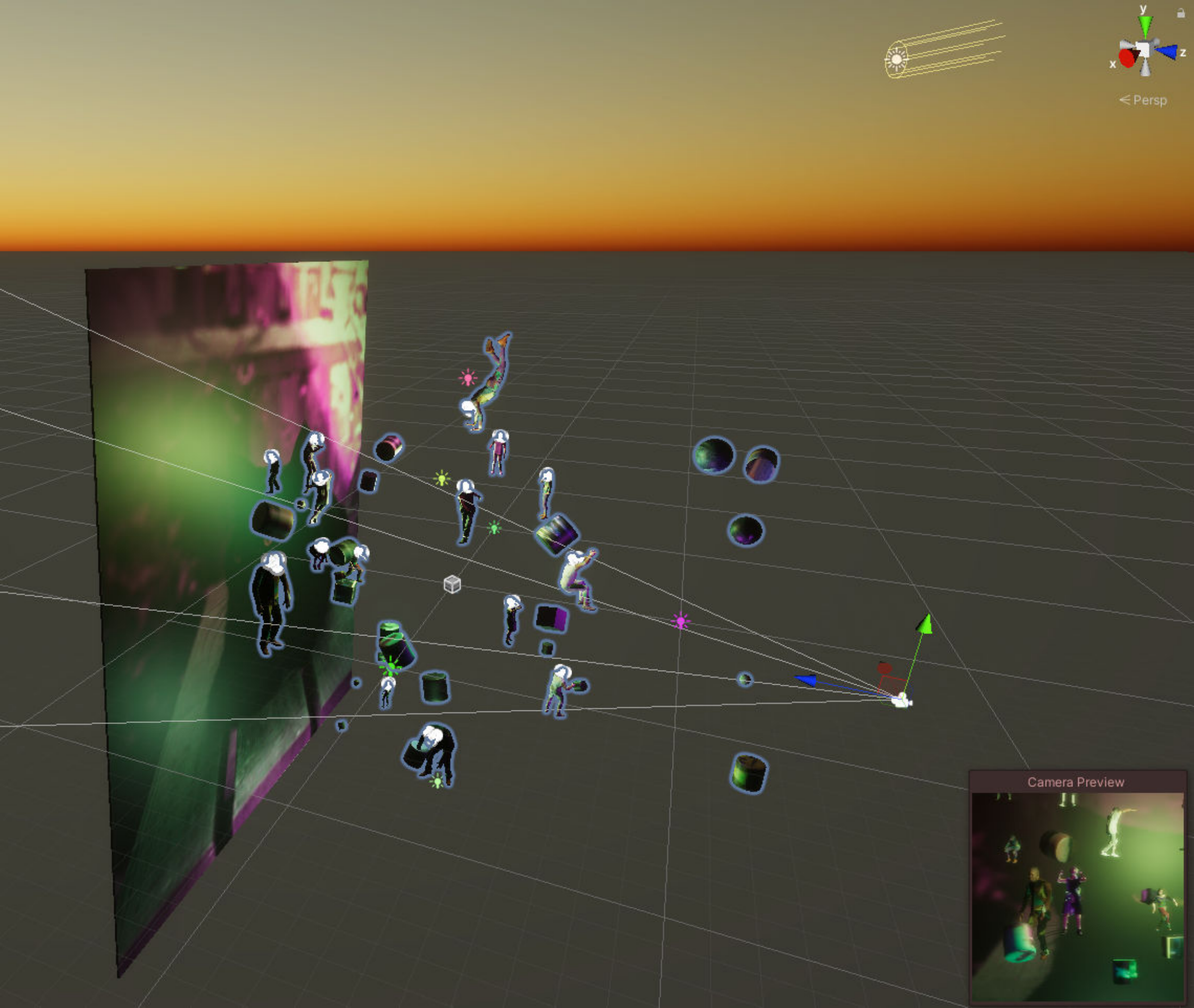}}
        \caption{ }
    \end{subfigure}
    \hfill
    \\
    \vspace{0.1cm}
    \begin{subfigure}[t]{0.49\textwidth}
        \raisebox{-\height}{\includegraphics[width=\textwidth]{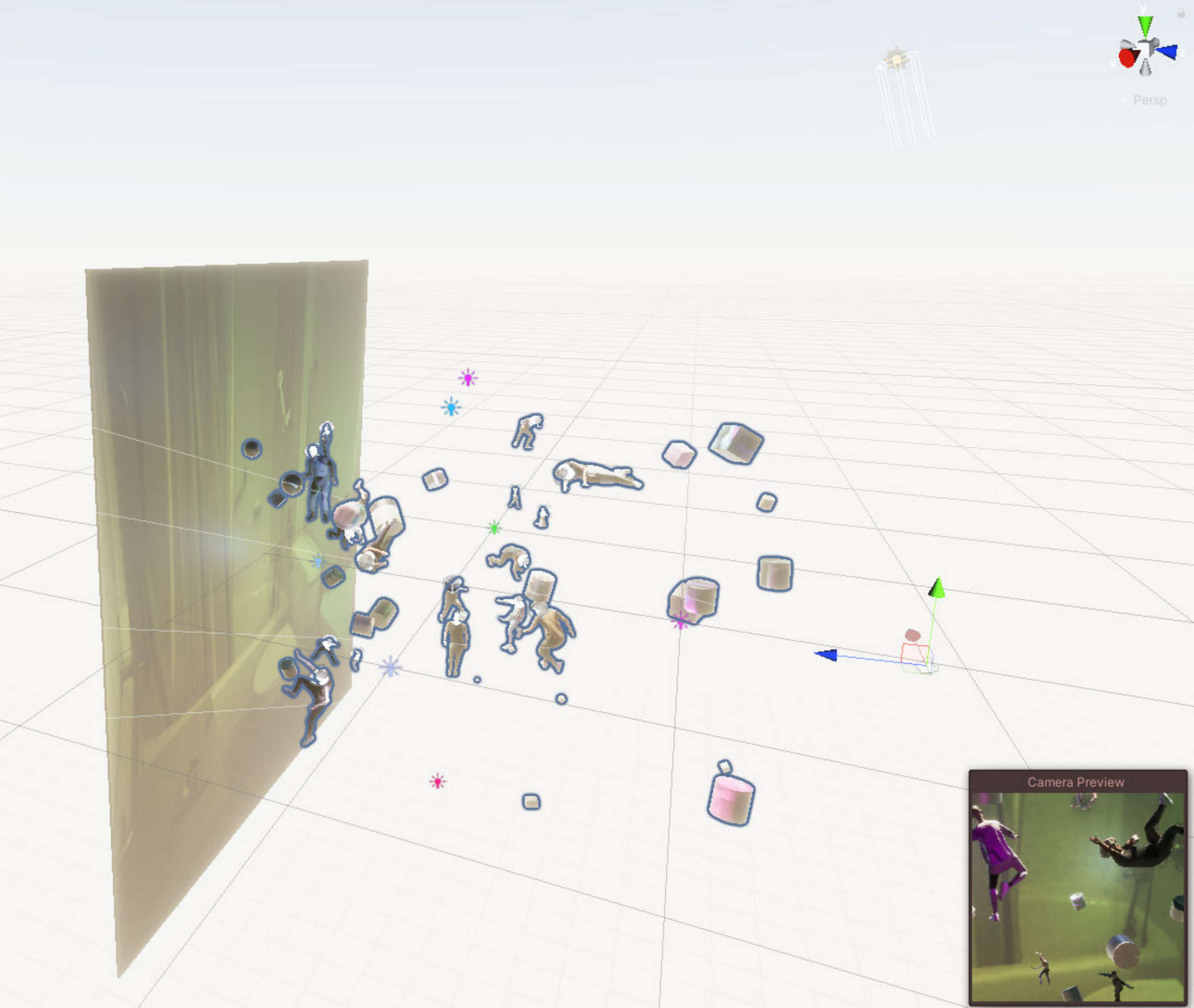}}
        \caption{ }
    \end{subfigure}
    \begin{subfigure}[t]{0.49\textwidth}
        \raisebox{-\height}{\includegraphics[width=\textwidth]{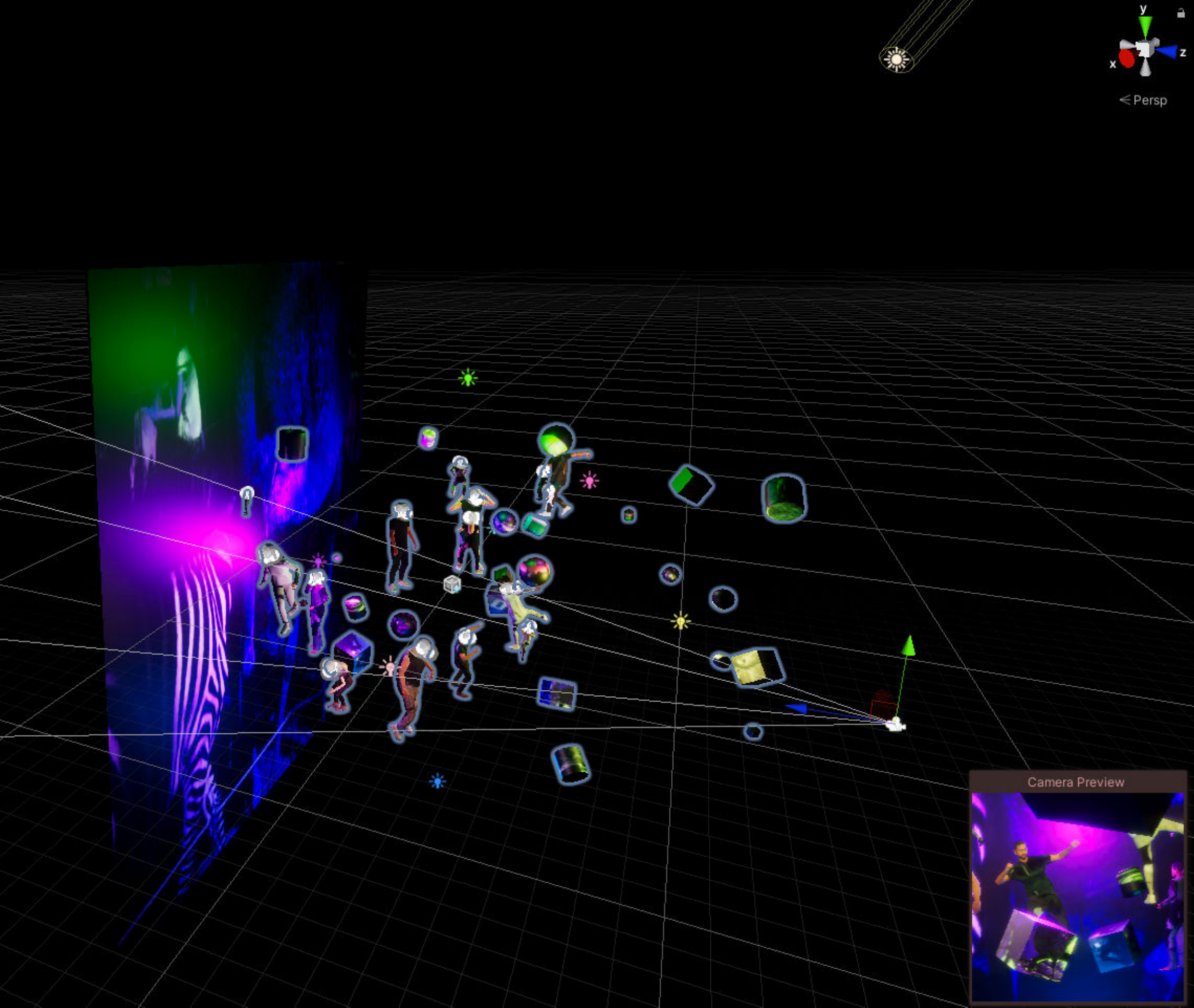}}
        \caption{ }
    \end{subfigure}
\caption{\textbf{\psp{} Design.} (a) \textbf{Scene Setup.} The scene has a background wall, a Perception camera, one directional light (the Sun), one moving point light, and six stationary point lights. (b), (c), and (d) \textbf{Example Simulations.}
The small camera preview pane on the bottom right of each figure shows a render preview from the perception camera.
We change the wall background texture, point lights color, intensity, and position, and the directional sunlight in each frame. We also change the field of view, focal length, position, and orientation of the camera. 
We spawn human assets in the scene in front of the wall with different scales, poses, clothing texture, and rotations around the $Y$-axis. Additionally, we spawn primitive occluder objects in the scene with different orientations, scales, and textures. 
}
     \label{fig:unityenv}%
\end{figure}
\label{subsec:environment}

We use Unity version \texttt{2020.3.20f1} and Unity's Perception package \texttt{0.9.0-preview.2} to develop \psp{}. In Fig.~\ref{fig:unityenv} we show our Unity environment setup. Our 3D scene comprises a background wall, a Perception camera, one directional light (the Sun), one moving point light, and six stationary scene point lights. 

\paragraph{Scene Background and Lighting}
We randomly chose a background wall texture from a set of 1600 natural images taken from the COCO unlabeled 2017 dataset. We ensured that no pictures of humans (even framed pictures of humans hung on a wall in an image) appear in these natural images. We also change the hue offset of textures.  We alter the color, intensity, and on/off state of six Point Lights and one Directional Light in our scene. In addition, we have one moving Point Light that changes position and rotation. This set of eight lights in the scene produce diverse lighting, shadows, and looks for the scene (Fig.~\ref{fig:unityenv}). 

\paragraph{Perception Camera} 
The Perception camera extends the rendering process to generate annotation labels. 
In \psp{}, for our benchmark experiments we have one object class (person) for which we produce a 2D bounding box and human keypoint labels. Using the Unity Perception package~\citep{borkman2021unity}, we can include semantic and instance segmentation masks and 3D bounding boxes.
Our 2D bounding box and human keypoints follow the COCO dataset standard.
Visibility states for our keypoint labels are; $v=0$ not labeled, $v=1$ labeled but not visible, and $v=2$ labeled and visible, similar to COCO. 
However, we do not use the $iscrowd=1$ tag since we can generate sub-pixel-perfect labels even in the most crowded scenes.

Keypoint labeling in the Unity Perception package enables fine-grained control to match the desired labeling strategy. 
This fine control is vital when handling self-occlusion. 
In the Unity Perception package, self-occlusion is determined by comparing the distance between the keypoint and the closest visible part of the object with a threshold, where keypoints too far from the front of the mesh are considered occluded. 
For the \psp{} generator with the RenderPeople assets, we empirically chose the self-occlusion distance per keypoint per model to best approximate the labeling rules for the COCO dataset. 
For visibility state ($v=0$), we set the keypoint coordinates to $(0, 0)$. When keypoints are occluded ($v=1$) and fully visible ($v=2$) we provide the keypoint coordinates. However, for the sake of simplicity, we do not use the self-occlusion labeler in our benchmark experiments, and only mark keypoints occluded by other objects to have state $v=1$. In our provided binaries and template environment, the self-occlusion labeler is enabled.


Human keypoint labeling (as with other manual labeling tasks) is more or less a subjective feat down to the opinion and accuracy of the human annotator, and the results tend to vary from one annotator to the next. Keypoint labeling in the Perception package accounts for this with fine-grained controls for tuning the labeling to match the labeling strategy used in the real dataset.

Self-occlusion is especially prone to variability. 
In the Perception Package, we determine self-occlusion by comparing the distance between the keypoint and the closest visible part of the object with a threshold, where keypoints too far from the front of the mesh are considered occluded. The user is provided several ways to set this self-occlusion distance for each keypoint, including setting a global default, specifying a certain value for a keypoint on a model, and by scaling all of the keypoints for a particular model. Through a combination of these techniques a user can tune the visibility calculations to their preferences. To approximate a realistic labeling job, we empirically chose a set of self-occlusion distances tuned for each keypoint on our human assets. 


In total the Perception camera provides the user with three choices of labeling schemes:
\begin{itemize}[leftmargin=*]
    \item \textbf{Visible objects:} this is the annotation behaviour described above.
    \item \textbf{Visible and occluded objects:} in this case if a human is occluded by itself or another object, it is annotated as visible ($v=2$). 
    \item \textbf{All objects:} in this case even objects that fall fully behind another object are also annotated. This is specifically useful for human tracking and activity recognition.
\end{itemize}

\paragraph{Objects in the Scene}
We use a set of primitive 3D game objects, e.g., cubes, cylinders, and spheres, to act as background or occluder/distractor objects. We can spawn these objects at arbitrary positions, scales, orientations, textures, and hue offsets in the scene. We use the same COCO unlabeled 2017 textures that we used for the background wall for these objects.

The last component in our 3D scene is the human assets. As with our background/occluder objects, we spawn these assets at different positions in front of the background wall and Perception camera with different poses, scales, clothing textures and hue offsets, and rotations around the $Y$-axis. In the next section, we describe how we achieve randomization across frames for these components.
\subsection{Domain Randomization}
\label{subsec:domain_randomization}
To train models using synthetic data that can generalize to the real domain, we rely on \textit{Domain Randomization}~\citep{tobin2017domain} where aspects of the simulation environment are randomized to introduce variation into the synthetic data.
The Unity Perception package provides a domain randomization framework~\citep{borkman2021unity}. 
At each frame, randomizers act on predefined Unity scene components. We first provide a parametric definition of the component that we want to randomize. Then we define how we would like those parameters distributed. We provided normal, uniform, and binomial distributions, though custom distributions can also be defined. For simplicity, all the randomizer values in \psp{} use a uniform distribution. 

In brief, we randomize aspects of the 3D object placement and pose, the texture and colors of the 3D objects in the scene, the configuration and color of the lighting, the camera parameters, and some post-processing effects. 
Certain types of domain randomizations, such as the lighting, hue offset, camera rotation/Field of View/Focal Length, mimic standard data augmentations' behavior. Hence, we do not use data augmentations during synthetic data training.
Tab.~\ref{tab:randomizers} outlines the statistical distributions for our randomizer parameters.


\subsection{Data Generation}
\label{generation}

We provided binary builds of \psp{} for macOS and Linux systems. On a MacBook Pro (16-inch, 2019) with 2.3 GHz 8-Core Intel Core i9, AMD Radeon Pro 5500M 4 GB, Intel UHD Graphics 630 1536 MB, and 32 GB 2667 MHz DDR4 Memory, \psp{} generates $10\times10^3$ images, bounding boxes, and keypoint labels in approximately $3$ minutes. This time includes the time to write the data to disk.


\section{Experiments}
\label{sec:experiments}
We analyzed the dataset statistics of our domain randomized synthetic data, generated using na\"ive parameters and compared them to the COCO-person train dataset. We then used this synthetic data to train a Dectectron2 Keypoint R-CNN variant on person and keypoint detection. We then trained the model on various amounts of
real-world data to establish a 
set of baselines for simulation-to-real (sim2real) transfer learning for human-centric computer vision.

\subsection{Dataset Statistics}
\label{subsec:statistics}


We generate a synthetic dataset using domain randomization parameters that we chose na\"ively to cover a wide range of variations for our 3D scene components. 
We performed an extensive statistical analysis to understand how these parameters affect the generated data and compare those statistics to real-world data. 
We considered the following categories: high-level dataset features; bounding box placement, size, and number in the generated images; keypoint number per image and instance; and lastly, the variations in the human pose.

For our benchmark experiments we generated a training dataset with \num[group-separator={,}]{490000} images with bounding box and human keypoint annotations. There are more than \num[group-separator={,}]{3070000} person instances in our dataset out of which approximately \num[group-separator={,}]{2900000} have annotated keypoints.
The entire COCO person dataset has \num[group-separator={,}]{64115} images with \num[group-separator={,}]{262465} person instances, out of which \num[group-separator={,}]{149813} have keypoint annotations. The JTA train dataset~\citep{fabbri2018jta} (which is a dataset of human characters walking in the GTA V game) has \num[group-separator={,}]{230400} images with \num[group-separator={,}]{5176685} person instances, out of which \num[group-separator={,}]{5176685} have keypoint annotations.

\begin{figure}[htbp] 
    \centering
    \hspace{1.2cm}
    \hfill
    \begin{subfigure}[t]{0.25\textwidth}
        \includegraphics[width=\textwidth, trim={0.2cm 0.3cm 0.1cm 0.1cm}, clip]{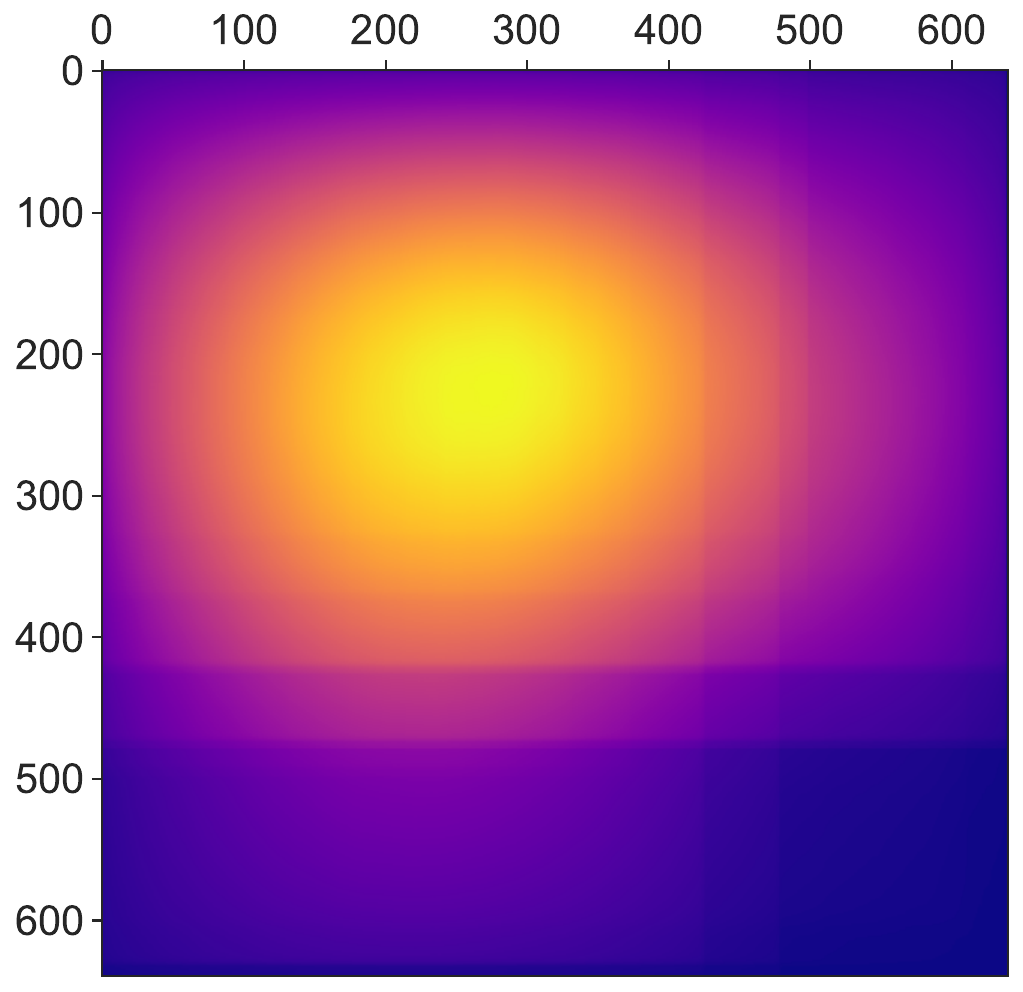}
        \caption{COCO}
    \end{subfigure}
    \hfill
    \begin{subfigure}[t]{0.25\textwidth}
        \includegraphics[width=\textwidth, trim={0.2cm 0.3cm 0.1cm 0.1cm}, clip]{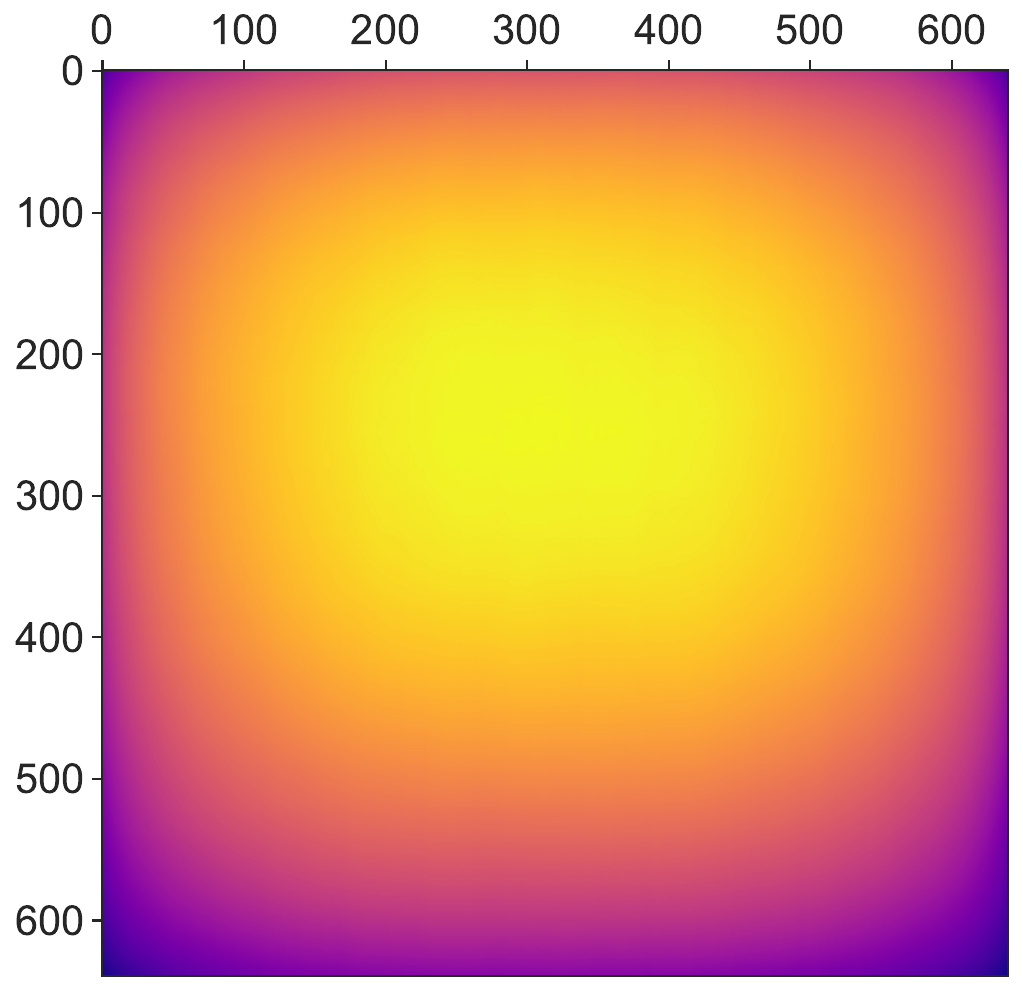}
    \caption{Synthetic} 
    \end{subfigure}
    \hfill
    \begin{subfigure}[t]{0.18\textwidth}
        \includegraphics[width=0.2\textwidth, trim={0.2cm 0.2cm 0.2cm 0.2cm}, clip]{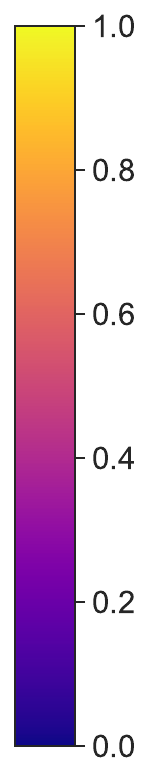}
    \end{subfigure}
    \hfill
    \caption{\textbf{Bounding Box Occupancy Heatmap.} For our benchmark experiments, we use an image size of $640\times640$. We overlay all the bounding boxes, using filled boxes, on the image to compute the bounding box occupancy map for both COCO-person and synthetic data. We use the Foreground Object Placement Randomizer (Tab.~\ref{tab:randomizers}) to control the placement of our 3D human assets.}
    \label{fig:bboxheatmap}
\end{figure}

To quantify the effect of our human and occluder placement and camera randomizers on the produced annotations, we plotted heatmaps of bounding box locations for both the synthetic data and COCO data (Fig.~\ref{fig:bboxheatmap}). Note that our human, occluder, and camera placements are sampling from a uniform distribution in 3D space. Additionally, our camera randomizers sample the camera focal length and field of view from a uniform distribution. All of these parameters affect the final visibility of the instances in the 2D image space. We observe more bounding boxes in the center of the final images; however, the instances spread to the very edge of the images. 
For the COCO dataset, we overlay all the bounding boxes on a $640\times640$ image frame. Since there are many portrait and landscape images in COCO, we observe oblong bounding box distributions tailing along with the image's height and width. We find the majority of the boxes near the center of most images and less spreading to the edges of the images.

\begin{figure}[htb] 
    \centering
    \begin{subfigure}[t]{0.49\textwidth}
        \includegraphics[width=\textwidth, trim={0.4cm 0.4cm 0.4cm 0.4cm}, clip]{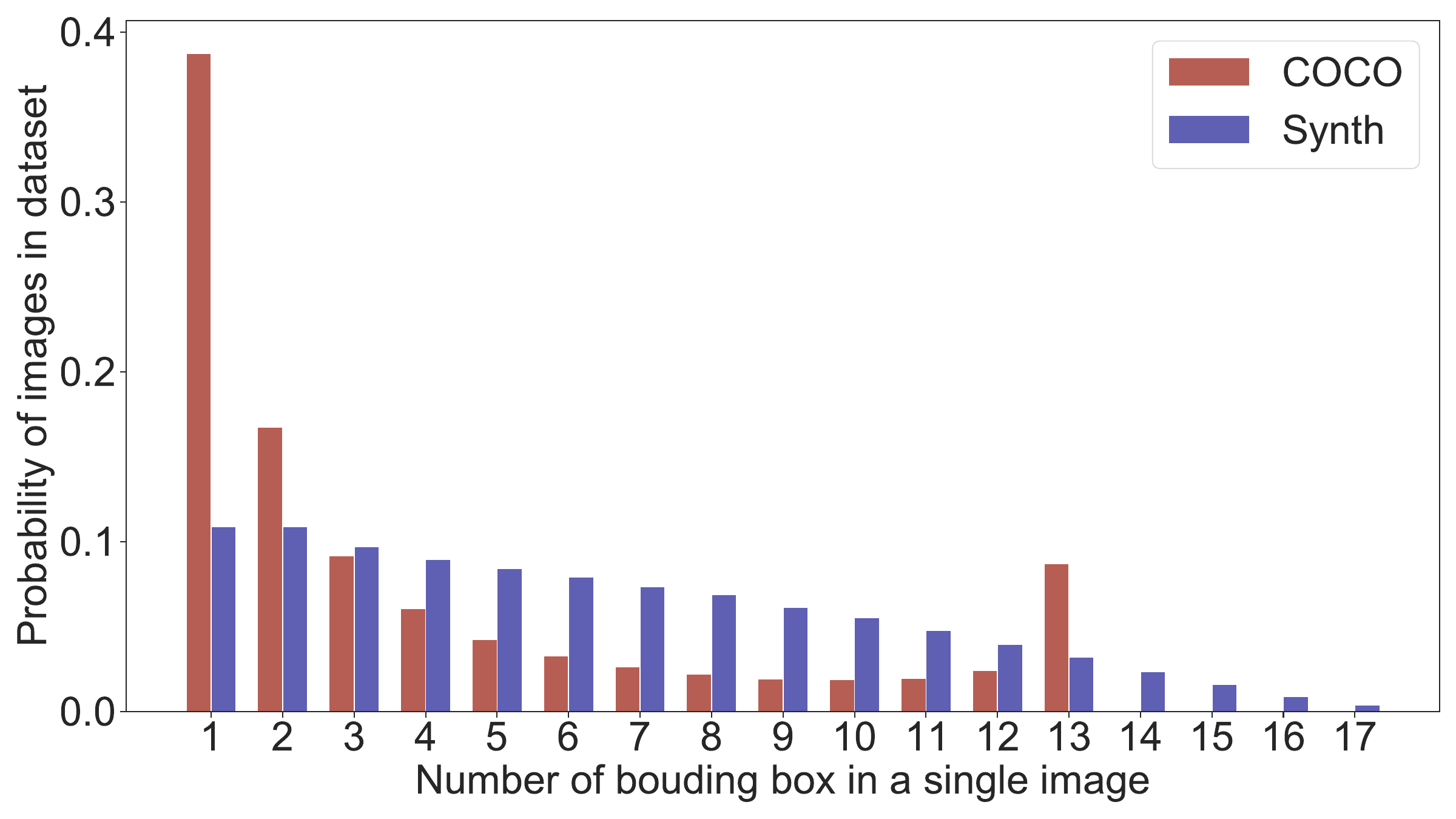}
        \caption{}
        \label{subfig:bbox-on-single-img}
    \end{subfigure}
    \begin{subfigure}[t]{0.49\textwidth}
        \includegraphics[width=1\textwidth, trim={0.4cm 0.4cm 0.4cm 0.4cm}, clip]{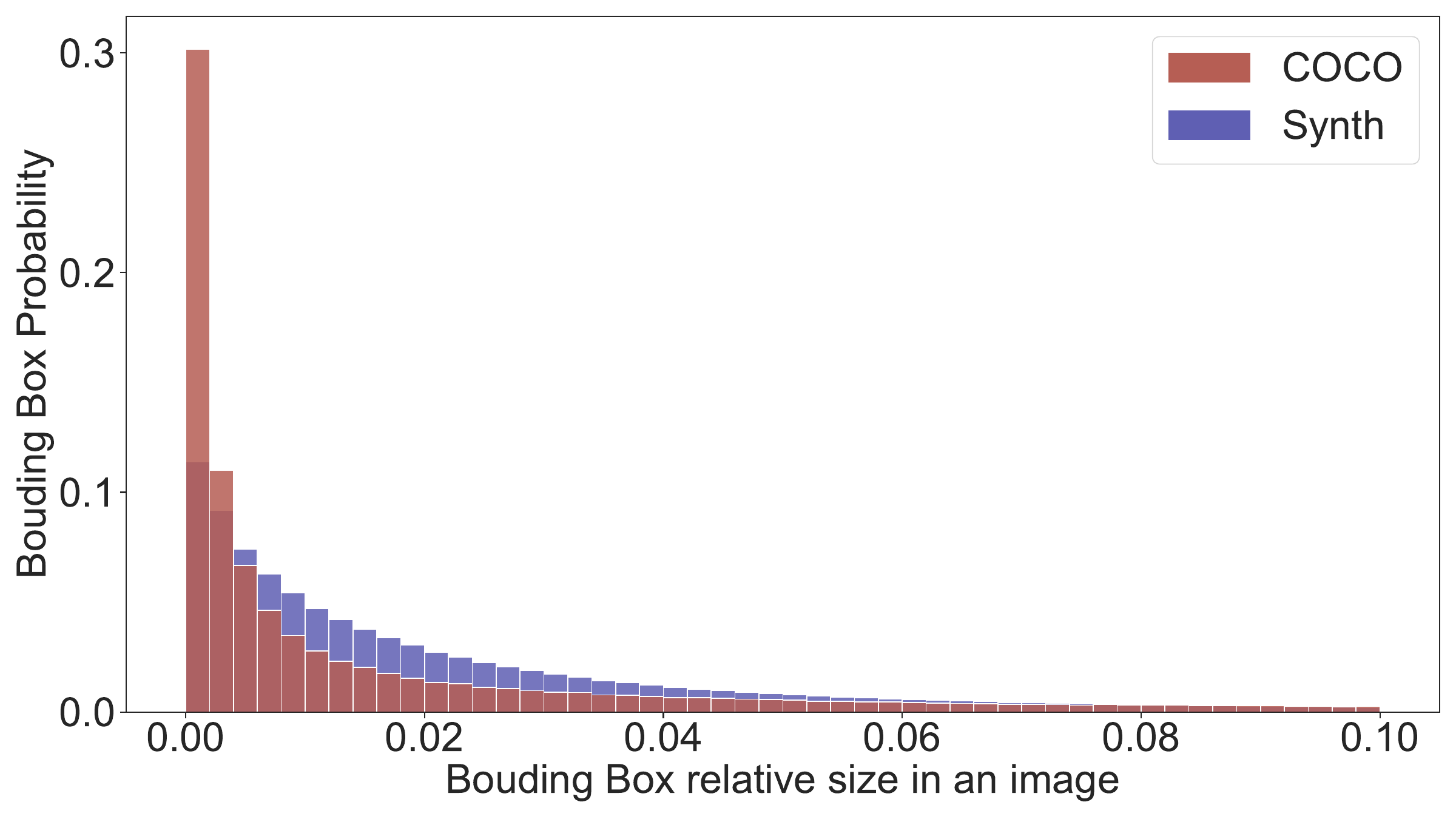}
        \caption{}
        \label{subfig:bbox-size-dist}
    \end{subfigure}
    \hfill
    \\
    \vspace{0.1cm}
    \begin{subfigure}[t]{0.49\textwidth}
        \includegraphics[width=1\textwidth, trim={0.4cm 0.4cm 0.4cm 0.4cm}, clip]{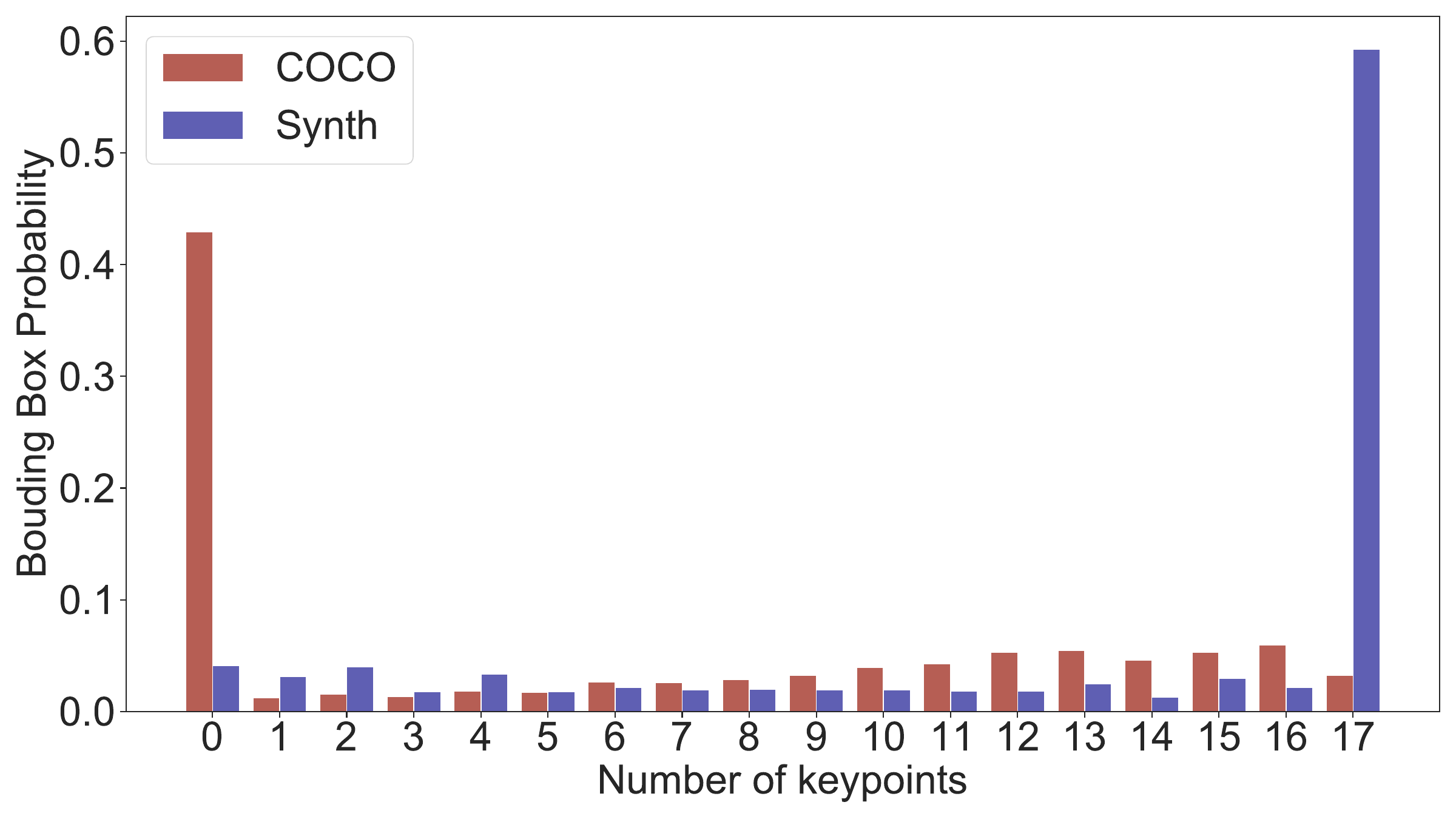}
        \caption{}
        \label{subfig:kp-per-bbox}
    \end{subfigure}
    \begin{subfigure}[t]{0.49\textwidth}
        \includegraphics[width=1\textwidth, trim={0.4cm 0.4cm 0.4cm 0.4cm}, clip]{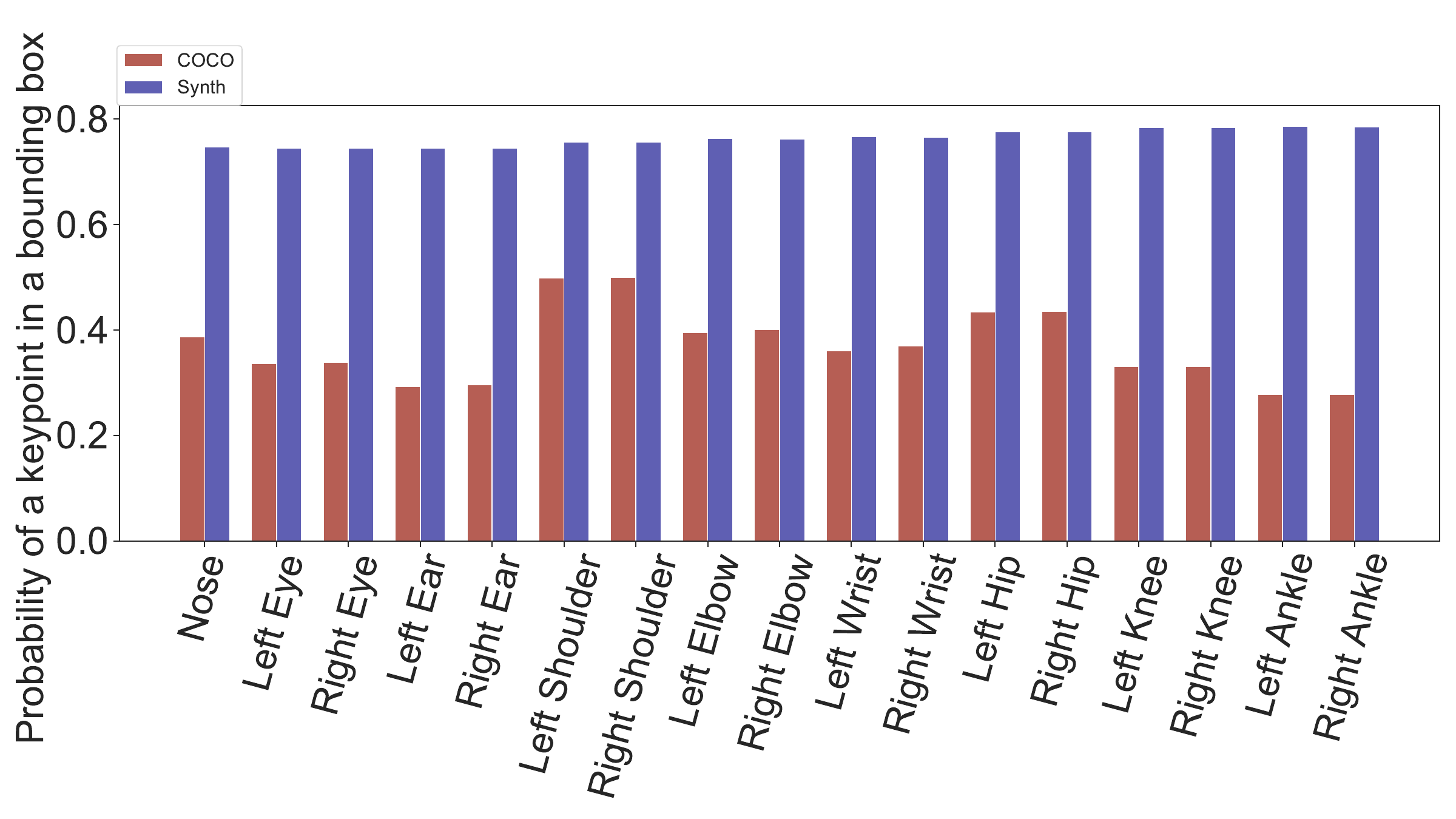}
        \caption{}
        \label{subfig:vis-per-keypoints}
    \end{subfigure}

    
    \caption{\textbf{Bounding Box and Keypoint Statistics.} All COCO statistics computed for COCO-person only; all Synth data generated with \psp{} using default parameters (Tab.~\ref{tab:randomizers}) \subref{subfig:bbox-on-single-img}) \textbf{Number of Bounding Boxes per Image.} \subref{subfig:bbox-size-dist}) \textbf{Bounding Box Size Relative to Image Size}. Here, $\text{relative size} = \sqrt{\frac{\text{bounding box occupied pixels}}{\text{total image pixels}}}$. 
   \subref{subfig:kp-per-bbox}) \textbf{Annotated Keypoints per Bounding Box.} \subref{subfig:vis-per-keypoints}) \textbf{Fraction of Keypoints Per Bounding Box.} The likelihood that a keypoint is annotated for a given bounding box. All annotated keypoints are counted whether visible or occluded.}
   \label{fig:bbox_kpt_compare}
\end{figure}

Next, we analyzed the bounding box and keypoint annotations (Fig.~\ref{fig:bbox_kpt_compare}. 
We see that our synthetic dataset contains more instances (bounding boxes) per image than COCO (Fig~\ref{fig:bbox_kpt_compare}a). 
Image sizes vary a lot in the COCO dataset; therefore, we used the largest COCO image size ($640\times640$ pixels) for all images in our synthetic dataset. 
As a result, the relative bounding box size in the image for our dataset appears to be smaller; however, the bounding boxes in COCO tend to occupy less of the total image area than our synthetic data (Fig.~\ref{fig:bbox_kpt_compare}b). 
Also, more keypoints are annotated per bounding box instances in the synthetic dataset than COCO (Fig.~\ref{fig:bbox_kpt_compare}c), and they are more likely to have an annotation for a specific keypoint in the synthetic data. Lastly, the distribution of individual keypoint annotations is more homogeneous than in the COCO-person dataset (Fig.~\ref{fig:bbox_kpt_compare}d). 

To compare between \psp{} and JTA dataset, see Fig.~\ref{fig:bbox_kpt_compare_jta} and Fig.~\ref{fig:bboxheatmapJTA}. In brief, we find that the JTA dataset has similar diversity of bounding box placements, smaller bounding boxes, and more bounding boxes per image than data generated with na\"ive parameters with \psp{}.

\begin{figure}[htb] 
    \centering
    \begin{subfigure}[t]{0.18\textwidth}
        {\includegraphics[height=2.5cm]{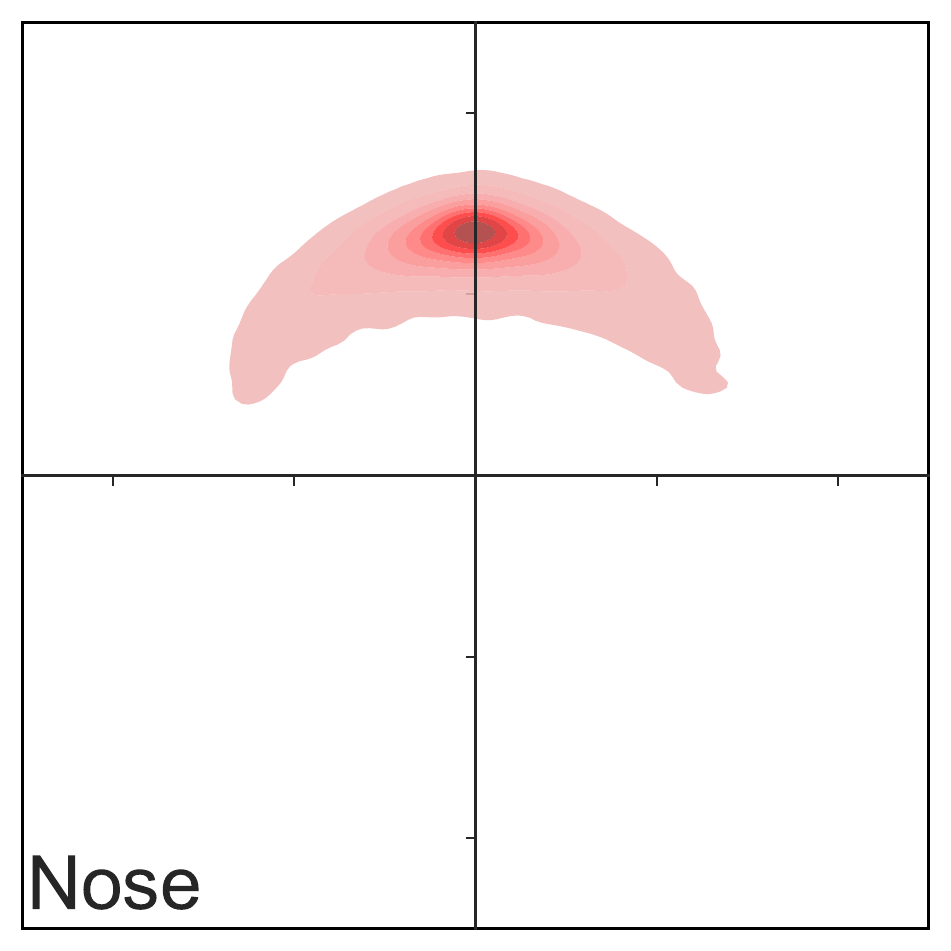}}
    \end{subfigure}
        \begin{subfigure}[t]{0.18\textwidth}
        {\includegraphics[height=2.5cm]{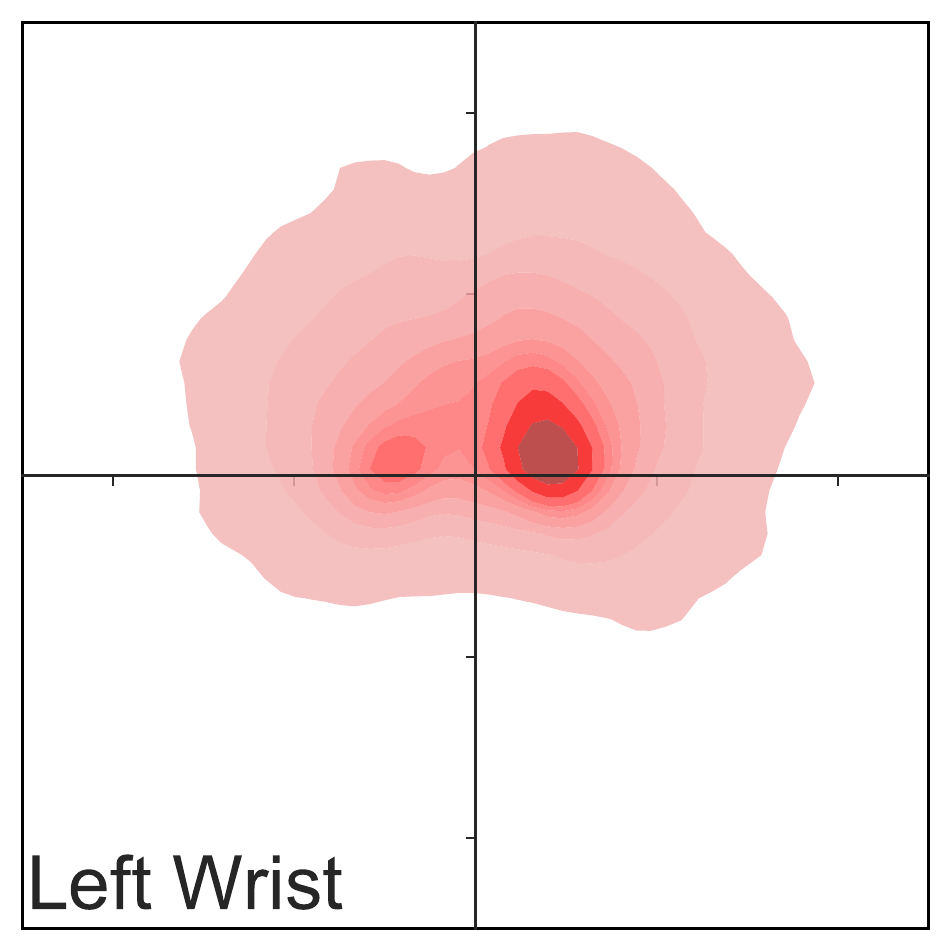}}
    \end{subfigure}
        \begin{subfigure}[t]{0.18\textwidth}
        {\includegraphics[height=2.5cm]{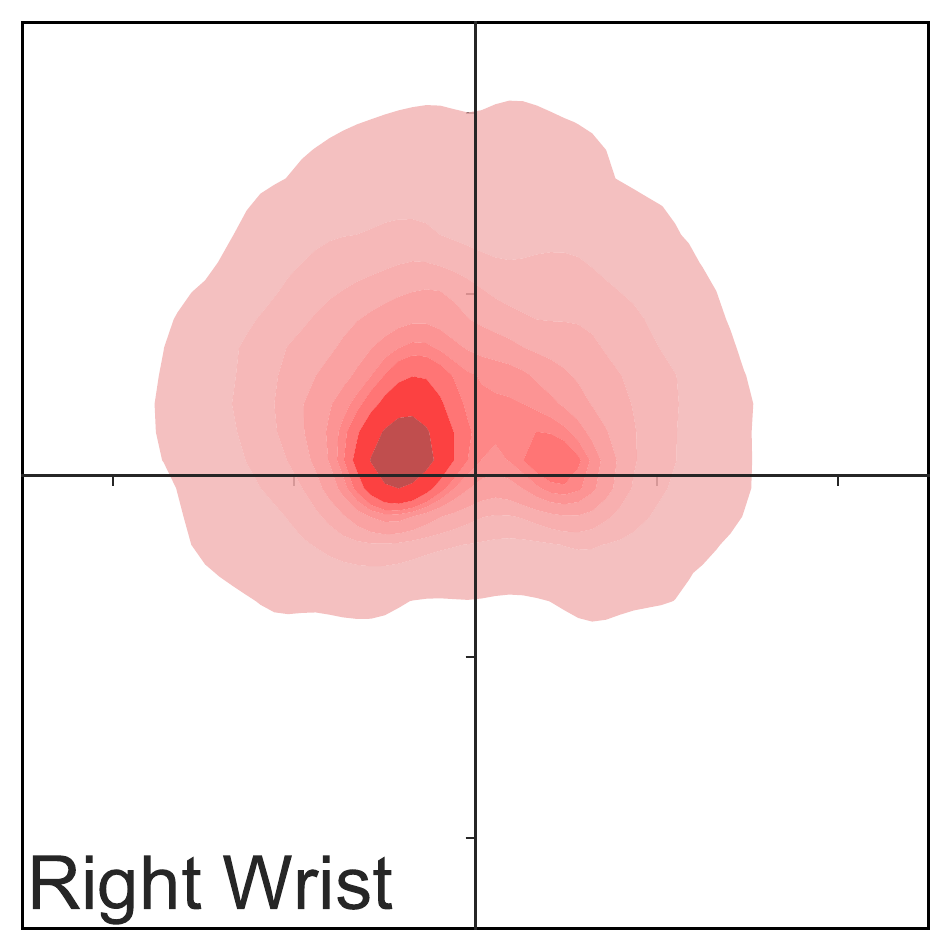}}
    \end{subfigure}
        \begin{subfigure}[t]{0.18\textwidth}
       {\includegraphics[height=2.5cm]{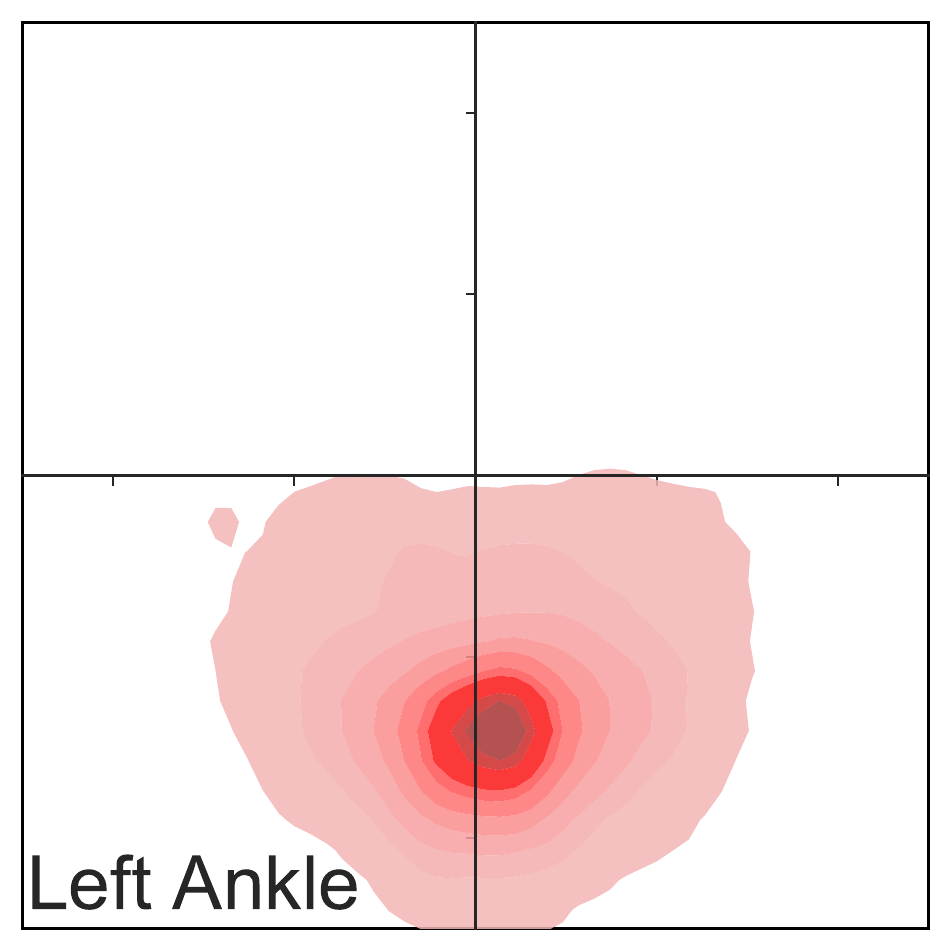}}
    \end{subfigure}
        \begin{subfigure}[t]{0.18\textwidth}
        {\includegraphics[height=2.5cm]{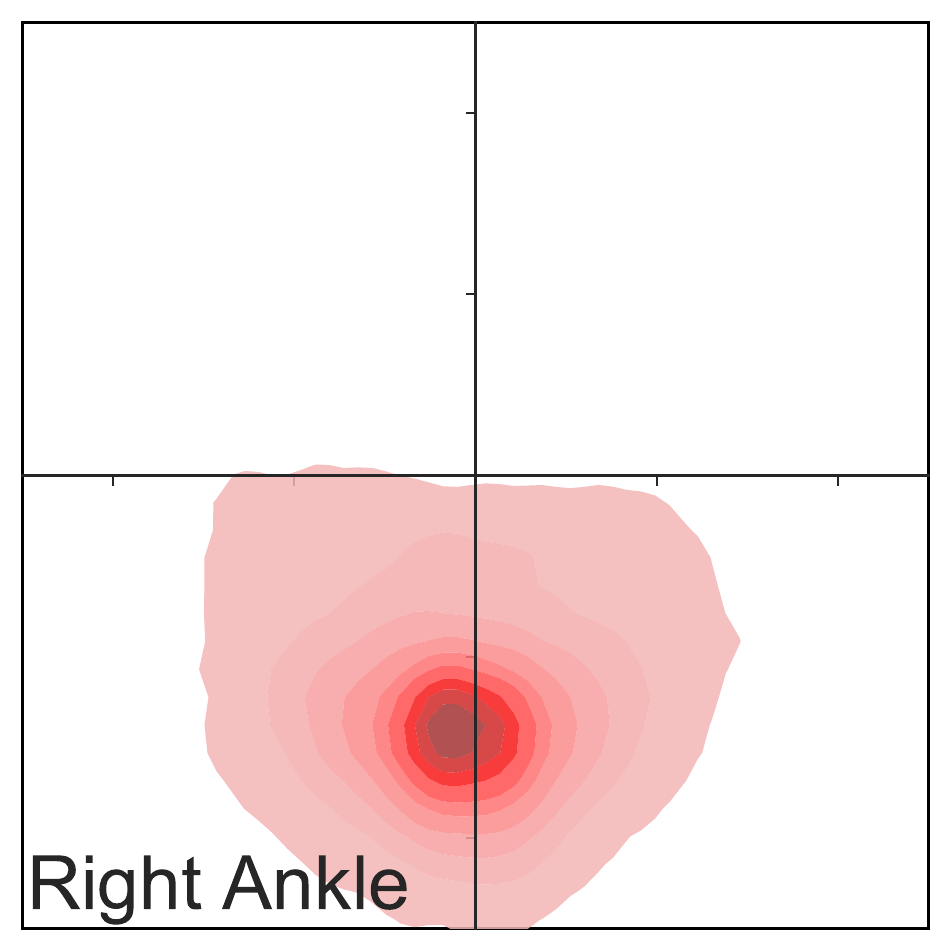}}
    \end{subfigure}
    \begin{subfigure}[t]{0.05\textwidth}
        {\includegraphics[height=2.5cm]{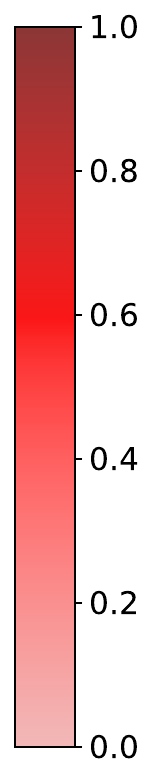}}
    \end{subfigure}
    \\
    \begin{subfigure}[t]{0.18\textwidth}
        {\includegraphics[height=2.5cm]{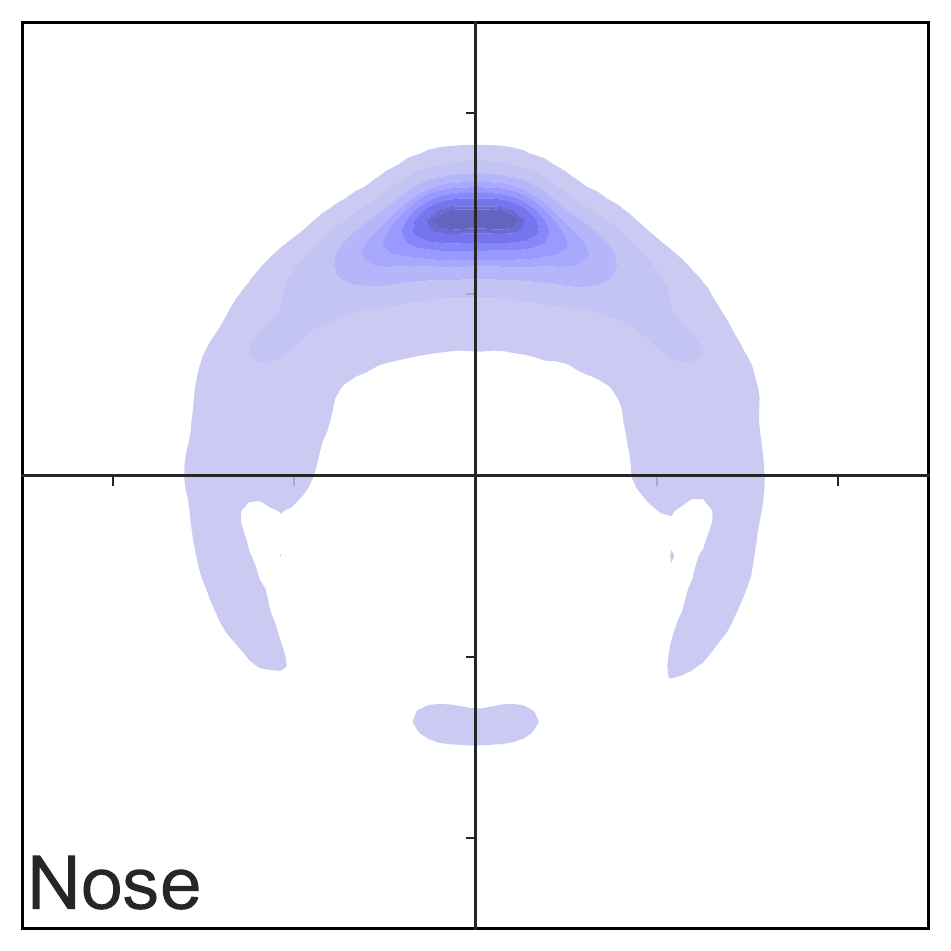}}
    \end{subfigure}
        \begin{subfigure}[t]{0.18\textwidth}
        {\includegraphics[height=2.5cm]{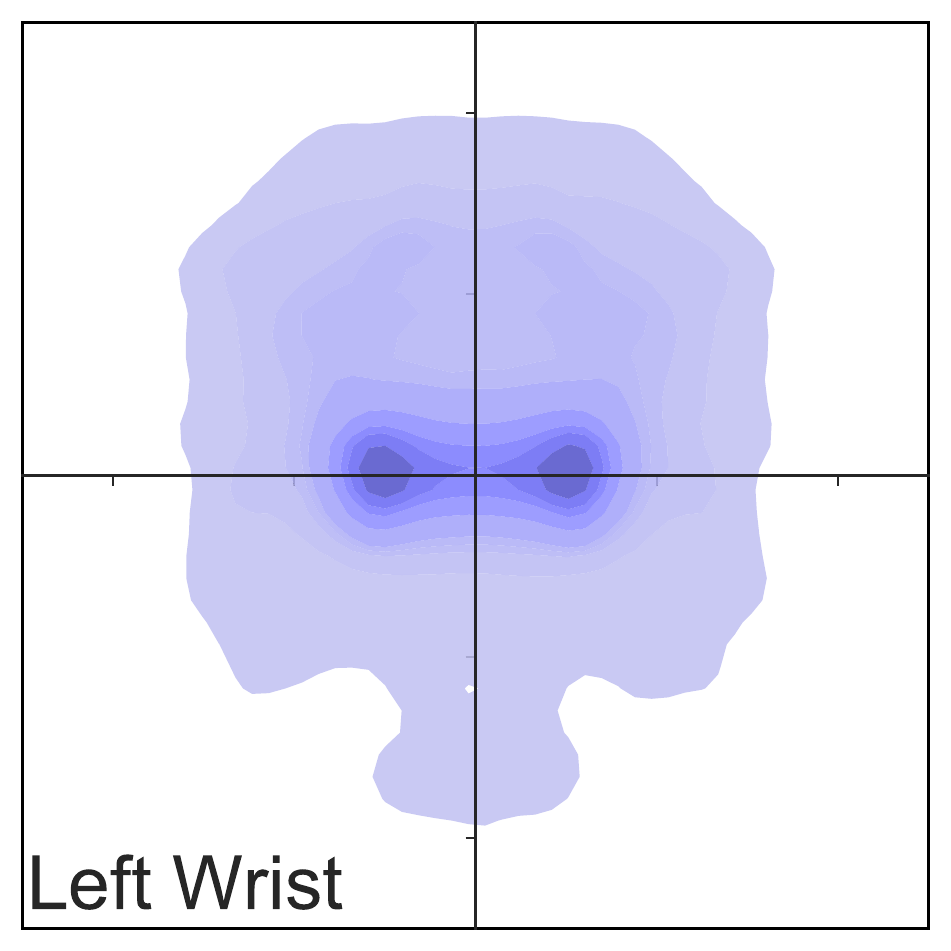}}
    \end{subfigure}
        \begin{subfigure}[t]{0.18\textwidth}
        {\includegraphics[height=2.5cm]{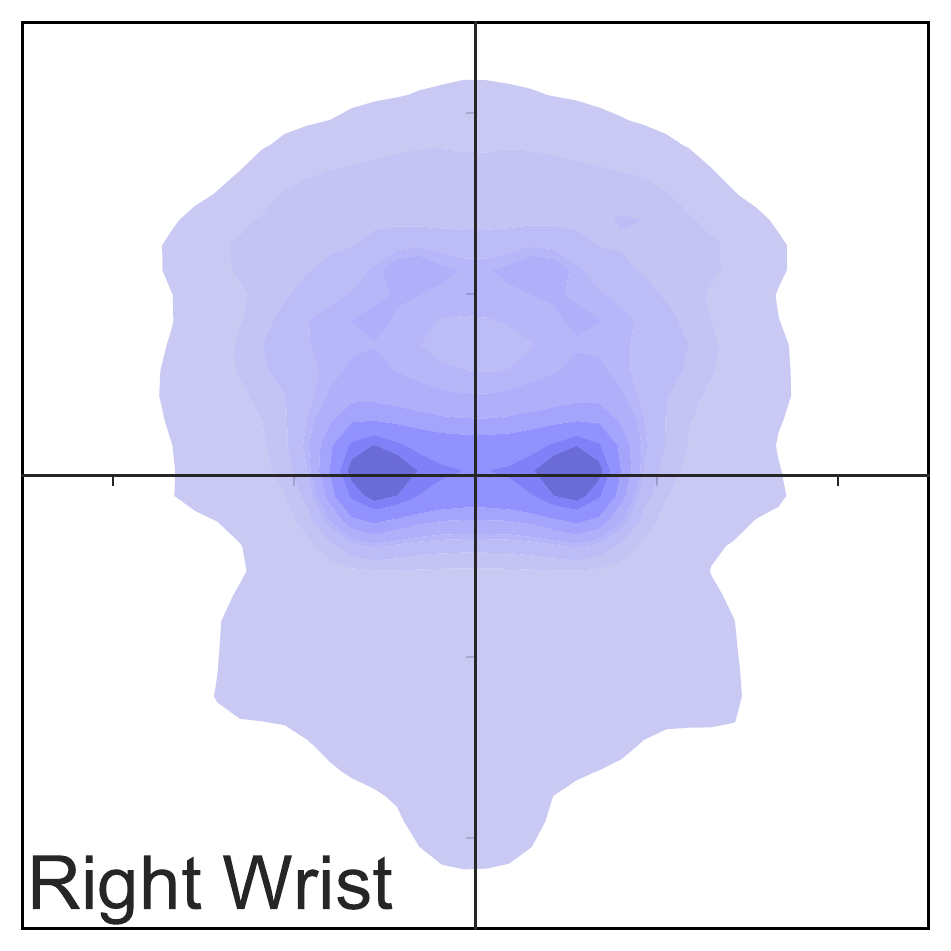}}
    \end{subfigure}
        \begin{subfigure}[t]{0.18\textwidth}
        {\includegraphics[height=2.5cm]{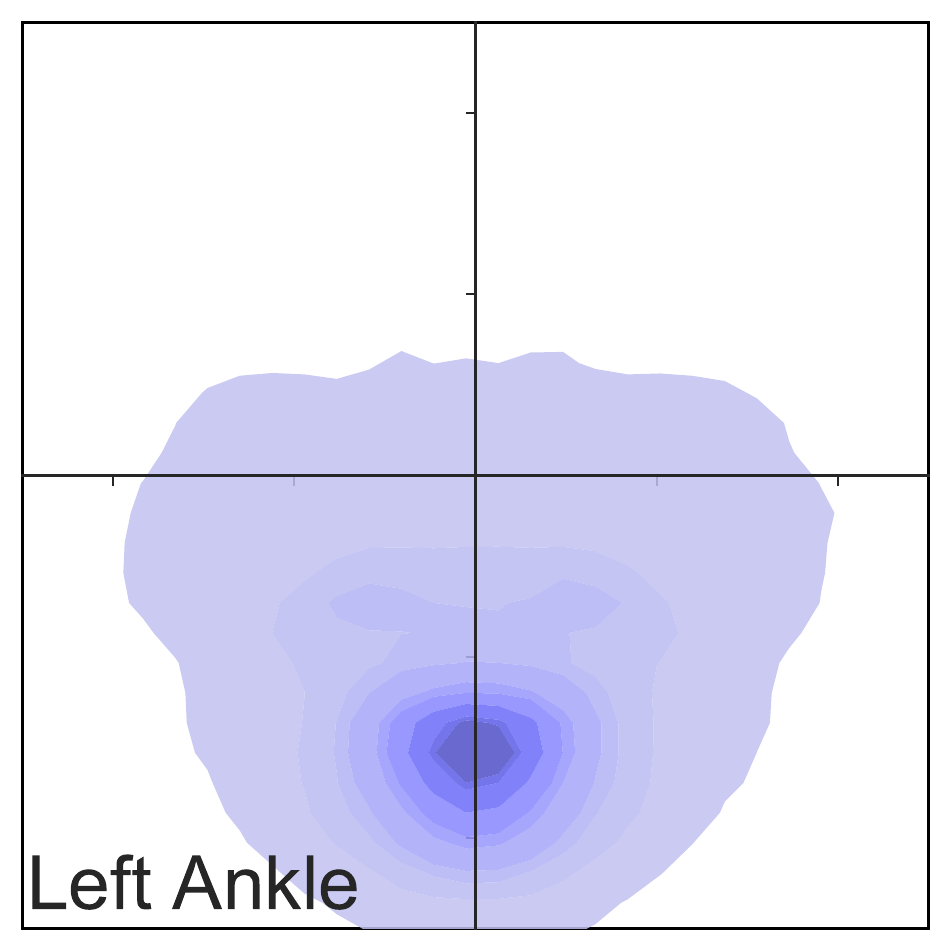}}
    \end{subfigure}
    \begin{subfigure}[t]{0.18\textwidth}
        {\includegraphics[height=2.5cm]{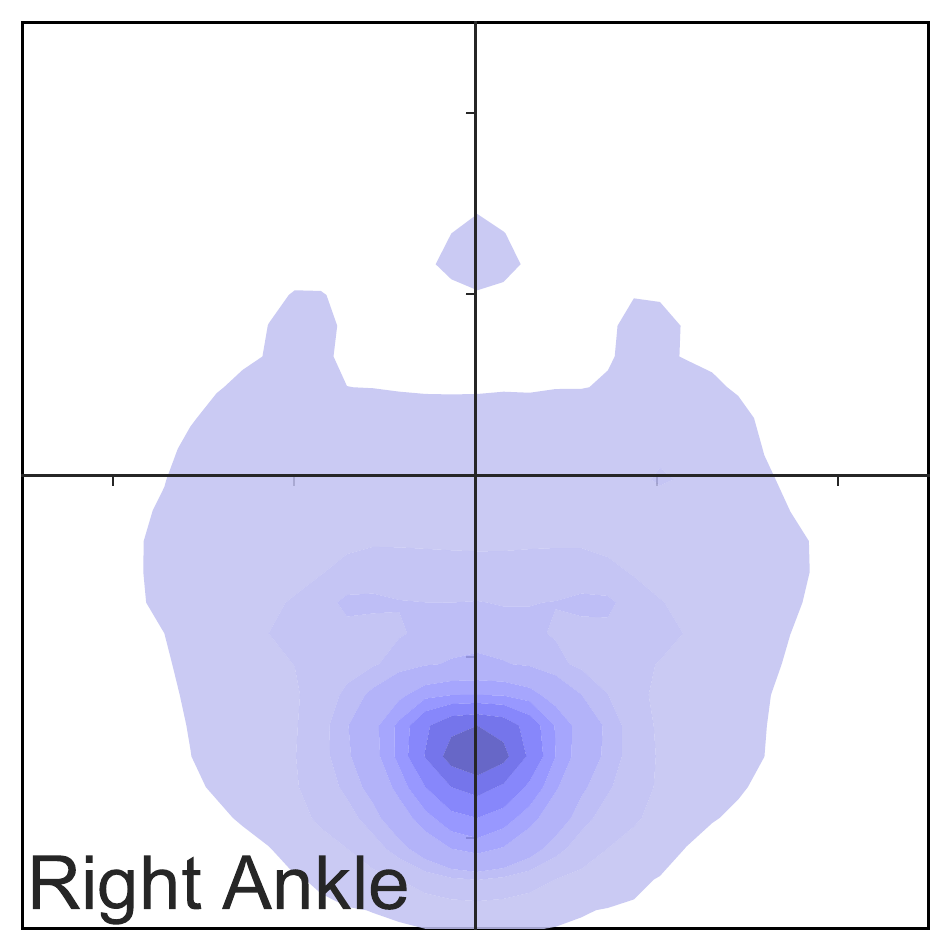}}
    \end{subfigure}
    \begin{subfigure}[t]{0.05\textwidth}
        {\includegraphics[height=2.5cm]{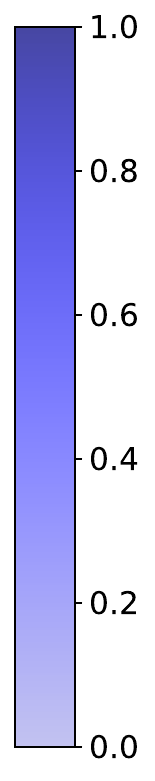}}
    \end{subfigure}
    \caption{\textbf{Five Representative Keypoint Location Heatmaps}. Top row: COCO-person. Bottom row: synthetic data generated with default parameters (Tab.~\ref{tab:randomizers}). We aligned all keypoints according Alg.~\ref{alg:pose_analysis} to produce normalized keypoint locations. We use the animation randomization to control the generated human pose diversity. 
    For heatmaps of all of the keypoints refer to Fig.~\ref{fig:posestatsall}.}
    \label{fig:posestatselect}
\end{figure}

To vary the pose of each human model, we chose a set of animations derived from motion capture clips to create a reasonably diverse set of poses. To quantify the pose diversity created with the provided animations, we use the keypoint annotations from all the instances in the COCO and synthetic datasets, with annotated hip and shoulder keypoints, whether occluded or visible. For these instances, we calculate the mid-hip point and translate all points such that the mid-hip falls at $(0, 0)$. Then we measure the distances between the left-hip and left-shoulder and the right-hip and right-shoulder and use their average to scale all other keypoints (Alg.~\ref{alg:pose_analysis}), giving all the person instances roughly the same skeletal distances. We used the translated and scaled keypoints to create the heatmap plots of each keypoint (Fig~\ref{fig:posestatselect} and~\ref{fig:posestatsall}). Heatmaps are created from the entire datasets and normalized according to the size of the datasets for comparison. From these heatmaps, we see that: 1) the distribution of synthetic dataset poses encompass the distribution of poses in COCO; 2) the distribution of our synthetic poses is more extensive than for COCO, and 3) in COCO, most people are front-facing, leading to an asymmetry with \textit{``handedness''} in the density of points which is absent in the synthetic data.

To compare between the \psp{} and JTA dataset pose heatmaps see Fig.~\ref{fig:posestatsjta}. We find that the JTA dataset has no ``handedness". However, we can see that the poses present in the JTA dataset are less diverse than COCO and much less diverse than what \psp{} can achieve. The lack of pose diversity in JTA dataset is not surprising since it is a dataset of people walking in GTA V game. Additionally, the JTA dataset is a fixed set of sequences, so we cannot easily create additional pose diversity.

\subsection{Training}
\label{subsection:training}
For our benchmarking experiments, we use the Detectron2 Keypoint R-CNN \texttt{R50-FPN} variant~\citep{he2017mask} with ResNet-50~\citep{he2016deep} plus Feature Pyramid Network (FPN)~\citep{lin2017feature} backbones\footnote{Model configuration taken from \url{https://github.com/facebookresearch/detectron2/blob/master/configs/COCO-Keypoints/keypoint_rcnn_R_50_FPN_3x.yaml}}.
We trained our models from scratch (without using the pre-trained ImageNet~\citep{deng2009imagenet} or pre-trained COCO~\citep{lin2014microsoftcoco} weights) and followed the recommendations for training from scratch~\citep{wu2019detectron2} including training with Group Normalization (GN) \citep{wu2018group}. 

Motivated by previous work~\citep{he2019rethinking} we use a learning rate annealing strategy for all our models, where we reduce the learning rate when the validation keypoint AP metric has stopped improving. Our models benefited from reducing the learning rate by a factor of $10\times$ once learning has stagnated based on a certain threshold (epsilon) for a certain number of epochs (patience period). We reduce the learning rate every time the patience period ends, and halve epsilon and the next patience period. 
We perform the learning rate reduction three times for all our models. Every time we reduce the learning rate, we revert the model iteration to the checkpoint that achieves the highest metrics on the validation. Thus we ensure that the last model checkpoint is also the best performing model.

We set the initial learning rate for all models to $0.02$, the initial patience to $38$, and the initial epsilon to $5$. The weight decay is $0.0001$, and momentum is $0.9$. We perform a $linear$ warm-up period of $1000$ iterations at the start of training (both for training from scratch and transfer learning), where we slowly increase the learning rate to the initial learning rate. We use $8$ NVIDIA Tesla V100 GPUs using synchronized SGD with a mini-batch size of $2$ images per GPU. We use the mean pixel value and standard deviation from ImageNet for our image normalization in the model. We do not change the default augmentations used by Detectron2. We perform the evaluation every two epochs. This affects the total number of iterations, the patience period, and learning rate scheduling periods. We also fix the model seed to improve reproducibility. 

For our real-world dataset, we use the COCO 2017 person keypoints training and validation sets~\citep{lin2014microsoftcoco}. We split the COCO training set into overlapping subsets that contain 
$641$, 
$6411$, $32057$,
, and $64115$ 
images, that contain $1\%$, $10\%$, $50\%$, and $100\%$ of the COCO train set respectively to study few-shot transfer. 
The smaller set is a subset of the larger set.
We use the person COCO validation set for evaluation during training. We report our final model performance for all the COCO training data experiments with the COCO test-dev2017 dataset. We generated $3$ datasets of $500\times10^3$ images from $3$ random seeds for our synthetic datasets. We split them into $490\times10^3$ training and $10\times10^3$ validation sets. We use the synthetic validation set to evaluate the model during training from scratch on purely synthetic data. After training, we report the performance of these models using the person COCO validation and test-dev2017 sets. 

For our benchmark experiments, we first train our models from scratch and evaluate their performance on the COCO validation set (See Appendix) and COCO test-dev2017 set. Second, we use the weights of the models trained on synthetic data and fine-tune them on limited COCO (real) data subsets for few-shot transfer. For a complete comparison we also use ImageNet pre-trained weights and fine-tune on COCO data subsets. During the few-shot transfer learning training, we re-train all the network layers. The hyperparameters and learning rate schedules are the same for both models trained from scratch and few-shot transfer learning.
\section{Results}
\label{results}

\begin{table}[htb]
\caption{\textbf{Keypoint Test Metrics for Models Trained from Scratch.}
We trained all models from randomly-initialized weights and evaluated them on the COCO test-dev2017 set.
We report the mean and standard deviation of the results from three synthetic datasets generated from three different seeds. 
The highest metrics in each category are in boldface. 
}
\begin{adjustbox}{max width=\textwidth}
\centering
\begin{tabular}{crrrlllll}
\toprule
              data &     dataset size &                         training steps & no. of epochs &               AP &              $\textrm{AP}^{\textrm{\textit{IoU=.50}}}$ &             $\textrm{AP}^{\textrm{\textit{IoU=.75}}}$ &               $\textrm{AP}^{\textrm{\textit{large}}}$ &              $\textrm{AP}^{\textrm{\textit{medium}}}$                \\
\midrule
\parbox[t]{2mm}{\multirow{4}{*}{\rotatebox[origin=c]{90}{COCO}}} & 641 &                         5280 & 132 &             6.40 &             20.30 &             2.40 &             7.90 &             5.60      \\
              & 6411 &                        40000 & 100 &            37.30 &             67.60 &            35.60 &            43.80 &            33.30      \\
              & 32057 &                       168252 & 84 &            55.80 &             82.00 &            60.60 &            64.20 &            50.70      \\
              & 64115 &                       577008 & 144 &            \textbf{62.00} &             \textbf{86.20} &            \textbf{68.10} &            \textbf{70.50} &            \textbf{56.70}      \\
\midrule
\parbox[t]{2mm}{\multirow{4}{*}{\rotatebox[origin=c]{90}{Synth}}} & $\text{4.9}\times10^3$ &        [38556, 53244, 38556] & [126, 174, 126] &  1.83 $\pm$ 0.17 &   4.13 $\pm$ 0.34 &  1.30 $\pm$ 0.28 &  2.17 $\pm$ 0.12 &  2.07 $\pm$ 0.21      \\
              & $\text{49}\times10^3$ &     [391936, 594028, 398060] & [128, 194, 130] &  \textbf{4.87 $\pm$ 0.09} &  \textbf{10.20 $\pm$ 0.08} &  \textbf{4.13 $\pm$ 0.21} &  \textbf{5.40 $\pm$ 0.49} &  \textbf{5.77 $\pm$ 0.45}      \\
              & $\text{245}\times10^3$ &  [2143680, 1745568, 2388672] & [140, 114, 156] &  4.33 $\pm$ 0.34 &   8.77 $\pm$ 0.60 &  3.83 $\pm$ 0.26 &  4.70 $\pm$ 0.24 &  5.40 $\pm$ 0.41      \\
              & $\text{490}\times10^3$ &  [3920000, 4471250, 4042500] & [128, 146, 132] &  3.70 $\pm$ 0.57 &   7.53 $\pm$ 1.20 &  3.17 $\pm$ 0.52 &  4.17 $\pm$ 0.82 &  4.40 $\pm$ 0.50      \\
\bottomrule
\end{tabular}
\end{adjustbox}
\label{tab:scratch_results_testdev}
\end{table}
\begin{table}[htb]
\caption{\textbf{Keypoint Test Metrics for Transfer-Learning with Real Data from Pre-Trained Synthetic and ImageNet Weights.}
For all models, we report the results on the COCO test-dev2017 set. We report the mean and standard deviation of the results from three synthetic datasets generated from three different seeds. The highest metrics in each category are in boldface.}
\begin{adjustbox}{max width=\textwidth}
\centering
\begin{tabular}{crrrlllll}
\toprule
      \begin{tabular}{@{}c@{}}fine-tune\\real size\end{tabular} &     pre-training data &               training steps & no. of epochs &                AP &              $\textrm{AP}^{\textrm{\textit{IoU=.50}}}$ &             $\textrm{AP}^{\textrm{\textit{IoU=.75}}}$ &               $\textrm{AP}^{\textrm{\textit{large}}}$ &              $\textrm{AP}^{\textrm{\textit{medium}}}$                \\
\midrule
\parbox[t]{2mm}{\multirow{6}{*}{\rotatebox[origin=c]{90}{641}}} & - &                      5280 & 132 &              6.40 &             20.30 &              2.40 &              7.90 &              5.60      \\
 & ImageNet &                      6480 & 162 &              21.90 &             50.90 &              15.90 &              26.90 &              18.80      \\
      & $\text{4.9}\times10^3$ synth &        [2960, 2240, 3840] & [74, 56, 96] &  23.80 $\pm$ 0.51 &  50.77 $\pm$ 0.74 &  19.17 $\pm$ 0.52 &  27.80 $\pm$ 0.54 &  21.53 $\pm$ 0.48      \\
      & $\text{49}\times10^3$ synth &          [880, 720, 2480] & [22, 18, 62] &  39.63 $\pm$ 1.23 &  67.43 $\pm$ 0.66 &  39.90 $\pm$ 1.70 &  45.37 $\pm$ 0.97 &  36.43 $\pm$ 1.38      \\
      & \textbf{$\text{245}\times\text{10}^\text{3}$ synth} &          \textbf{[1040, 960, 800]} & \textbf{[26, 24, 20]} &  \textbf{44.43 $\pm$ 0.17} &  \textbf{71.43 $\pm$ 0.12} &  \textbf{46.27 $\pm$ 0.12} &  \textbf{50.47 $\pm$ 0.12} &  \textbf{41.13 $\pm$ 0.31}      \\
      & $\text{490}\times\text{10}^\text{3}$ synth &        [1040, 2240, 1120] & [26, 56, 28] &  42.93 $\pm$ 2.80 &  70.43 $\pm$ 2.16 &  44.20 $\pm$ 3.77 &  49.07 $\pm$ 2.76 &  39.57 $\pm$ 2.88      \\
\midrule
\parbox[t]{2mm}{\multirow{6}{*}{\rotatebox[origin=c]{90}{6411}}} & - &                     40000 & 100 &             37.30 &             67.60 &             35.60 &             43.80 &             33.30      \\
 & ImageNet &                     36800 & 92 &             44.20 &             73.90 &             45.00 &             52.40 &             38.80      \\
      & $\text{4.9}\times10^3$ synth &     [27200, 15200, 21600] & [68, 38, 54] &  42.03 $\pm$ 0.48 &  71.50 $\pm$ 0.24 &  42.40 $\pm$ 0.64 &  49.10 $\pm$ 0.33 &  37.83 $\pm$ 0.54      \\
      & $\text{49}\times10^3$ synth &     [13600, 13600, 14400] & [34, 34, 36] &  51.10 $\pm$ 0.41 &  78.53 $\pm$ 0.29 &  54.47 $\pm$ 0.66 &  58.57 $\pm$ 0.37 &  46.73 $\pm$ 0.53      \\
      & $\text{245}\times10^3$ synth &     [16000, 13600, 16000] & [40, 34, 40] &  52.40 $\pm$ 0.57 &  79.40 $\pm$ 0.36 &  56.10 $\pm$ 0.78 &  60.03 $\pm$ 0.52 &  47.90 $\pm$ 0.59      \\
      & \textbf{$\text{490}\times\text{10}^\text{3}$ synth} &     \textbf{[12800, 12800, 13600]} & \textbf{[32, 32, 34]} &  \textbf{52.70 $\pm$ 0.36} &  \textbf{79.70 $\pm$ 0.28} &  \textbf{56.47 $\pm$ 0.40} &  \textbf{60.27 $\pm$ 0.40} &  \textbf{48.17 $\pm$ 0.34}      \\
\midrule 
\parbox[t]{2mm}{\multirow{6}{*}{\rotatebox[origin=c]{90}{32057}}} & - &                    168252 & 84 &             55.80 &             82.00 &             60.60 &             64.20 &             50.70      \\
 & ImageNet &                    160240 & 80 &             57.50 &             83.60 &             62.40 &             66.40 &             51.70      \\
      & $\text{4.9}\times10^3$ synth &  [168252, 184276, 200300] & [84, 92, 100] &  56.17 $\pm$ 0.17 &  82.80 $\pm$ 0.22 &  60.83 $\pm$ 0.09 &  64.57 $\pm$ 0.17 &  50.90 $\pm$ 0.22      \\
      & $\text{49}\times10^3$ synth &   [100150, 96144, 100150] & [50, 48, 50] &  59.30 $\pm$ 0.22 &  84.50 $\pm$ 0.14 &  64.77 $\pm$ 0.17 &  67.60 $\pm$ 0.22 &  54.17 $\pm$ 0.25      \\
      & $\text{245}\times10^3$ synth &     [64096, 68102, 68102] & [32, 34, 34] &  60.23 $\pm$ 0.24 &  84.90 $\pm$ 0.08 &  65.90 $\pm$ 0.33 &  68.77 $\pm$ 0.26 &  54.90 $\pm$ 0.22      \\
      & \textbf{$\text{490}\times\text{10}^\text{3}$ synth} &    \textbf{[84126, 100150, 80120]} & \textbf{[42, 50, 40]} &  \textbf{60.37 $\pm$ 0.48} &  \textbf{85.03 $\pm$ 0.33} &  \textbf{66.10 $\pm$ 0.59} &  \textbf{68.83 $\pm$ 0.52} &  \textbf{55.13 $\pm$ 0.54}      \\
\midrule
\parbox[t]{2mm}{\multirow{6}{*}{\rotatebox[origin=c]{90}{64115}}} & - &                    577008 & 144 &             62.00 &             86.20 &             68.10 &             70.50 &             56.70      \\
 & ImageNet &                    352616 & 88 &             62.40 &             86.60 &             68.60 &             71.20 &             56.80      \\
      & $\text{4.9}\times10^3$ synth &  [416728, 432756, 472826] & [104, 108, 118] &  61.90 $\pm$ 0.28 &  86.17 $\pm$ 0.12 &  67.97 $\pm$ 0.54 &  70.30 $\pm$ 0.22 &  56.60 $\pm$ 0.36      \\
      & $\text{49}\times10^3$ synth &  [384672, 400700, 376658] & [96, 100, 94] &  62.57 $\pm$ 0.17 &  86.37 $\pm$ 0.09 &  68.70 $\pm$ 0.16 &  71.00 $\pm$ 0.08 &  57.33 $\pm$ 0.19      \\
      & $\text{245}\times10^3$ synth &  [400700, 264462, 296518] & [100, 66, 74] &  63.13 $\pm$ 0.19 &  86.80 $\pm$ 0.08 &  69.37 $\pm$ 0.25 &  71.53 $\pm$ 0.12 &  57.87 $\pm$ 0.31      \\
      & \textbf{$\text{490}\times\text{10}^\text{3}$ synth} &  \textbf{[232406, 216378, 248434]} & \textbf{[58, 54, 62]} &  \textbf{63.47 $\pm$ 0.19} &  \textbf{87.03 $\pm$ 0.09} &  \textbf{69.83 $\pm$ 0.38} &  \textbf{72.03 $\pm$ 0.17} &  \textbf{58.10 $\pm$ 0.28}      \\
\bottomrule
\end{tabular}
\end{adjustbox}
\label{tab:transfer_results_testdev}
\end{table}
To obtain a set of benchmark results on simulation to real transfer learning, we trained on various synthetic and real dataset sizes and combinations for person bounding box (bbox) and keypoint detection. We report our results on the COCO person validation and test-dev2017 using Average Precision (AP) as the primary metric for model performance. 

We started our benchmarks by training models from random initialization on various sizes of real data alone. Unsurprisingly, we saw model performance improved with the amount of real data used (Tab.~\ref{tab:scratch_results_testdev} and~\ref{tab:scratch_results}).
We continued our benchmarks by training the model from random initialization on synthetic data generated by \psp{} using na\"ively-chosen parameters. As before, we report the model performance on COCO-person validation and test-dev2017 sets (Tab.~\ref{tab:scratch_results_testdev} and~\ref{tab:scratch_results}). The benchmarks indicate that a model trained solely on synthetic data generated with na\"ive domain randomization struggles to generalize on the real domain.

To complete our benchmarks, we took the models trained from scratch on synthetic data and fine-tune them on various amounts of real COCO data, which has shown excellent results in other sim2real problems~\citep{tremblay2018domain}. For comparison, we also use ImageNet pre-trained weights and fine-tune them on various amounts of real COCO data. We find that the pre-training on the synthetic data improves model performance after fine-tuning the model on real data for all dataset sizes used. 
Our best models achieve a keypoint AP of $63.47 \pm 0.19$ on the COCO test-dev2017 dataset, outperforming the same model pre-trained on ImageNet and fine-tuned on the entire COCO (keypoint AP of $62.40$).
The effects of synthetic pre-training for few-shot transfer to real domain are more significant; whilst training on $641$ real images alone yields a keypoint AP of $6.40$, ImageNet pre-training increases this to $21.90$; with synthetic pre-training, we observe more than double the performance boost with keypoint AP of $44.43 \pm 0.17$.

The bounding box metrics are reported in Tab.~\ref{tab:scratch_results_bbox_appendix} and \ref{tab:transfer_results_bbox_appendix}. In Tab.~\ref{tab:delta_ap} we show the gains of synthetic pre-training over training from scratch and ImageNet pre-training for the best models. These results show that synthetic pre-training helps more with localization of keypoints than bounding box detection which is the classification of a region of the image. 

\begin{table}[htb]
\centering
\caption{\textbf{Comparison of Gains Obtained from Pre-Training on Synthetic and Fine-Tuning on COCO over Training from Scratch on COCO.} For each dataset size we show the results of the best performing model from Tab.~\ref{tab:transfer_results_testdev},~\ref{tab:transfer_results}, and~\ref{tab:transfer_results_bbox_appendix}.}
\label{tab:delta_ap}
\resizebox{\textwidth}{!}{%
\begin{tabular}{rcccrr|cccrr}
\toprule
 & \multicolumn{5}{c|}{bbox AP (person val2017)} & \multicolumn{5}{c}{keypoint AP (person val2017)} \\ 
 \midrule
\multicolumn{1}{r|}{COCO} & scratch & w/ ImageNet & w/ Synth & $\Delta/_{scratch}$ & $\Delta/_{ImageNet}$ & scratch & w/ ImageNet & w/ Synth & $\Delta/_{scratch}$ & $\Delta/_{ImageNet}$ \\ 
\midrule
\multicolumn{1}{r|}{641} & 13.82 & 27.61 & 42.58 & +28.76 & +14.97 & 7.47 & 23.51 & 46.40 & +38.93 & +22.89 \\ 
\multicolumn{1}{r|}{6411} & 37.82 & 42.53 & 49.04 & +11.22 & +6.51 & 39.48 & 45.99 & 55.21 & +15.73 & +9.22 \\ 
\multicolumn{1}{r|}{32057} & 52.15 & 52.75 & 55.04 & +2.89 & +2.29 & 58.68 & 60.28 & 63.38 & +4.70 & +3.10 \\ 
\multicolumn{1}{r|}{64115} & 56.73 & 56.09 & 57.44 & +0.71 & +1.35 & 65.12 & 65.10 & 66.83 & +1.71 & +1.73 \\ 
\bottomrule
\end{tabular}%
}

\resizebox{0.535\textwidth}{!}{%
\begin{tabular}{rcccrr}
\\
\toprule
 & \multicolumn{5}{c}{keypoint AP (test-dev2017)} \\ 
 \midrule
\multicolumn{1}{r|}{COCO} & scratch & w/ ImageNet & w/ Synth & $\Delta/_{scratch}$ & $\Delta/_{ImageNet}$  \\ 
\midrule
\multicolumn{1}{r|}{641} & 6.40 & 21.90 & 44.43 & +38.03 & +22.53 \\
\multicolumn{1}{r|}{6411} & 37.30 & 44.20 & 52.70 & +15.40 & +8.50 \\
\multicolumn{1}{r|}{32057} & 55.80 & 57.50 & 60.37 & +4.57 & +2.87 \\
\multicolumn{1}{r|}{64115} & 62.00 & 62.40 & 63.47 & +1.47 & +1.07 \\ 
\bottomrule
\end{tabular}%
}
\end{table}

\subsection{Discussion}
We did \emph{not} perform any grid search for model or data generation hyper-parameters in any of our benchmarks, nor did we vary model seed, initialization, or model configuration. Additionally, we used the same learning rate scheduling strategy for all model training. We made these choices to focus the benchmarks on the role of the 
synthetic data
pre-training on 
model performance. These benchmarks show that we can improve model performance with synthetic data using na\"ive domain randomization parameters and the same model training strategy approach. 

However, the na\"ive approach did generate models with a significant zero-shot performance gap on real data (Tab~\ref{tab:scratch_results_testdev}).  We also found that when using synthetic data for pre-training and fine-tuning on real data, the number of iterations needed to converge on the final model performance varies significantly (Tab~\ref{tab:transfer_results_testdev}) with the random seed used for data generation. These variations in the training iterations across multiple dataset seeds indicate that there might be some \textit{``right''} set of synthetic data that is best for fine-tuning and can train a model faster. \psp{} comes with highly-parameterized randomizers, and it is straightforward to integrate custom randomizers into it. Therefore, we expect that \psp{} will enable research into data hyper-parameter tuning to optimize for both zero-shot real-world data performance and optimal fine-tuning performance. 

Another interesting finding from the benchmarks using the Detectron2 Keypoint R-CNN variant was that synthetic data improved keypoint AP (Tab.~\ref{tab:transfer_results_testdev} and~\ref{tab:transfer_results}) more than the bounding box AP (Tab.~\ref{tab:transfer_results_bbox_appendix}). Superficially, this could be because of the improved keypoint labeling (Fig.~\ref{fig:bbox_kpt_compare}) provided by synthetic data. Since \psp{} comes with a range of labelers, we expect \psp{} to enable research in understanding what human-centric tasks synthetic data can best improve.

Our encouraging results open further research into hyper-parameter search, optimization strategy, training schedule, and alternative training strategies to bridge the simulation to reality gap. We envisage that the most exciting line of research will involve generating synthetic data that bridges the simulation to real transfer learning and addresses the domain gap between the synthetic and the real data. We anticipate that \psp{} can expedite this type of research and facilitate other human-centric computer vision research, such as semantic and instance segmentation and 3D bounding box localization. \psp{} can also facilitate meta-learning approaches where data generation is a function of model performance in a feedback loop.
\section{Conclusion and Limitations}
\label{conclusion}
In this work, we introduce \psp{} a highly-parameterized synthetic data generator to enable and accelerate research into the usefulness of synthetic data for human-centric computer vision. \psp{} contains a range of 3D human models with variable appearance and poses. We also provide a set of object primitives to act as distractors and occluders. All 3D assets also allow for programmatic placement. Furthermore, we provide fine control over the lighting, camera settings, and post-processing effects. Additionally, \psp{} generates a wide range of labeling methods for human-centric computer vision. 

To improve the research speed, we provide a fully functional Unity binary capable of generating large amounts of domain-randomized data using a simple JSON configuration. We also provide a Unity template environment with example assets and full functionality of PeopleSansPeople. However, due to RenderPeople redistribution and licensing policies, we do not provide direct access to the 3D human assets; instead, we provide detailed instructions and examples for sourcing and making the human assets simulation-ready. Although the pre-made \psp{} binary does not enable complex structured placement of assets, researchers can update the provided randomizers to allow for different strategies. 
To validate and benchmark the provided parameterization of \psp{} we conducted a set of benchmarks. These benchmarks showed that model performance is improved using synthetic data. We expect that \psp{} and these benchmarks will enable a wide range of research into the simulation to reality domain gap, including but not limited to model training strategies, data hyper-parameter search, and alternative data generation manipulation strategies. 


\begin{ack}
The authors would like to thank Alex Thaman, Maciek Chociej, Priyesh Wani, Steven Leal, Mohsen Kamalzadeh, Wesley Mareovich Smith, Charles Metze, and Ruiyu Zhang for their valuable contributions to this project.
\end{ack}


\bibliography{bib}
\bibliographystyle{unsrtnat}
\newpage
\renewcommand\thefigure{\thesection.\arabic{figure}}  
\renewcommand\thetable{\thesection.\arabic{table}}  

\appendix
\section{Appendix}
\setcounter{figure}{0} 
\setcounter{table}{0} 

\begin{table}[htb]
\caption{\textbf{Keypoint Evaluation Metrics for Models Trained from Scratch.}
We trained all models from randomly-initialized weights and evaluated them on the COCO-person validation set.
We report the mean and standard deviation of the results from three synthetic datasets generated from three different seeds. 
The highest metrics in each category are in boldface. 
}
\begin{adjustbox}{max width=\textwidth}
\centering
\begin{tabular}{crrrlllll}
\toprule
              data &     dataset size &                         training steps & no. of epochs &               AP &              $\textrm{AP}^{\textrm{\textit{IoU=.50}}}$ &             $\textrm{AP}^{\textrm{\textit{IoU=.75}}}$ &               $\textrm{AP}^{\textrm{\textit{large}}}$ &              $\textrm{AP}^{\textrm{\textit{medium}}}$                \\
\midrule
\parbox[t]{2mm}{\multirow{4}{*}{\rotatebox[origin=c]{90}{COCO}}} & 641 &                         5280 & 132 &             7.47 &             23.26 &             3.10 &              8.85 &             6.88                   \\
              & 6411 &                        40000 & 100 &            39.48 &             69.04 &            38.66 &             44.87 &            36.36                   \\
              & 32057 &                       168252 & 84 &            58.68 &             83.51 &            63.10 &             65.41 &            55.15                   \\
              & 64115 &                       577008 & 144 &            \textbf{65.12} &             \textbf{86.73} &            \textbf{70.97} &             \textbf{72.64} &            \textbf{61.15}                   \\
\midrule
\parbox[t]{2mm}{\multirow{4}{*}{\rotatebox[origin=c]{90}{Synth}}} & $\text{4.9}\times10^3$ &        [38556, 53244, 38556] & [126, 174, 126] &  1.89 $\pm$ 0.29 &   4.64 $\pm$ 0.31 &  1.26 $\pm$ 0.29 &   1.92 $\pm$ 0.30 &  2.22 $\pm$ 0.15                   \\
              & $\text{49}\times10^3$ &     [391936, 594028, 398060] & [128, 194, 130] &  \textbf{5.47 $\pm$ 0.10} &  \textbf{11.77 $\pm$ 0.28} &  \textbf{4.54 $\pm$ 0.19} &   \textbf{5.50 $\pm$ 0.44} &  \textbf{6.42 $\pm$ 0.16}                   \\
              & $\text{245}\times10^3$ &  [2143680, 1745568, 2388672] & [140, 114, 156] &  4.78 $\pm$ 0.24 &   9.74 $\pm$ 0.55 &  4.05 $\pm$ 0.13 &   4.86 $\pm$ 0.23 &  5.74 $\pm$ 0.32                   \\
              & $\text{490}\times10^3$ &  [3920000, 4471250, 4042500] & [128, 146, 132] &  3.97 $\pm$ 0.61 &   8.23 $\pm$ 1.37 &  3.33 $\pm$ 0.50 &   4.27 $\pm$ 0.75 &  4.69 $\pm$ 0.54                   \\
\bottomrule
\end{tabular}
\end{adjustbox}
\label{tab:scratch_results}
\end{table}
\begin{table}[t!]
\caption{\textbf{Keypoint Evaluation Metrics for Transfer-Learning with Real Data from Pre-Trained Synthetic and ImageNet Weights.} For all models, we report the results on the COCO person validation set. We report the mean and standard deviation of the results from three synthetic datasets generated from three different seeds. The highest metrics in each category are in boldface.}
\begin{adjustbox}{max width=\textwidth}
\centering
\begin{tabular}{crrrllllll}
\toprule
     \begin{tabular}{@{}c@{}}fine-tune\\real size\end{tabular} &     pre-training data &               training steps & no. of epochs &                AP &              $\textrm{AP}^{\textrm{\textit{IoU=.50}}}$ &             $\textrm{AP}^{\textrm{\textit{IoU=.75}}}$ &               $\textrm{AP}^{\textrm{\textit{large}}}$ &              $\textrm{AP}^{\textrm{\textit{medium}}}$                \\
\midrule
\parbox[t]{2mm}{\multirow{6}{*}{\rotatebox[origin=c]{90}{641}}} & - &                      5280 & 132 &              7.47 &             23.26 &              3.10 &              8.85 &              6.88                    \\
 & ImageNet &                      6480 & 162 &              23.51 &             52.57 &              17.32 &              27.12 &              21.30                    \\
      & $\text{4.9}\times10^3$ synth &        [2960, 2240, 3840] & [74, 56, 96] &  25.02 $\pm$ 0.53 &  53.20 $\pm$ 0.65 &  20.14 $\pm$ 0.77 &  27.13 $\pm$ 0.48 &  24.05 $\pm$ 0.55                    \\
      & $\text{49}\times10^3$ synth &          [880, 720, 2480] & [22, 18, 62] &  41.53 $\pm$ 1.46 &  69.03 $\pm$ 1.17 &  41.95 $\pm$ 1.84 &  44.52 $\pm$ 1.20 &  40.28 $\pm$ 1.69                    \\
      & $\text{245}\times10^3$ synth &          [1040, 960, 800] & [26, 24, 20] &  \textbf{46.40 $\pm$ 0.04} &  \textbf{73.00 $\pm$ 0.16} &  \textbf{48.41 $\pm$ 0.07} &  \textbf{49.77 $\pm$ 0.10} &  \textbf{44.88 $\pm$ 0.17}                    \\
      & $\text{490}\times10^3$ synth &        [1040, 2240, 1120] & [26, 56, 28] &  44.90 $\pm$ 2.84 &  71.98 $\pm$ 2.29 &  45.91 $\pm$ 3.76 &  48.46 $\pm$ 2.95 &  43.20 $\pm$ 2.90                    \\
\midrule
\parbox[t]{2mm}{\multirow{6}{*}{\rotatebox[origin=c]{90}{6411}}} & - &                     40000 & 100 &             39.48 &             69.04 &             38.66 &             44.87 &             36.36                    \\
 & ImageNet &                     36800 & 92 &             45.99 &             74.09 &             47.65 &             52.94 &             41.85                    \\
      & $\text{4.9}\times10^3$ synth &     [27200, 15200, 21600] & [68, 38, 54] &  44.22 $\pm$ 0.39 &  72.84 $\pm$ 0.41 &  45.36 $\pm$ 0.53 &  49.75 $\pm$ 0.36 &  41.35 $\pm$ 0.59                    \\
      & $\text{49}\times10^3$ synth &     [13600, 13600, 14400] & [34, 34, 36] &  53.23 $\pm$ 0.55 &  79.36 $\pm$ 0.44 &  56.69 $\pm$ 0.46 &  58.83 $\pm$ 0.43 &  50.37 $\pm$ 0.64                    \\
      & $\text{245}\times10^3$ synth &     [16000, 13600, 16000] & [40, 34, 40] &  54.76 $\pm$ 0.45 &  80.33 $\pm$ 0.31 &  58.93 $\pm$ 0.67 &  60.67 $\pm$ 0.38 &  51.67 $\pm$ 0.63                    \\
      & $\text{490}\times10^3$ synth &     [12800, 12800, 13600] & [32, 32, 34] &  \textbf{55.21 $\pm$ 0.44} &  \textbf{80.87 $\pm$ 0.26} &  \textbf{59.46 $\pm$ 0.78} &  \textbf{61.35 $\pm$ 0.32} &  \textbf{52.05 $\pm$ 0.55}                    \\
\midrule
\parbox[t]{2mm}{\multirow{6}{*}{\rotatebox[origin=c]{90}{32057}}} & - &                    168252 & 84 &             58.68 &             83.51 &             63.10 &             65.41 &             55.15                    \\
 & ImageNet &                    160240 & 80 &             60.28 &             84.38 &             65.16 &             68.01 &             55.80                    \\
      & $\text{4.9}\times10^3$ synth &  [168252, 184276, 200300] & [84, 92, 100] &  58.91 $\pm$ 0.21 &  83.63 $\pm$ 0.09 &  63.65 $\pm$ 0.38 &  66.07 $\pm$ 0.30 &  54.99 $\pm$ 0.30                    \\
      & $\text{49}\times10^3$ synth &   [100150, 96144, 100150] & [50, 48, 50] &  62.29 $\pm$ 0.23 &  85.19 $\pm$ 0.17 &  67.79 $\pm$ 0.19 &  69.42 $\pm$ 0.32 &  58.42 $\pm$ 0.35                    \\
      & $\text{245}\times10^3$ synth &     [64096, 68102, 68102] & [32, 34, 34] &  63.37 $\pm$ 0.26 &  86.13 $\pm$ 0.11 &  69.06 $\pm$ 0.19 &  70.42 $\pm$ 0.21 &  59.57 $\pm$ 0.37                    \\
      & $\text{490}\times10^3$ synth &    [84126, 100150, 80120] & [42, 50, 40] &  \textbf{63.38 $\pm$ 0.42} &  \textbf{85.89 $\pm$ 0.34} &  \textbf{68.92 $\pm$ 0.77} &  \textbf{70.24 $\pm$ 0.69} &  \textbf{59.66 $\pm$ 0.25}                    \\
\midrule
\parbox[t]{2mm}{\multirow{6}{*}{\rotatebox[origin=c]{90}{64115}}} & - &                    577008 & 144 &             65.12 &             86.73 &             70.97 &             72.64 &             61.15                    \\
 & ImageNet &                    352616 & 88 &             65.10 &             86.72 &             70.39 &             72.89 &             60.73                    \\
      & $\text{4.9}\times10^3$ synth &  [416728, 432756, 472826] & [104, 108, 118] &  65.19 $\pm$ 0.12 &  87.28 $\pm$ 0.01 &  70.79 $\pm$ 0.42 &  72.52 $\pm$ 0.31 &  61.24 $\pm$ 0.11                    \\
      & $\text{49}\times10^3$ synth &  [384672, 400700, 376658] & [96, 100, 94] &  65.81 $\pm$ 0.19 &  87.39 $\pm$ 0.32 &  71.81 $\pm$ 0.43 &  73.20 $\pm$ 0.05 &  61.76 $\pm$ 0.29                    \\
      & $\text{245}\times10^3$ synth &  [400700, 264462, 296518] & [100, 66, 74] &  66.52 $\pm$ 0.25 &  87.61 $\pm$ 0.14 &  72.72 $\pm$ 0.52 &  73.81 $\pm$ 0.19 &  62.47 $\pm$ 0.30                    \\
      & $\text{490}\times10^3$ synth &  [232406, 216378, 248434] & [58, 54, 62] &  \textbf{66.83 $\pm$ 0.11} &  \textbf{87.89 $\pm$ 0.08} &  \textbf{72.76 $\pm$ 0.08} &  \textbf{74.05 $\pm$ 0.12} &  \textbf{62.95 $\pm$ 0.22}                    \\
\bottomrule
\end{tabular}
\end{adjustbox}
\label{tab:transfer_results}
\end{table}
\begin{table}[t!] 
\caption{\textbf{Bounding Box Evaluation Metrics for Models Trained from Scratch.}
We trained all models from randomly-initialized weights and evaluated them on the COCO-person validation set.
We report the mean and standard deviation of the results from three synthetic datasets generated from three different seeds. 
The highest metrics in each category are in boldface.
}
\begin{adjustbox}{max width=\textwidth}
\centering
\begin{tabular}{crrrllllll}
\toprule
             data &     dataset size &                         training steps & no. of epochs &               AP &              $\textrm{AP}^{\textrm{\textit{IoU=.50}}}$ &             $\textrm{AP}^{\textrm{\textit{IoU=.75}}}$ &               $\textrm{AP}^{\textrm{\textit{large}}}$ &              $\textrm{AP}^{\textrm{\textit{medium}}}$ &              $\textrm{AP}^{\textrm{\textit{small}}}$ \\
\midrule
\parbox[t]{2mm}{\multirow{4}{*}{\rotatebox[origin=c]{90}{COCO}}} & 641 &                         5280 & 132 &            13.82 &             36.49 &             7.01 &             21.98 &            17.64 &             6.83 \\
              & 6411 &                        40000 & 100 &            37.82 &             69.08 &            37.18 &             54.14 &            44.22 &            22.21 \\
              & 32057 &                       168252 & 84 &            52.15 &             81.89 &            56.09 &             68.84 &            58.78 &            35.09 \\
              & 64115 &                       577008 & 144 &            \textbf{56.73} &             \textbf{85.20} &            \textbf{61.49} &             \textbf{73.23} &            \textbf{63.56} &            \textbf{39.14} \\
\midrule
\parbox[t]{2mm}{\multirow{4}{*}{\rotatebox[origin=c]{90}{Synth}}} & $\text{4.9}\times10^3$ &        [38556, 53244, 38556] & [126, 174, 126] &  4.34 $\pm$ 0.12 &   8.46 $\pm$ 0.29 &  3.82 $\pm$ 0.11 &   8.61 $\pm$ 0.20 &  5.94 $\pm$ 0.02 &  1.29 $\pm$ 0.13 \\
              & $\text{49}\times10^3$ &     [391936, 594028, 398060] & [128, 194, 130] &  \textbf{7.61 $\pm$ 0.12} &  \textbf{14.65 $\pm$ 0.33} &  \textbf{7.00 $\pm$ 0.09} &  \textbf{14.48 $\pm$ 1.30} &  \textbf{9.86 $\pm$ 0.27} &  2.36 $\pm$ 0.10 \\
              & $\text{245}\times10^3$ &  [2143680, 1745568, 2388672] & [140, 114, 156] &  6.86 $\pm$ 0.29 &  13.11 $\pm$ 0.65 &  6.37 $\pm$ 0.31 &  12.47 $\pm$ 0.28 &  8.95 $\pm$ 0.52 &  \textbf{2.47 $\pm$ 0.26} \\
              & $\text{490}\times10^3$ &  [3920000, 4471250, 4042500] & [128, 146, 132] &  6.02 $\pm$ 0.61 &  11.57 $\pm$ 1.10 &  5.42 $\pm$ 0.55 &  11.38 $\pm$ 1.02 &  7.66 $\pm$ 0.99 &  2.15 $\pm$ 0.18  \\
\bottomrule
\end{tabular}
\end{adjustbox}
\label{tab:scratch_results_bbox_appendix}
\end{table}
\begin{table}[t!]
\caption{\textbf{Bounding Box Evaluation Metrics for Transfer-Learning with Real Data from Pre-Trained Synthetic and ImageNet Weights.} For all models, we report the results on the COCO person validation set. We report the mean and standard deviation of the results from three synthetic datasets generated from three different seeds. The highest metrics in each category are in boldface.}
\begin{adjustbox}{max width=\textwidth}
\centering
\begin{tabular}{crrrllllll}
\toprule
     \begin{tabular}{@{}c@{}}fine-tune\\real size\end{tabular} &     pre-training data &               training steps & no. of epochs &                AP &              $\textrm{AP}^{\textrm{\textit{IoU=.50}}}$ &             $\textrm{AP}^{\textrm{\textit{IoU=.75}}}$ &               $\textrm{AP}^{\textrm{\textit{large}}}$ &              $\textrm{AP}^{\textrm{\textit{medium}}}$ &              $\textrm{AP}^{\textrm{\textit{small}}}$ \\
\midrule
\parbox[t]{2mm}{\multirow{6}{*}{\rotatebox[origin=c]{90}{641}}} & - &                      5280 & 132 &             13.82 &             36.49 &              7.01 &             21.98 &             17.64 &              6.83 \\
 & ImageNet &                      6480 & 162 &              27.61 &             57.56 &              23.43 &              38.98 &              34.01      &  15.54         \\
      & $\text{4.9}\times10^3$ synth &        [2960, 2240, 3840] & [74, 56, 96] &  30.32 $\pm$ 0.48 &  56.58 $\pm$ 0.75 &  28.90 $\pm$ 0.53 &  43.86 $\pm$ 0.36 &  37.00 $\pm$ 0.47 &  16.47 $\pm$ 0.51 \\
      & $\text{49}\times10^3$ synth &          [880, 720, 2480] & [22, 18, 62] &  39.65 $\pm$ 0.64 &  66.92 $\pm$ 1.03 &  40.36 $\pm$ 0.74 &  54.16 $\pm$ 0.52 &  47.74 $\pm$ 0.73 &  23.65 $\pm$ 0.92 \\
      & $\text{245}\times10^3$ synth &          [1040, 960, 800] & [26, 24, 20] &  \textbf{42.58 $\pm$ 0.08} &  \textbf{69.80 $\pm$ 0.16} &  \textbf{43.84 $\pm$ 0.18} &  \textbf{57.20 $\pm$ 0.04} &  \textbf{50.79 $\pm$ 0.14} &  \textbf{26.34 $\pm$ 0.14} \\
      & $\text{490}\times10^3$ synth &        [1040, 2240, 1120] & [26, 56, 28] &  41.24 $\pm$ 2.07 &  68.21 $\pm$ 2.30 &  42.46 $\pm$ 2.14 &  56.22 $\pm$ 2.32 &  49.42 $\pm$ 2.08 &  24.93 $\pm$ 1.73 \\
\midrule
\parbox[t]{2mm}{\multirow{6}{*}{\rotatebox[origin=c]{90}{6411}}} & - &                     40000 & 100 &             37.82 &             69.08 &             37.18 &             54.14 &             44.22 &             22.21 \\
 & ImageNet &                     36800 & 92 &             42.53 &             73.57 &             43.47 &             58.59 &             49.40         &   25.96        \\
      & $\text{4.9}\times10^3$ synth &     [27200, 15200, 21600] & [68, 38, 54] &  42.83 $\pm$ 0.37 &  72.85 $\pm$ 0.24 &  44.01 $\pm$ 0.37 &  58.93 $\pm$ 0.37 &  49.92 $\pm$ 0.39 &  26.34 $\pm$ 0.59 \\
      & $\text{49}\times10^3$ synth &     [13600, 13600, 14400] & [34, 34, 36] &  48.46 $\pm$ 0.29 &  77.48 $\pm$ 0.21 &  51.22 $\pm$ 0.45 &  64.18 $\pm$ 0.37 &  56.00 $\pm$ 0.13 &  31.38 $\pm$ 0.58 \\
      & $\text{245}\times10^3$ synth &     [16000, 13600, 16000] & [40, 34, 40] &  \textbf{49.04 $\pm$ 0.29} &  77.97 $\pm$ 0.30 &  \textbf{52.06 $\pm$ 0.42} &  64.90 $\pm$ 0.35 &  \textbf{56.40 $\pm$ 0.40} &  \textbf{31.95 $\pm$ 0.37} \\
      & $\text{490}\times10^3$ synth &     [12800, 12800, 13600] & [32, 32, 34] &  48.97 $\pm$ 0.17 &  \textbf{78.10 $\pm$ 0.22} &  51.89 $\pm$ 0.27 &  \textbf{65.00 $\pm$ 0.10} &  56.25 $\pm$ 0.22 &  31.87 $\pm$ 0.29 \\
    \midrule
\parbox[t]{2mm}{\multirow{6}{*}{\rotatebox[origin=c]{90}{32057}}} & - &                    168252 & 84 &             52.15 &             81.89 &             56.09 &             68.84 &             58.78 &             35.09 \\
 & ImageNet &                    160240 & 80 &             52.75 &             82.56 &             56.76 &             69.51 &             59.25         &   35.46       \\
      & $\text{4.9}\times10^3$ synth &  [168252, 184276, 200300] & [84, 92, 100] &  52.97 $\pm$ 0.04 &  82.20 $\pm$ 0.04 &  56.95 $\pm$ 0.04 &  69.69 $\pm$ 0.05 &  59.83 $\pm$ 0.16 &  35.56 $\pm$ 0.17 \\
      & $\text{49}\times10^3$ synth &   [100150, 96144, 100150] & [50, 48, 50] &  54.74 $\pm$ 0.20 &  83.38 $\pm$ 0.06 &  59.22 $\pm$ 0.25 &  71.31 $\pm$ 0.34 &  61.51 $\pm$ 0.04 &  37.20 $\pm$ 0.15 \\
      & $\text{245}\times10^3$ synth &     [64096, 68102, 68102] & [32, 34, 34] &  \textbf{55.04 $\pm$ 0.08} &  \textbf{83.57 $\pm$ 0.09} &  \textbf{59.49 $\pm$ 0.23} &  \textbf{71.54 $\pm$ 0.18} &  \textbf{62.02 $\pm$ 0.28} &  37.32 $\pm$ 0.04 \\
      & $\text{490}\times10^3$ synth &    [84126, 100150, 80120] & [42, 50, 40] &  54.93 $\pm$ 0.15 &  83.51 $\pm$ 0.10 &  59.40 $\pm$ 0.30 &  71.20 $\pm$ 0.25 &  61.65 $\pm$ 0.13 &  \textbf{37.48 $\pm$ 0.28} \\
\midrule
\parbox[t]{2mm}{\multirow{6}{*}{\rotatebox[origin=c]{90}{64115}}} & - &                    577008 & 144 &             56.73 &             85.20 &             61.49 &             73.23 &             63.56 &             39.14 \\
 & ImageNet &                    352616 & 88 &             56.09 &             84.96 &             60.61 &             72.72 &             63.12         &    38.35      \\
      & $\text{4.9}\times10^3$ synth &  [416728, 432756, 472826] & [104, 108, 118] &  56.65 $\pm$ 0.02 &  85.15 $\pm$ 0.02 &  61.62 $\pm$ 0.30 &  73.07 $\pm$ 0.16 &  63.46 $\pm$ 0.06 &  39.19 $\pm$ 0.12 \\
      & $\text{49}\times10^3$ synth &  [384672, 400700, 376658] & [96, 100, 94] &  57.24 $\pm$ 0.06 &  85.34 $\pm$ 0.11 &  62.36 $\pm$ 0.18 &  73.79 $\pm$ 0.31 &  64.04 $\pm$ 0.05 &  39.66 $\pm$ 0.19 \\
      & $\text{245}\times10^3$ synth &  [400700, 264462, 296518] & [100, 66, 74] &  57.31 $\pm$ 0.13 &  85.38 $\pm$ 0.19 &  62.33 $\pm$ 0.02 &  \textbf{74.06 $\pm$ 0.14} &  63.78 $\pm$ 0.15 &  39.63 $\pm$ 0.15 \\
      & $\text{490}\times10^3$ synth &  [232406, 216378, 248434] & [58, 54, 62] &  \textbf{57.44 $\pm$ 0.11} &  \textbf{85.52 $\pm$ 0.19} &  \textbf{62.44 $\pm$ 0.04} &  73.84 $\pm$ 0.15 &  \textbf{64.39 $\pm$ 0.10} &  \textbf{39.69 $\pm$ 0.11} \\
\bottomrule
\end{tabular}
\end{adjustbox}
\label{tab:transfer_results_bbox_appendix}
\end{table}


\subsection{Bounding Box Evaluation Metrics and Gains Over Training from Scratch}
In this section we present the bounding box evaluation metrics for the benchmarks we presented in Tab.~\ref{tab:scratch_results_testdev} and \ref{tab:transfer_results_testdev}. Since COCO test-dev2017 does not provide bounding box analysis we use the validation set for evaluations. Therefore, for completness, we also provide the keypoint evaluation for the COCO validation set (Tab.~\ref{tab:scratch_results} and Tab.~\ref{tab:transfer_results}). 
In Tab.~\ref{tab:scratch_results_bbox_appendix} the bounding box evaluation metrics for models trained from scratch are shown. 
Tab.~\ref{tab:transfer_results_bbox_appendix} shows the bounding box evaluation metrics for transfer-learning with real data from pre-trained synthetic and ImageNet weights. We observe the same trends similar to Tab.~\ref{tab:transfer_results} with bounding box detection metrics as well.
Our model pre-trained with $490\times10^3$ synthetic images and fine-tuned on the $100\%$ of COCO-person (\num[group-separator={,}]{64115} images) achieves the highest bounding box AP of $57.44 \pm 0.11$ out-performing the best model pre-trained on ImageNet with bounding box AP of $56.09$ as shown in Tab.~\ref{tab:transfer_results_bbox_appendix}.




\subsection{\psp{} Label Annotations}
\begin{figure}[htb]
    \centering
    \begin{subfigure}[t]{0.245\textwidth}
        \raisebox{-\height}{\includegraphics[width=\textwidth]{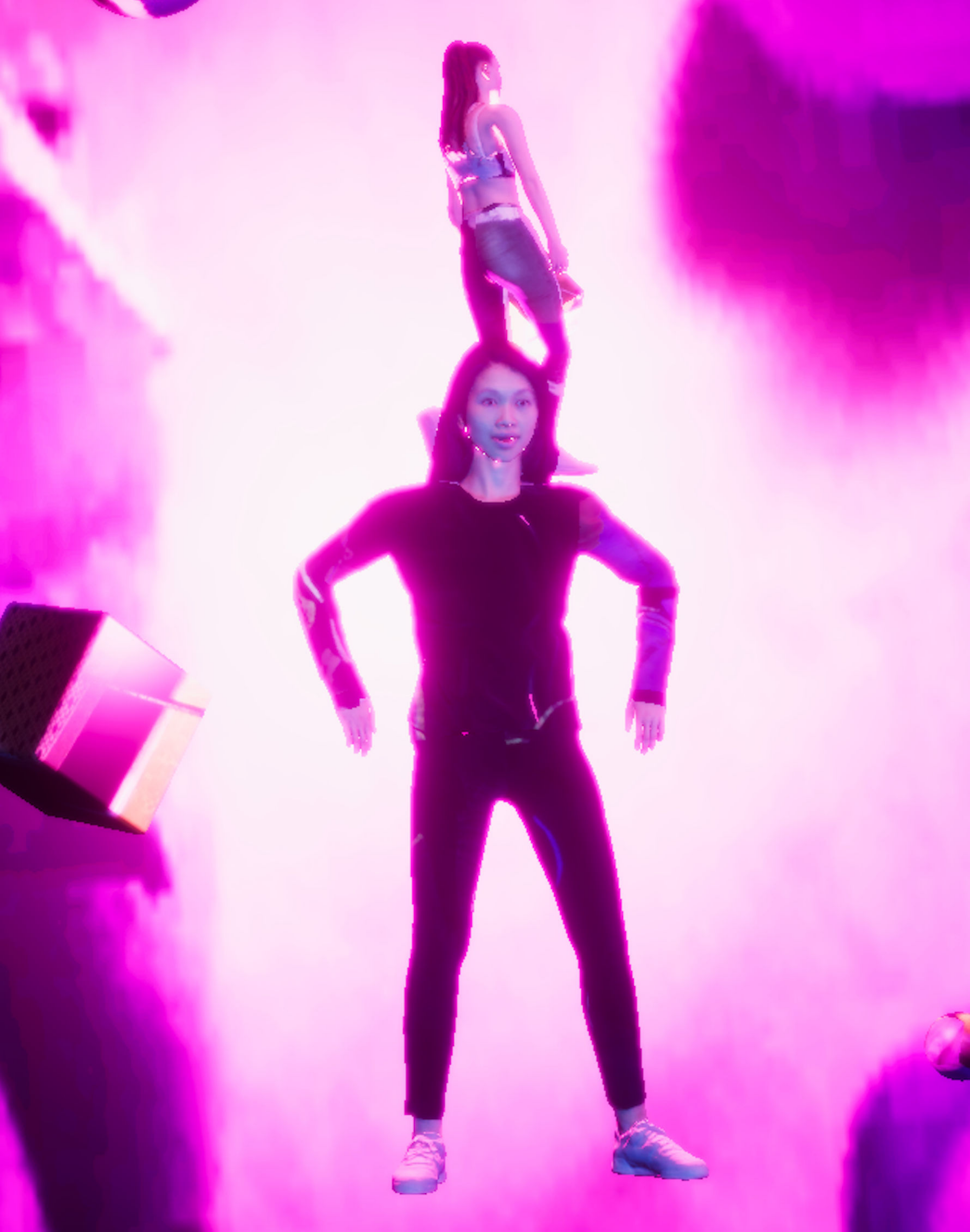}}
        \caption{ }
    \end{subfigure}
    \begin{subfigure}[t]{0.245\textwidth}
        \raisebox{-\height}{\includegraphics[width=\textwidth]{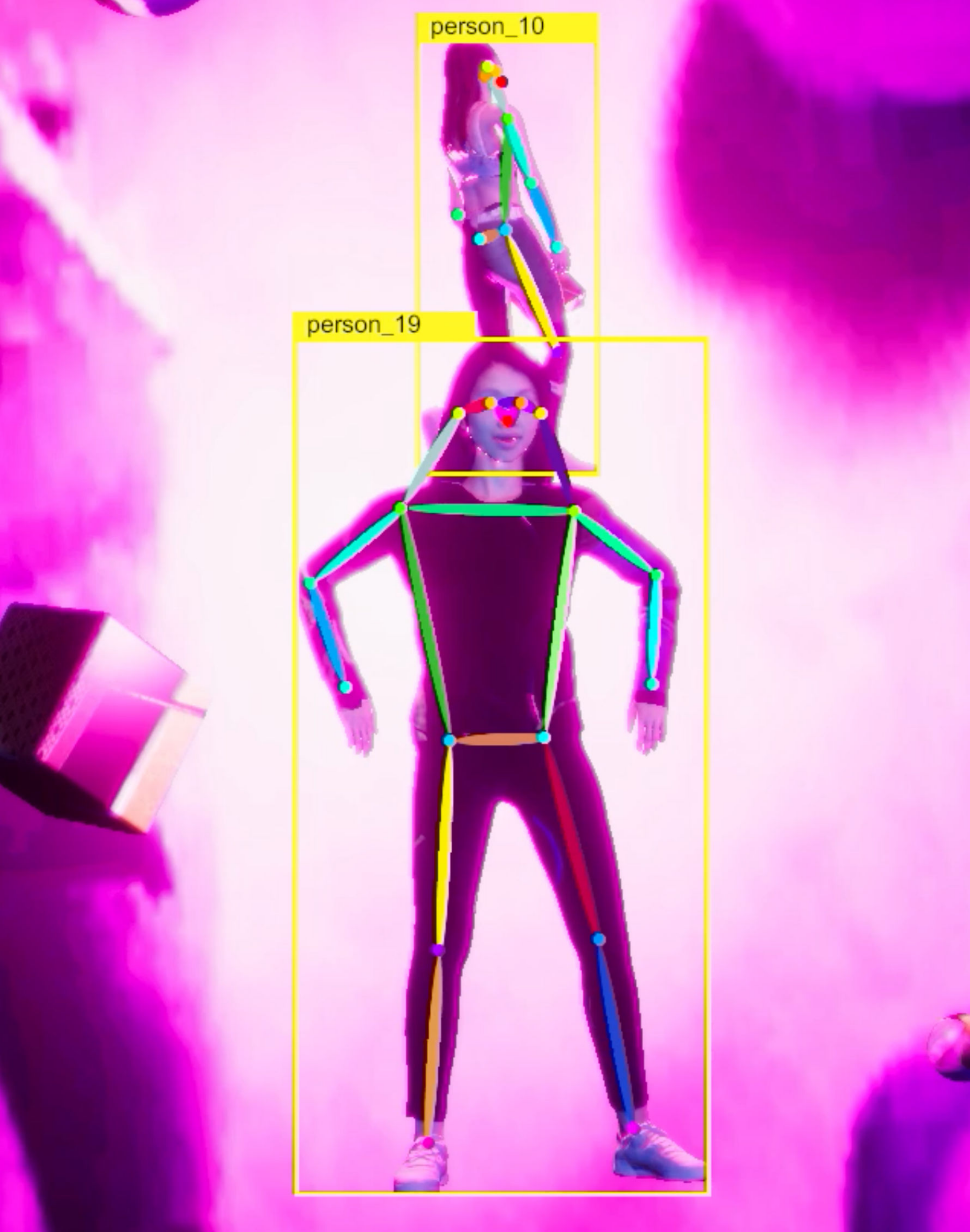}}
        \caption{ }
    \end{subfigure}
    \begin{subfigure}[t]{0.245\textwidth}
        \raisebox{-\height}{\includegraphics[width=\textwidth]{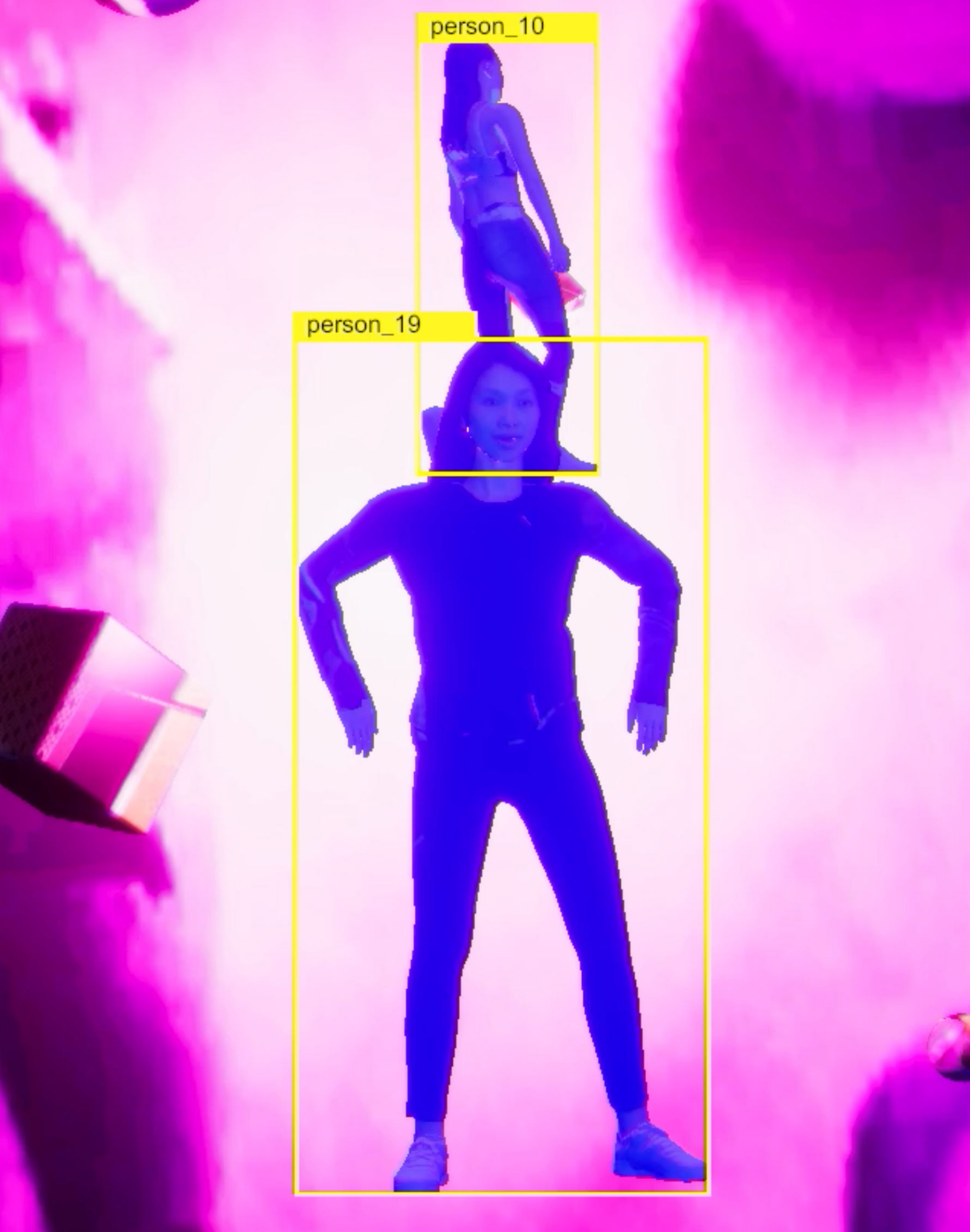}}
        \caption{ }
    \end{subfigure}
    \begin{subfigure}[t]{0.245\textwidth}
        \raisebox{-\height}{\includegraphics[width=\textwidth]{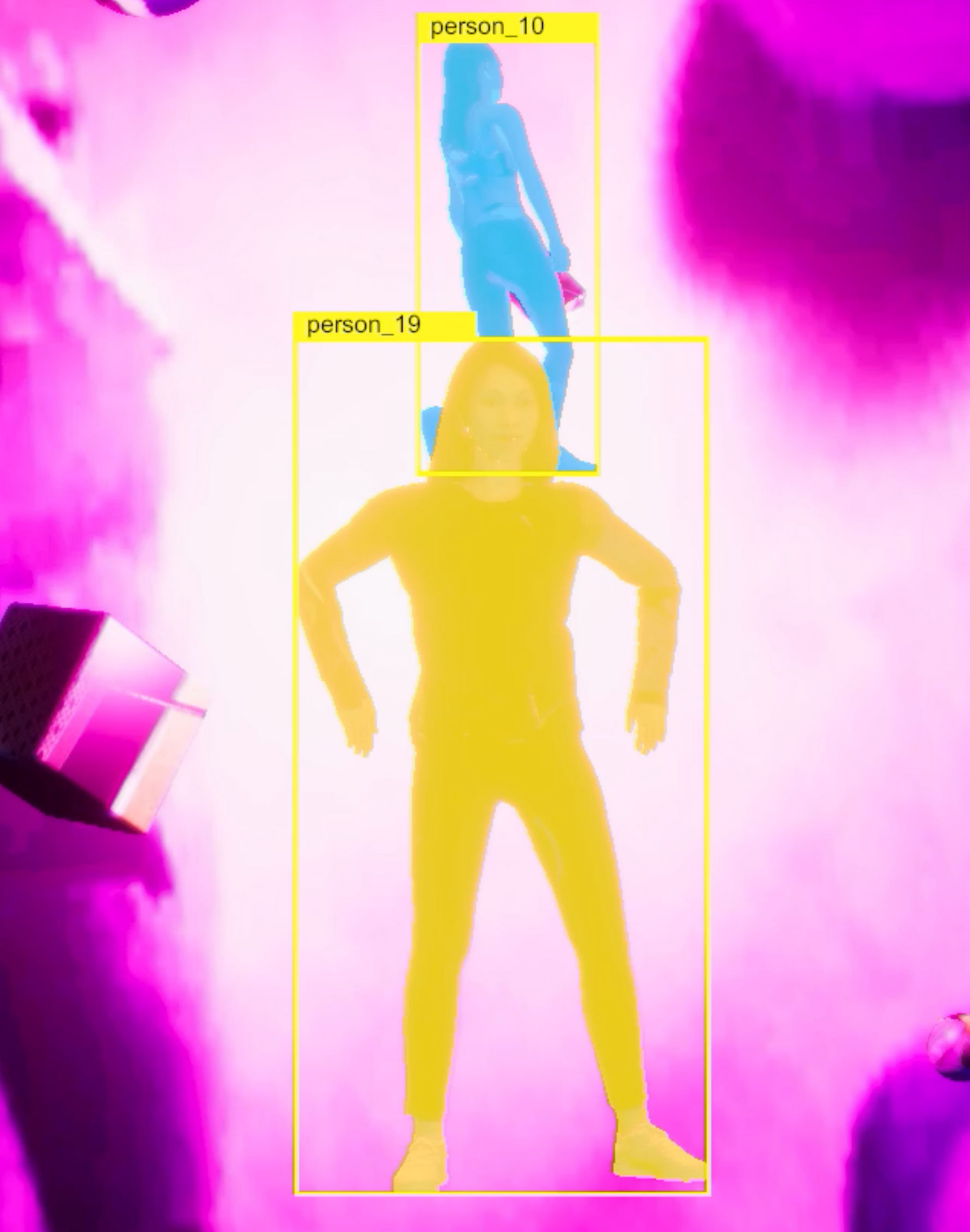}}
        \caption{ }
    \end{subfigure}
\caption{\textbf{Different annotation types produced by the Perception camera.} (a) rendered image, (b) bounding box and keypoint annotations, (c) bounding box and semantic segmentation annotations, (d) bounding box and instance segmentation annotations. Any combinations of these annotations types are possible; examples are provided with the aforementioned combinations for ease of demonstration.}
     \label{fig:labelfig}%
\end{figure}

In this section, we show examples of the types of label annotations \psp{} provides.
Our Unity scene environment includes one camera with an attached \textit{Perception camera} component, extending the rendering process to generate annotation labels for each frame. The Perception camera can produce sub-pixel-perfect annotations such as 2D/3D bounding box, human keypoints, semantic segmentation, and instance segmentation for as many object classes as required by the user.  
Fig.~\ref{fig:labelfig} shows different annotation types that are enabled in \psp{}. Although for our benchmarks, we only used the bounding box and keypoint labels, the users have the option to enable semantic and instance segmentation labeling -- shown in Fig.~\ref{fig:labelfig}c) and d) -- as well from the Unity Editor. For more information about the labeling and its schema, refer to \citet{borkman2021unity}.


\subsection{\psp{} Randomizers}
The Perception package comes with sample scene randomizers (e.g., random object placement, rotation, texture). As noted in \citet{borkman2021unity}, the randomizers are customize-able to fit the users' needs. The users are also able to create and append their randomizers to the simulation. In \psp{} we have used several Perception package's default randomizers, as well as our custom-designed ones. We regard certain types of our randomizers as data augmentation techniques, specifically, the Lighting, Hue Offset, Camera Rotation/Field of View/Focal Length, and Post-Process effects. Hence the resulting dataset will not require data augmentations during training that speeds up the training itself.
In all the randomizers, the values are sampled from a uniform distribution. Tab.~\ref{tab:randomizers} outlines the statistical distributions for our randomizer parameters.
Below we describe the randomizers used in \psp{}.

\paragraph{Background/Occluder Object Placement Randomizer.} Randomly spawns background/occluder objects within user-defined 3D volumes. The separation distance parameter can be adjusted to control the proximity of the objects to each other. It uses Poisson-Disk sampling \citep{bridson2007fast} to select random positions from a given area. The background and occluder objects are sourced from a set of primitive 3D game objects (cubes, cylinders, spheres, etc.) provided by Unity's Perception package. 

\paragraph{Background/Occluder Scale Randomizer.} Randomizes the scale of the background/occluder objects spawned in the scene.

\paragraph{Background/Occluder Rotation Randomizer.} Randomizes the 3D rotation of the objects in the scene.

\paragraph{Foreground Object Placement Randomizer.} Similar to the \textit{Background/Occluder Object Placement Randomizer}, this randomly spawns foreground objects -- chosen from our set of 3D human asset prefabs -- within a specified volume in the scene.

\paragraph{Foreground Scale Randomizer.} Similar to the \textit{Background/Occluder Scale Randomizer}, randomizes the scale of the foreground objects.

\paragraph{Foreground Rotation Randomizer.} Randomizes the rotation of the foreground objects around the $Y$-axis only. We decided that the 3D human assets do not need to be rotated around the $X$, $Z$-axis, because such orientations are rarely seen in the real data.

\paragraph{Animation Randomizer.} Randomizes the pose applied to the character. The pose is a randomly chosen frame from a randomly chosen animation from our database of human animations.

\paragraph{Texture Randomizer.} Randomizes the texture applied to predefined objects. The textures can be provided as JPEG or PNG images. For \psp{}, we used 1600 images from the COCO unlabeled 2017 set, and we ensured that no images of humans appear within this set. The random textures are applied to the background wall, as well as the background/occluder objects in the scene.

\paragraph{Hue Offset Randomizer.} Randomizes the hue offset applied to textures on the objects. This is applied to our background wall and the background/occluder objects.

\paragraph{Shader Graph Texture Randomizer.} Randomizes the clothing texture and hue offset for our human assets. We exposed rendering controls for our human assets to let us vary the Albedo, Normal, and Mask texture of the materials at runtime. We source these from the database of all Albedo, Normal, and Mask textures in our RenderPeople assets. These textures are randomly chosen and applied to the character materials during simulation.

\paragraph{Sun Angle Randomizer.} Randomizes the directional light's intensity, elevation, and orientation to mimic the time of the day and the day of the year.

\paragraph{Light Intensity and color Randomizer.} Randomizes the lights' intensity and color parameters (in the RGBA color model).
A light switcher with an on probability of $80\%$ controls the light's on/off states.

\paragraph{Light Position and Rotation Randomizer.} Randomizes the lights' global position and rotation in the scene.

\paragraph{Camera Randomizer.} Randomizes the extrinsic camera parameters, such as its global position and rotation; also randomizes the intrinsic camera parameters, such as Field of View (FoV) and Focal Length by mimicking a physical camera. The combination of change of FoV and Focal Length allows us to capture images ranging from extreme close-ups of subjects (telephoto)  to wide-angle views of the scene with varying focus points, adding camera bloom and lens blur around the objects that are out of focus. The camera's varying position and rotation captures unique and diverse perspectives from the scene.

\paragraph{Post Process Volume Randomizer.} Randomizes some post-processing effects on the rendered images deterministically: Vignette, Exposure, White Balance, Depth of Field, and Color Adjustments such as contrast and saturation. We left additional options for Lens Blur and Film Grain open for the user, although these will not yield deterministic behaviour across multiple simulations. The post-processing effects also increase the image appearance diversity, acting as a data augmentation technique.

\begin{table}[htb]
\caption{Domain Randomization Parameters in Our Data Generator.}
\label{tab:randomizers}
\resizebox{\textwidth}{!}{%
\bgroup
\def\arraystretch{1.75}%
\begin{tabular}{l|l|l|l}
\toprule
category & randomizer & parameters & distribution \\ \toprule
\multirow{11}{*}{3D Objects} & \multirow{2}{*}{\begin{tabular}[c]{@{}l@{}}Background/Occluder \\ Object Placement\end{tabular}} & object placement & \texttt{Cartesian{[}Uniform(-7.5, 7.5), Uniform(-7.5, 7.5), Uniform(-10, 14){]}} \\ \cline{3-4} 
 &  & separation distance & \texttt{Cartesian{[}Constant(2.5), Constant(2.5), Constant(2.5){]}} \\ \cline{2-4} 
 & Background/Occluder Scale & object scale range & \texttt{Cartesian{[}Uniform(1, 12), Uniform(1, 12), Uniform(1, 12){]}} \\ \cline{2-4} 
 & Background/Occluder Rotation & object rotation & \texttt{Euler{[}Uniform(0, 360), Uniform(0, 360), Uniform(0, 360){]}} \\ \cline{2-4} 
 & \multirow{2}{*}{Foreground Object Placement} & object placement & \texttt{Cartesian{[}Uniform(-7.5, 7.5), Uniform(-7.5, 7.5), Uniform(-9, 6){]}} \\ \cline{3-4} 
 &  & separation distance & \texttt{Cartesian{[}Constant(3), Constant(3), Constant(3){]}} \\ \cline{2-4} 
 & Foreground Scale & object scale range & \texttt{Cartesian{[}Uniform(0.5, 3), Uniform(0.5, 3), Uniform(0.5, 3){]}} \\ \cline{2-4} 
 & Foreground Rotation & object rotation & \texttt{Euler{[}Uniform(0, 0), Uniform(0, 360), Uniform(0, 0){]}} \\ \cline{2-4} 
 & Animation & animations & A set of FBX animation clips of arbitrary length \\ \midrule
\multirow{8}{*}{\begin{tabular}[c]{@{}l@{}}Textures\\ and\\ Colors\end{tabular}} & Texture & textures & A set of texture assets \\ \cline{2-4} 
 & Hue Offset & hue offset & \texttt{Uniform(-180, 180)} \\ \cline{2-4} 
 & \multirow{6}{*}{Shader Graph Texture} & albedo textures & A set of albedo texture assets \\ \cline{3-4} 
 &  & normal textures & A set of normal texture assets \\ \cline{3-4} 
 &  & mask textures & A set of mask texture assets \\ \cline{3-4} 
 &  & materials & A set of material assets \\ \cline{3-4} 
 &  & hue top clothing & \texttt{Uniform(0, 360)} \\ \cline{3-4} 
 &  & hue bottom clothing & \texttt{Uniform(0, 360)} \\ \midrule
\multirow{8}{*}{Lights} & \multirow{3}{*}{Sun Angle} & hour & \texttt{Uniform(0, 24)} \\ \cline{3-4} 
 &  & day of the year & \texttt{Uniform(0, 365)} \\ \cline{3-4} 
 &  & lattitude & \texttt{Uniform(-90, 90)} \\ \cline{2-4} 
 & \multirow{3}{*}{Light Intensity and Color} & intensity & \texttt{Uniform(5000, 50000)} \\ \cline{3-4} 
 &  & color & \texttt{RGBA{[}Uniform(0, 1), Uniform(0, 1), Uniform(0, 1), Constant(1){]}} \\ \cline{3-4} 
 &  & light switcher enabled probability & $P(enabled) = 0.8, \quad P(disabled) = 0.2$ \\ \cline{2-4} 
 & \multirow{2}{*}{Light Position and Rotation} & position offset from initial position & \texttt{Cartesian{[}Uniform(-3.65, 3.65), Uniform(-3.65, 3.65), Uniform(-3.65, 3.65){]}} \\ \cline{3-4} 
 &  & rotation offset from initial rotation & \texttt{Euler{[}Uniform(-50, 50), Uniform(-50, 50), Uniform(-50, 50){]}} \\ \midrule
\multirow{4}{*}{Camera} & \multirow{4}{*}{Camera} & field of view & \texttt{Uniform(5, 50)} \\ \cline{3-4} 
 &  & focal length & \texttt{Uniform(1, 23)} \\ \cline{3-4} 
 &  & position offset from initial position & \texttt{Cartesian{[}Uniform(-5, 5), Uniform(-5, 5), Uniform(-5, 5){]}} \\ \cline{3-4} 
 &  & rotation offset from initial rotation & \texttt{Euler{[}Uniform(-5, 5), Uniform(-5, 5), Uniform(-5, 5){]}} \\ \midrule
\multirow{6}{*}{Post-Processing} & \multirow{6}{*}{Post Process Volume} & vignette intensity & \texttt{Uniform(0, 0.5)} \\ \cline{3-4} 
 &  & fixed exposure & \texttt{Uniform(5, 10)} \\ \cline{3-4} 
 &  & white balance temperature & \texttt{Uniform(-20, 20)} \\ \cline{3-4} 
 &  & depth of field focus distance & \texttt{Uniform(0.1, 4)} \\ \cline{3-4} 
 &  & color adjustments: contrast & \texttt{Uniform(-30, 30)} \\ \cline{3-4} 
 &  & color adjustments: saturation & \texttt{Uniform(-30, 30)} \\ 
 \bottomrule
\end{tabular}%
\egroup
}
\end{table}

\subsection{Additional Pose Heatmaps Comparison}
As we described in Section \ref{subsec:3dassets} we gathered a set of animations that are created from motion capture clips, and we use those to vary the pose of our humans in each scene. The choice of these animations was subjective and based on the author's intuitions for what constitutes a sufficiently diverse set of poses that capture enough variations in human activity. Therefore, our animation data is neither guaranteed nor intended to capture all possible human poses, as it is arguably an impossible feat. We analyze the pose diversity of our dataset next to that of a real-world dataset to provide some insight into whether we have been able to capture the variations seen in the real data at the very least. To do so, we take the keypoint annotations from all the instances in the COCO and synthetic datasets, where the hip and shoulder keypoints (torso of human) are indeed annotated, whether occluded or visible. For these instances, we calculate the mid-hip point and translate all points such that the mid-hip falls on $(0, 0)$. Then we measure the distances between the left-hip and left-shoulder and the right-hip and right-shoulder and use their average to scale all other points. As a result, all the human keypoints will roughly have the same skeletal distances. Refer to Alg.~\ref{alg:pose_analysis} for details of these calculations. 

\begin{figure}[htb] 
    \centering
    \begin{subfigure}[t]{0.195\textwidth}
        \raisebox{-\height}{\includegraphics[width=1\textwidth]{neurips_data_2021/plots/fig11_pose_panel/coco/nose_heatmap.pdf}}
    \end{subfigure}
    \hfill
    \begin{subfigure}[t]{0.195\textwidth}
        \raisebox{-\height}{\includegraphics[width=1\textwidth]{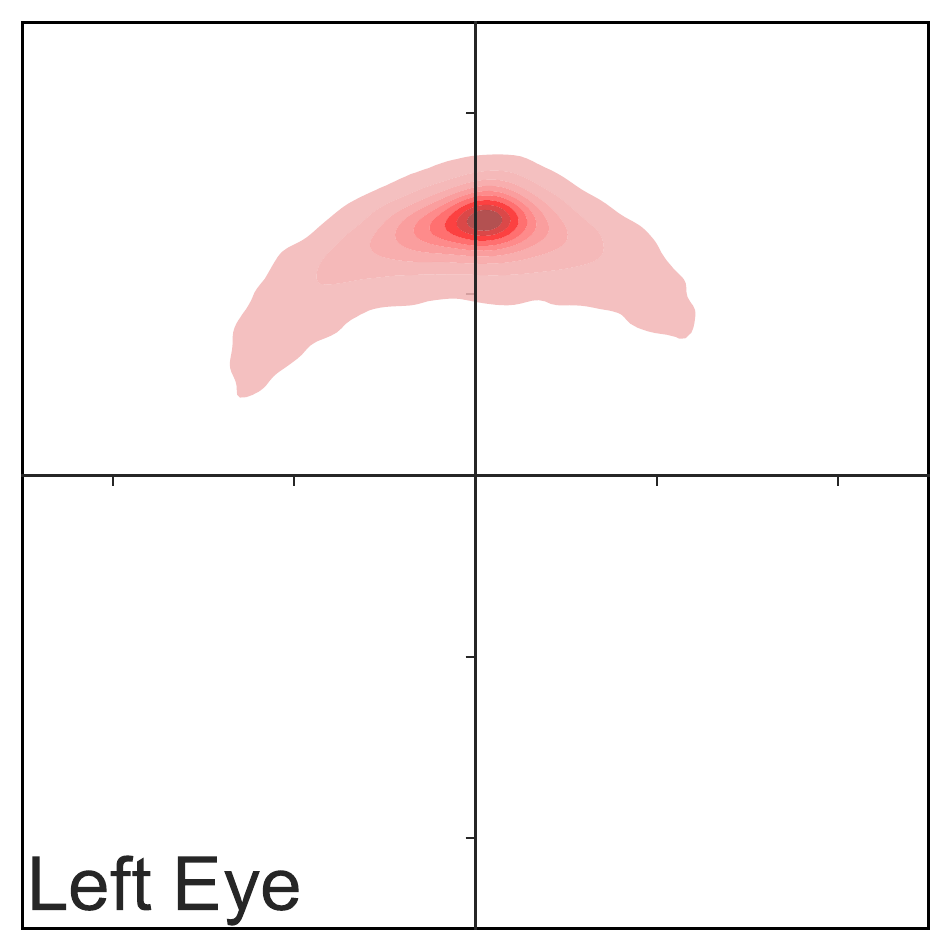}}
    \end{subfigure}
    \hfill
    \begin{subfigure}[t]{0.195\textwidth}
        \raisebox{-\height}{\includegraphics[width=1\textwidth]{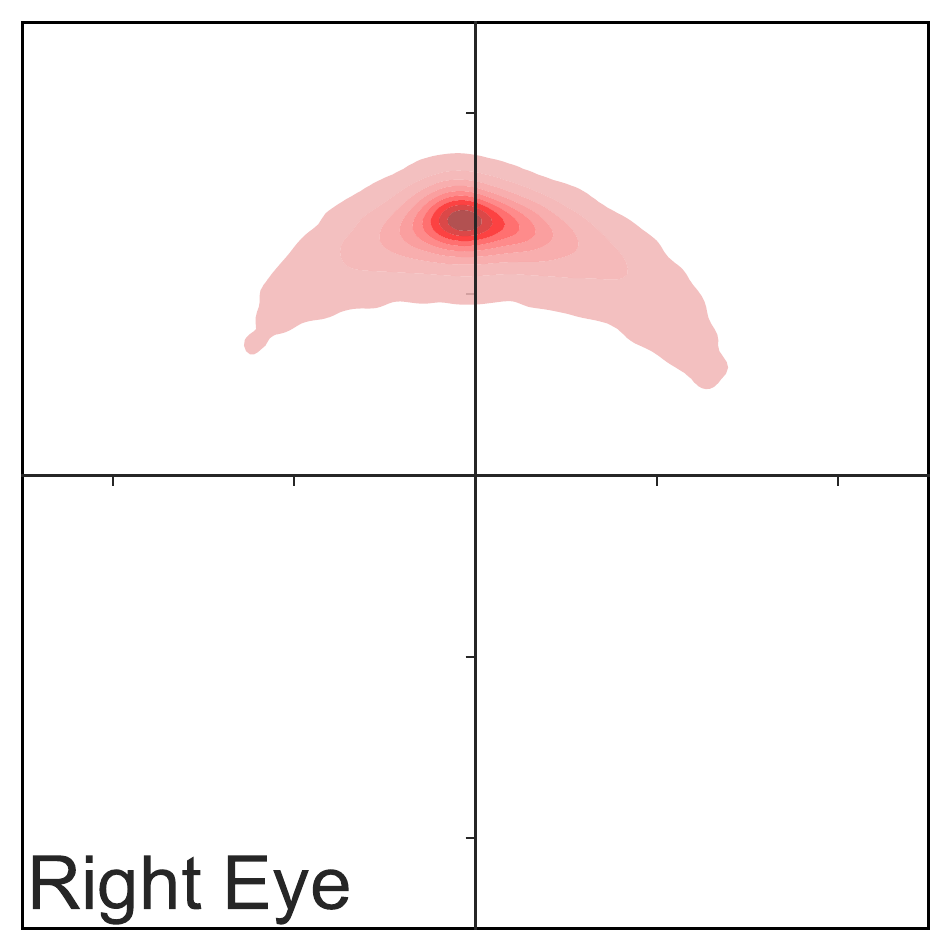}}
    \end{subfigure}
    \hfill
    \begin{subfigure}[t]{0.195\textwidth}
        \raisebox{-\height}{\includegraphics[width=1\textwidth]{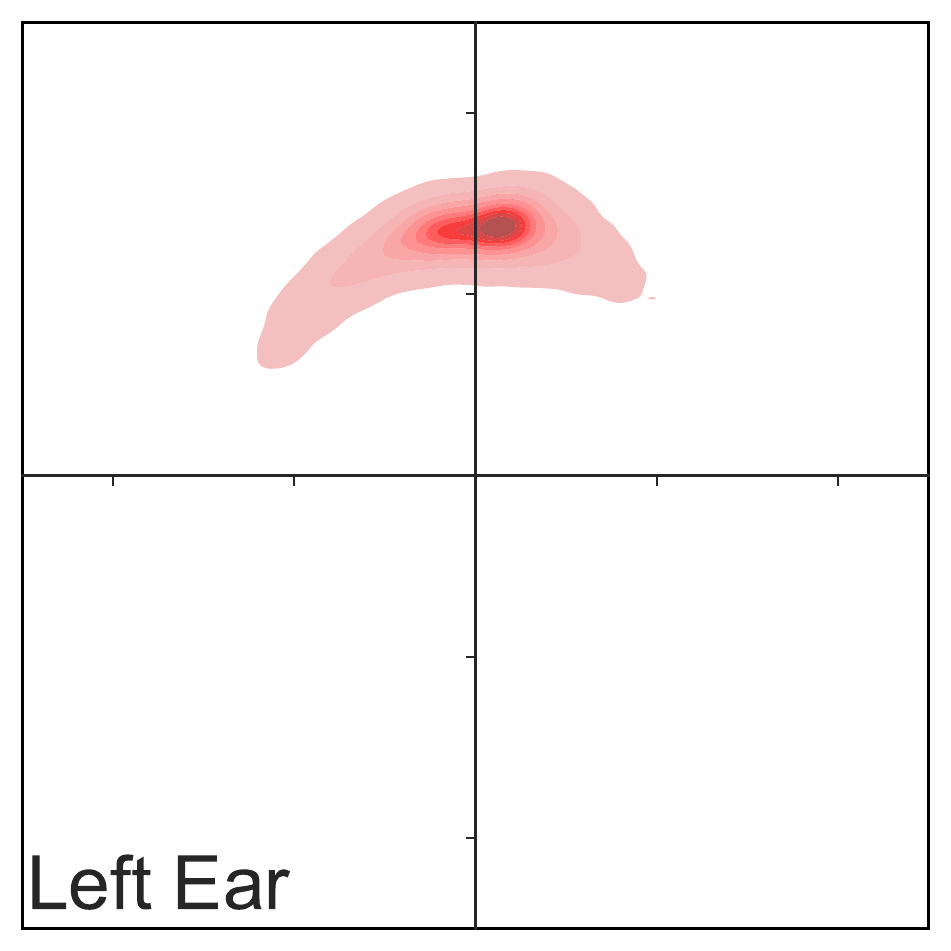}}
    \end{subfigure}
    \hfill
    \begin{subfigure}[t]{0.195\textwidth}
        \raisebox{-\height}{\includegraphics[width=1\textwidth]{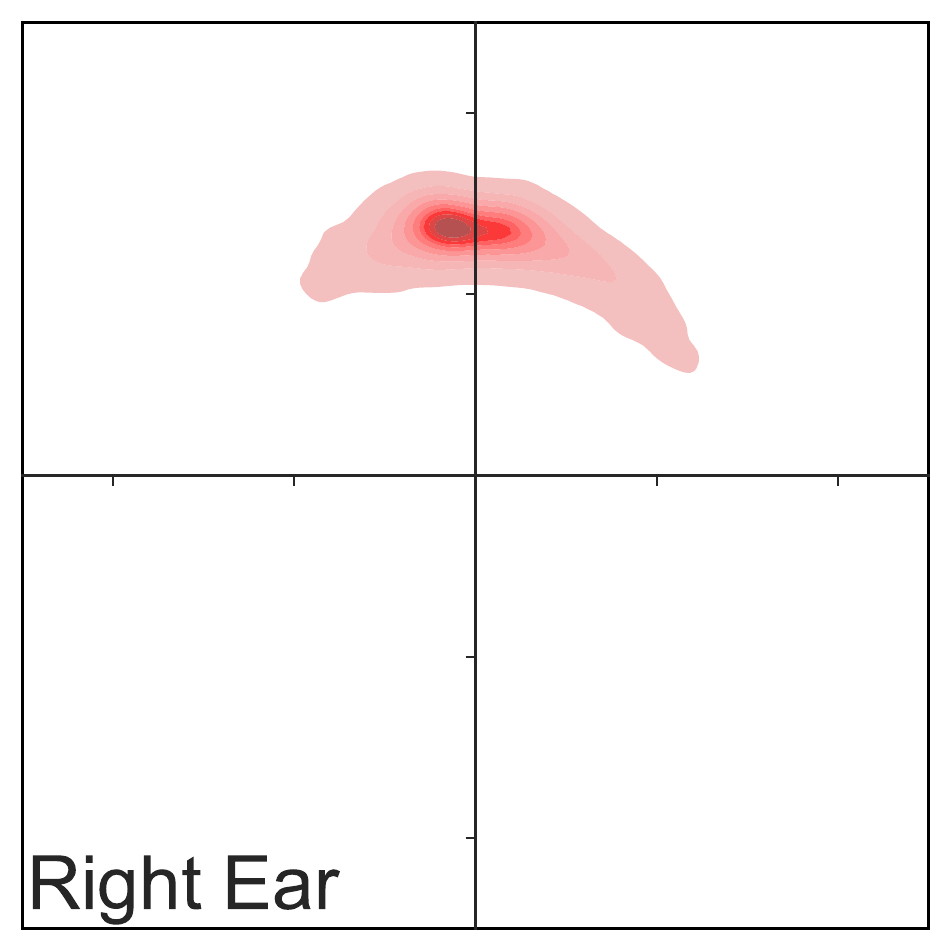}}
    \end{subfigure}
    \begin{subfigure}[t]{0.195\textwidth}
        \raisebox{-\height}{\includegraphics[width=1\textwidth]{neurips_data_2021/plots/fig11_pose_panel/synth/nose_heatmap.pdf}}
    \end{subfigure}
    \hfill
    \begin{subfigure}[t]{0.195\textwidth}
        \raisebox{-\height}{\includegraphics[width=1\textwidth]{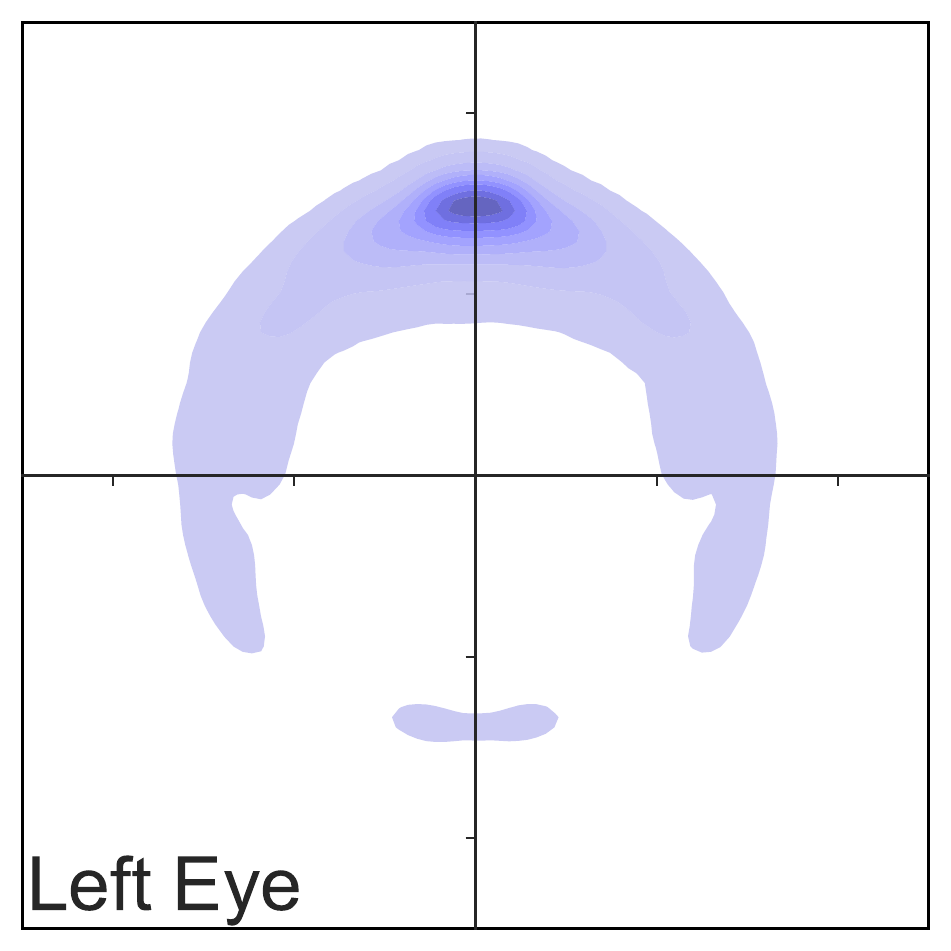}}
    \end{subfigure}
    \hfill
    \begin{subfigure}[t]{0.195\textwidth}
        \raisebox{-\height}{\includegraphics[width=1\textwidth]{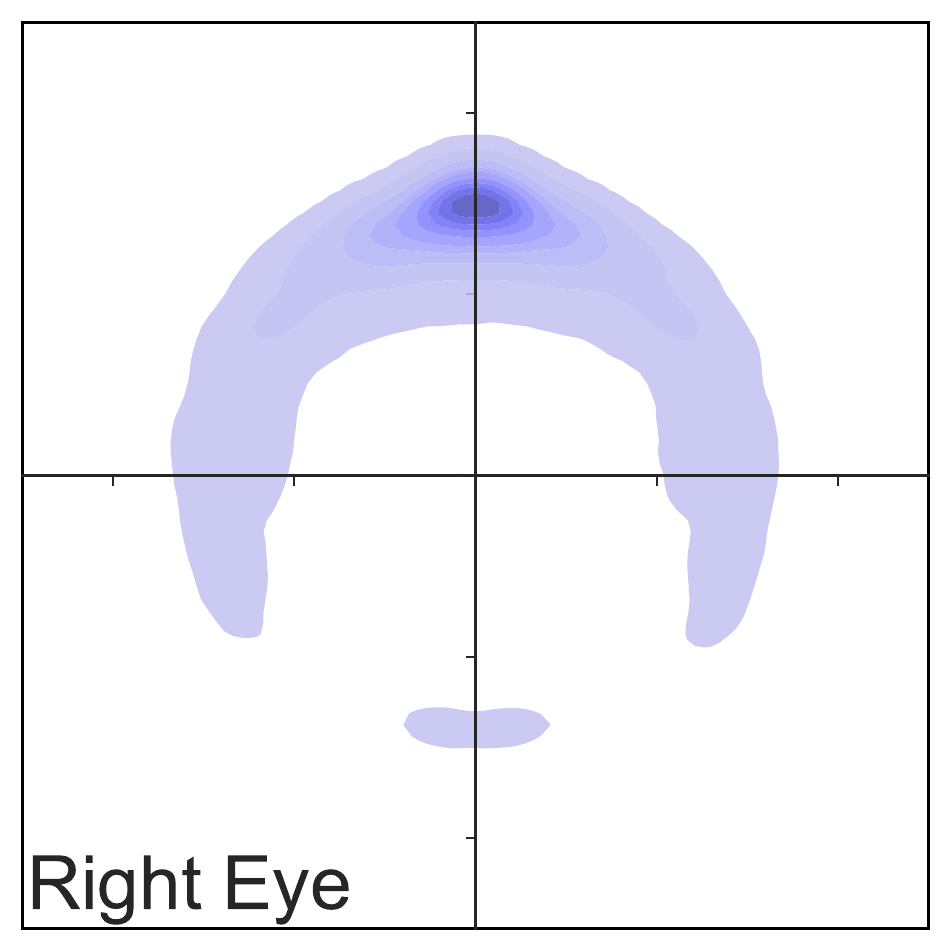}}
    \end{subfigure}
    \hfill
    \begin{subfigure}[t]{0.195\textwidth}
        \raisebox{-\height}{\includegraphics[width=1\textwidth]{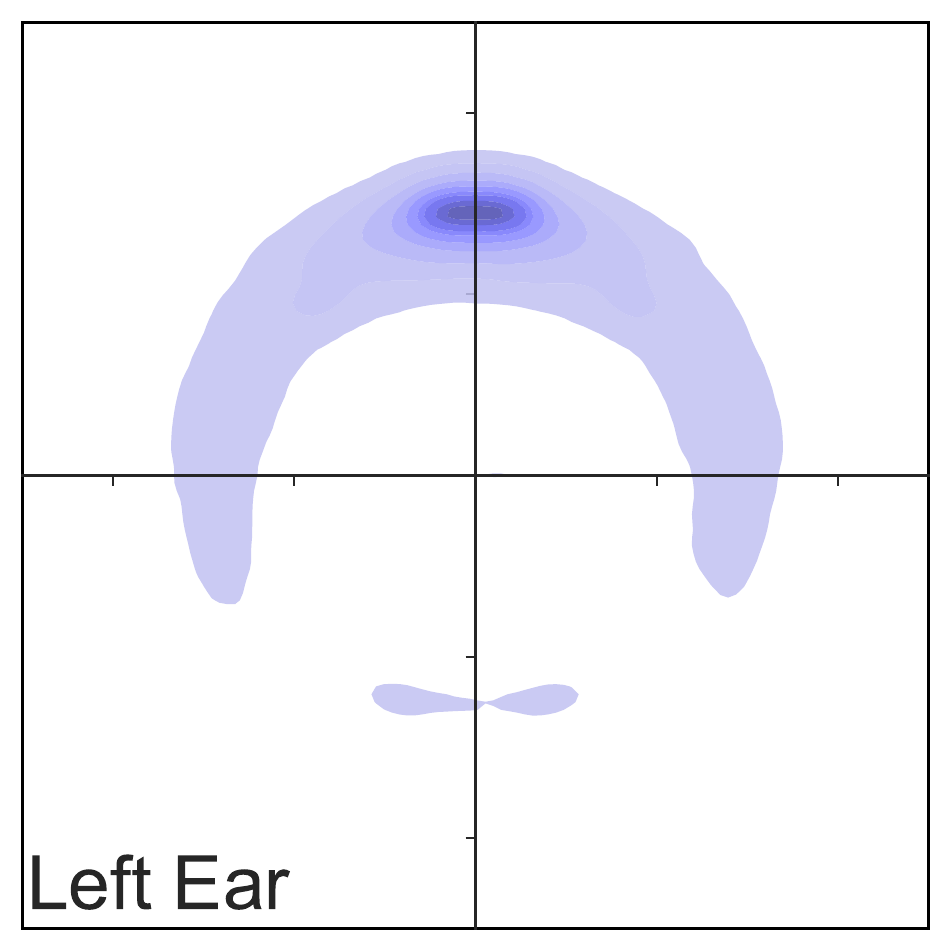}}
    \end{subfigure}
    \hfill
    \begin{subfigure}[t]{0.195\textwidth}
        \raisebox{-\height}{\includegraphics[width=1\textwidth]{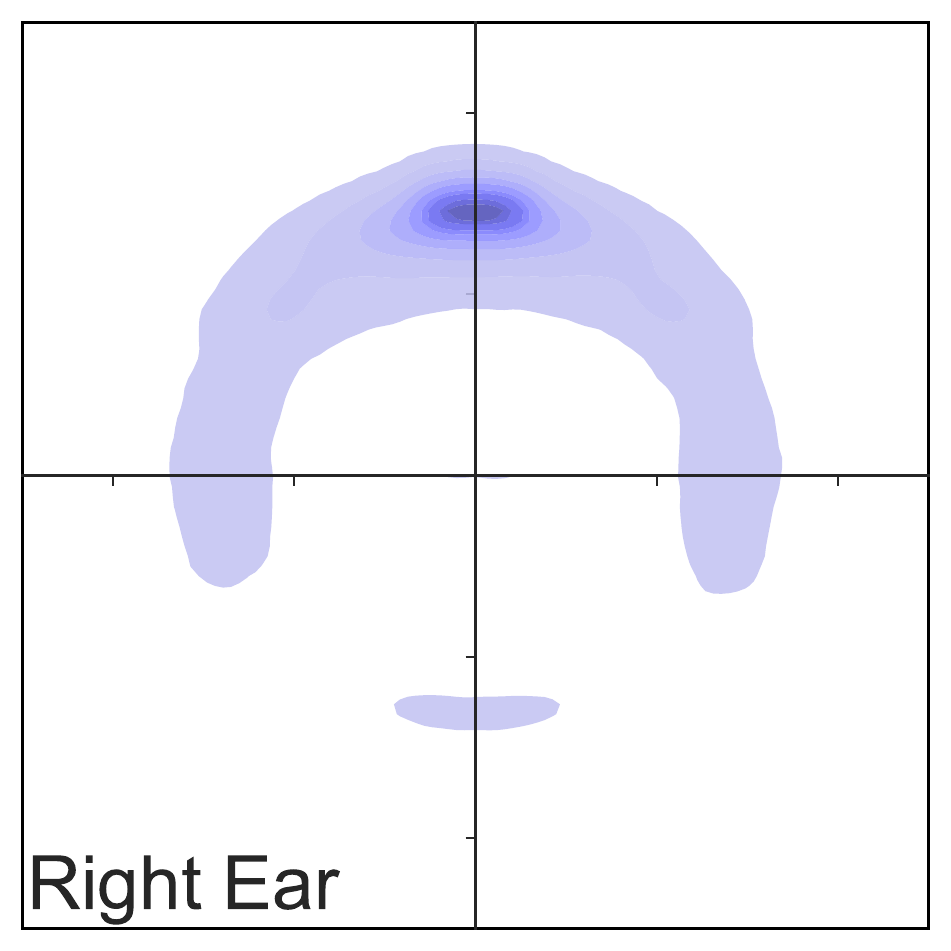}}
    \end{subfigure}
    \begin{subfigure}[t]{0.195\textwidth}
        \raisebox{-\height}{\includegraphics[width=1\textwidth]{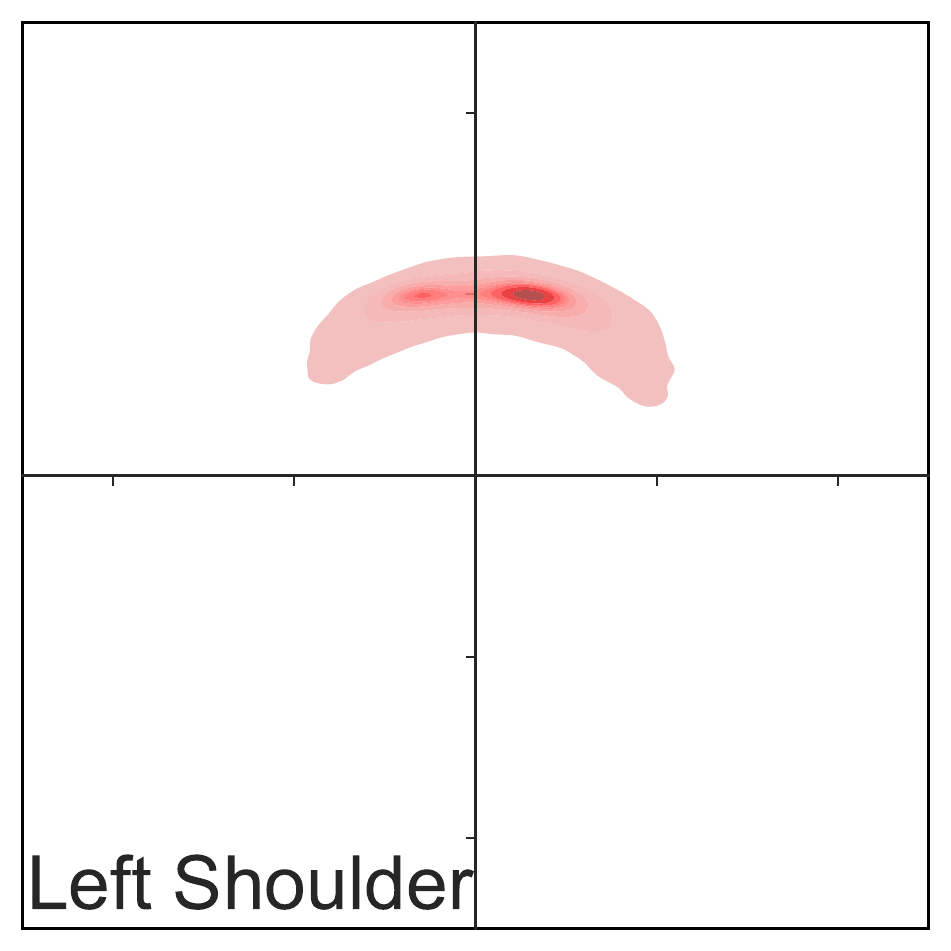}}
    \end{subfigure}
    \hfill
    \begin{subfigure}[t]{0.195\textwidth}
        \raisebox{-\height}{\includegraphics[width=1\textwidth]{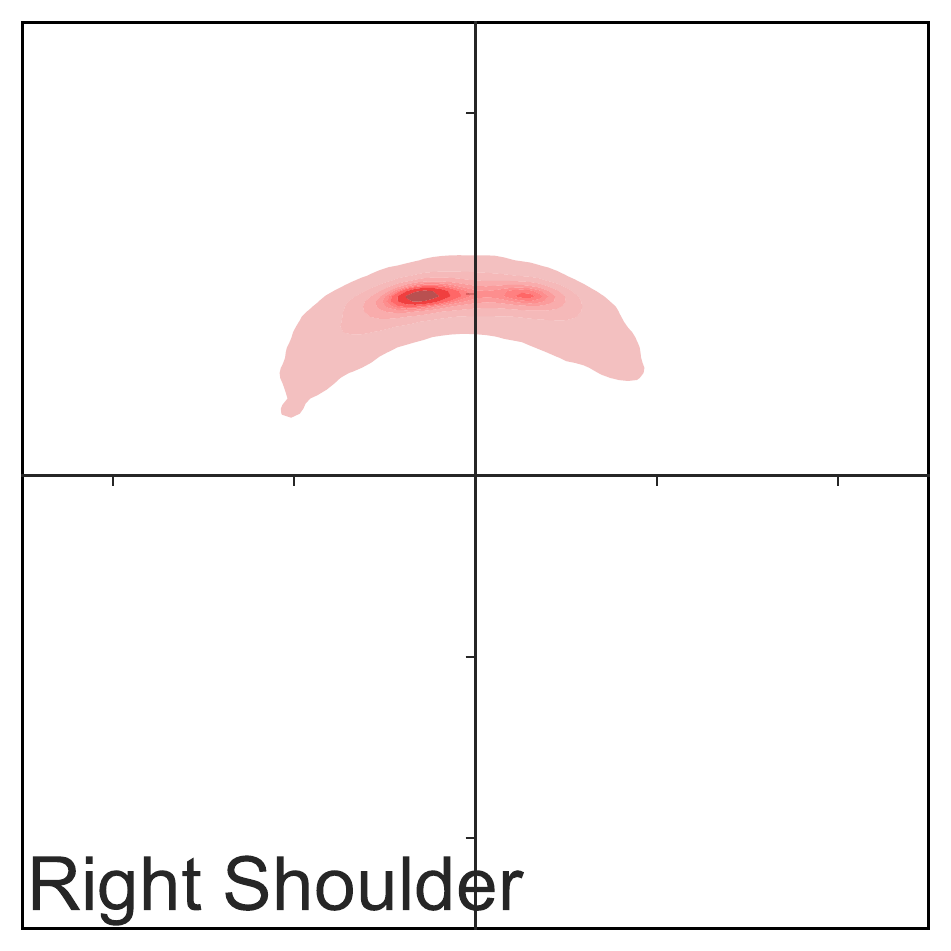}}
    \end{subfigure}
    \hfill
    \begin{subfigure}[t]{0.195\textwidth}
        \raisebox{-\height}{\includegraphics[width=1\textwidth]{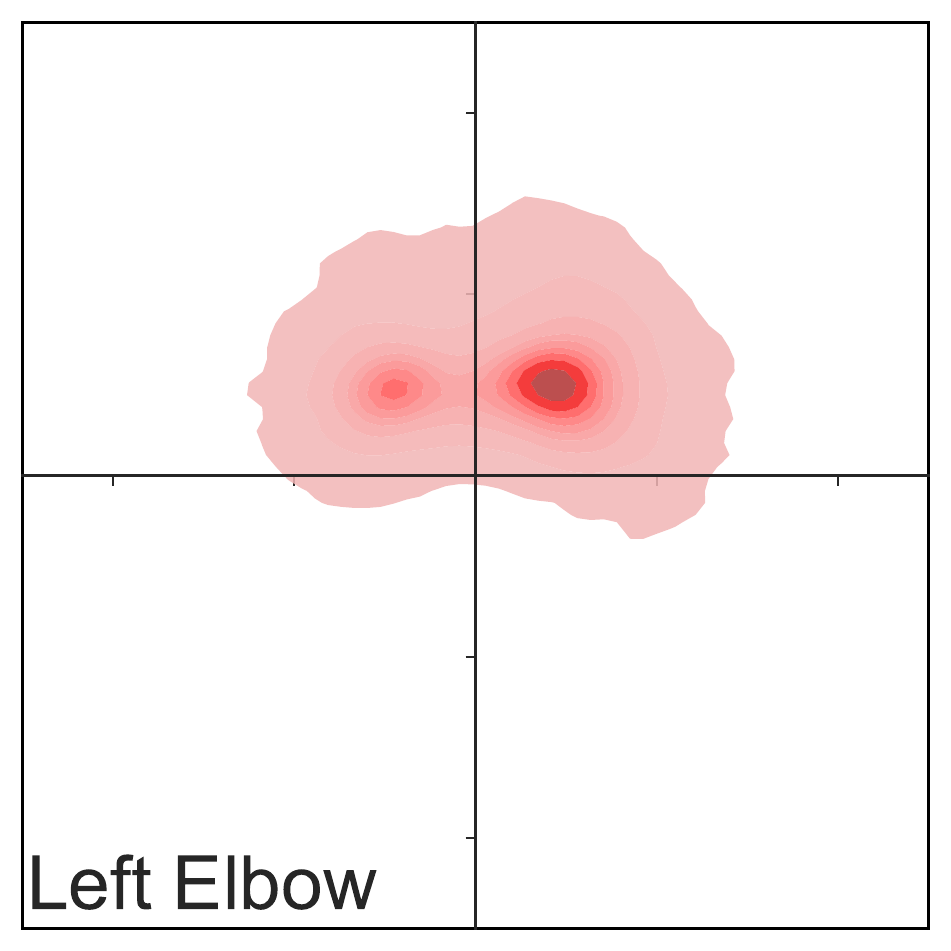}}
    \end{subfigure}
    \hfill
    \begin{subfigure}[t]{0.195\textwidth}
        \raisebox{-\height}{\includegraphics[width=1\textwidth]{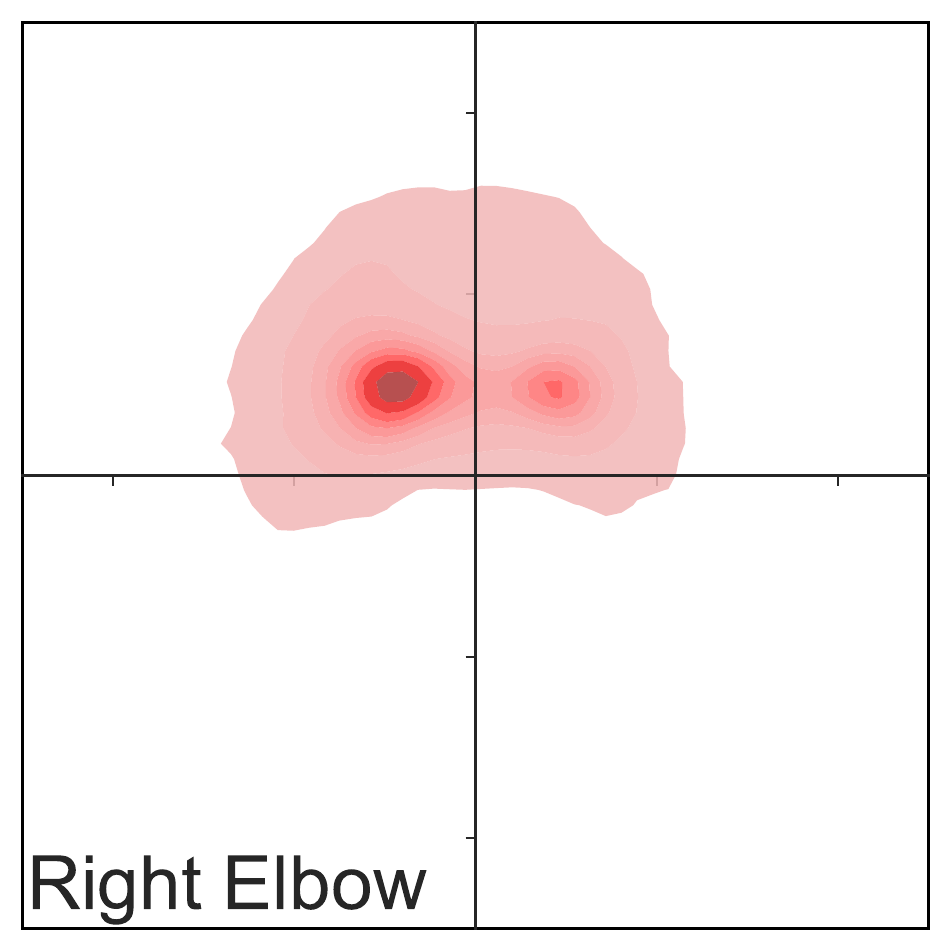}}
    \end{subfigure}
    \hfill
    \begin{subfigure}[t]{0.195\textwidth}
        \raisebox{-\height}{\includegraphics[width=1\textwidth]{neurips_data_2021/plots/fig11_pose_panel/coco/left_wrist_heatmap.pdf}}
    \end{subfigure}
    \begin{subfigure}[t]{0.195\textwidth}
        \raisebox{-\height}{\includegraphics[width=1\textwidth]{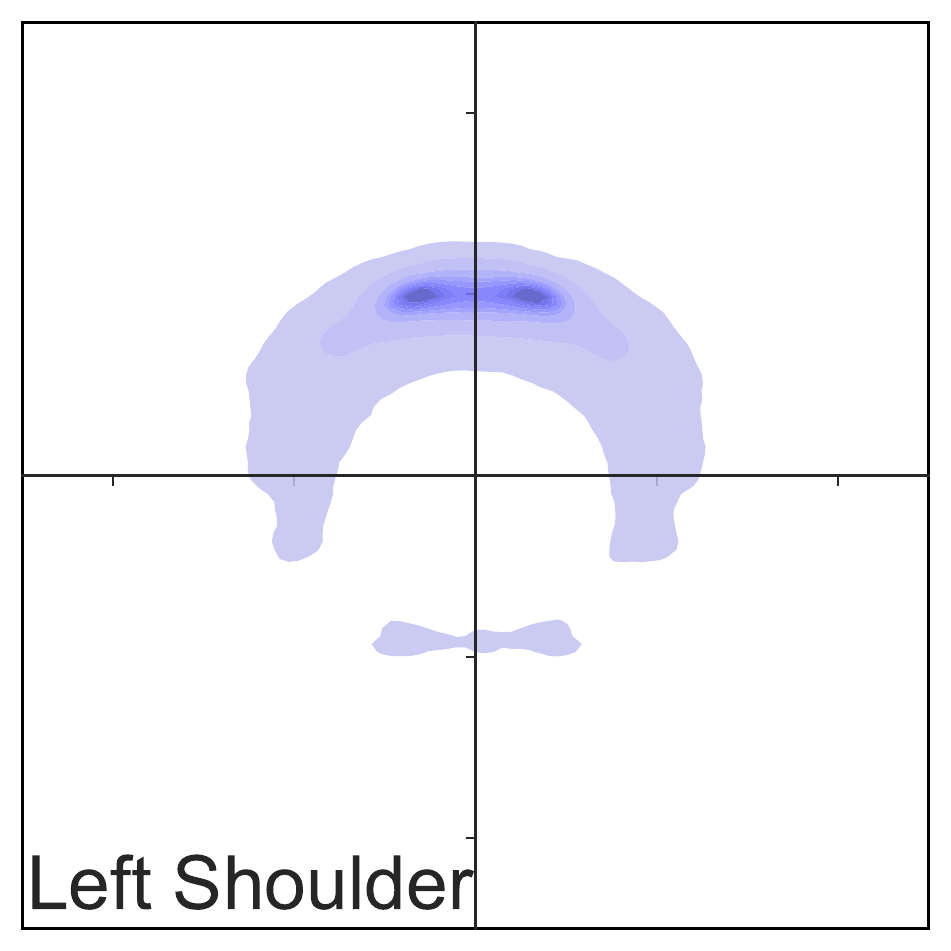}}
    \end{subfigure}
    \hfill
    \begin{subfigure}[t]{0.195\textwidth}
        \raisebox{-\height}{\includegraphics[width=1\textwidth]{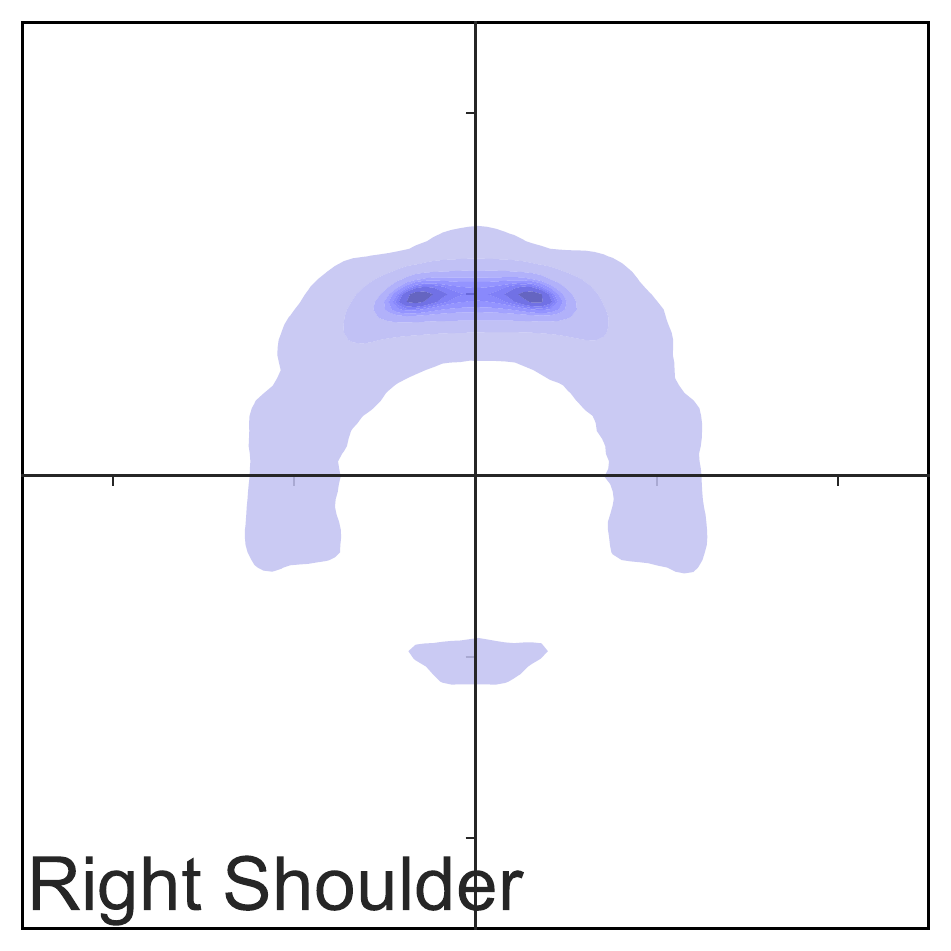}}
    \end{subfigure}
    \hfill
    \begin{subfigure}[t]{0.195\textwidth}
        \raisebox{-\height}{\includegraphics[width=1\textwidth]{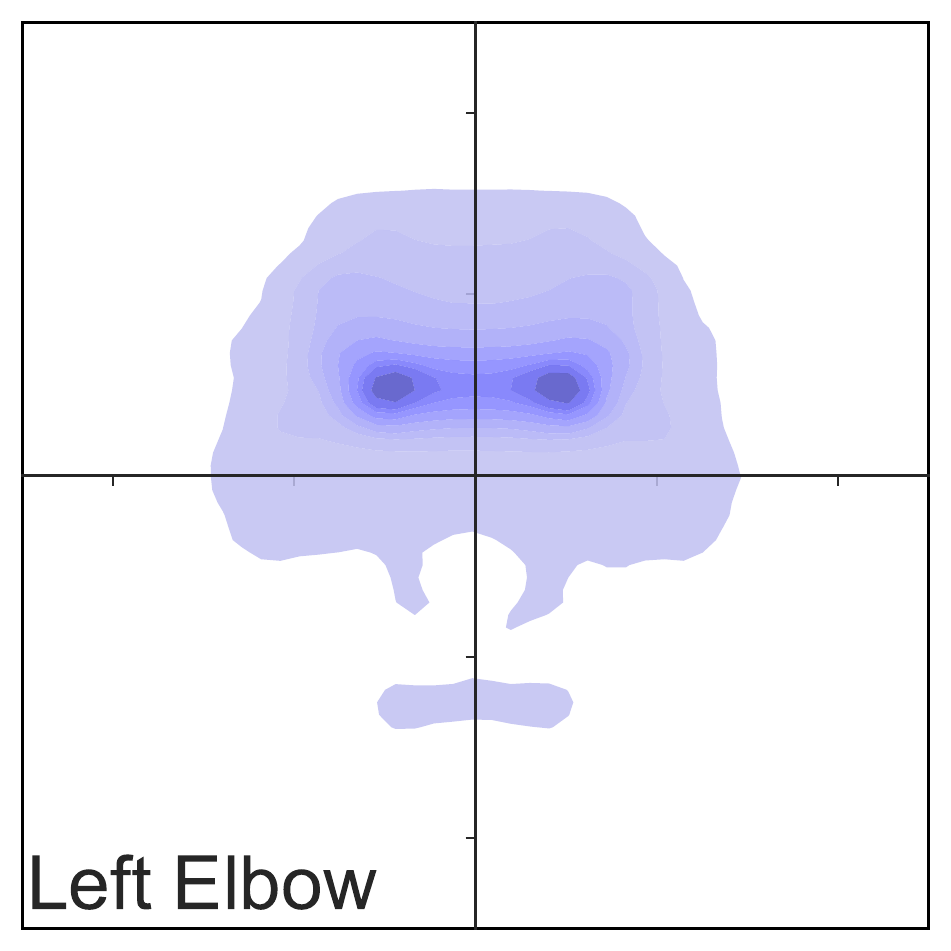}}
    \end{subfigure}
    \hfill
    \begin{subfigure}[t]{0.195\textwidth}
        \raisebox{-\height}{\includegraphics[width=1\textwidth]{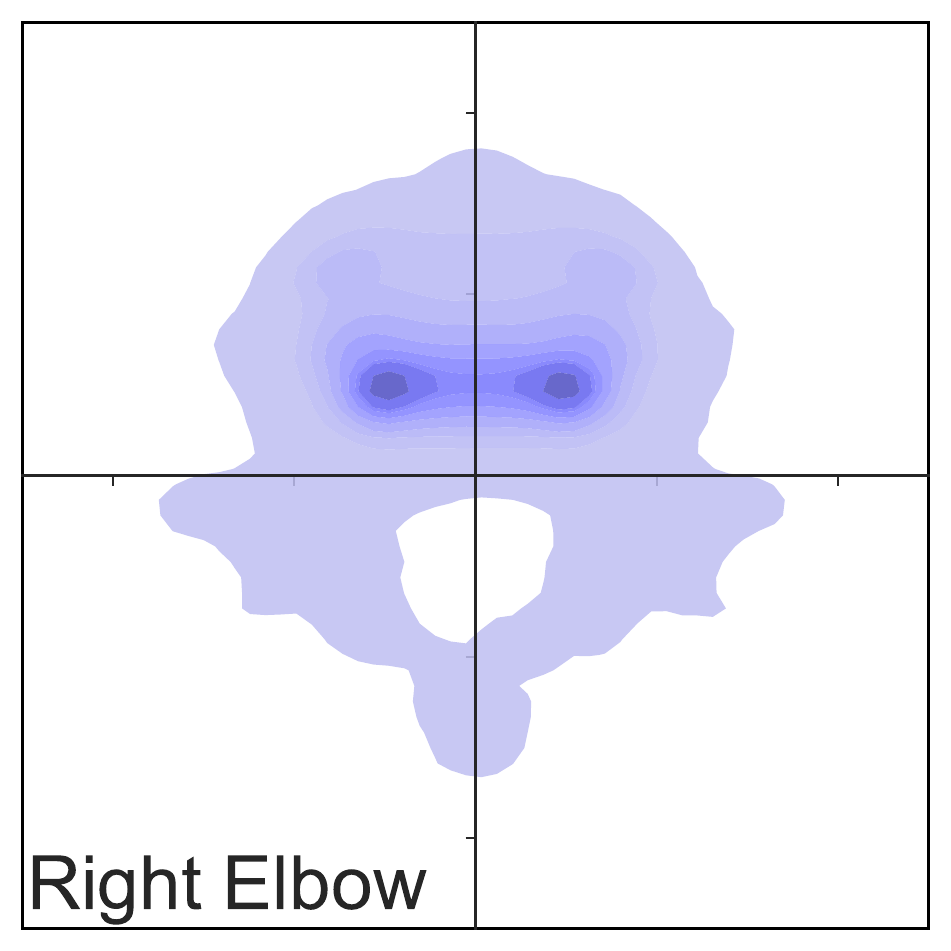}}
    \end{subfigure}
    \hfill
    \begin{subfigure}[t]{0.195\textwidth}
        \raisebox{-\height}{\includegraphics[width=1\textwidth]{neurips_data_2021/plots/fig11_pose_panel//synth/left_wrist_heatmap.pdf}}
    \end{subfigure}
    \begin{subfigure}[t]{0.195\textwidth}
        \raisebox{-\height}{\includegraphics[width=1\textwidth]{neurips_data_2021/plots/fig11_pose_panel/coco/right_wrist_heatmap.pdf}}
    \end{subfigure}
    \hfill
    \begin{subfigure}[t]{0.195\textwidth}
        \raisebox{-\height}{\includegraphics[width=1\textwidth]{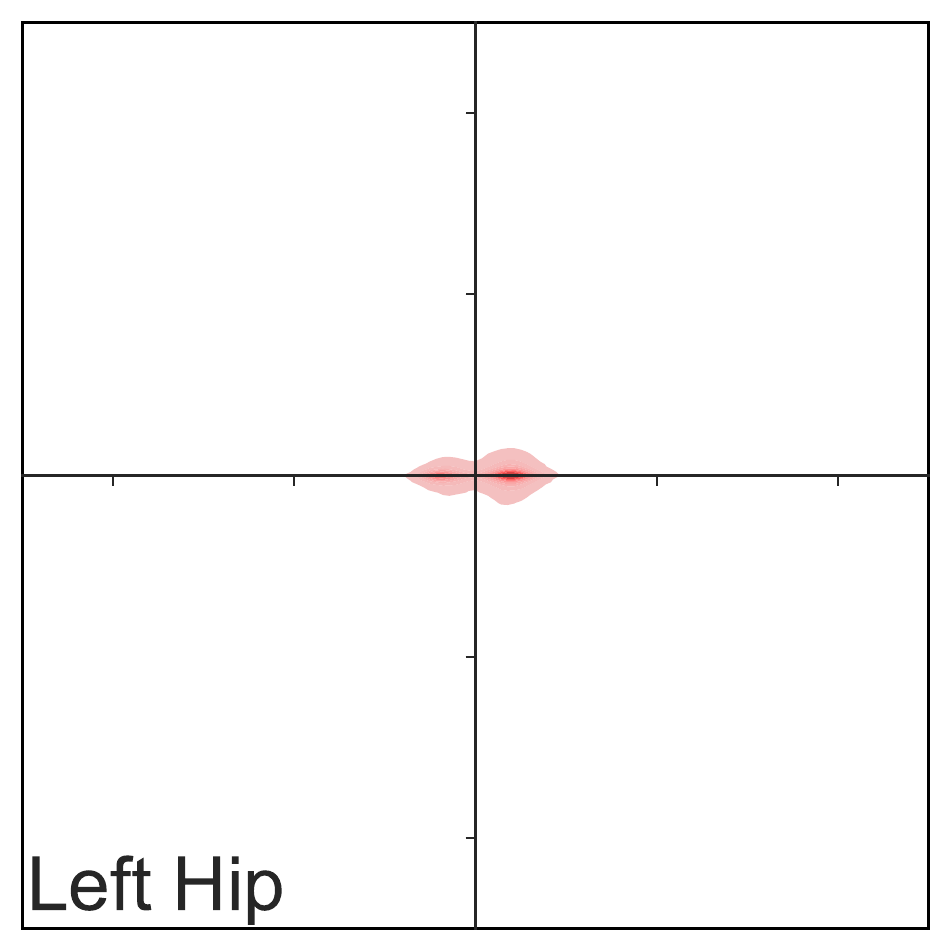}}
    \end{subfigure}
    \hfill
    \begin{subfigure}[t]{0.195\textwidth}
        \raisebox{-\height}{\includegraphics[width=1\textwidth]{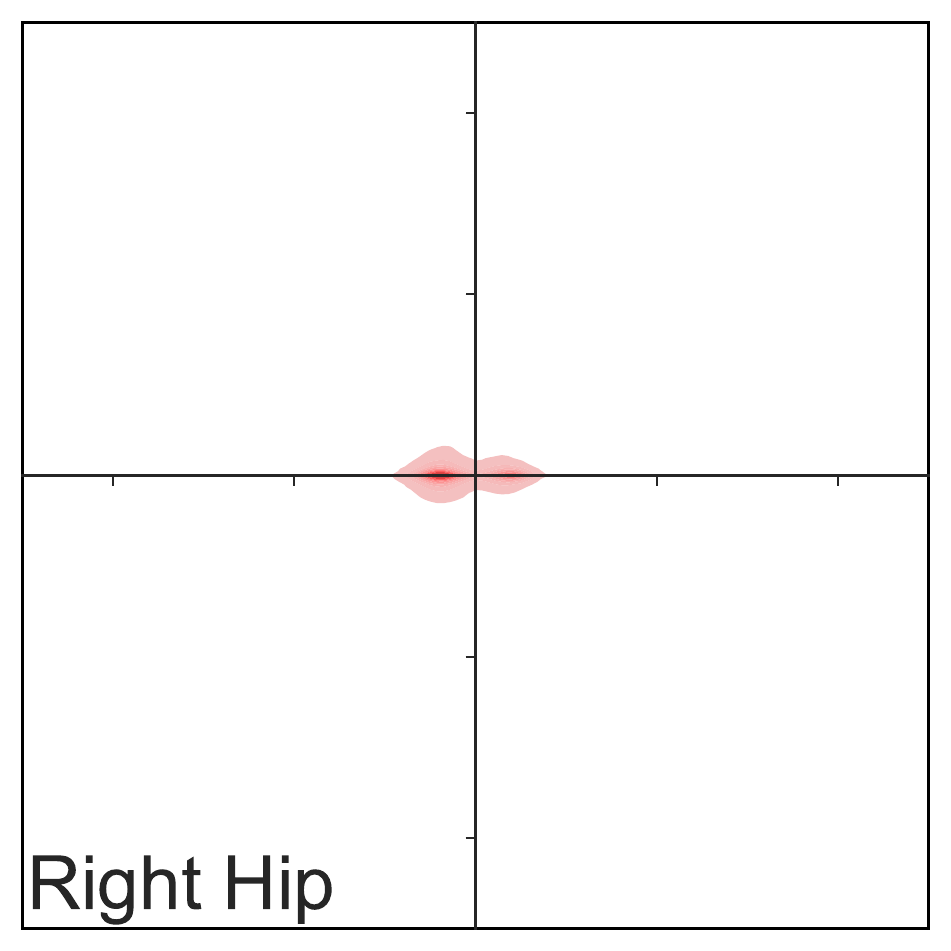}}
    \end{subfigure}
    \hfill
    \begin{subfigure}[t]{0.195\textwidth}
        \raisebox{-\height}{\includegraphics[width=1\textwidth]{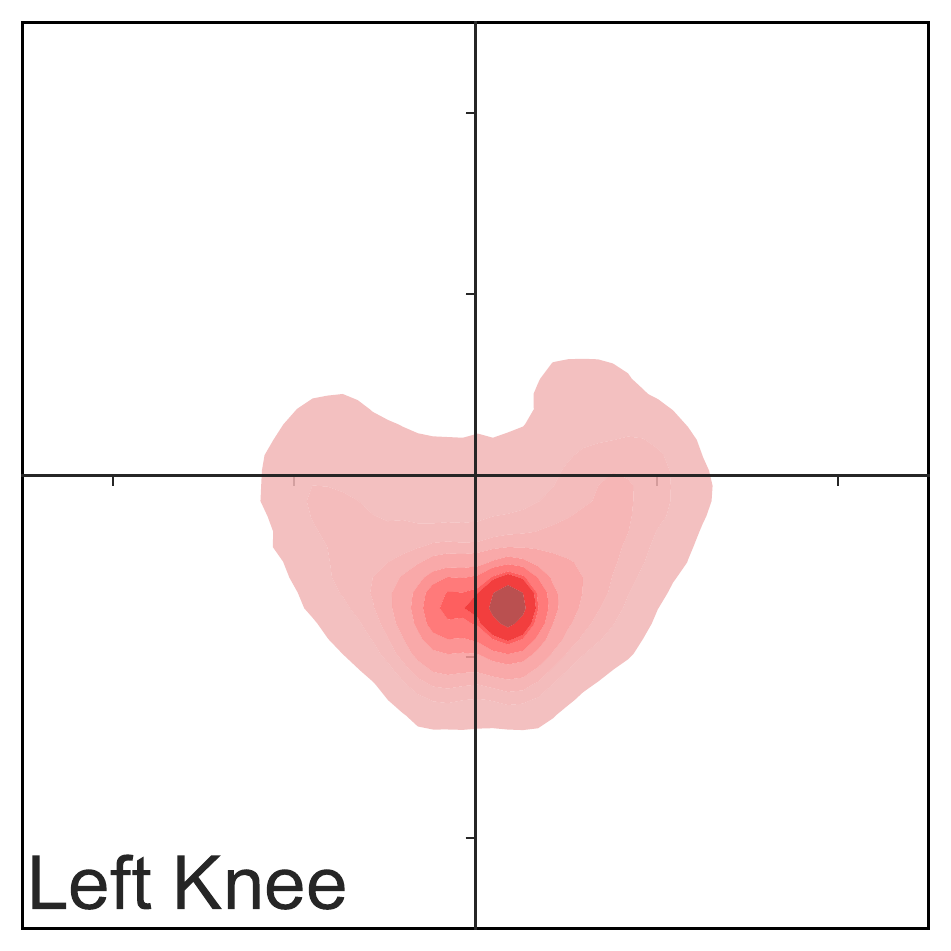}}
    \end{subfigure}
    \hfill
    \begin{subfigure}[t]{0.195\textwidth}
        \raisebox{-\height}{\includegraphics[width=1\textwidth]{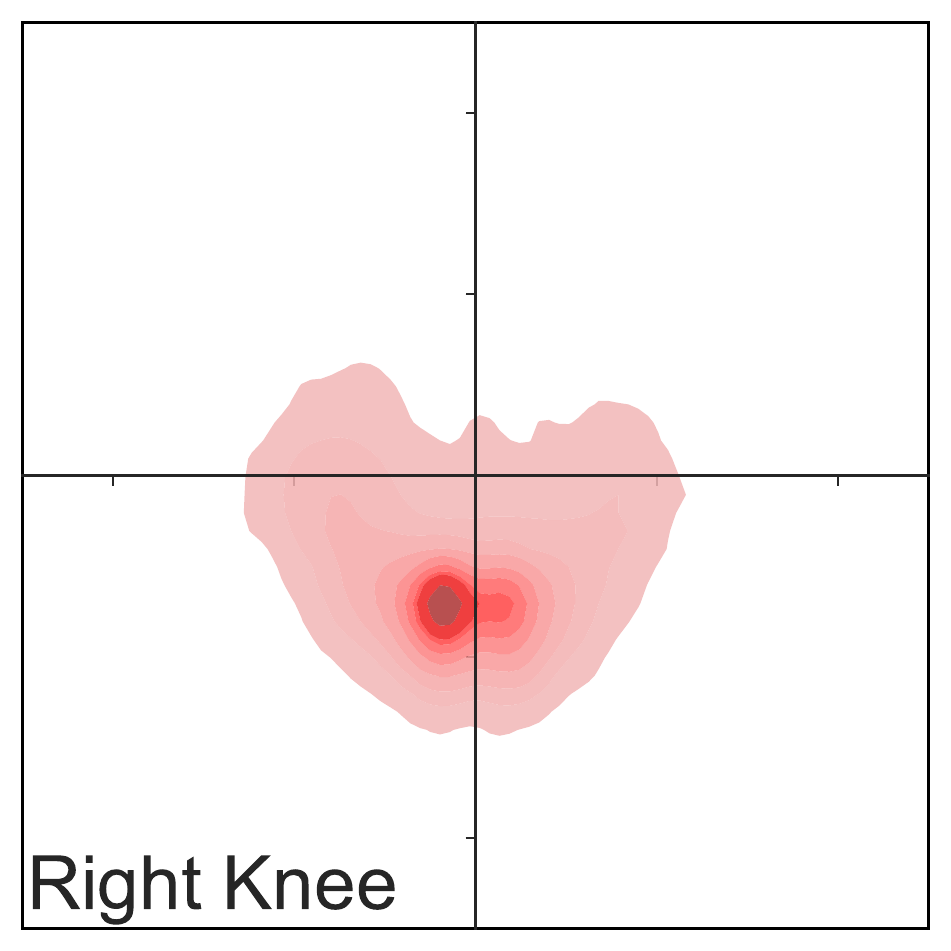}}
    \end{subfigure}
    \begin{subfigure}[t]{0.195\textwidth}
        \raisebox{-\height}{\includegraphics[width=1\textwidth]{neurips_data_2021/plots/fig11_pose_panel//synth/right_wrist_heatmap.pdf}}
    \end{subfigure}
    \hfill
    \begin{subfigure}[t]{0.195\textwidth}
        \raisebox{-\height}{\includegraphics[width=1\textwidth]{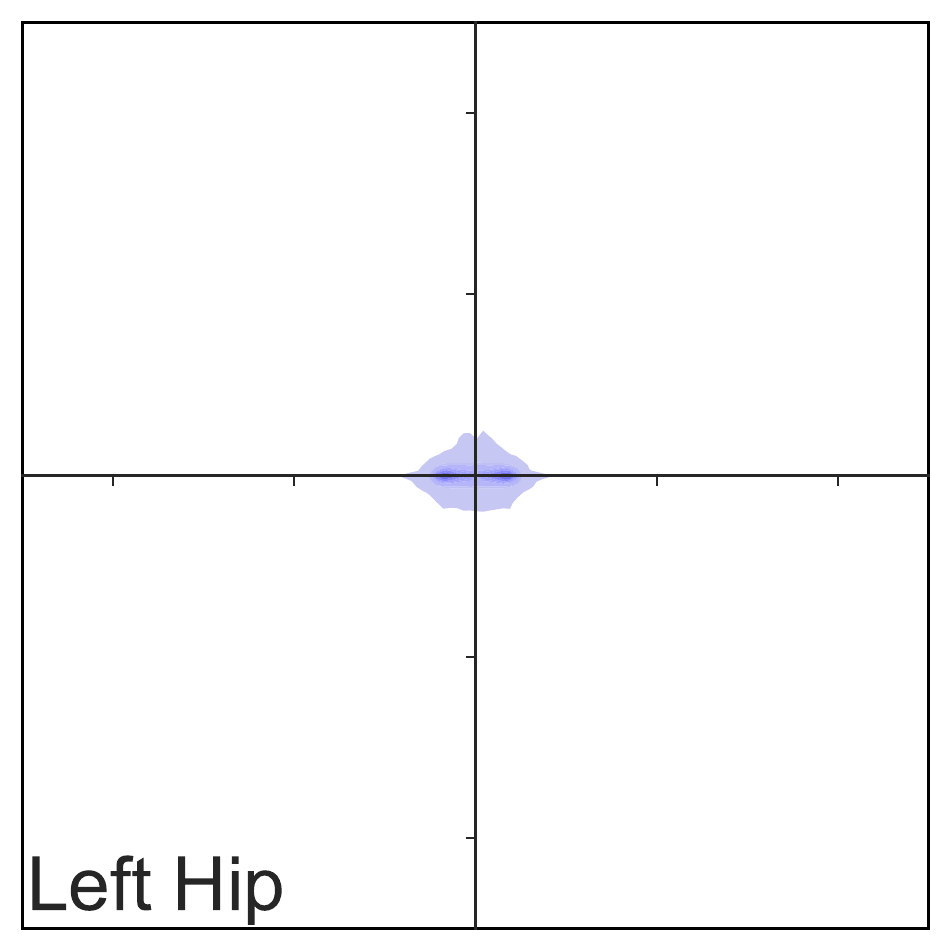}}
    \end{subfigure}
    \hfill
    \begin{subfigure}[t]{0.195\textwidth}
        \raisebox{-\height}{\includegraphics[width=1\textwidth]{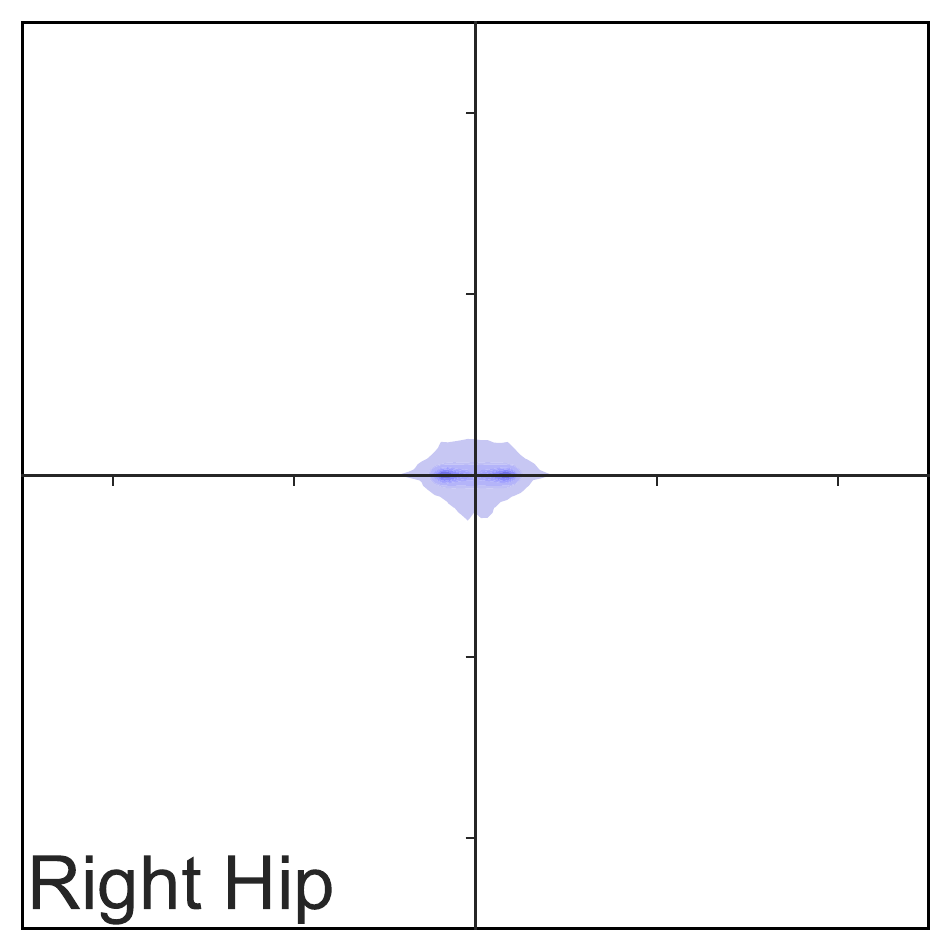}}
    \end{subfigure}
    \hfill
    \begin{subfigure}[t]{0.195\textwidth}
        \raisebox{-\height}{\includegraphics[width=1\textwidth]{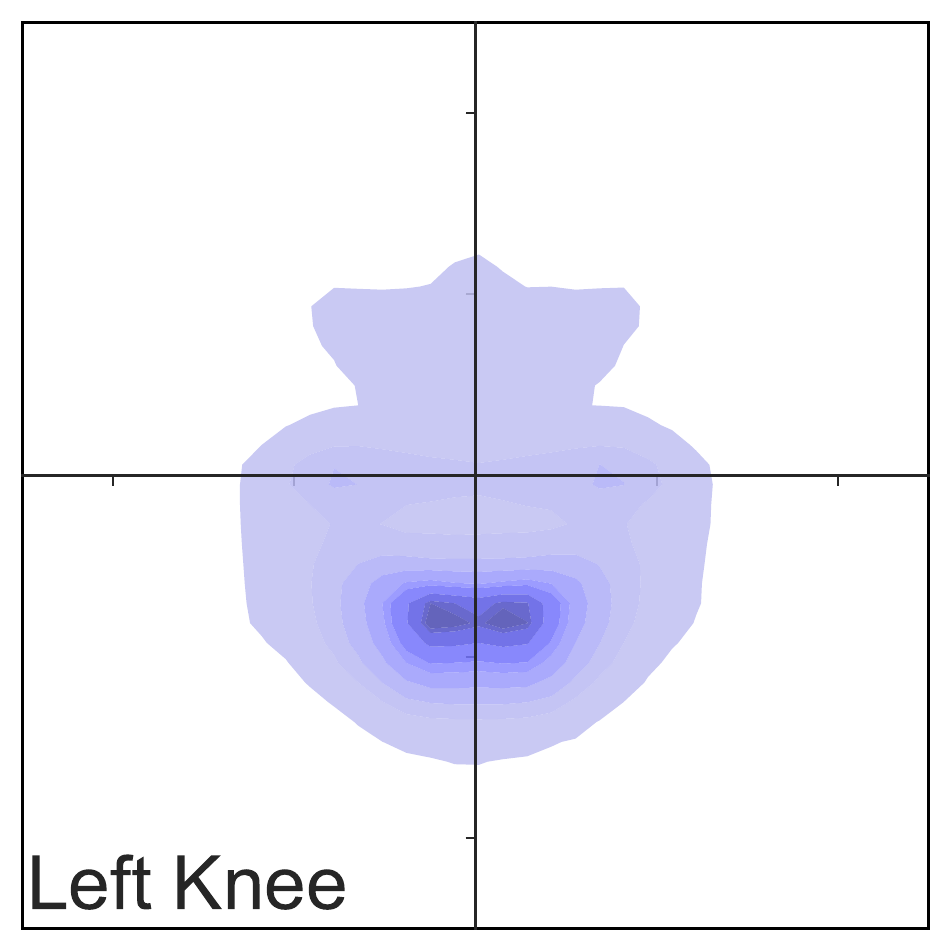}}
    \end{subfigure}
    \hfill
    \begin{subfigure}[t]{0.195\textwidth}
        \raisebox{-\height}{\includegraphics[width=1\textwidth]{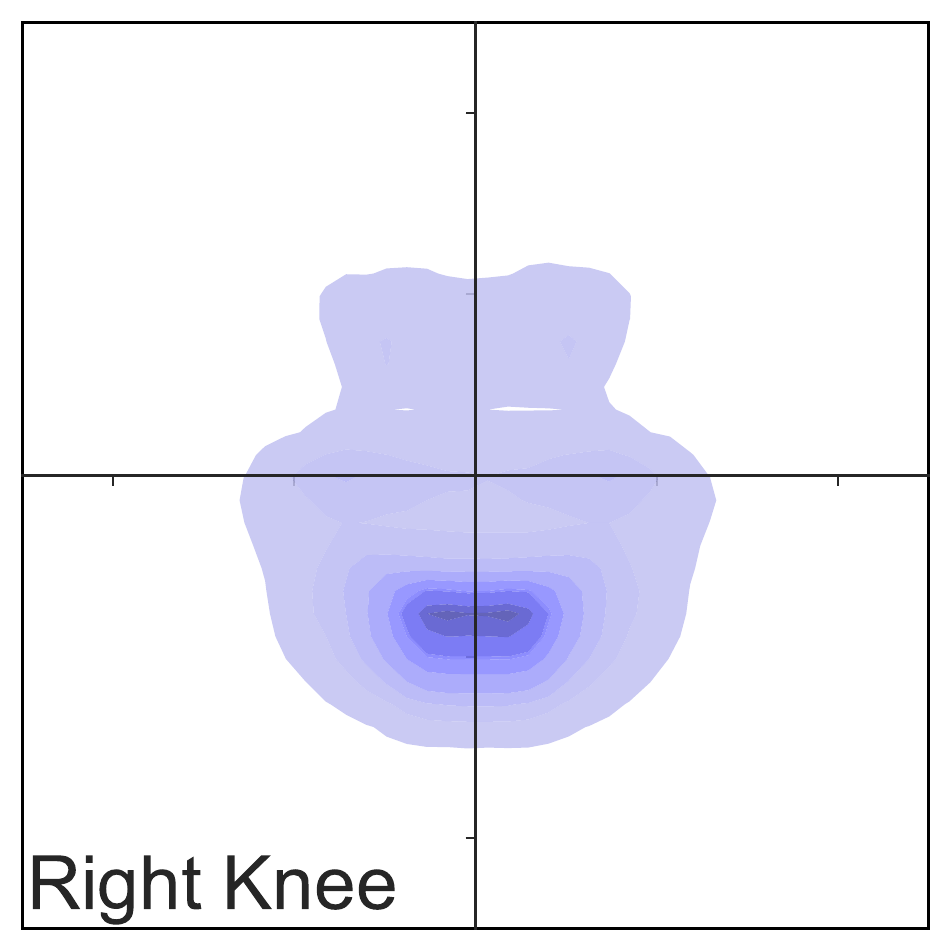}}
    \end{subfigure}
    \hfill
    \begin{subfigure}[t]{0.195\textwidth}
        \raisebox{-\height}{\includegraphics[width=1\textwidth]{neurips_data_2021/plots/fig11_pose_panel/coco/left_ankle_heatmap.pdf}}
    \end{subfigure}
    \begin{subfigure}[t]{0.195\textwidth}
        \raisebox{-\height}{\includegraphics[width=1\textwidth]{neurips_data_2021/plots/fig11_pose_panel/coco/right_ankle_heatmap.pdf}}
    \end{subfigure}
    \begin{subfigure}[t]{0.039\textwidth}
        \raisebox{-\height}{\includegraphics[width=1\textwidth]{neurips_data_2021/plots/fig11_pose_panel/colorbar_red.pdf}}
    \end{subfigure}
    \hfill
    \\
    \begin{subfigure}[t]{0.195\textwidth}
        \raisebox{-\height}{\includegraphics[width=1\textwidth]{neurips_data_2021/plots/fig11_pose_panel/synth/left_ankle_heatmap.pdf}}
    \end{subfigure}
    \begin{subfigure}[t]{0.195\textwidth}
        \raisebox{-\height}{\includegraphics[width=1\textwidth]{neurips_data_2021/plots/fig11_pose_panel/synth/right_ankle_heatmap.pdf}}
    \end{subfigure}
    \begin{subfigure}[t]{0.039\textwidth}
        \raisebox{-\height}{\includegraphics[width=1\textwidth]{neurips_data_2021/plots/fig11_pose_panel/colorbar_blue.pdf}}
    \end{subfigure}
    \caption{\textbf{Keypoint location heatmaps comparison between the COCO dataset (red) and our synthetic dataset (blue).} We aligned all keypoint labels according to \textit{mid-hip} joint, and scaled them proportional to distances between \textit{left shoulder}, \textit{left hip}, \textit{right shoulder}, and \textit{right hip} to produce normalized keypoint locations.}
    \label{fig:posestatsall}
\end{figure}

\begin{figure}[htb] 
    \centering
    \begin{subfigure}[t]{0.195\textwidth}
        \raisebox{-\height}{\includegraphics[width=1\textwidth]{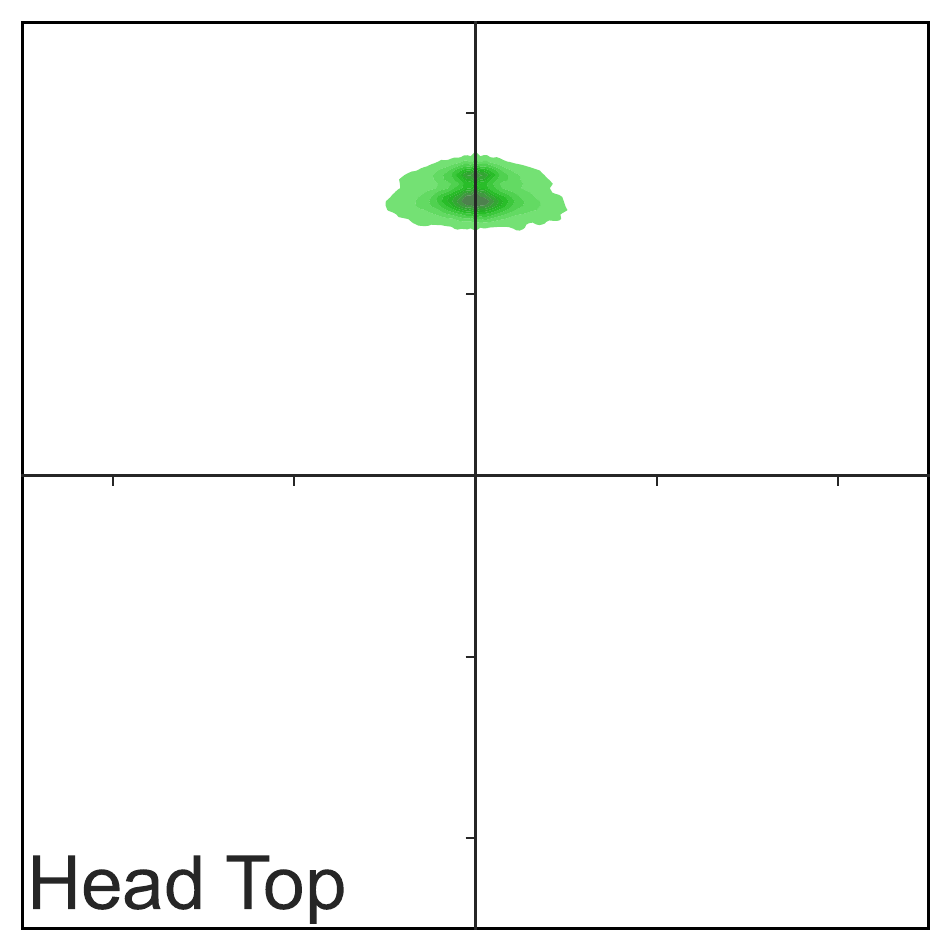}}
    \end{subfigure}
    \hfill
    \begin{subfigure}[t]{0.195\textwidth}
        \raisebox{-\height}{\includegraphics[width=1\textwidth]{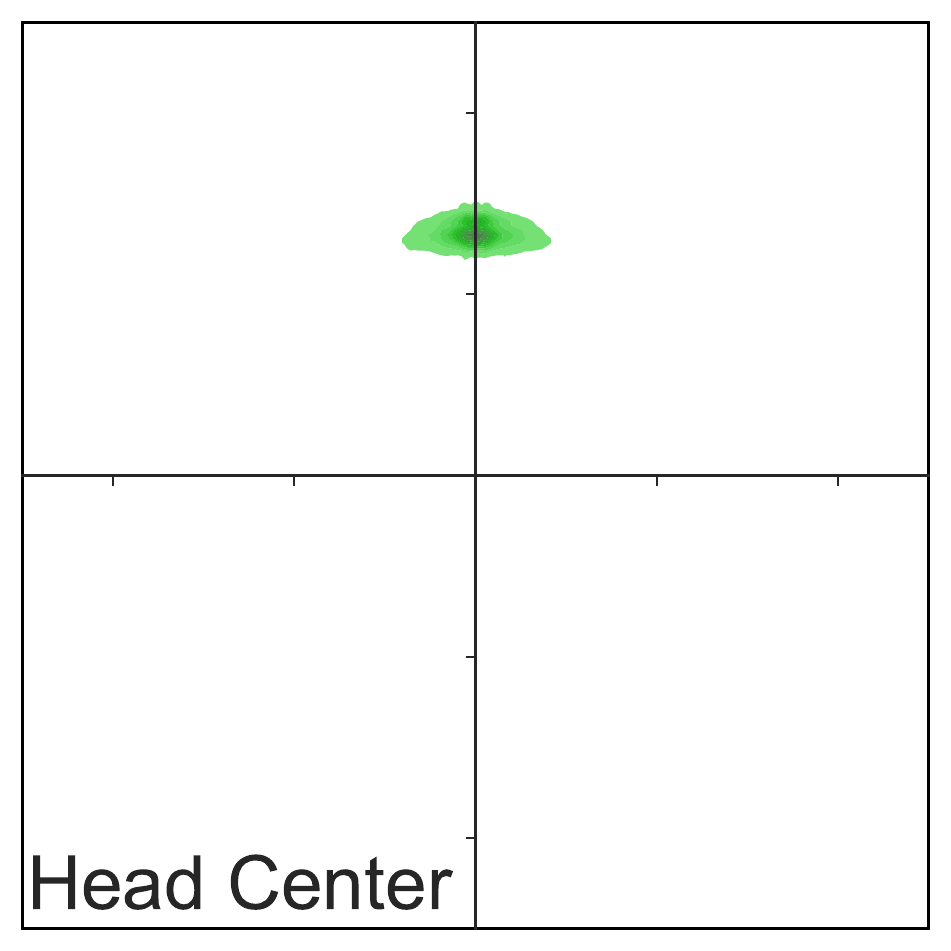}}
    \end{subfigure}
    \hfill
    \begin{subfigure}[t]{0.195\textwidth}
        \raisebox{-\height}{\includegraphics[width=1\textwidth]{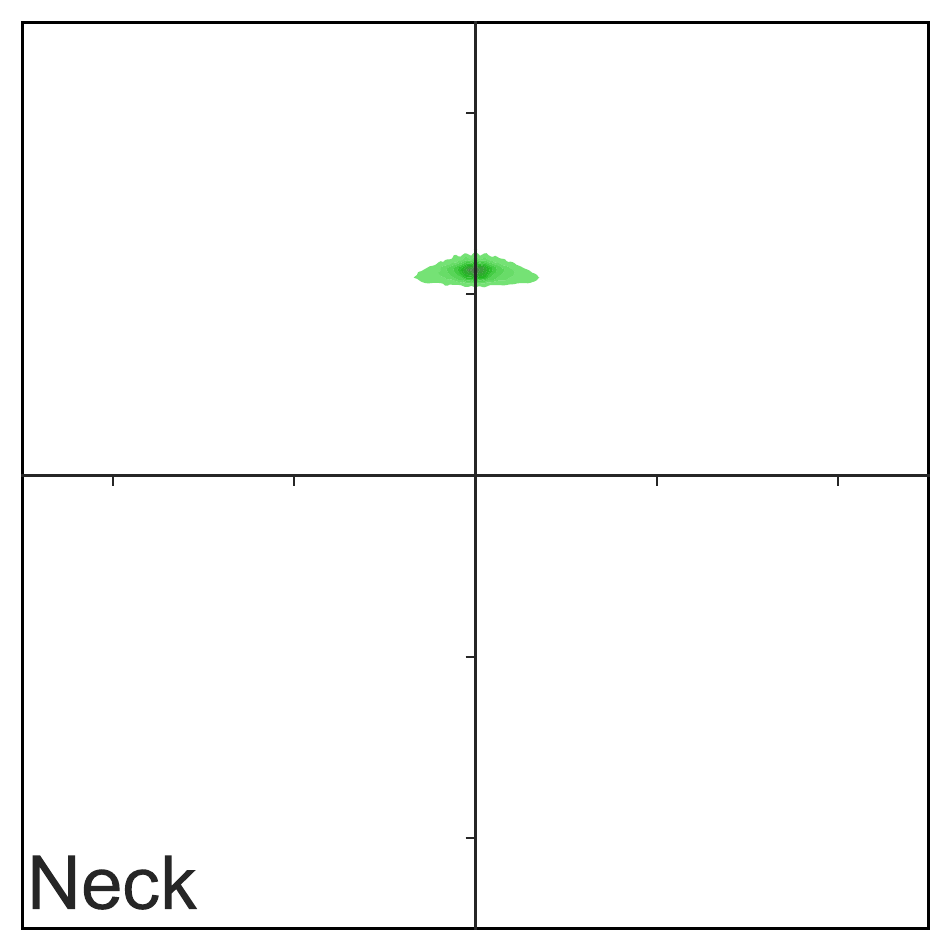}}
    \end{subfigure}
    \hfill
    \begin{subfigure}[t]{0.195\textwidth}
        \raisebox{-\height}{\includegraphics[width=1\textwidth]{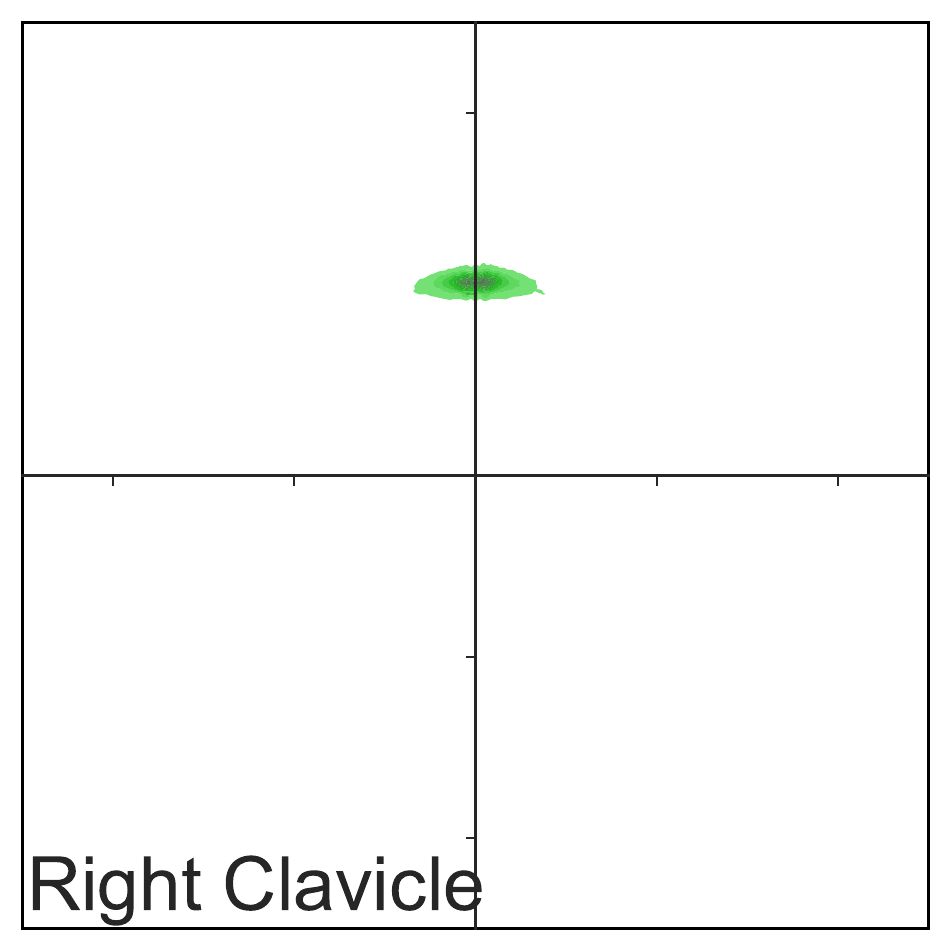}}
    \end{subfigure}
    \hfill
    \begin{subfigure}[t]{0.195\textwidth}
        \raisebox{-\height}{\includegraphics[width=1\textwidth]{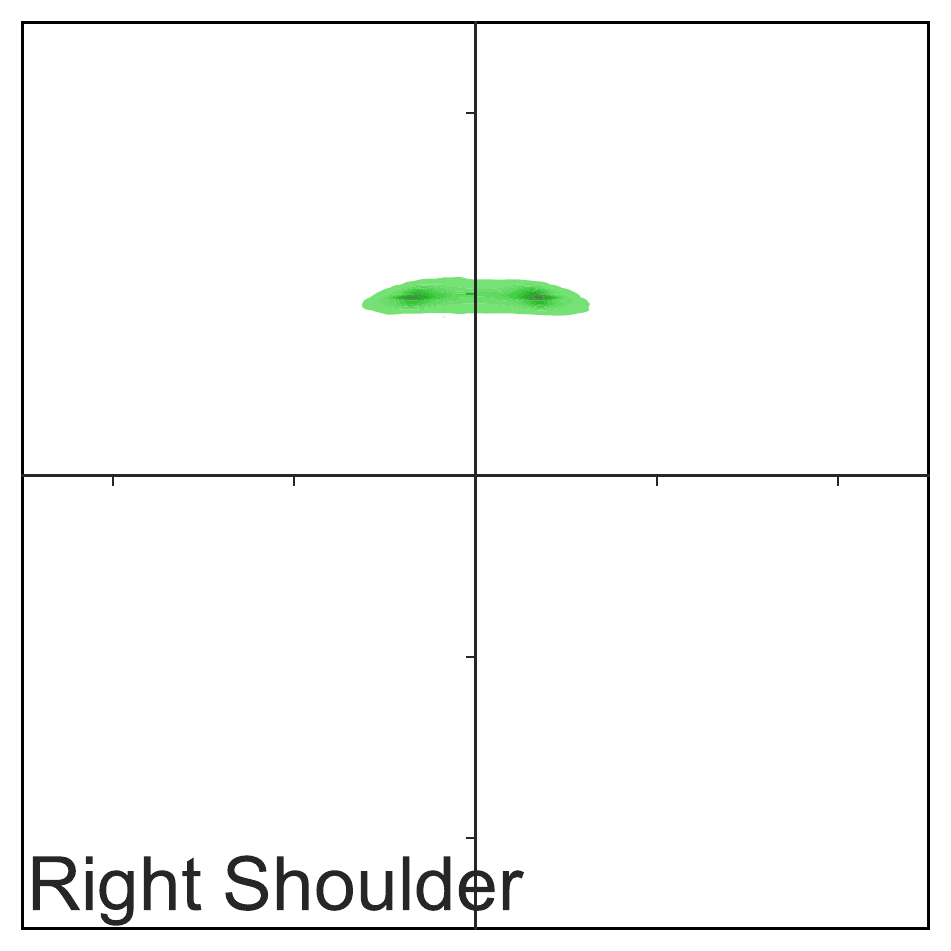}}
    \end{subfigure}
    \begin{subfigure}[t]{0.195\textwidth}
        \raisebox{-\height}{\includegraphics[width=1\textwidth]{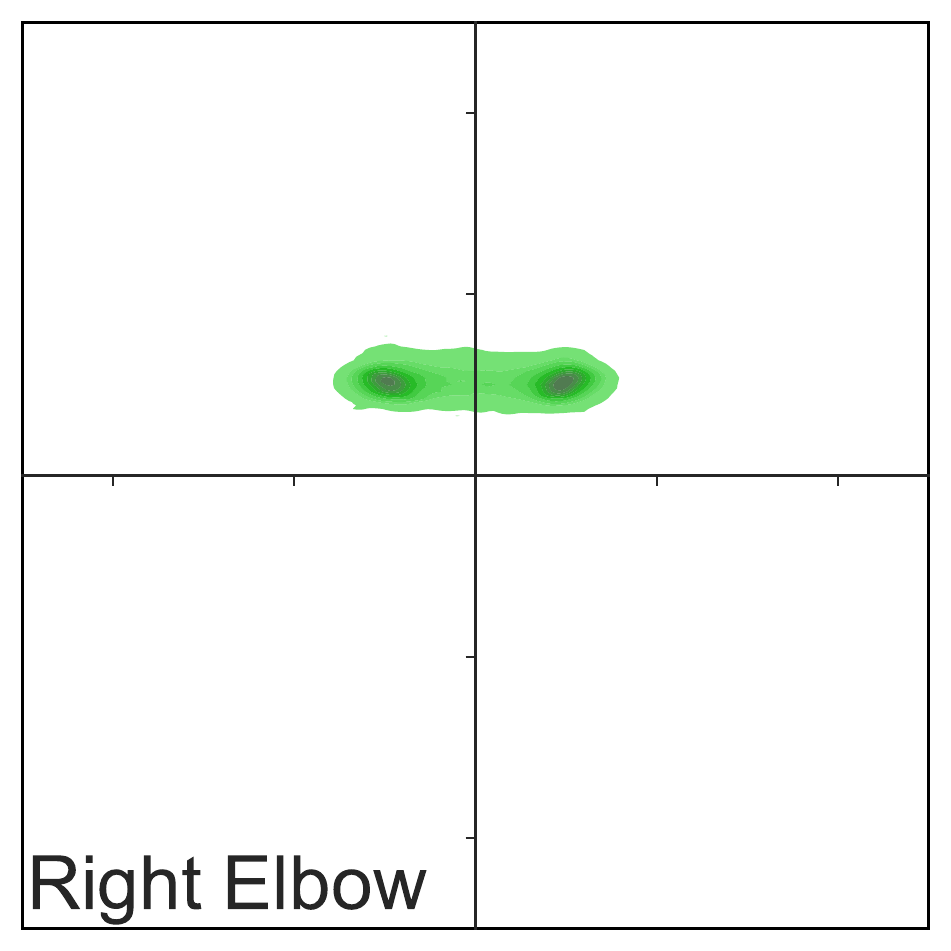}}
    \end{subfigure}
    \hfill
    \begin{subfigure}[t]{0.195\textwidth}
        \raisebox{-\height}{\includegraphics[width=1\textwidth]{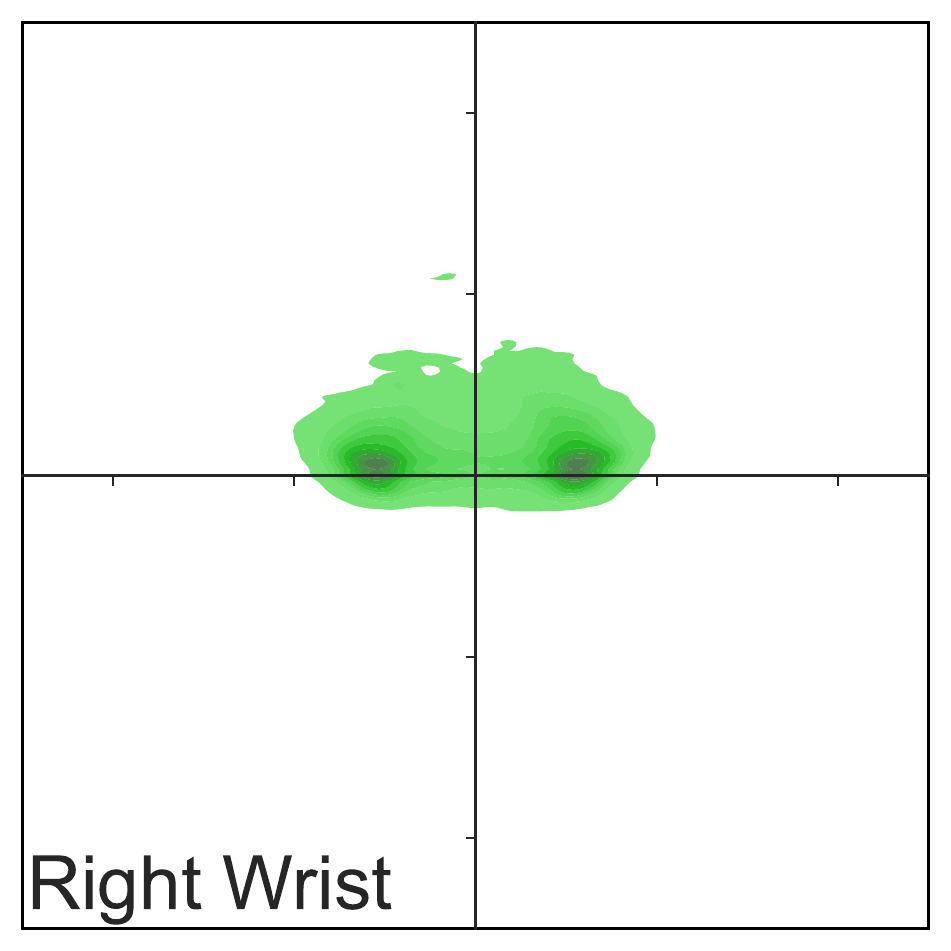}}
    \end{subfigure}
    \hfill
    \begin{subfigure}[t]{0.195\textwidth}
        \raisebox{-\height}{\includegraphics[width=1\textwidth]{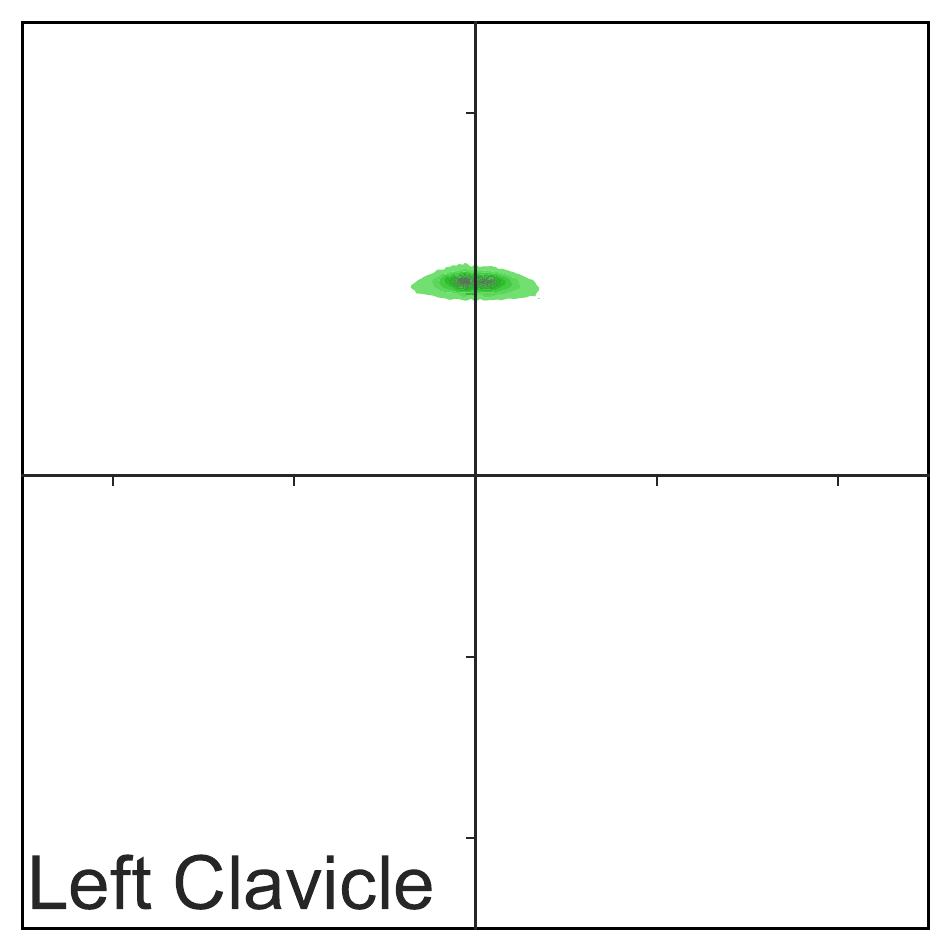}}
    \end{subfigure}
    \hfill
    \begin{subfigure}[t]{0.195\textwidth}
        \raisebox{-\height}{\includegraphics[width=1\textwidth]{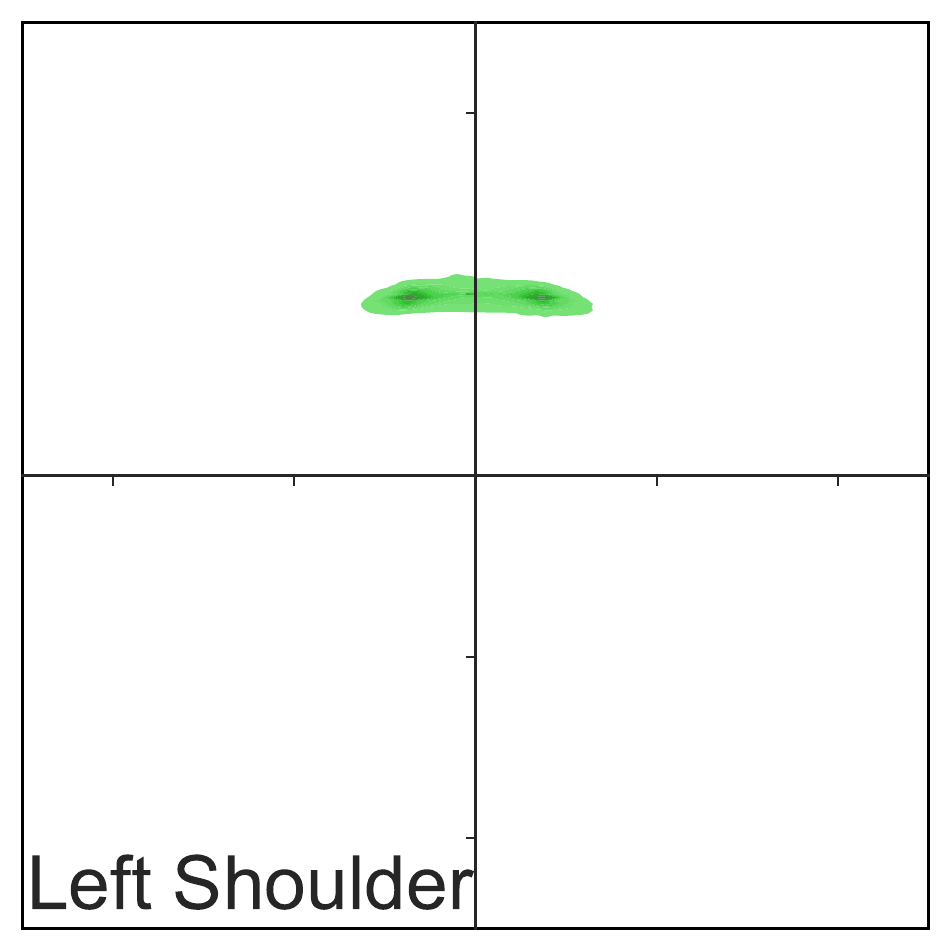}}
    \end{subfigure}
    \hfill
    \begin{subfigure}[t]{0.195\textwidth}
        \raisebox{-\height}{\includegraphics[width=1\textwidth]{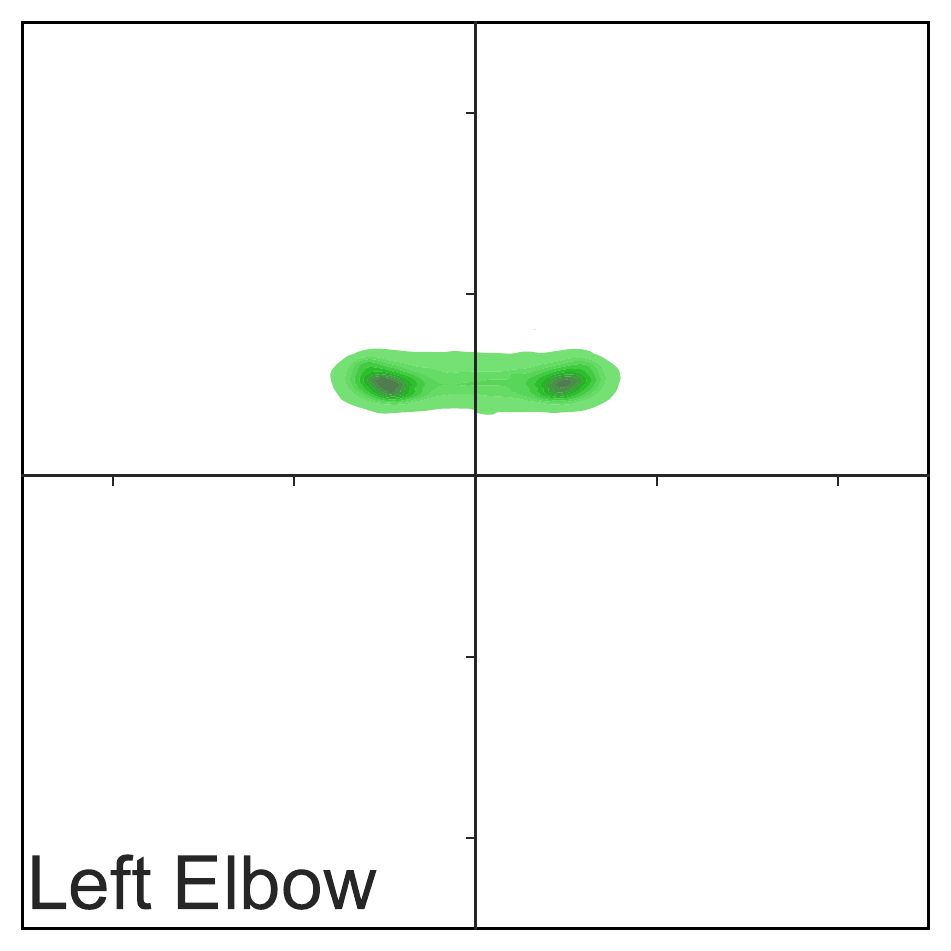}}
    \end{subfigure}
    \begin{subfigure}[t]{0.195\textwidth}
        \raisebox{-\height}{\includegraphics[width=1\textwidth]{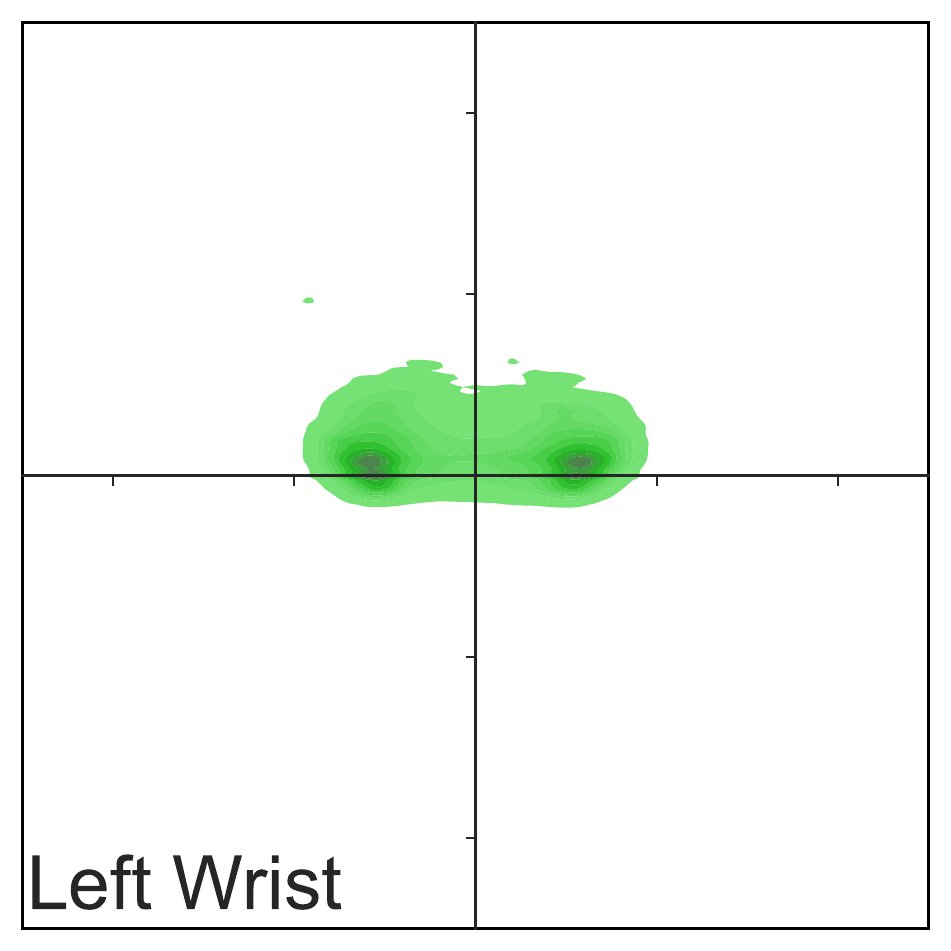}}
    \end{subfigure}
    \hfill
    \begin{subfigure}[t]{0.195\textwidth}
        \raisebox{-\height}{\includegraphics[width=1\textwidth]{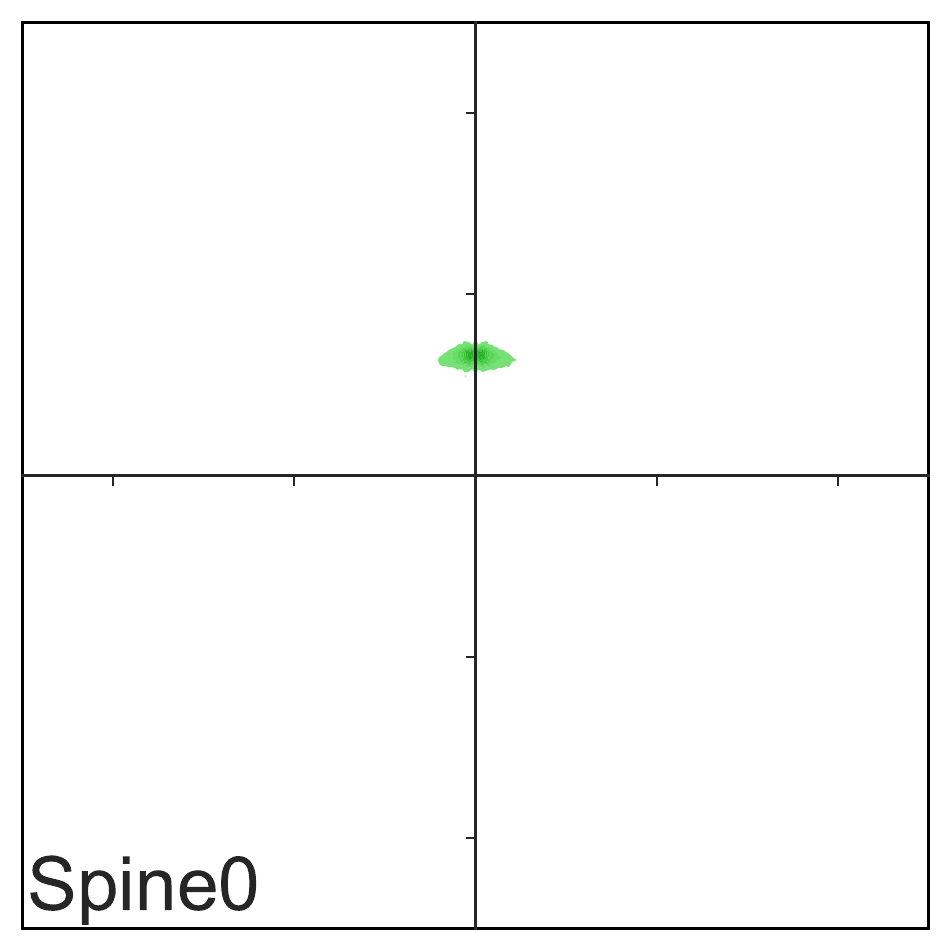}}
    \end{subfigure}
    \hfill
    \begin{subfigure}[t]{0.195\textwidth}
        \raisebox{-\height}{\includegraphics[width=1\textwidth]{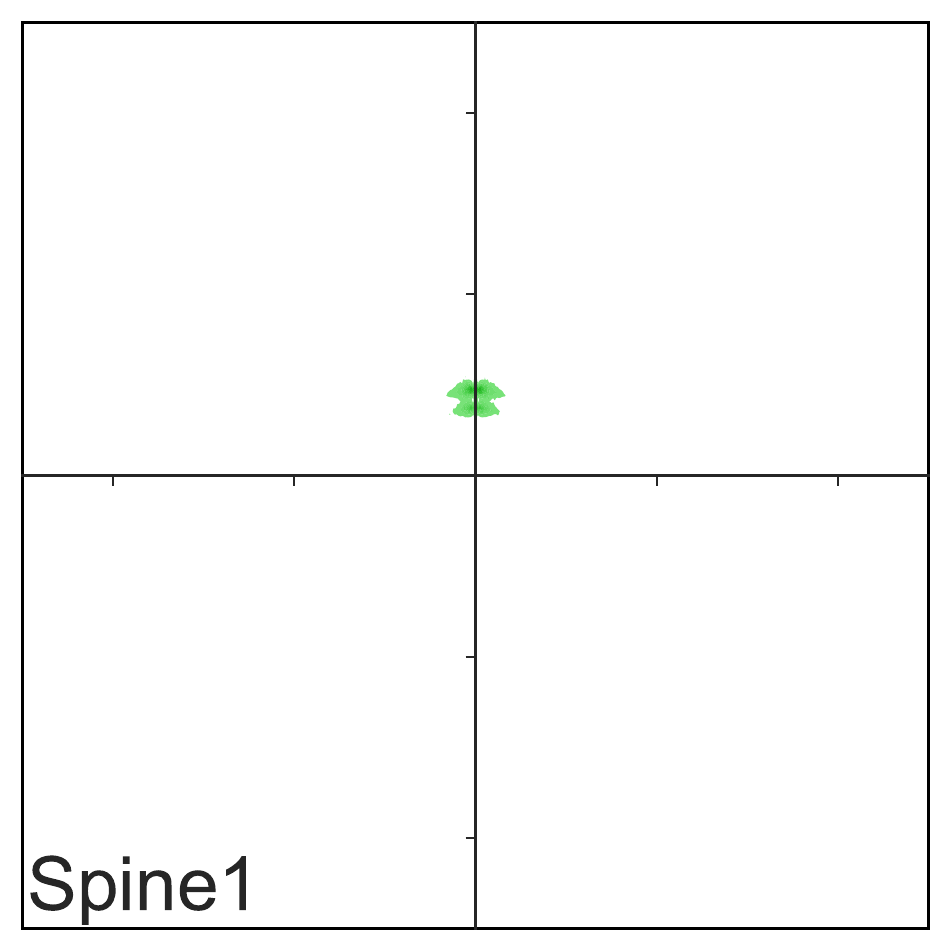}}
    \end{subfigure}
    \hfill
    \begin{subfigure}[t]{0.195\textwidth}
        \raisebox{-\height}{\includegraphics[width=1\textwidth]{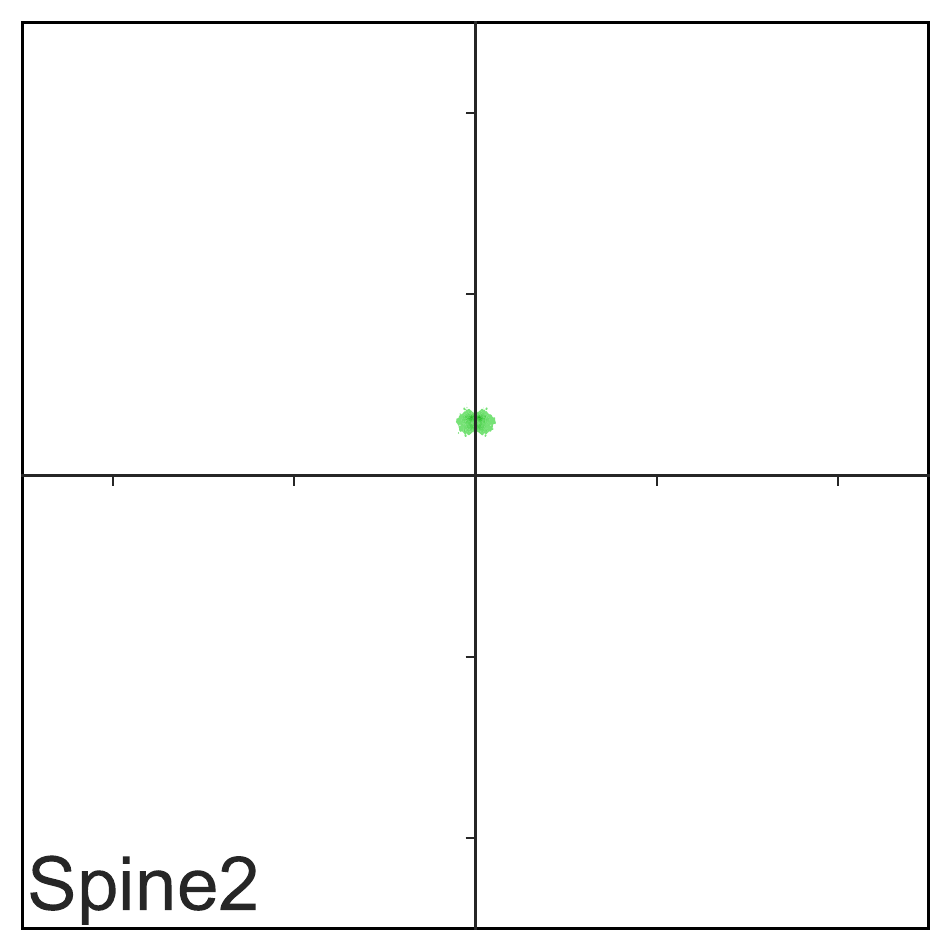}}
    \end{subfigure}
    \hfill
    \begin{subfigure}[t]{0.195\textwidth}
        \raisebox{-\height}{\includegraphics[width=1\textwidth]{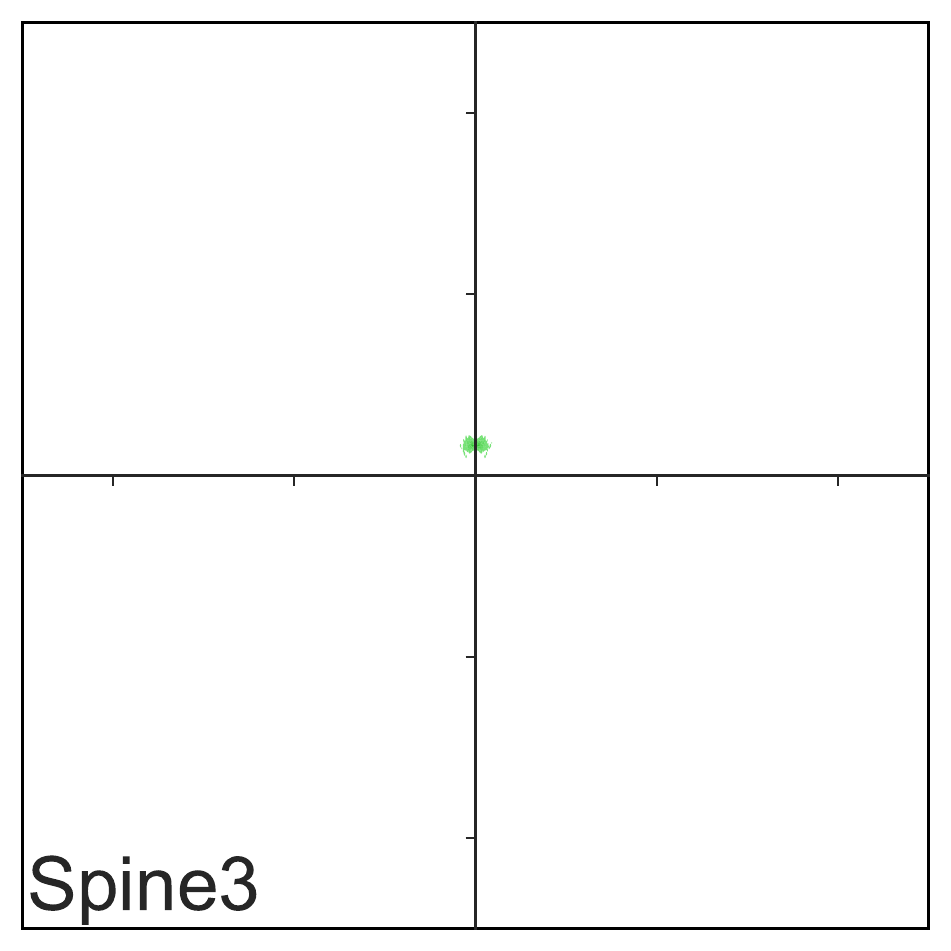}}
    \end{subfigure}
    \begin{subfigure}[t]{0.195\textwidth}
        \raisebox{-\height}{\includegraphics[width=1\textwidth]{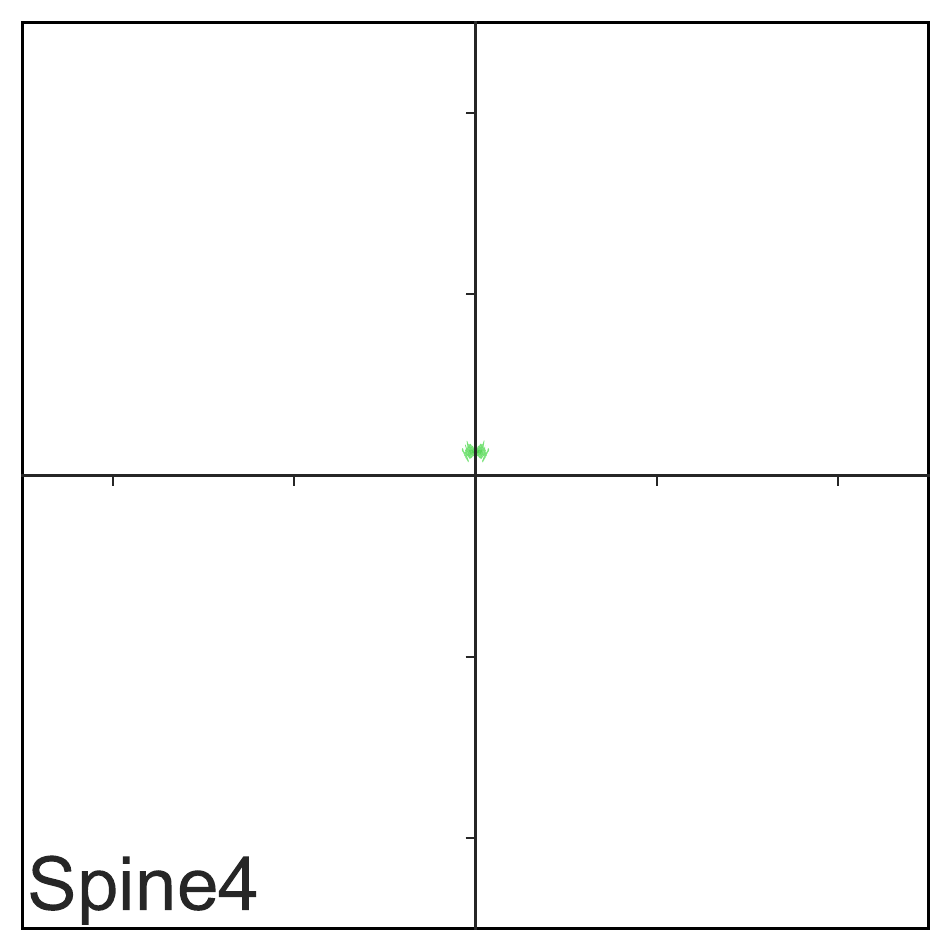}}
    \end{subfigure}
    \hfill
    \begin{subfigure}[t]{0.195\textwidth}
        \raisebox{-\height}{\includegraphics[width=1\textwidth]{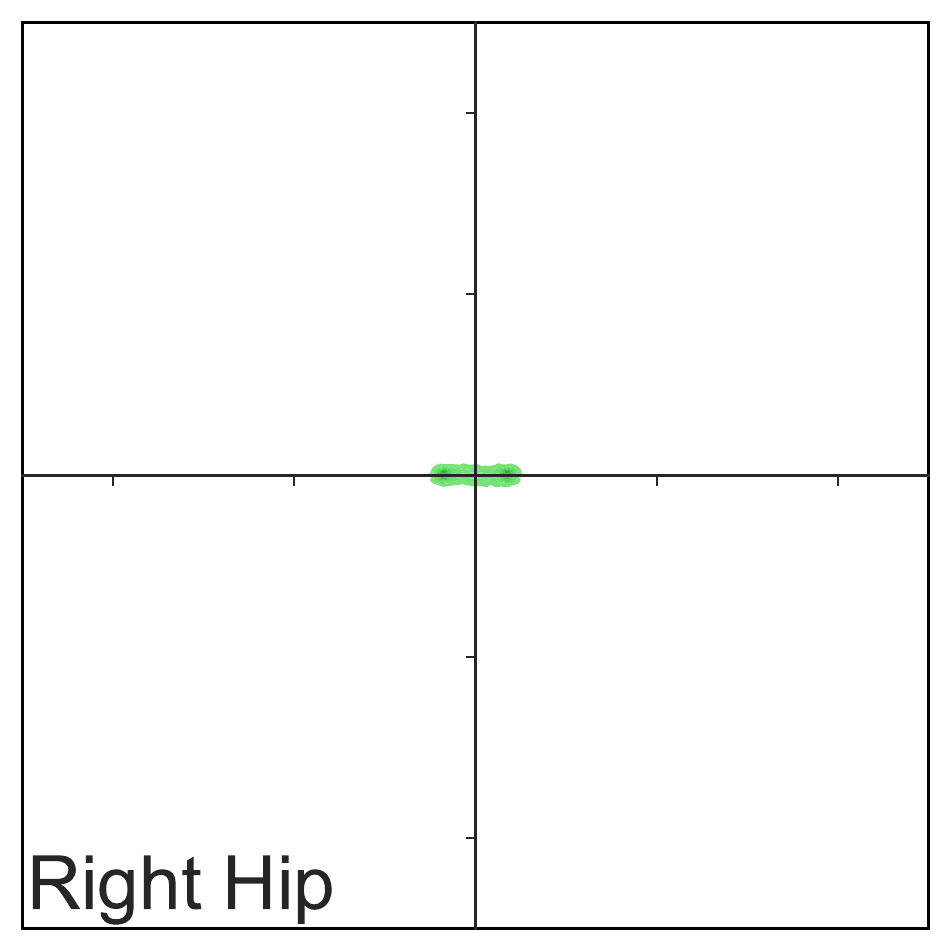}}
    \end{subfigure}
    \hfill
    \begin{subfigure}[t]{0.195\textwidth}
        \raisebox{-\height}{\includegraphics[width=1\textwidth]{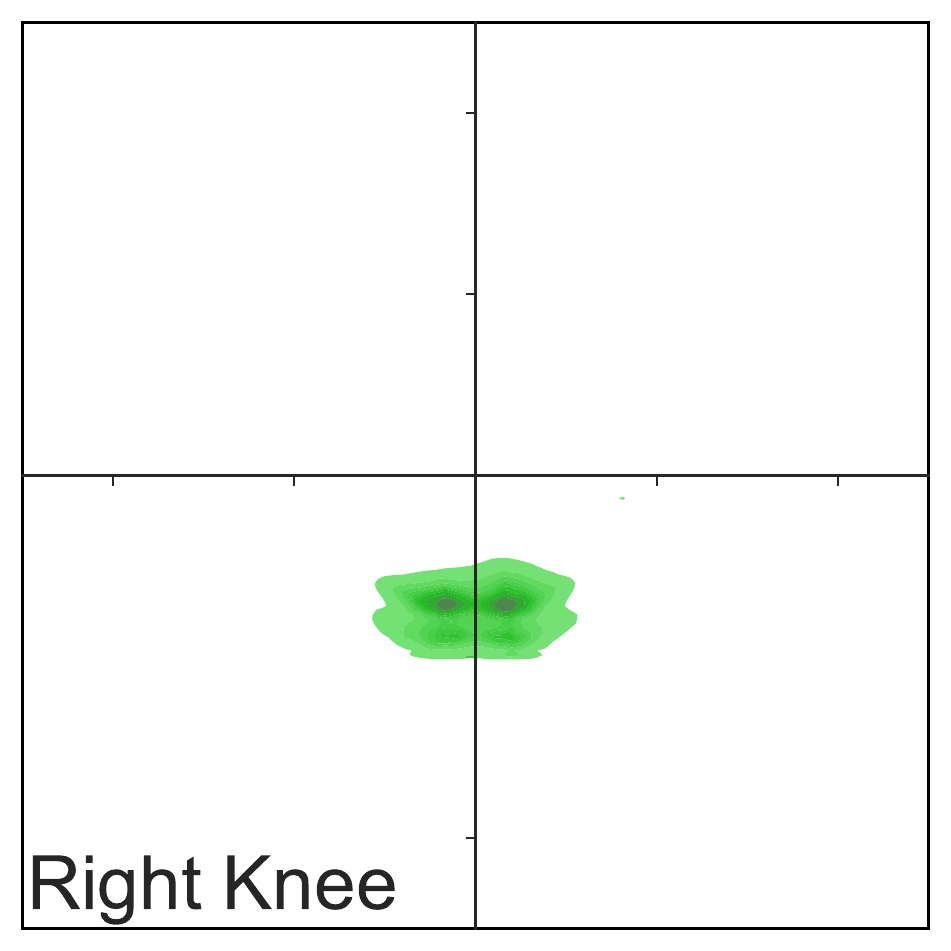}}
    \end{subfigure}
    \hfill
    \begin{subfigure}[t]{0.195\textwidth}
        \raisebox{-\height}{\includegraphics[width=1\textwidth]{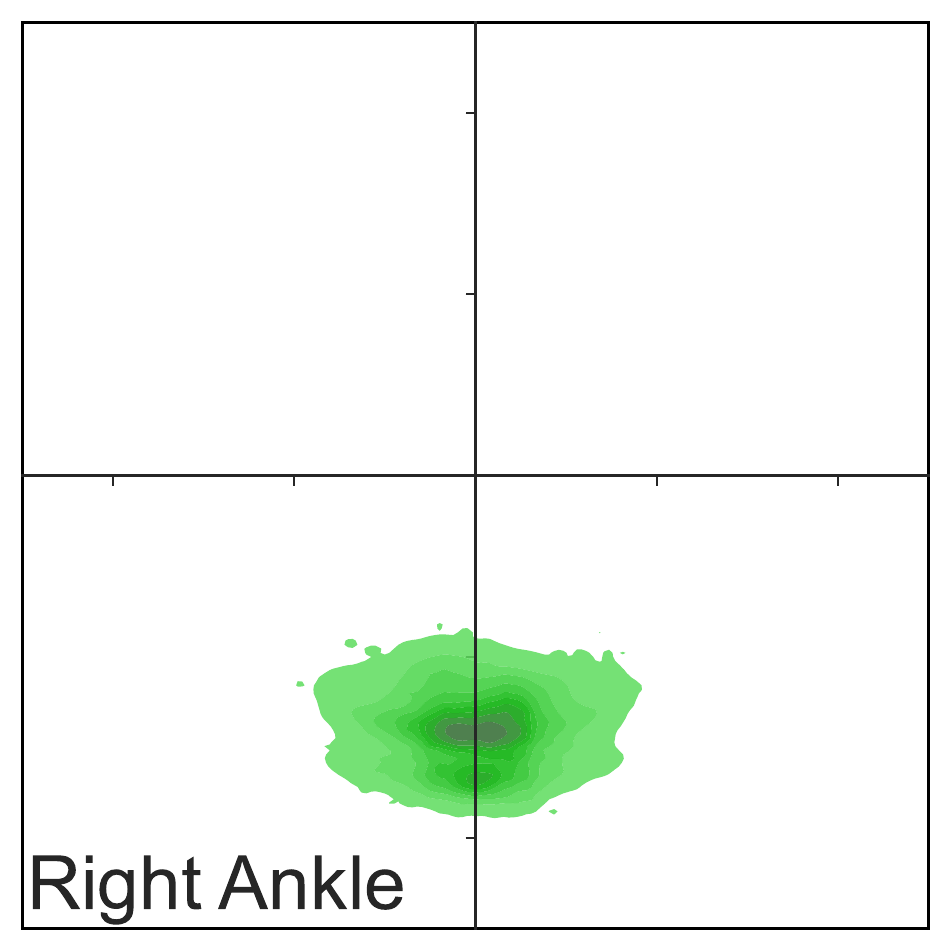}}
    \end{subfigure}
    \hfill
    \begin{subfigure}[t]{0.195\textwidth}
        \raisebox{-\height}{\includegraphics[width=1\textwidth]{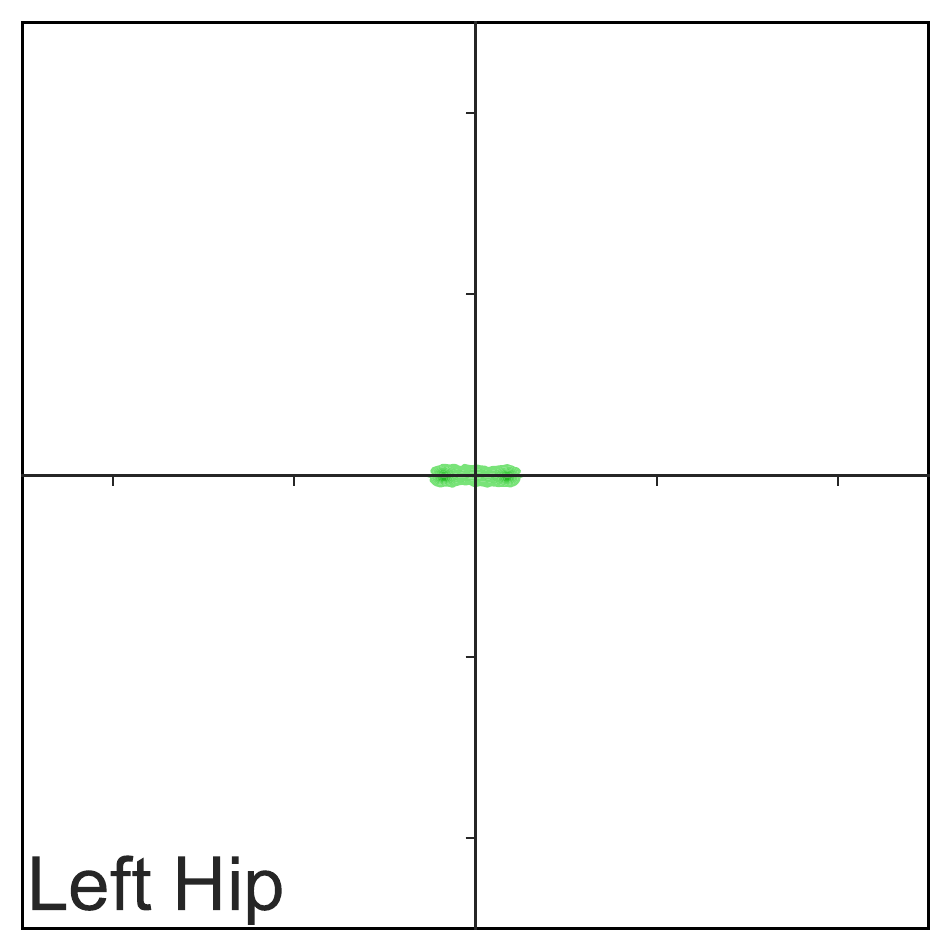}}
    \end{subfigure}


    \begin{subfigure}[t]{0.195\textwidth}
        \raisebox{-\height}{\includegraphics[width=1\textwidth]{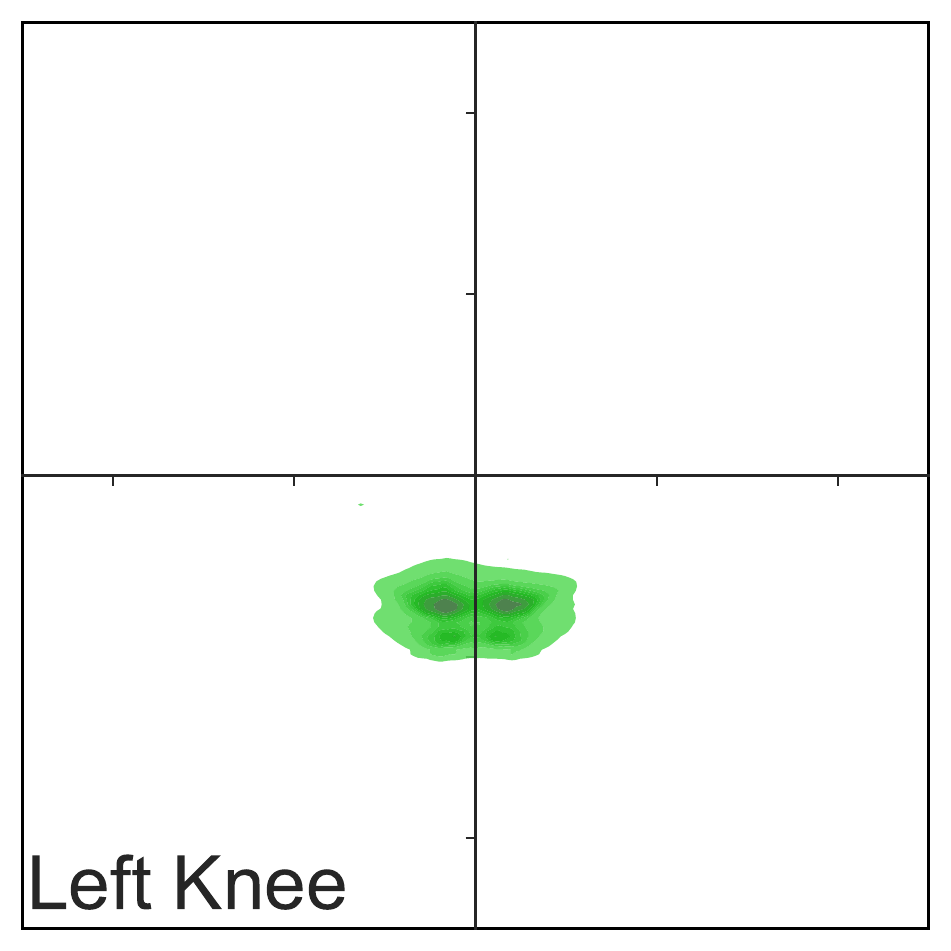}}
    \end{subfigure}
    \begin{subfigure}[t]{0.195\textwidth}
        \raisebox{-\height}{\includegraphics[width=1\textwidth]{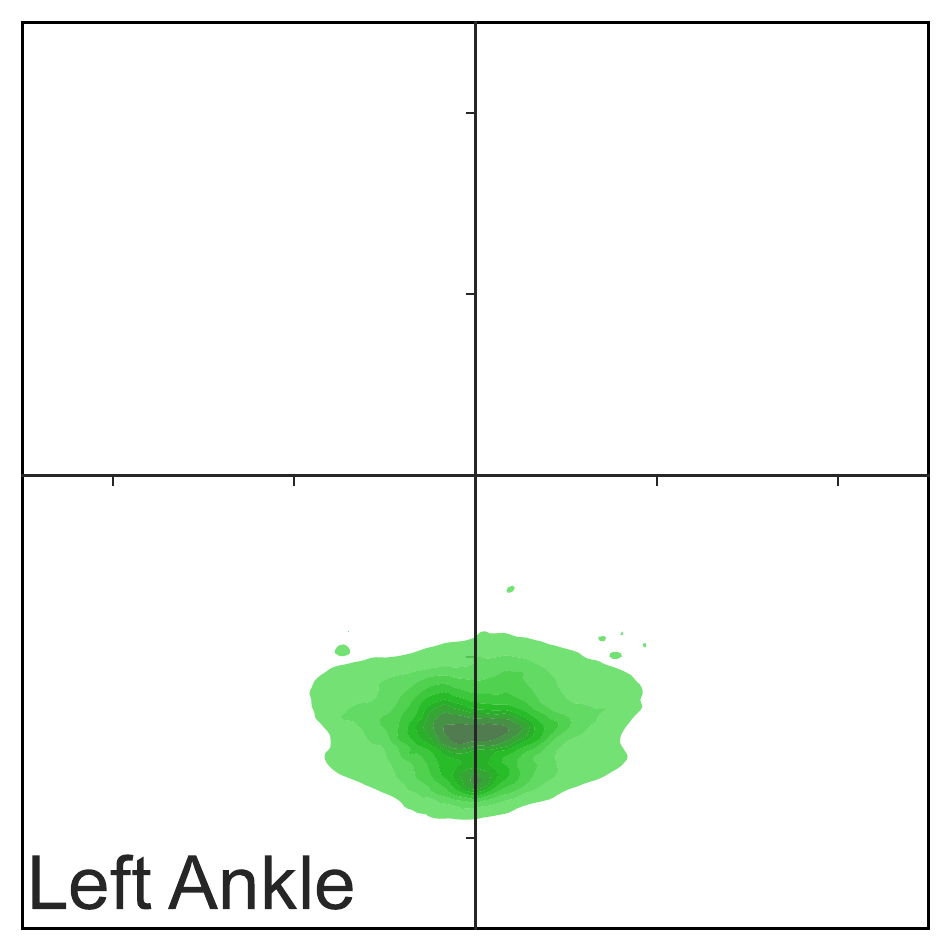}}
    \end{subfigure}
    \begin{subfigure}[t]{0.039\textwidth}
        \raisebox{-\height}{\includegraphics[width=1\textwidth]{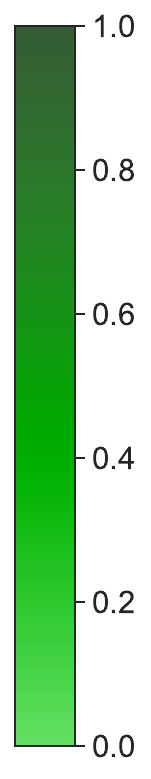}}
    \end{subfigure}
    \caption{\textbf{Keypoint location heatmaps for JTA dataset.} We aligned all keypoint labels according to \textit{mid-hip} joint, and scaled them proportional to distances between \textit{left shoulder}, \textit{left hip}, \textit{right shoulder}, and \textit{right hip} to produce normalized keypoint locations. The JTA dataset uses 22 keypoints instead of the standard COCO 17 keypoints. Also no facial keypoints are annotated in this dataset.}
    \label{fig:posestatsjta}
\end{figure}

The translated and scaled keypoints are then used to create the heatmap plots of keypoints as in Fig.~\ref{fig:posestatselect},~\ref{fig:posestatsall}, and~\ref{fig:posestatsjta}. The heatmaps are created using entire datasets and normalized according to the dataset size for better comparison. In Fig.~\ref{fig:posestatselect} we showed five representative keypoint location heatmaps. These keypoints are the nose, which encapsulates head positioning and orientation -- and the extremity points of wrists and ankles exhibit the largest translations among all keypoints. From these heatmaps, we can conclude the following: 1) the distribution of synthetic dataset poses encompasses the distribution of poses in COCO; 2) the distribution of our poses is larger than that of COCO's as the heatmaps have a larger \textit{``footprint''}; 3) In COCO most people are captured from the frontal view; hence the \textit{``handedness''} of the density of points for right and left body parts and the asymmetrical patterns arising from this. In contrast, the synthetic data has no \textit{``handedness''} bias and is more or less symmetrical for both body sides. 
We believe that our pose diversity helps train more performant models.

\begin{figure}[htb] 
    \centering
    \hspace{1.2cm}
    \hfill
    \begin{subfigure}[t]{0.23\textwidth}
        {\includegraphics[width=\textwidth, trim={0.2cm 0.3cm 0.1cm 0.1cm}, clip]{neurips_data_2021/plots/fig10_bbox_hm/coco.pdf}}
        \caption{COCO}
    \end{subfigure}
    \hfill
    \begin{subfigure}[t]{0.23\textwidth}
        \includegraphics[width=\textwidth, trim={0.2cm 0.3cm 0.1cm 0.1cm}, clip]{neurips_data_2021/plots/fig10_bbox_hm/synth.pdf}
    \caption{Synthetic} 
    \end{subfigure}
    \hfill
    \begin{subfigure}[t]{0.23\textwidth}
        \includegraphics[width=\textwidth, trim={0.2cm 0.3cm 0.1cm 0.1cm}, clip]{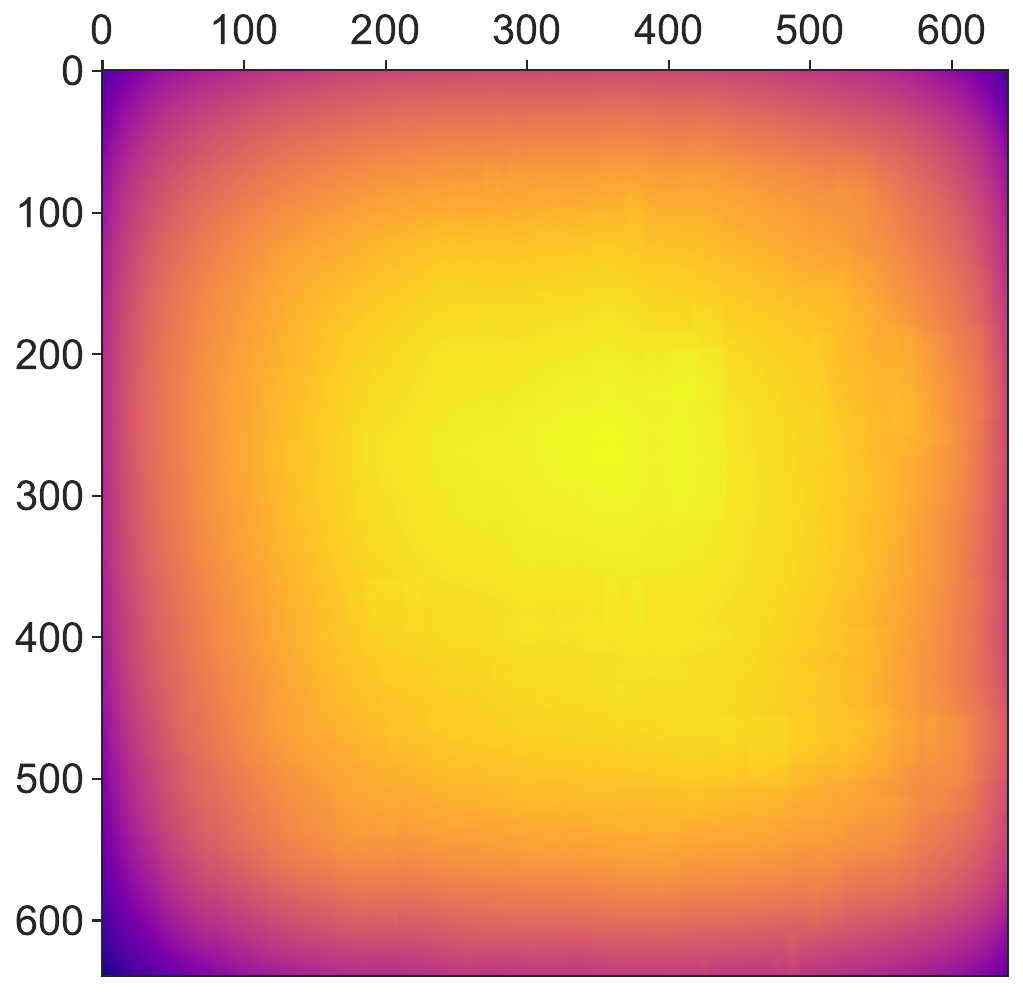}
    \caption{JTA \citep{fabbri2018jta}} 
    \end{subfigure}
    \hfill
    \begin{subfigure}[t]{0.18\textwidth}
        \includegraphics[width=0.2\textwidth, trim={0.2cm 0.2cm 0.2cm 0.2cm}, clip]{neurips_data_2021/plots/fig10_bbox_hm/colorbar.pdf}
    \end{subfigure}
    \hfill
    \caption{\textbf{Bounding Box Occupancy Heatmap.} For our benchmark experiments, we use an image size of $640\times640$. We overlay all the bounding boxes, using filled boxes, on the image to compute the bounding box occupancy map for all datasets.}
    \label{fig:bboxheatmapJTA}
\end{figure}

\begin{figure}[htb] 
    \centering
    \begin{subfigure}[t]{0.7\textwidth}
        \includegraphics[width=\textwidth, trim={0.4cm 0.4cm 0.4cm 0.4cm}, clip]{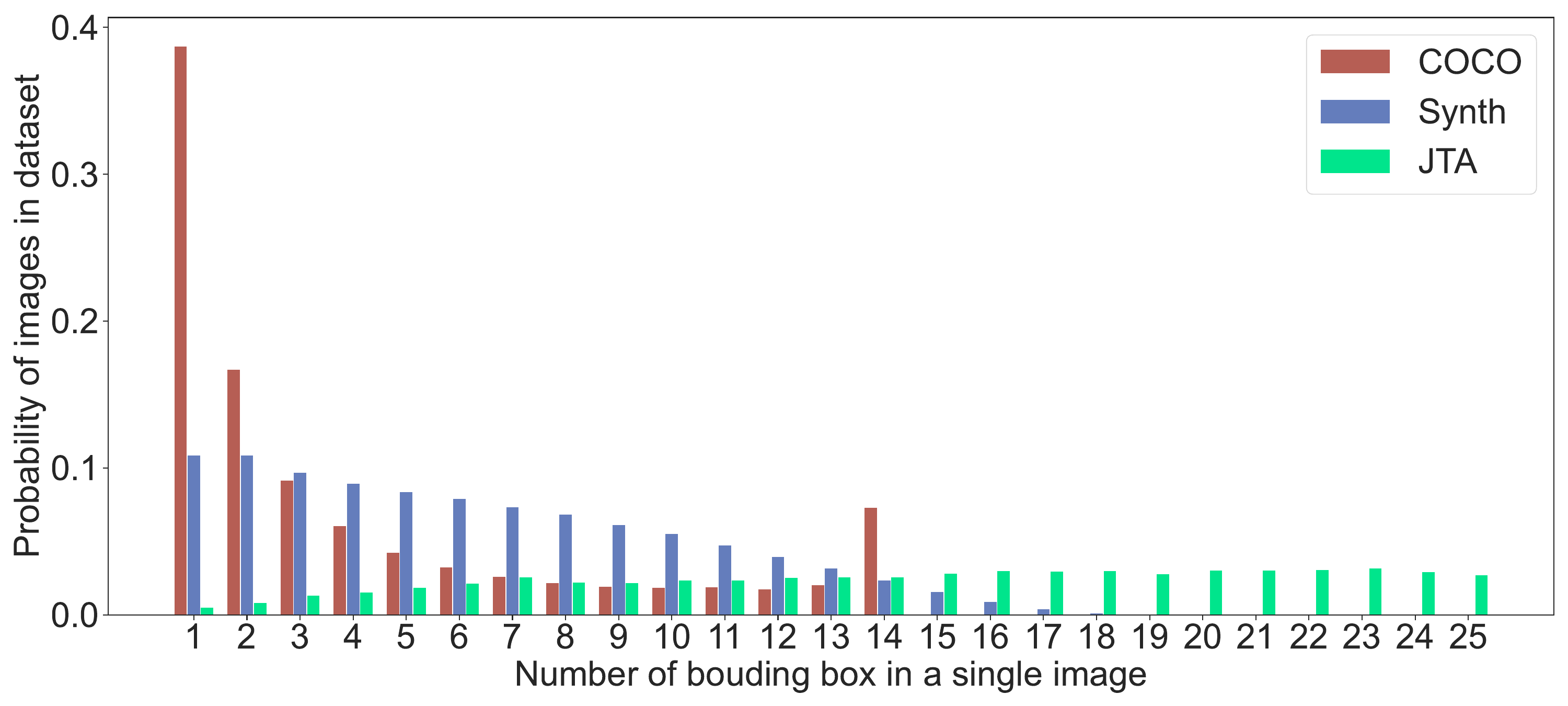}
        \caption{}
        \label{subfig:jta-bbox-on-single-img}
    \end{subfigure}
    \hfill
    \\
    \vspace{0.1cm}
    \begin{subfigure}[t]{0.7\textwidth}
        \includegraphics[width=1\textwidth, trim={0.4cm 0.4cm 0.4cm 0.4cm}, clip]{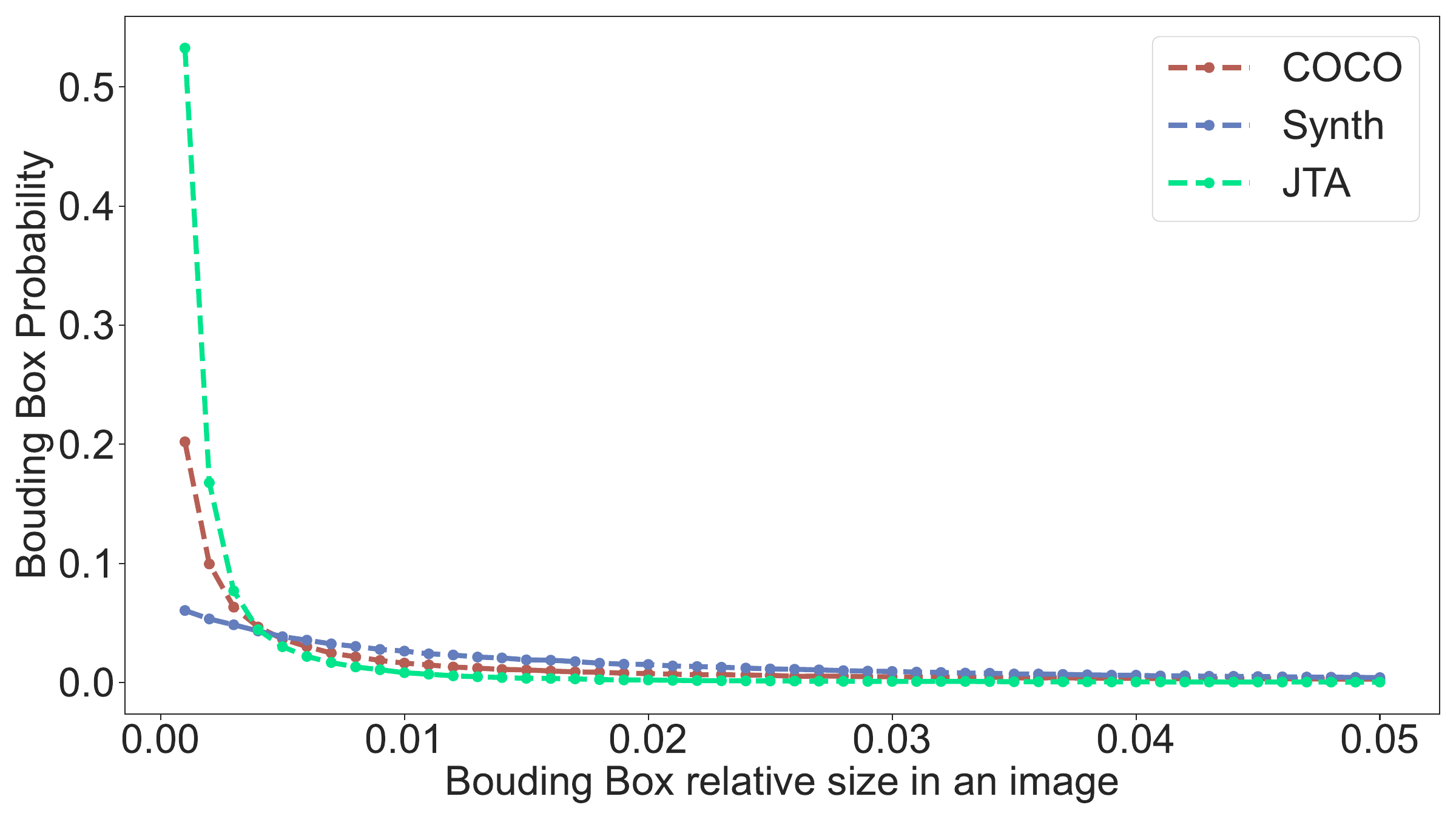}
        \caption{}
        \label{subfig:jta-bbox-size-dist}
    \end{subfigure}
    \hfill
    \\
    \vspace{0.1cm}
    \begin{subfigure}[t]{0.7\textwidth}
        \includegraphics[width=1\textwidth, trim={0.4cm 0.4cm 0.4cm 0.4cm}, clip]{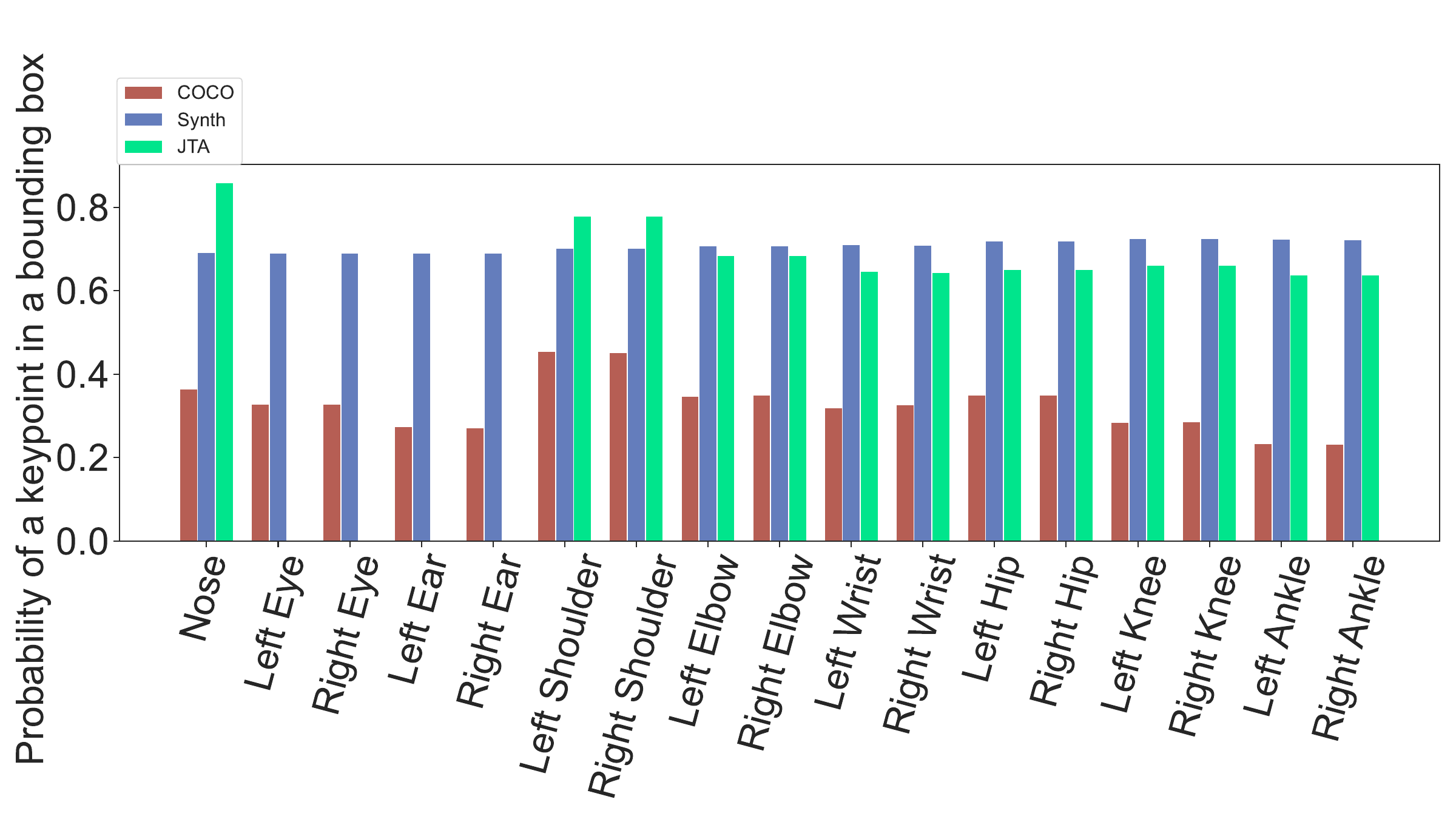}
        \caption{}
        \label{subfig:jta-vis-per-keypoints}
    \end{subfigure}

    
    \caption{\textbf{Bounding Box and Keypoint Statistics.} All COCO statistics computed for COCO-person only, all Synth data generated with \psp{} using default parameters (Tab.~\ref{tab:randomizers}) and using JTA train data \subref{subfig:jta-bbox-on-single-img}) \textbf{Number of Bounding Boxes per Image.} Here the x-axis is clipped at 25, although the JTA dataset has as many as 79 bounding boxes per image, and like \psp{} they have not used $iscrowd=1$.  \subref{subfig:jta-bbox-size-dist}) \textbf{Bounding Box Size Relative to Image Size}. Here, $\text{relative size} = \sqrt{\frac{\text{bounding box occupied pixels}}{\text{total image pixels}}}$. 
    \subref{subfig:jta-vis-per-keypoints}) \textbf{Fraction of Keypoints Per Bounding Box.} The likelihood that a keypoint is visible and labeled for a given bounding box.}
   \label{fig:bbox_kpt_compare_jta}
\end{figure}

In Fig.~\ref{fig:bboxheatmapJTA} we show bounding box occupancy heatmap comparisons with the JTA dataset. We observe that JTA has similar box placement to our synthetic dataset, with some patchy impulses on some regions of the image frame. Also, Fig.~\ref{fig:bbox_kpt_compare_jta} shows statistical analysis for COCO, \psp{}, and JTA datasets. We observe from Fig.~\ref{fig:bbox_kpt_compare_jta}a that JTA contains mostly crowded scenes, hence it has many more boxes per image. In Fig.~\ref{fig:bbox_kpt_compare_jta}b we observe that JTA has higher number of small boxes per image, and fewer large boxes per image, whilst our synthetic dataset provides more diverse box sizes across the dataset. Finally, Fig.~\ref{fig:bbox_kpt_compare_jta}c shows that JTA lacks facial COCO keypoints (we assumed that their \textit{Head Center} keypoint corresponds with nose), but for the annotated keypoints they are more or less as likely to appear within each instance box as our synthetic dataset.

\algnewcommand\algorithmicforeach{\textbf{for each}}
\algdef{S}[FOR]{ForEach}[1]{\algorithmicforeach\ #1\ \algorithmicdo}
\newcommand{\stt}[1]{\text{\tiny #1}} 
\begin{algorithm}[htb]
\caption{Keypoint alignment algorithm}
\label{alg:pose_analysis}
\textbf{Input:} Keypoints $K = \{k_i \}$, $i \in \{\text{nose}, \text{left shoulder}, \dots\}$ \\
\textbf{Output:} Translated and scaled keypoints $\hat{K} = \{\hat{k}_i \}$, $i \in \{\text{nose}, \text{left shoulder}, \dots\}$
\begin{algorithmic}
    \Require Keypoints $K$ with both hip and shoulder keypoints annotated ($v = 1 \lor v = 2$)
    \State $m \gets (\frac{k_{\stt{left hip}} + k_{\stt{right hip}})}{2}$
    \State $s \gets (\frac{d(k_{\stt{left hip}},k_{\stt{left shoulder}}) + d(k_{\stt{right hip}},k_{\stt{right shoulder}}))}{2}$ \Comment{where $d(p, q) = \sqrt{(p-q)^2}$}
    \State $\hat{k}_i \gets \frac{(k_i - m)}{s}$
\end{algorithmic}
\end{algorithm}


\subsection{Additional Examples from Generated Data}
\begin{figure}[htb]
    \centering
    \begin{subfigure}[t]{0.270\textwidth}
        \raisebox{-\height}{\includegraphics[width=\textwidth]{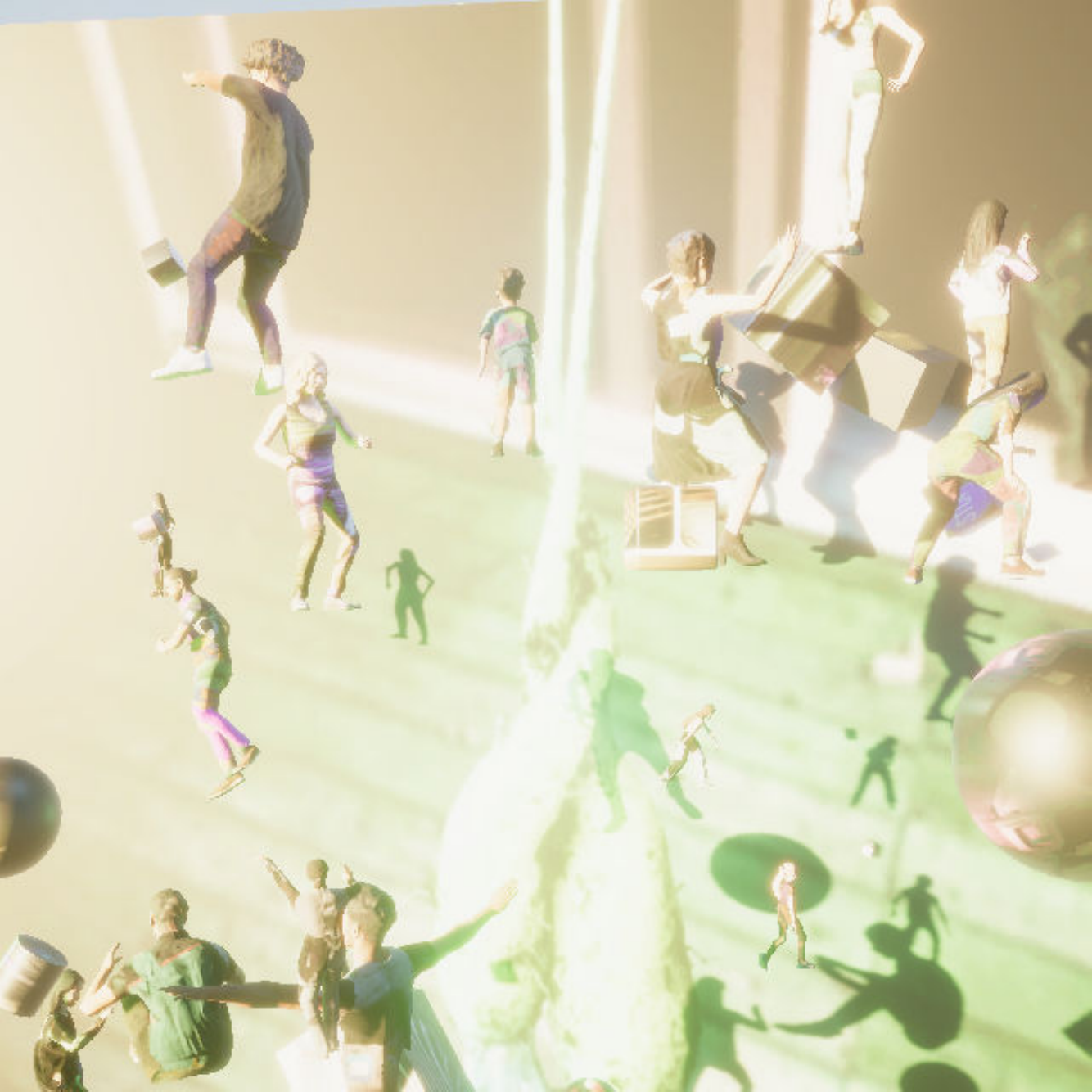}}
    \end{subfigure}
    \begin{subfigure}[t]{0.270\textwidth}
        \raisebox{-\height}{\includegraphics[width=\textwidth]{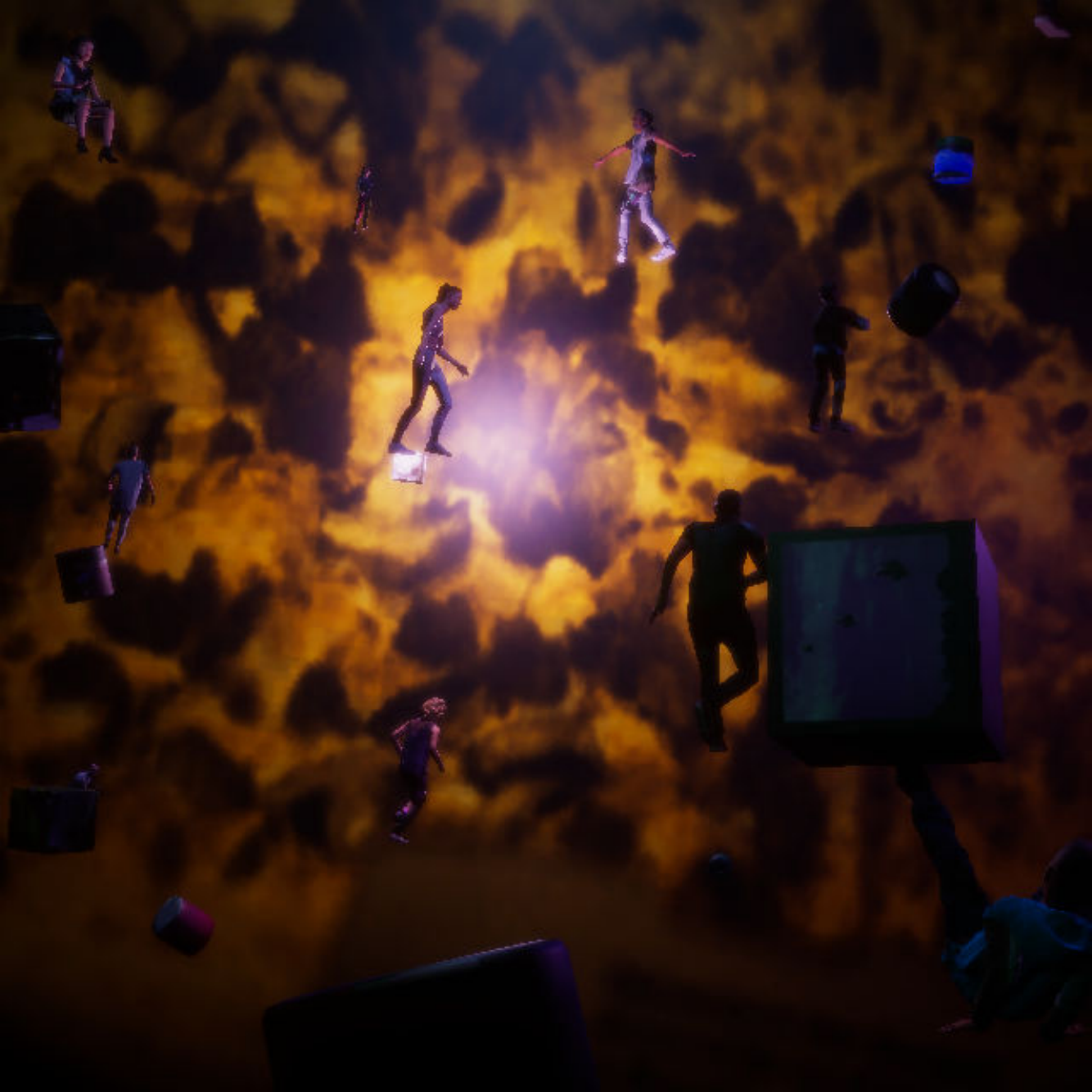}}
    \end{subfigure}
    \begin{subfigure}[t]{0.270\textwidth}
        \raisebox{-\height}{\includegraphics[width=\textwidth]{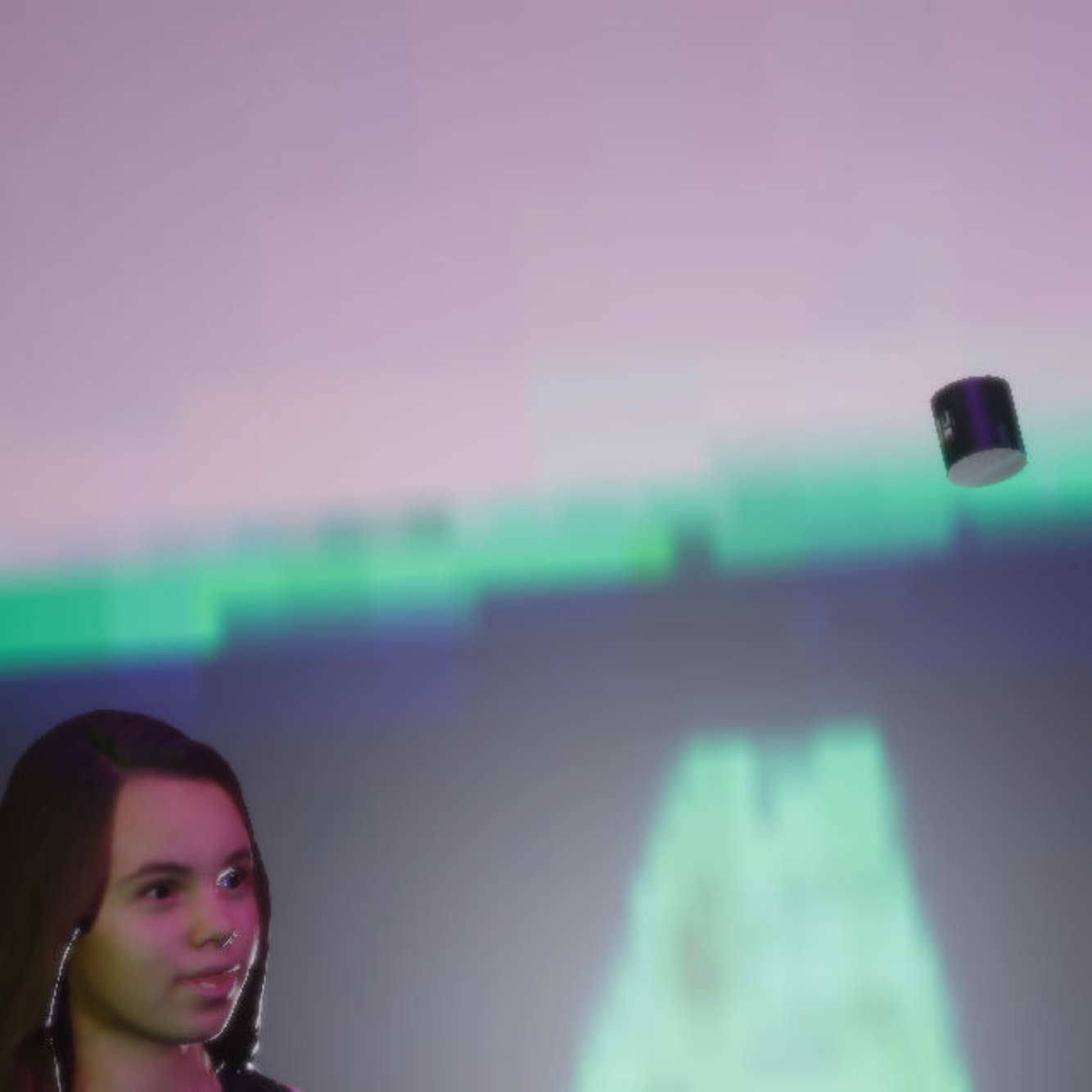}}
    \end{subfigure}
    \hfill
    \\
    \begin{subfigure}[t]{0.270\textwidth}
        \raisebox{-\height}{\includegraphics[width=\textwidth]{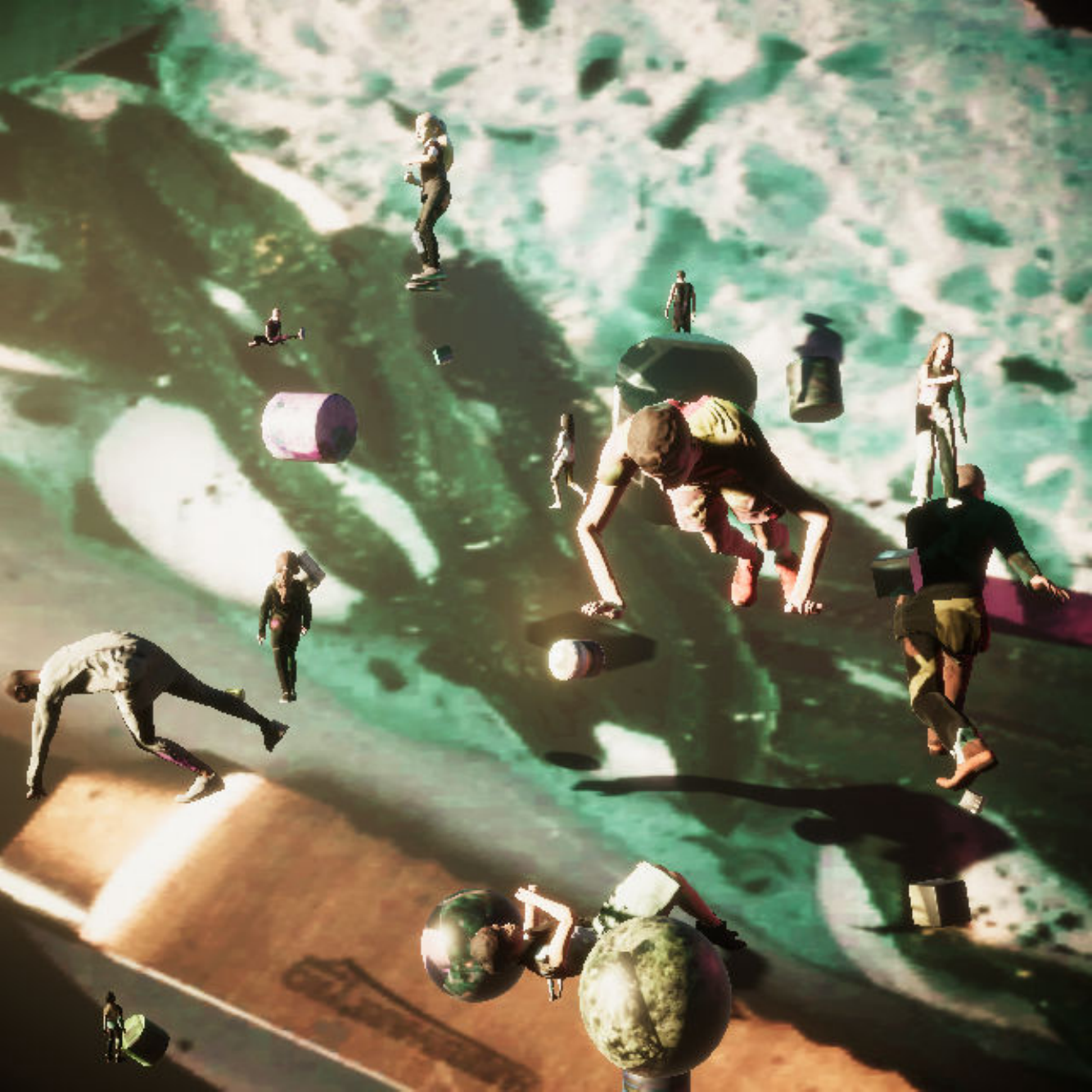}}
    \end{subfigure}
    \begin{subfigure}[t]{0.270\textwidth}
        \raisebox{-\height}{\includegraphics[width=\textwidth]{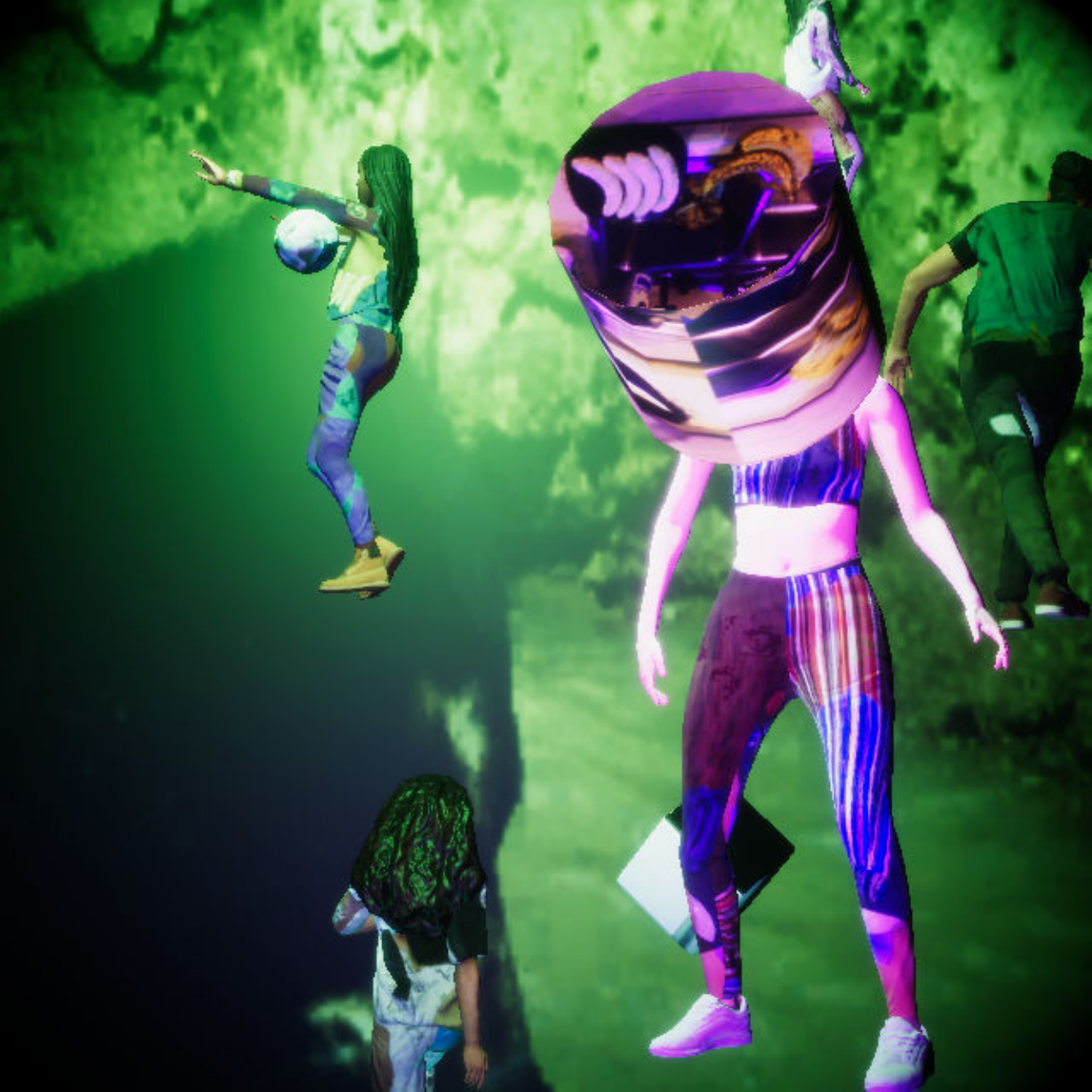}}
    \end{subfigure}
    \begin{subfigure}[t]{0.270\textwidth}
        \raisebox{-\height}{\includegraphics[width=\textwidth]{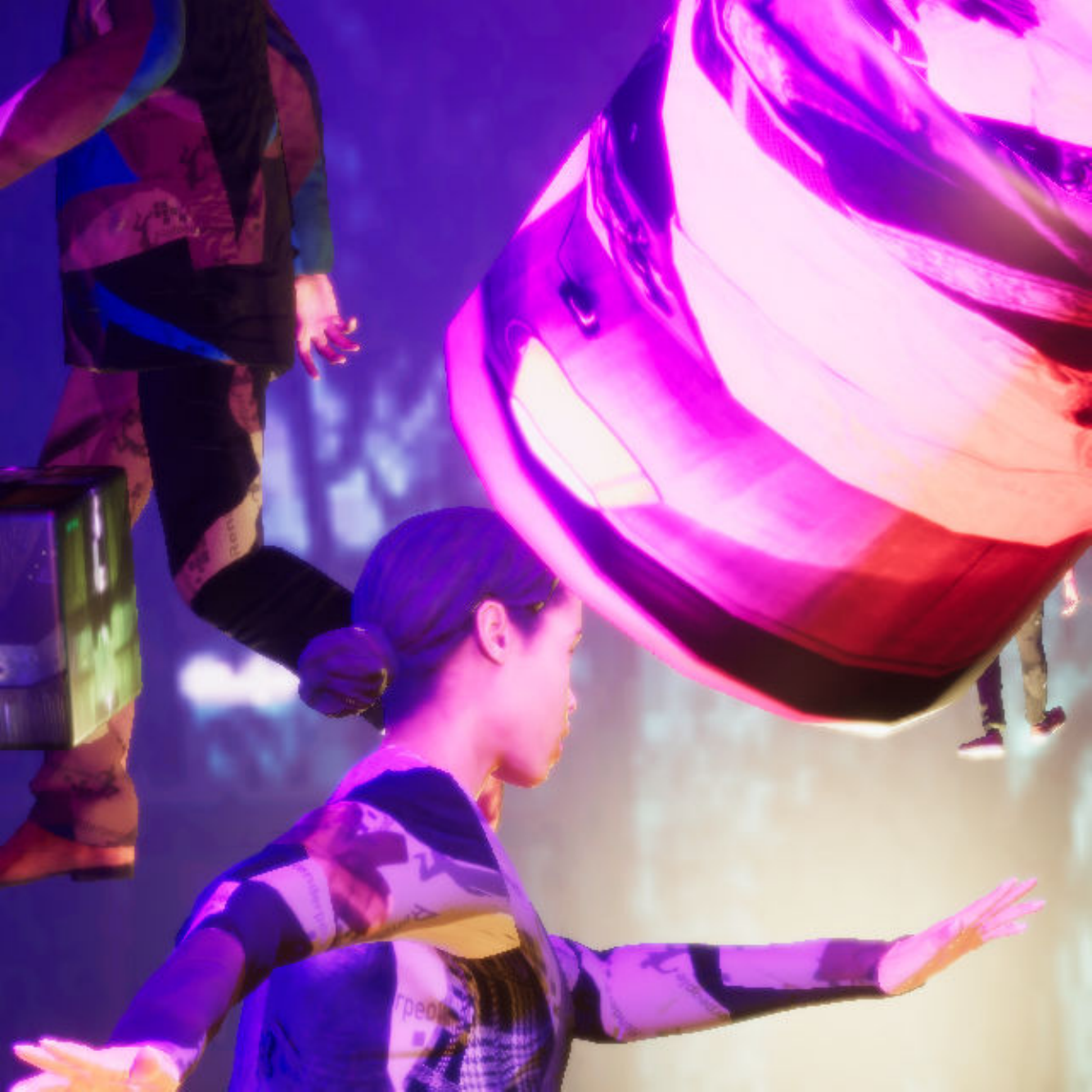}}
    \end{subfigure}
    \hfill
    \\
    \begin{subfigure}[t]{0.270\textwidth}
        \raisebox{-\height}{\includegraphics[width=\textwidth]{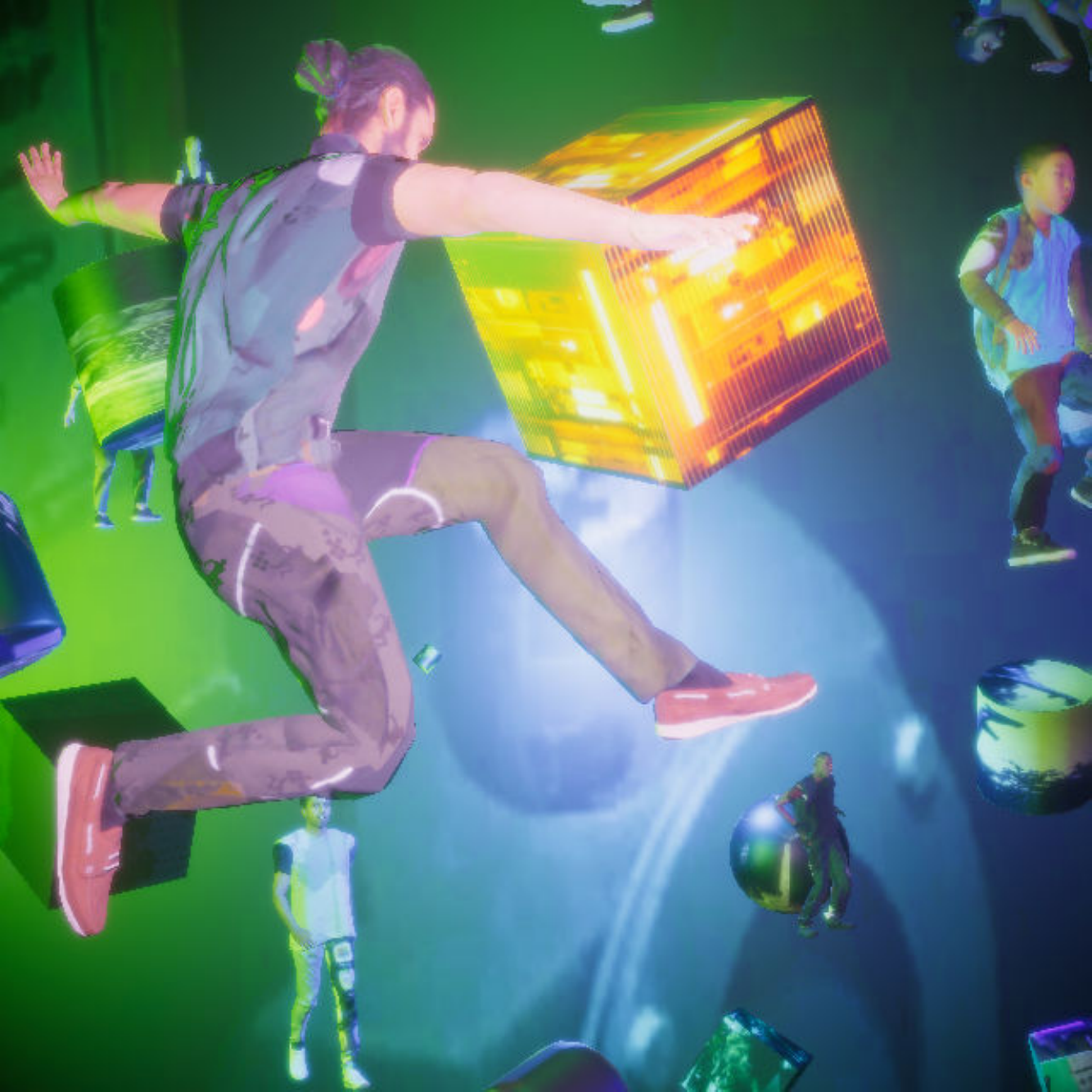}}
    \end{subfigure}
    \begin{subfigure}[t]{0.270\textwidth}
        \raisebox{-\height}{\includegraphics[width=\textwidth]{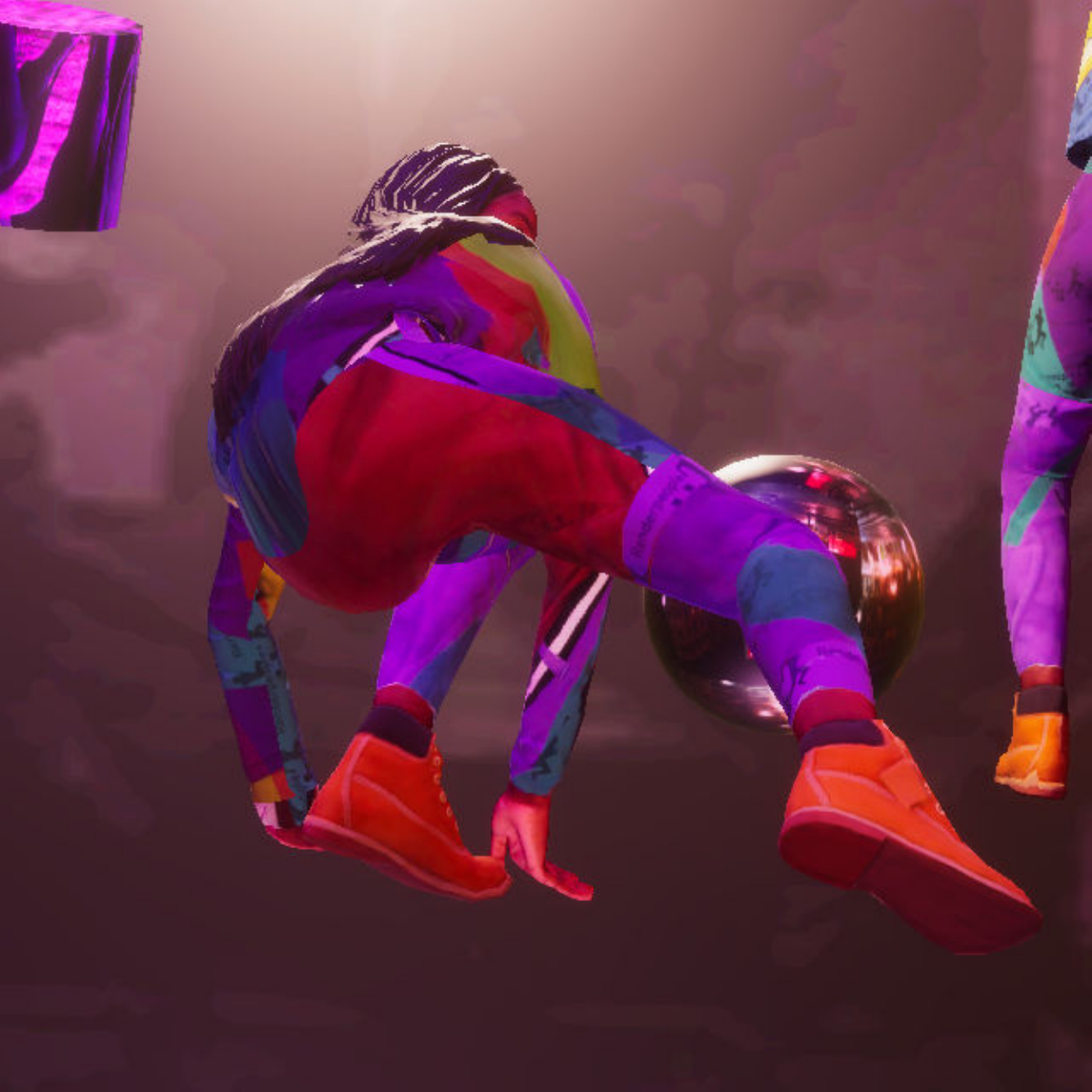}}
    \end{subfigure}
    \begin{subfigure}[t]{0.270\textwidth}
        \raisebox{-\height}{\includegraphics[width=\textwidth]{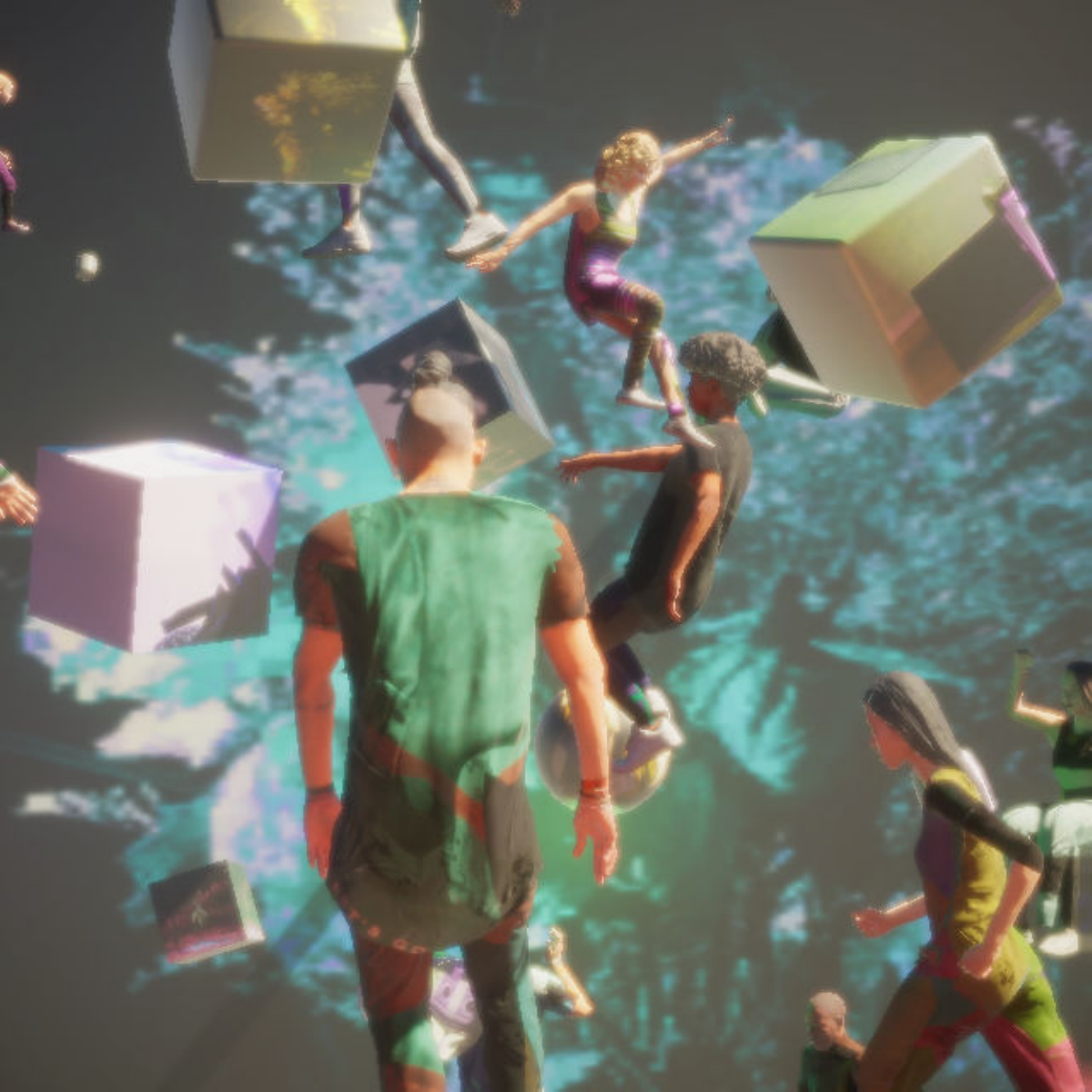}}
    \end{subfigure}
    \hfill
    \\
    \begin{subfigure}[t]{0.270\textwidth}
        \raisebox{-\height}{\includegraphics[width=\textwidth]{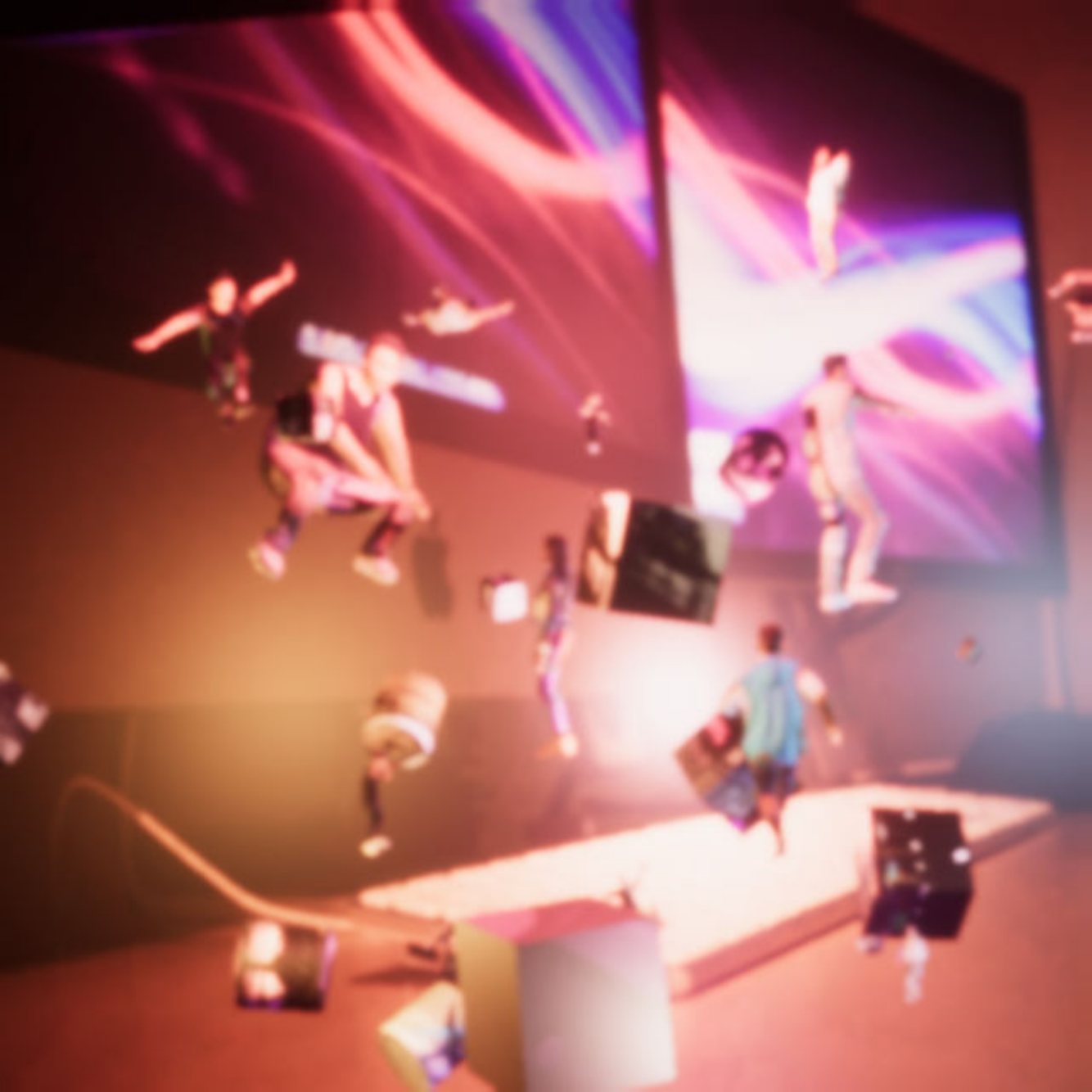}}
    \end{subfigure}
    \begin{subfigure}[t]{0.270\textwidth}
        \raisebox{-\height}{\includegraphics[width=\textwidth]{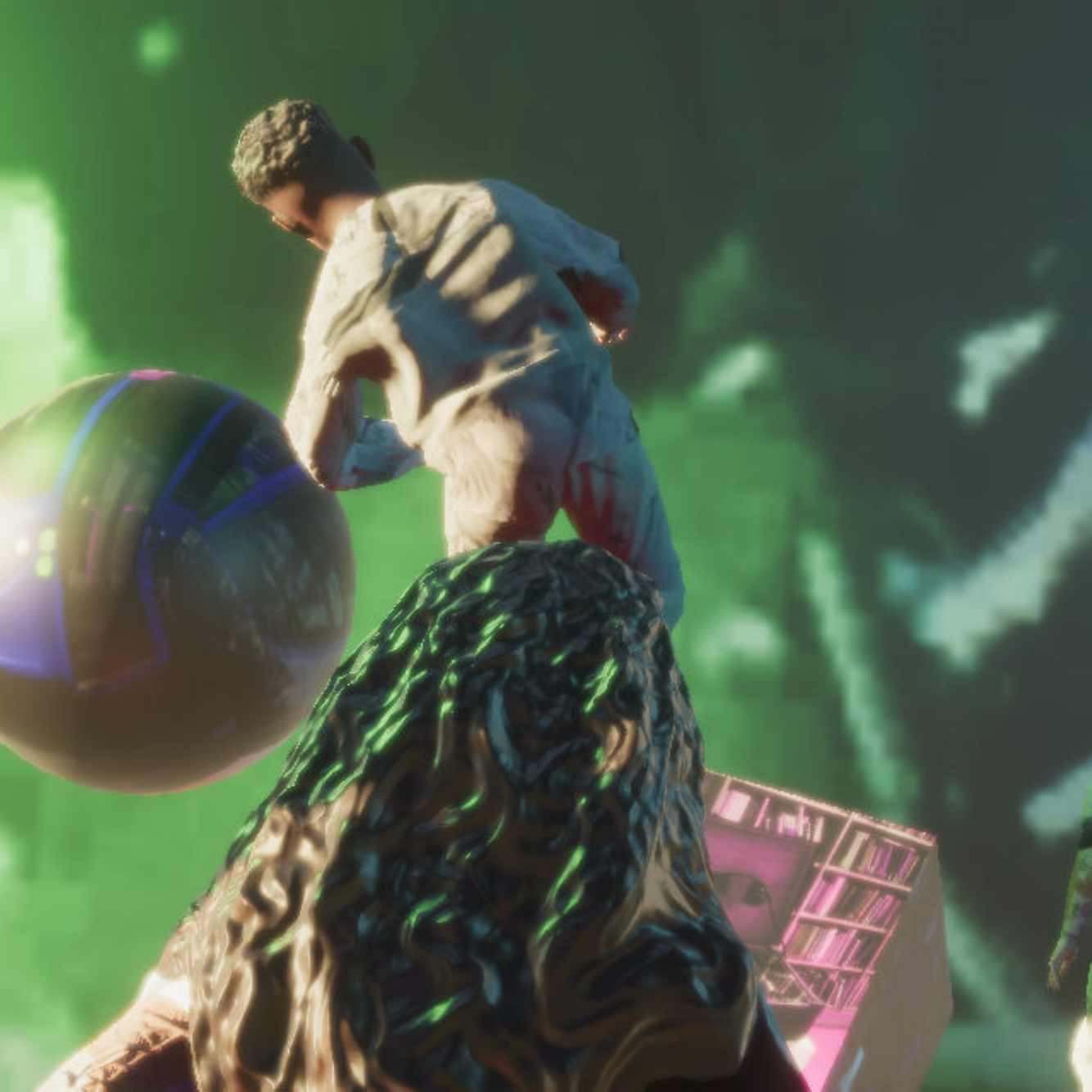}}
    \end{subfigure}
    \begin{subfigure}[t]{0.270\textwidth}
        \raisebox{-\height}{\includegraphics[width=\textwidth]{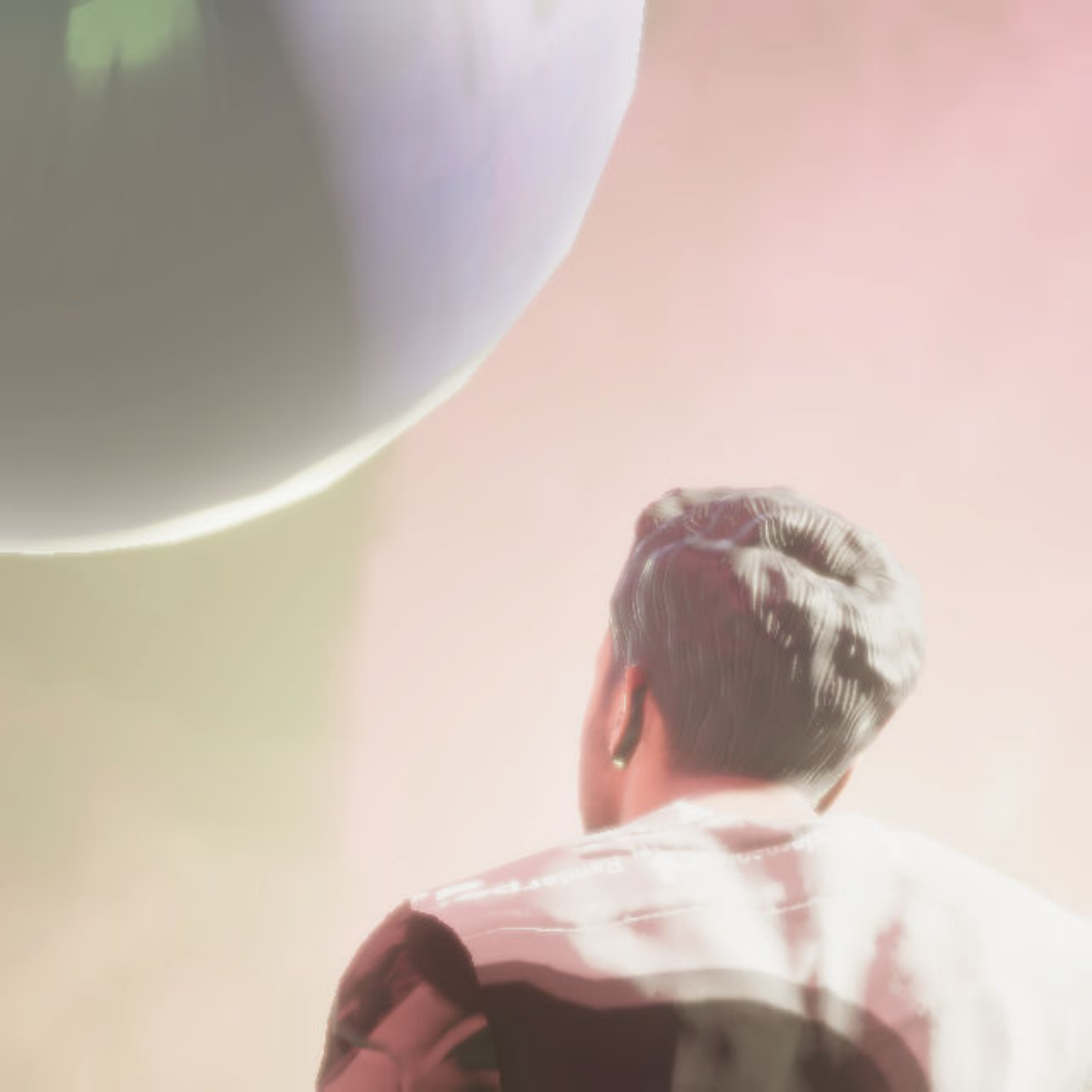}}
    \end{subfigure}
    \hfill
    \\
    \begin{subfigure}[t]{0.270\textwidth}
        \raisebox{-\height}{\includegraphics[width=\textwidth]{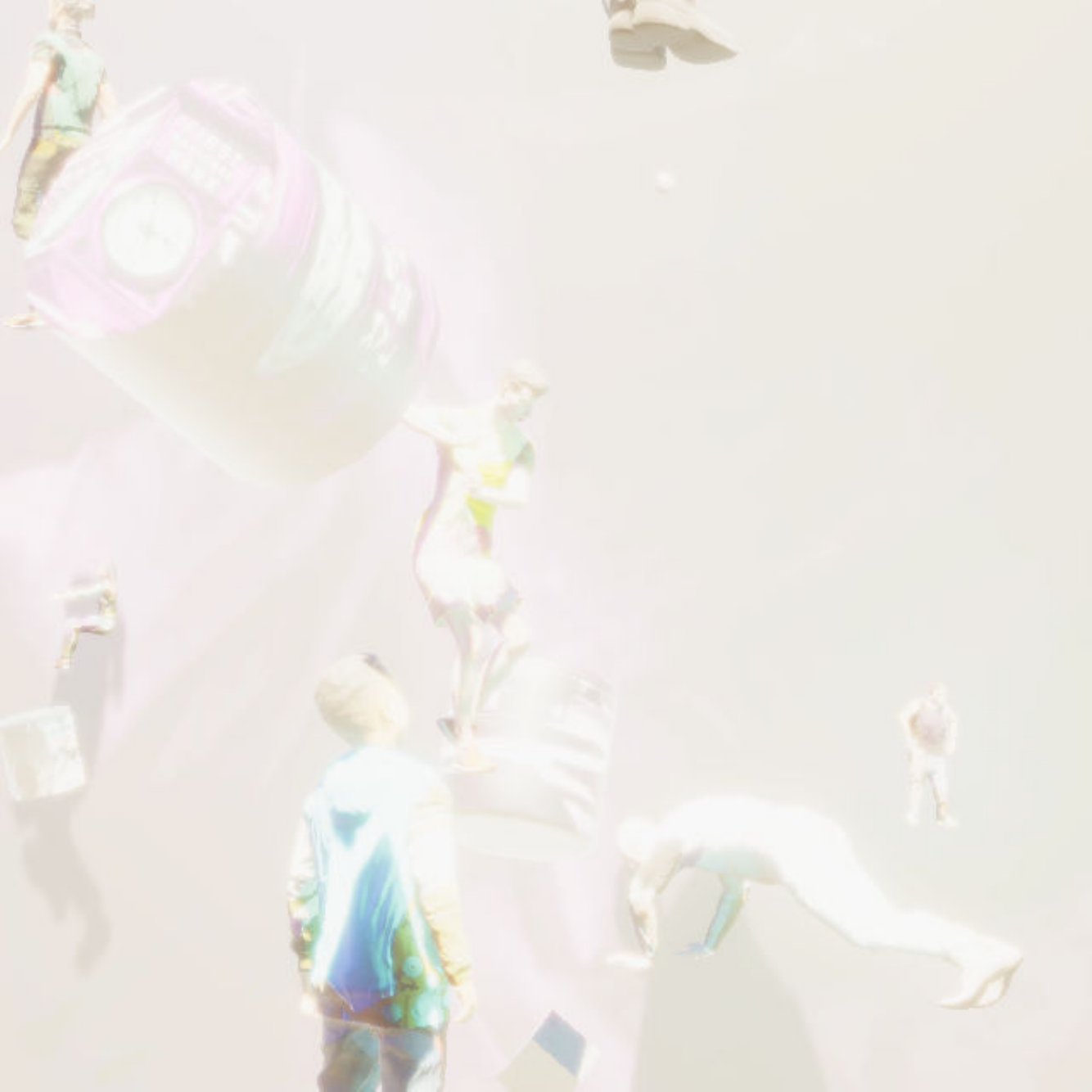}}
    \end{subfigure}
    \begin{subfigure}[t]{0.270\textwidth}
        \raisebox{-\height}{\includegraphics[width=\textwidth]{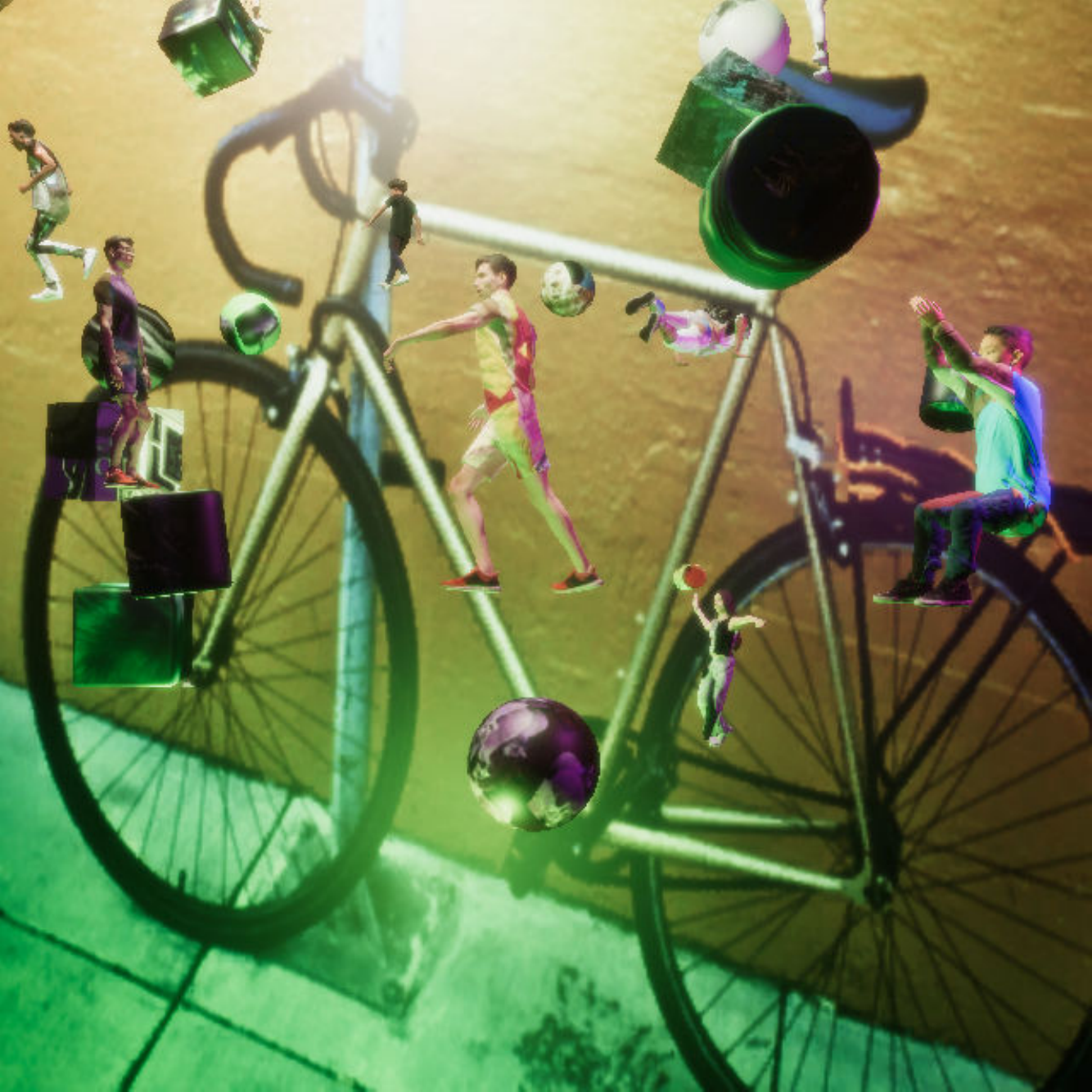}}
    \end{subfigure}
    \begin{subfigure}[t]{0.270\textwidth}
        \raisebox{-\height}{\includegraphics[width=\textwidth]{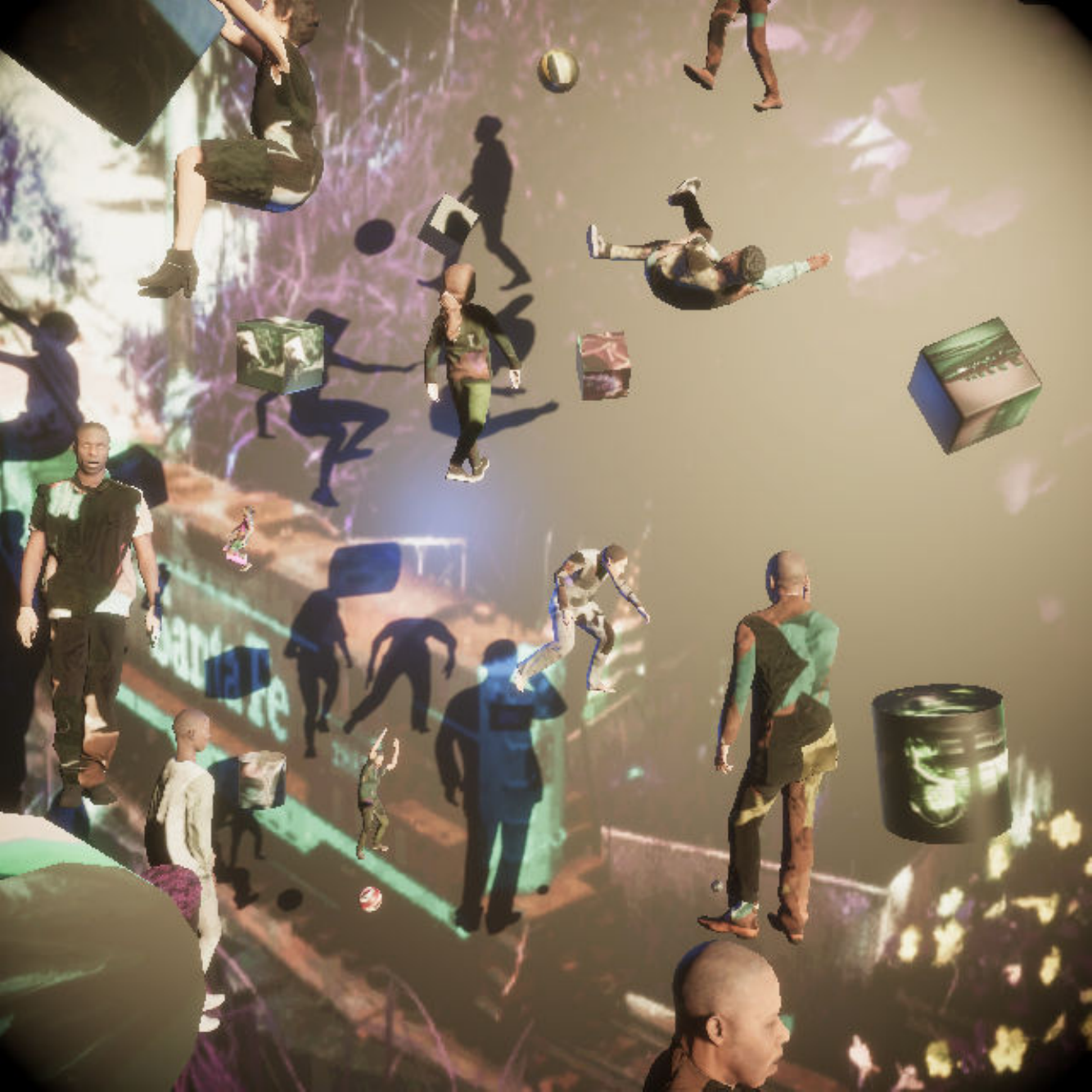}}
    \end{subfigure}
    \hfill
    \\
    \begin{subfigure}[t]{0.270\textwidth}
        \raisebox{-\height}{\includegraphics[width=\textwidth]{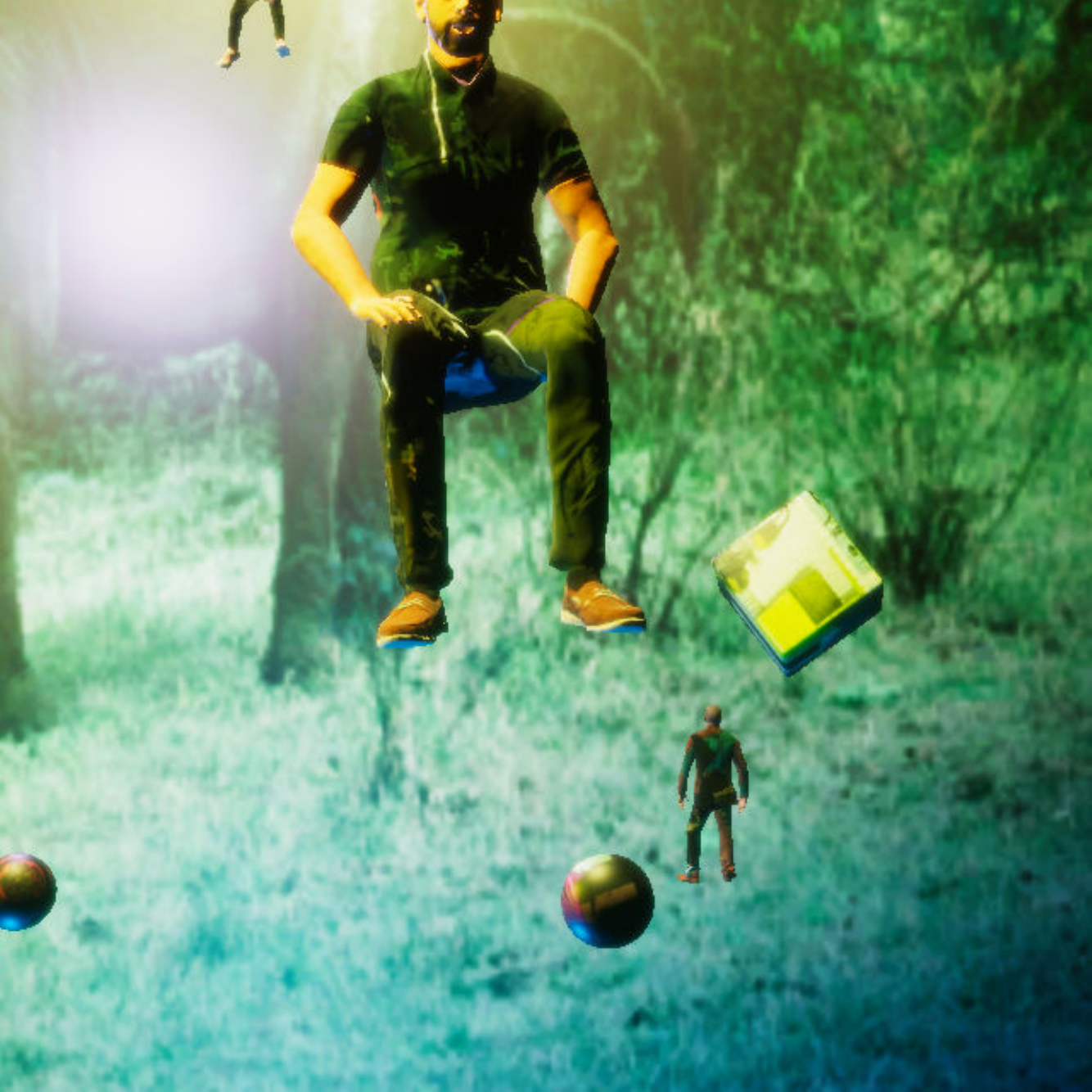}}
    \end{subfigure}
    \begin{subfigure}[t]{0.270\textwidth}
        \raisebox{-\height}{\includegraphics[width=\textwidth]{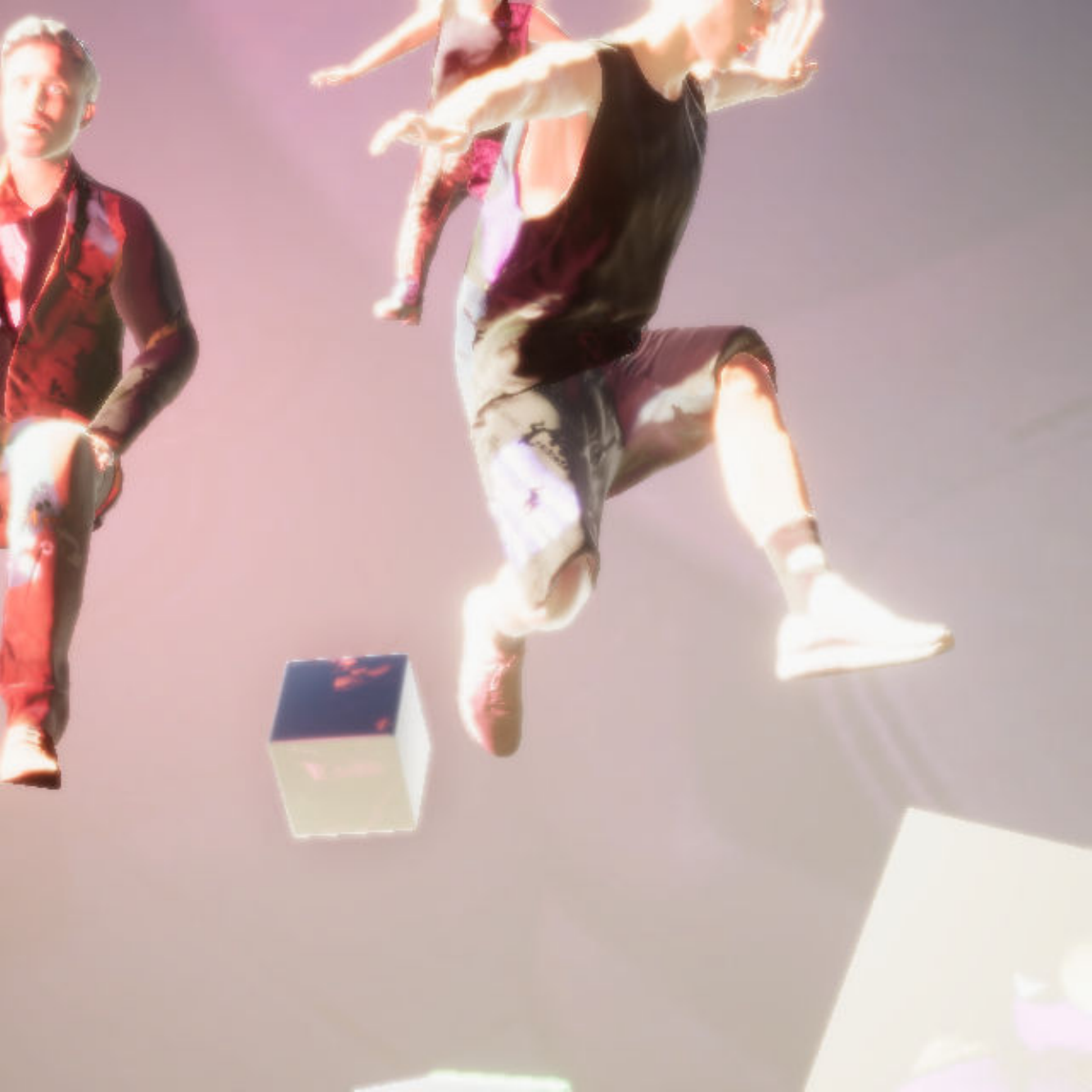}}
    \end{subfigure}
    \begin{subfigure}[t]{0.270\textwidth}
        \raisebox{-\height}{\includegraphics[width=\textwidth]{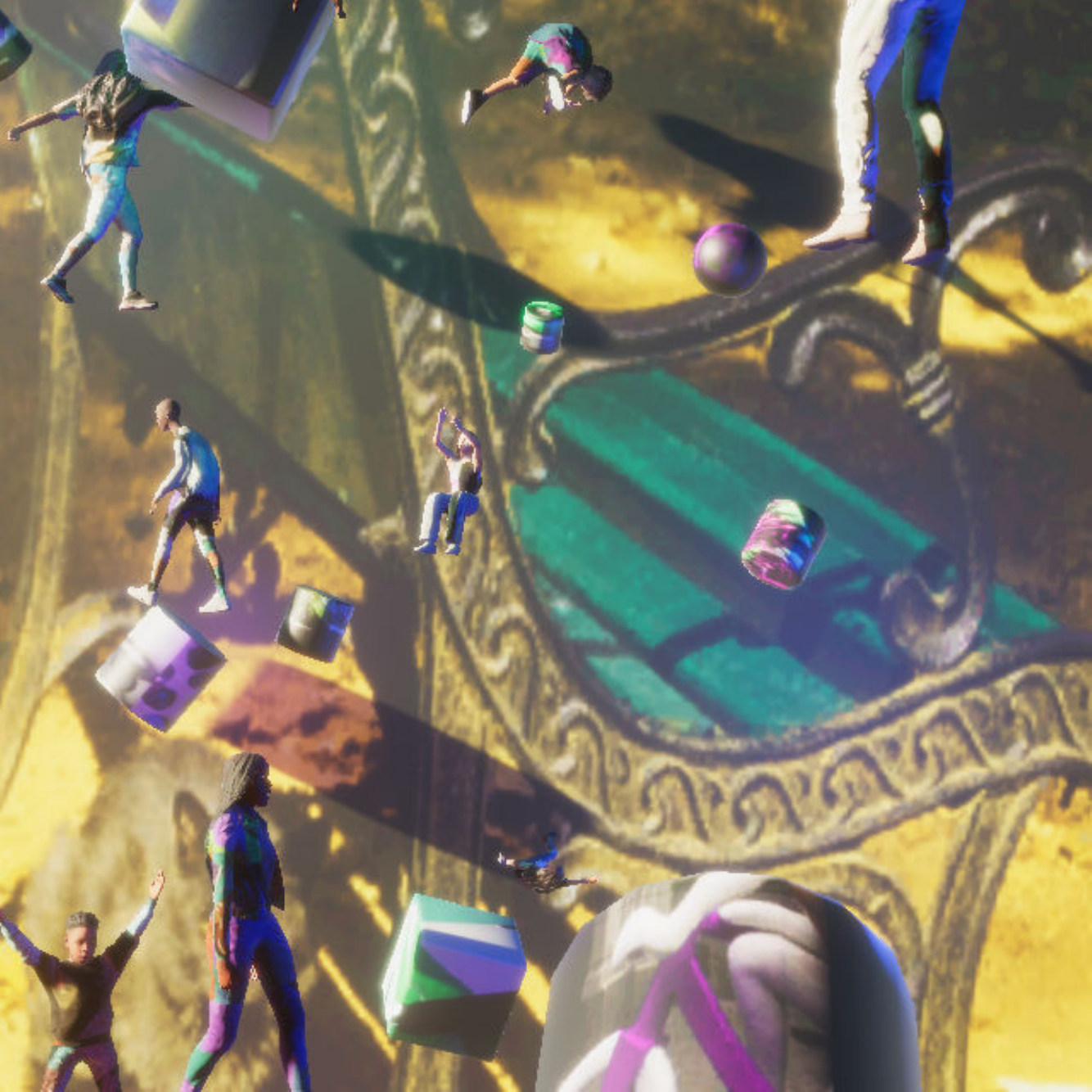}}
    \end{subfigure}
    \hfill
\caption{More examples of generated data 1/2.}
\label{fig:fig:moreteaser1}%
\end{figure}

\begin{figure}[htb]
    \centering
    \begin{subfigure}[t]{0.270\textwidth}
        \raisebox{-\height}{\includegraphics[width=\textwidth]{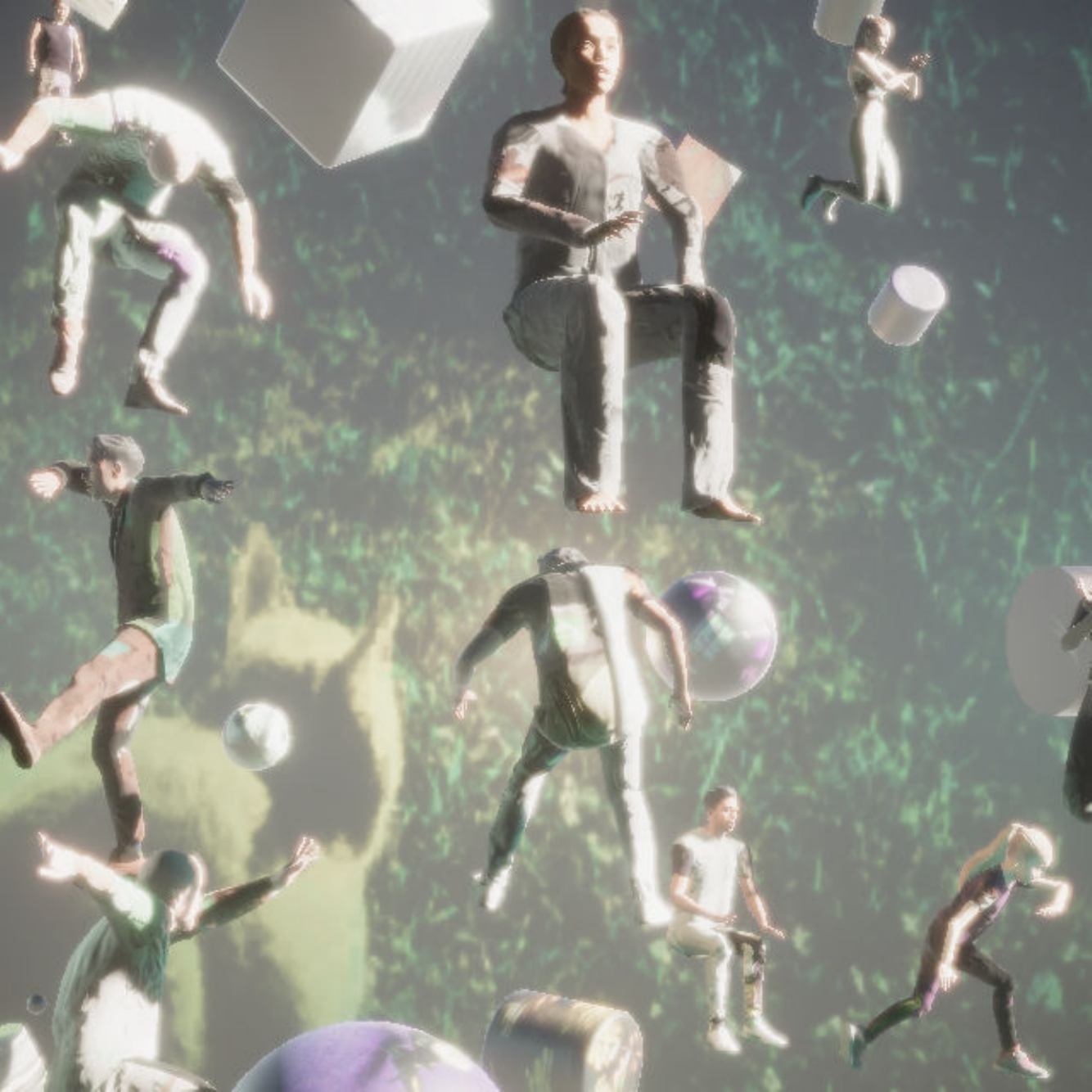}}
    \end{subfigure}
    \begin{subfigure}[t]{0.270\textwidth}
        \raisebox{-\height}{\includegraphics[width=\textwidth]{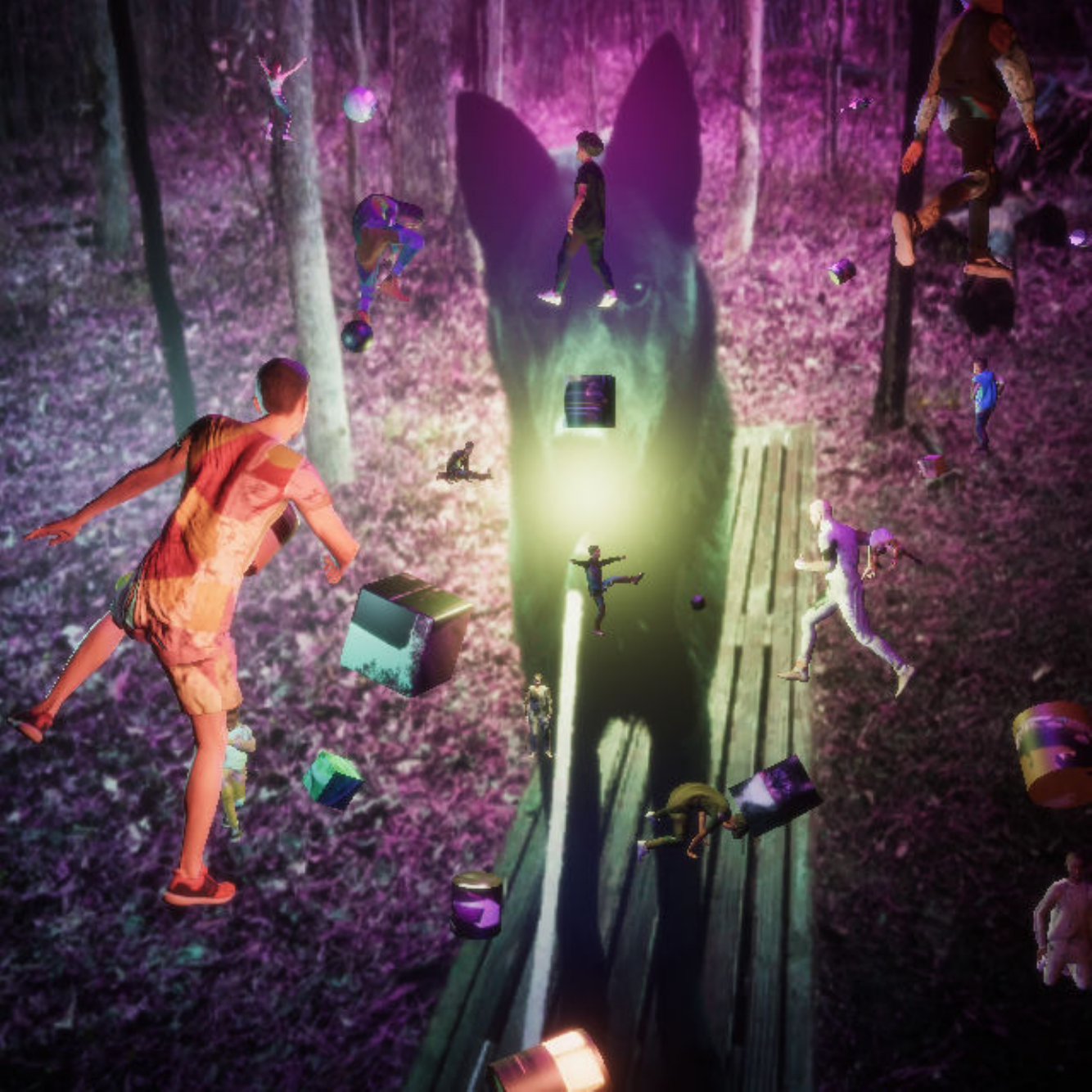}}
    \end{subfigure}
    \begin{subfigure}[t]{0.270\textwidth}
        \raisebox{-\height}{\includegraphics[width=\textwidth]{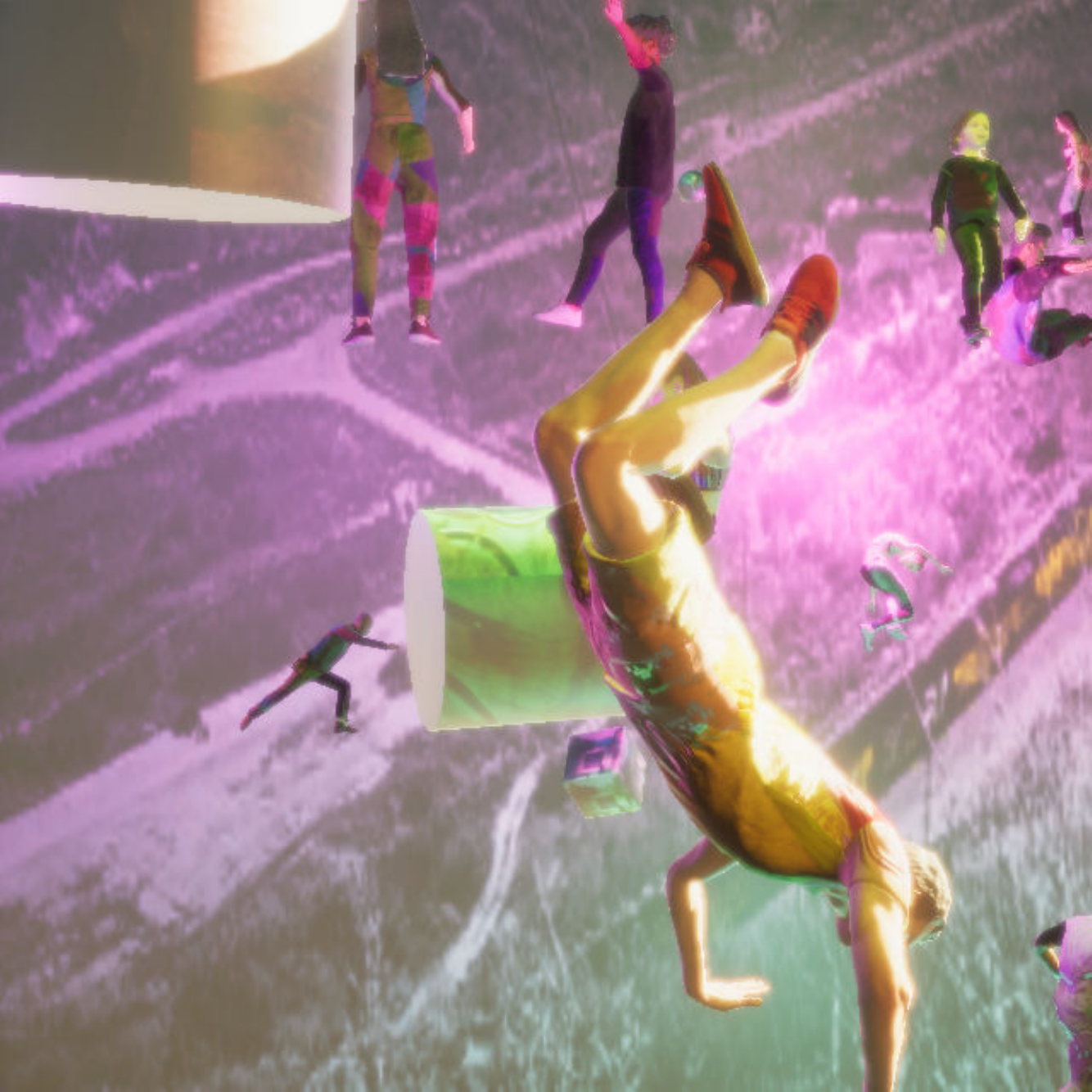}}
    \end{subfigure}
    \hfill
    \\
    \begin{subfigure}[t]{0.270\textwidth}
        \raisebox{-\height}{\includegraphics[width=\textwidth]{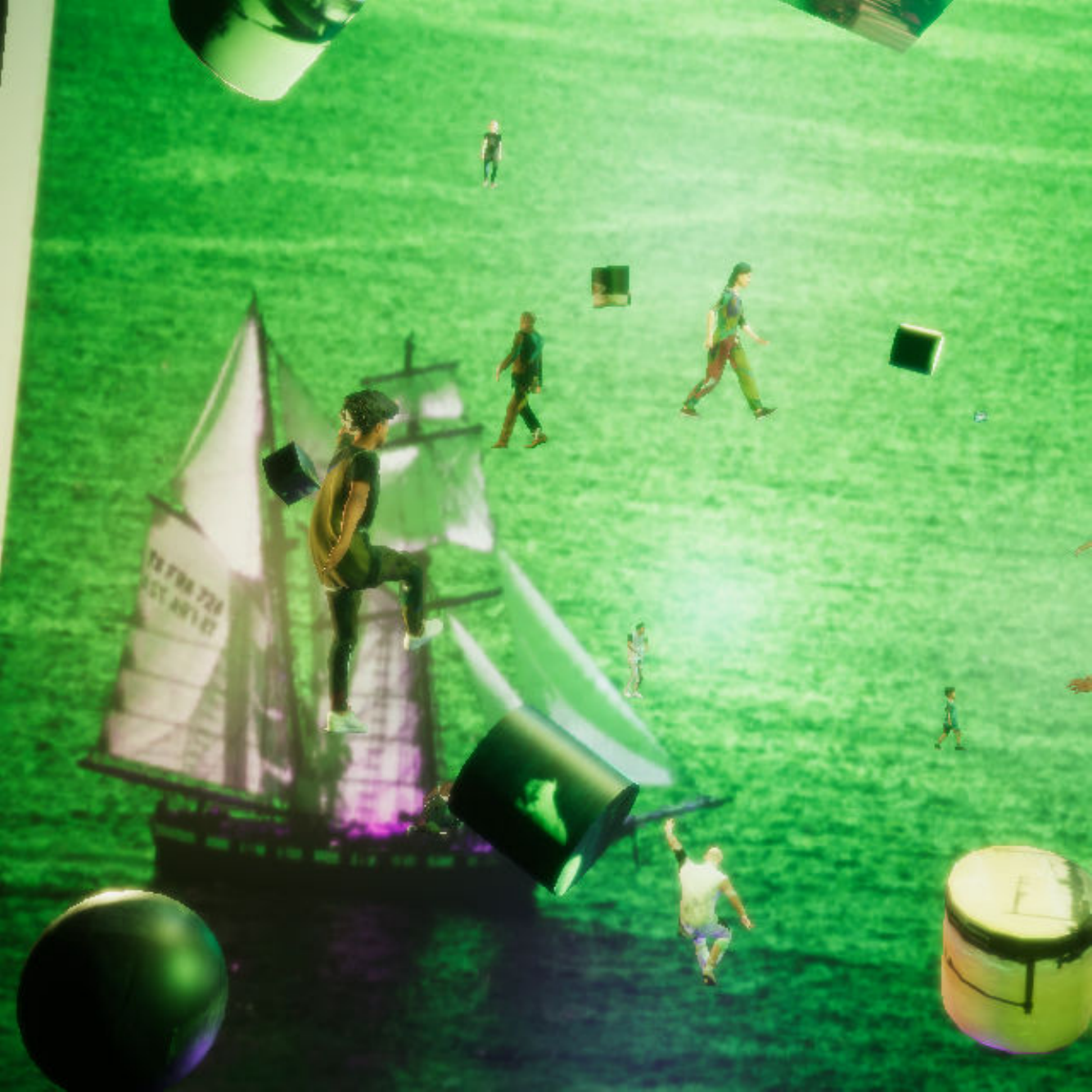}}
    \end{subfigure}
    \begin{subfigure}[t]{0.270\textwidth}
        \raisebox{-\height}{\includegraphics[width=\textwidth]{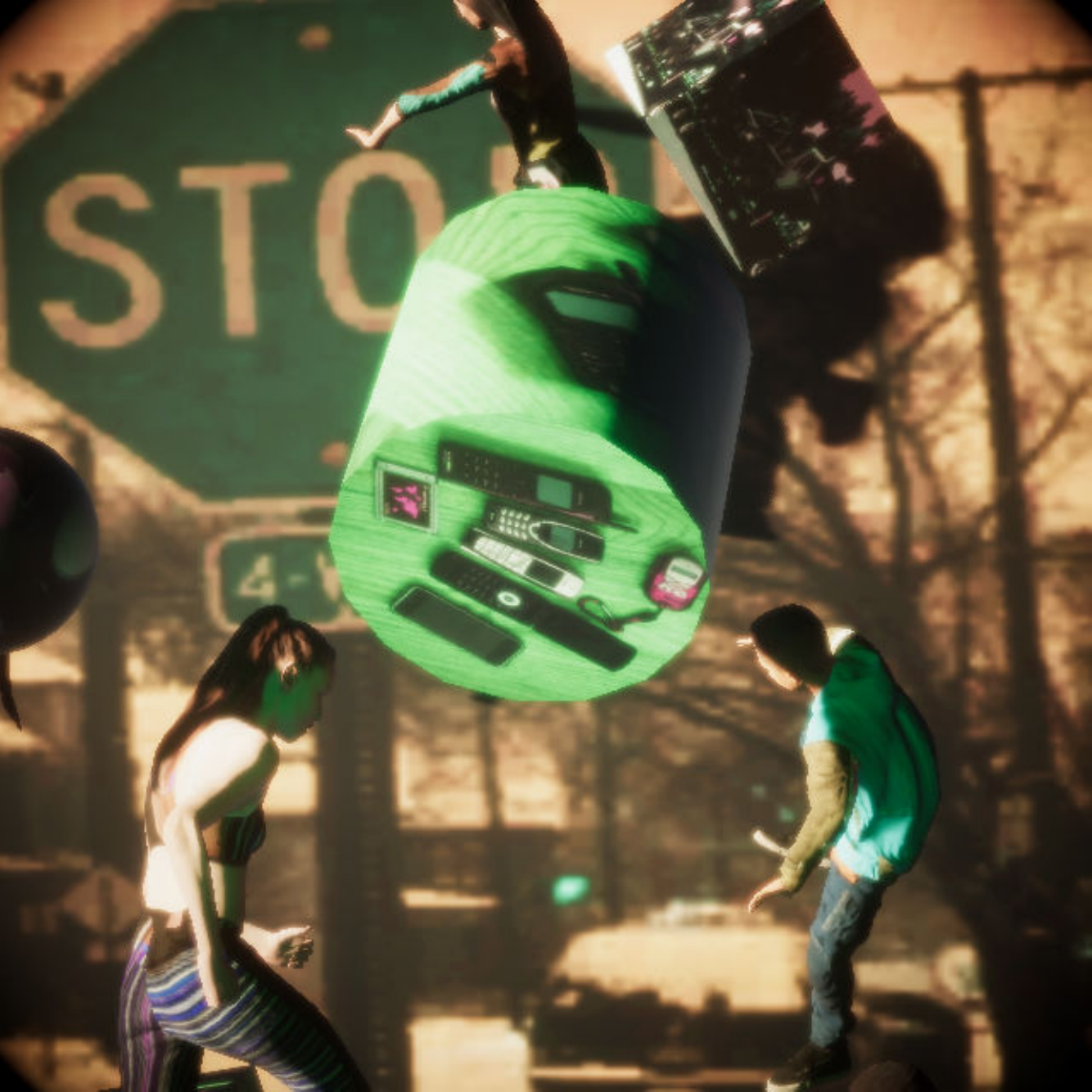}}
    \end{subfigure}
    \begin{subfigure}[t]{0.270\textwidth}
        \raisebox{-\height}{\includegraphics[width=\textwidth]{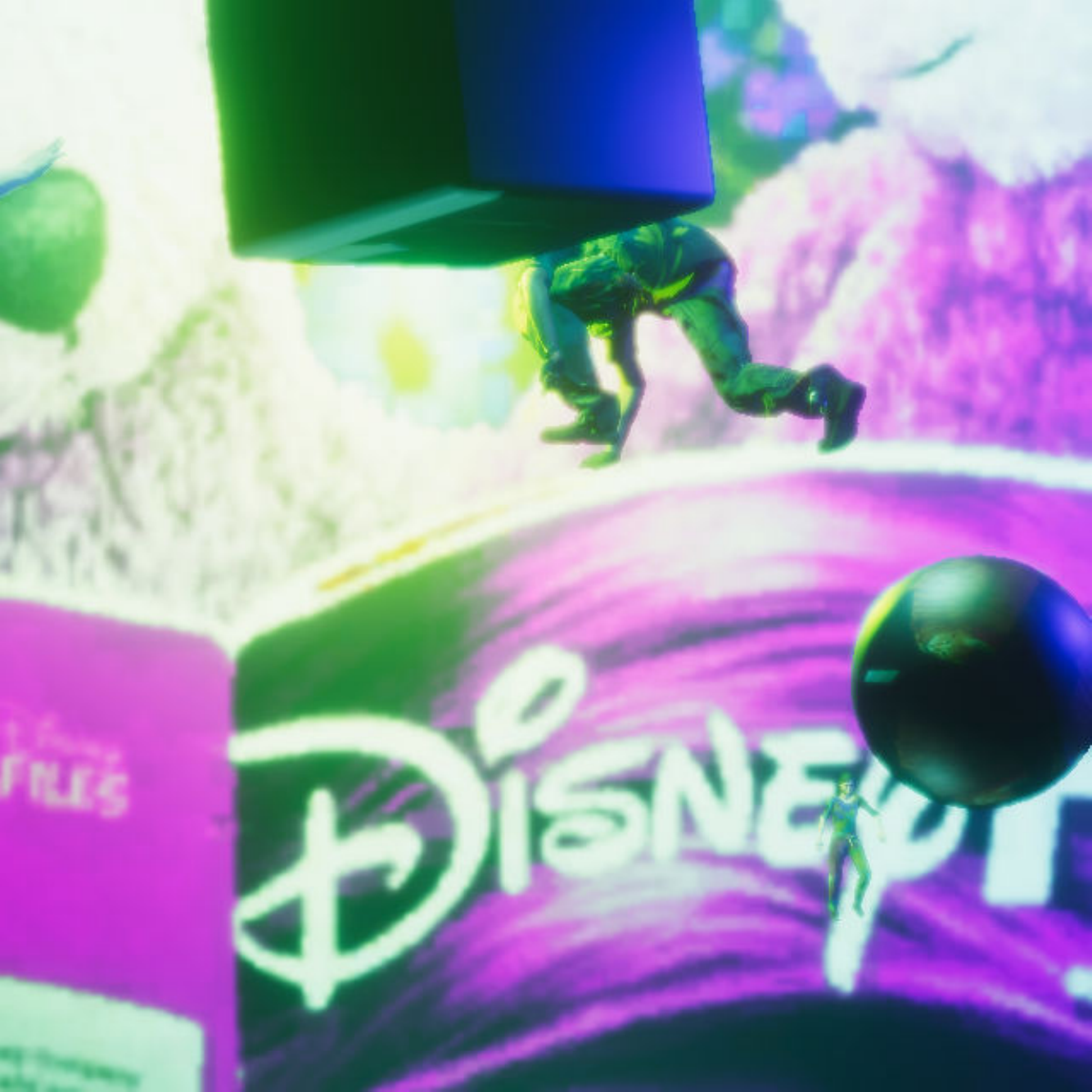}}
    \end{subfigure}
    \hfill
    \\
    \begin{subfigure}[t]{0.270\textwidth}
        \raisebox{-\height}{\includegraphics[width=\textwidth]{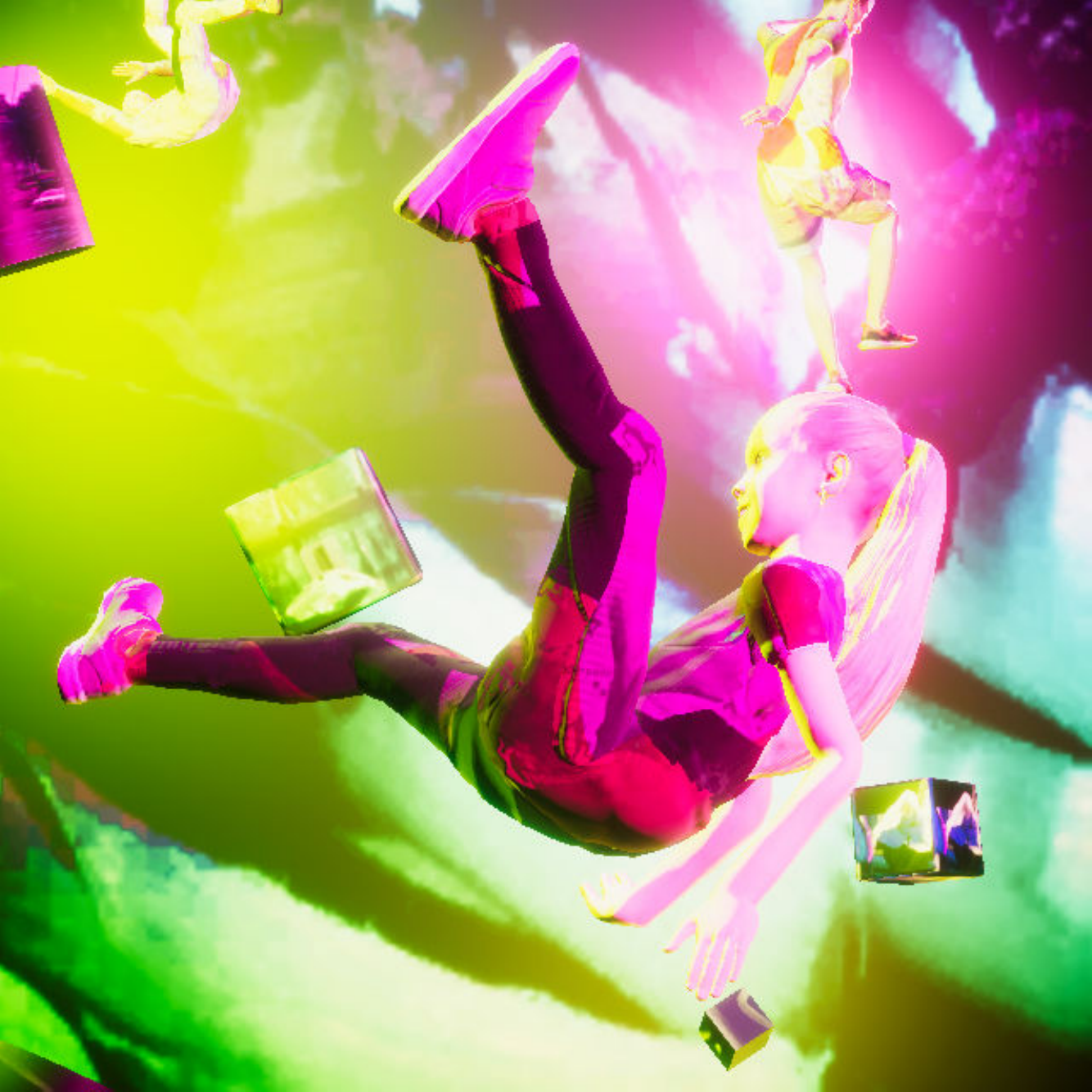}}
    \end{subfigure}
    \begin{subfigure}[t]{0.270\textwidth}
        \raisebox{-\height}{\includegraphics[width=\textwidth]{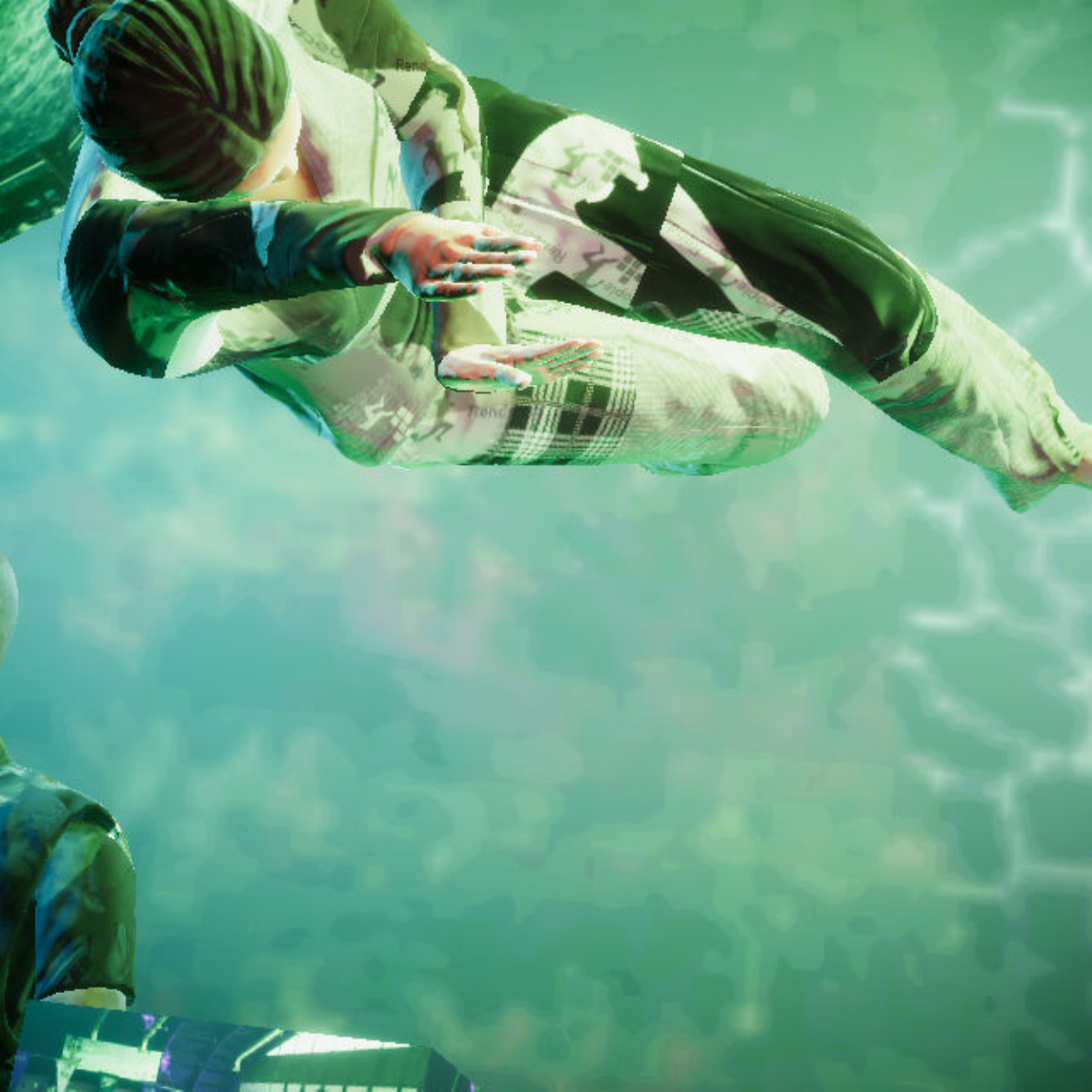}}
    \end{subfigure}
    \begin{subfigure}[t]{0.270\textwidth}
        \raisebox{-\height}{\includegraphics[width=\textwidth]{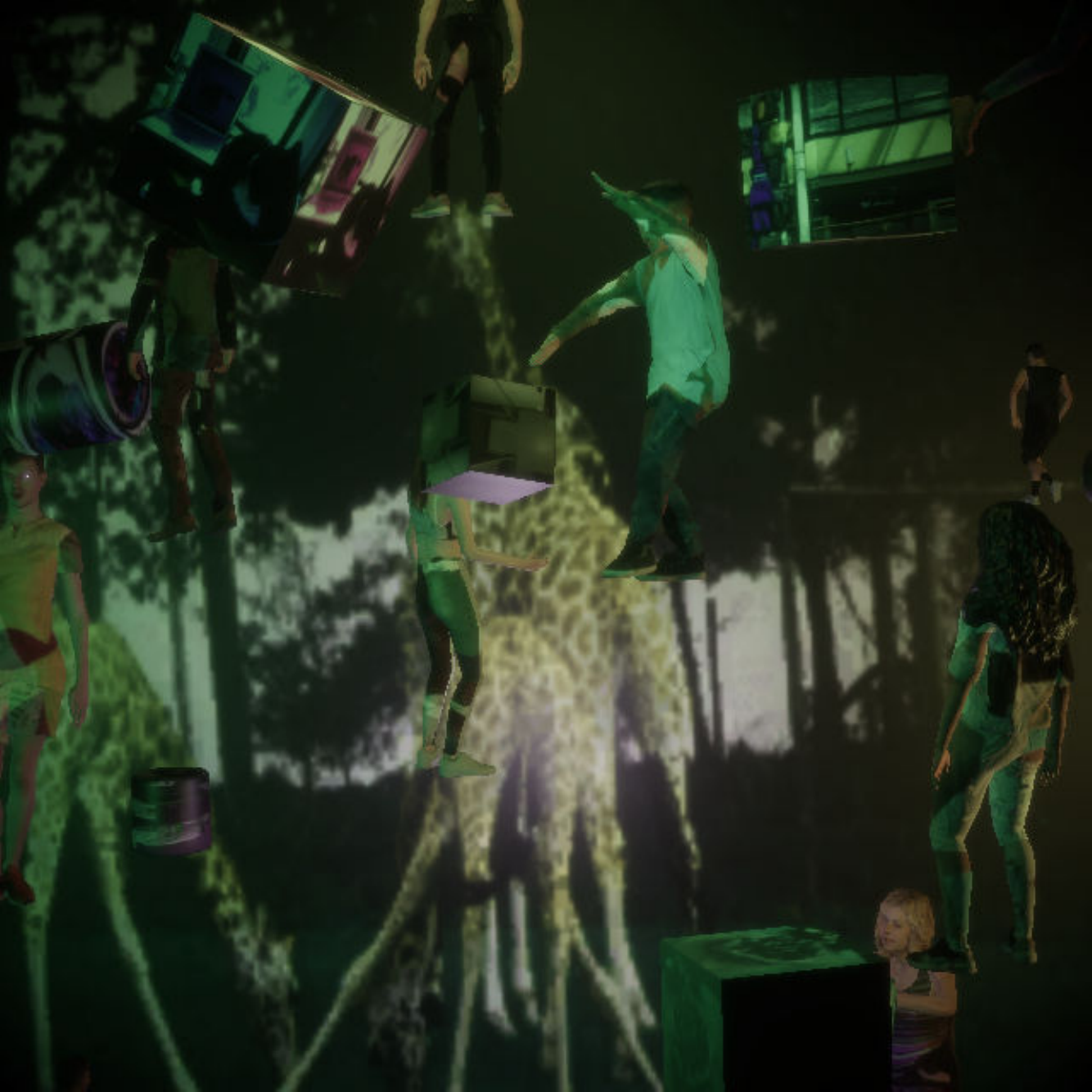}}
    \end{subfigure}
    \hfill
    \\
    \begin{subfigure}[t]{0.270\textwidth}
        \raisebox{-\height}{\includegraphics[width=\textwidth]{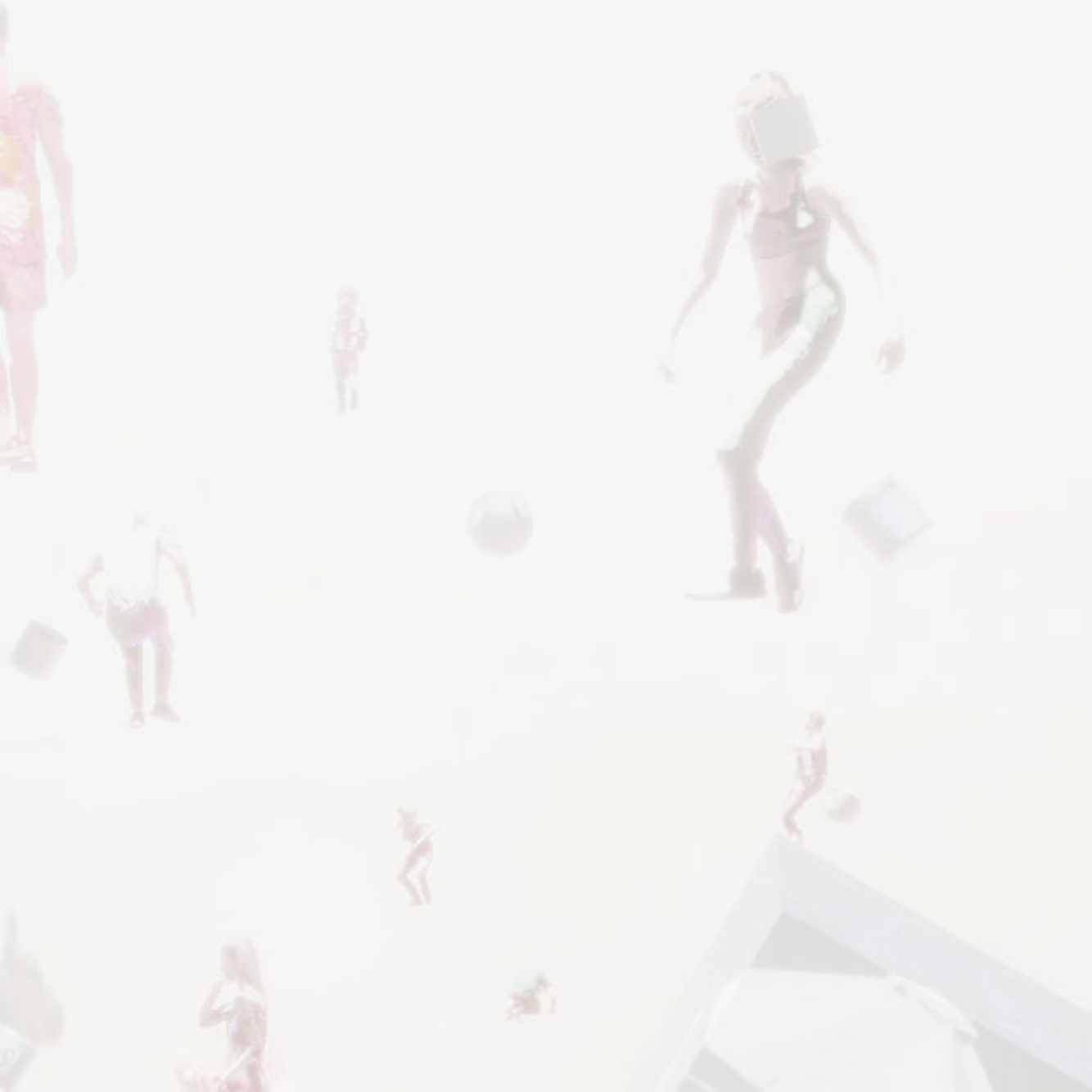}}
    \end{subfigure}
    \begin{subfigure}[t]{0.270\textwidth}
        \raisebox{-\height}{\includegraphics[width=\textwidth]{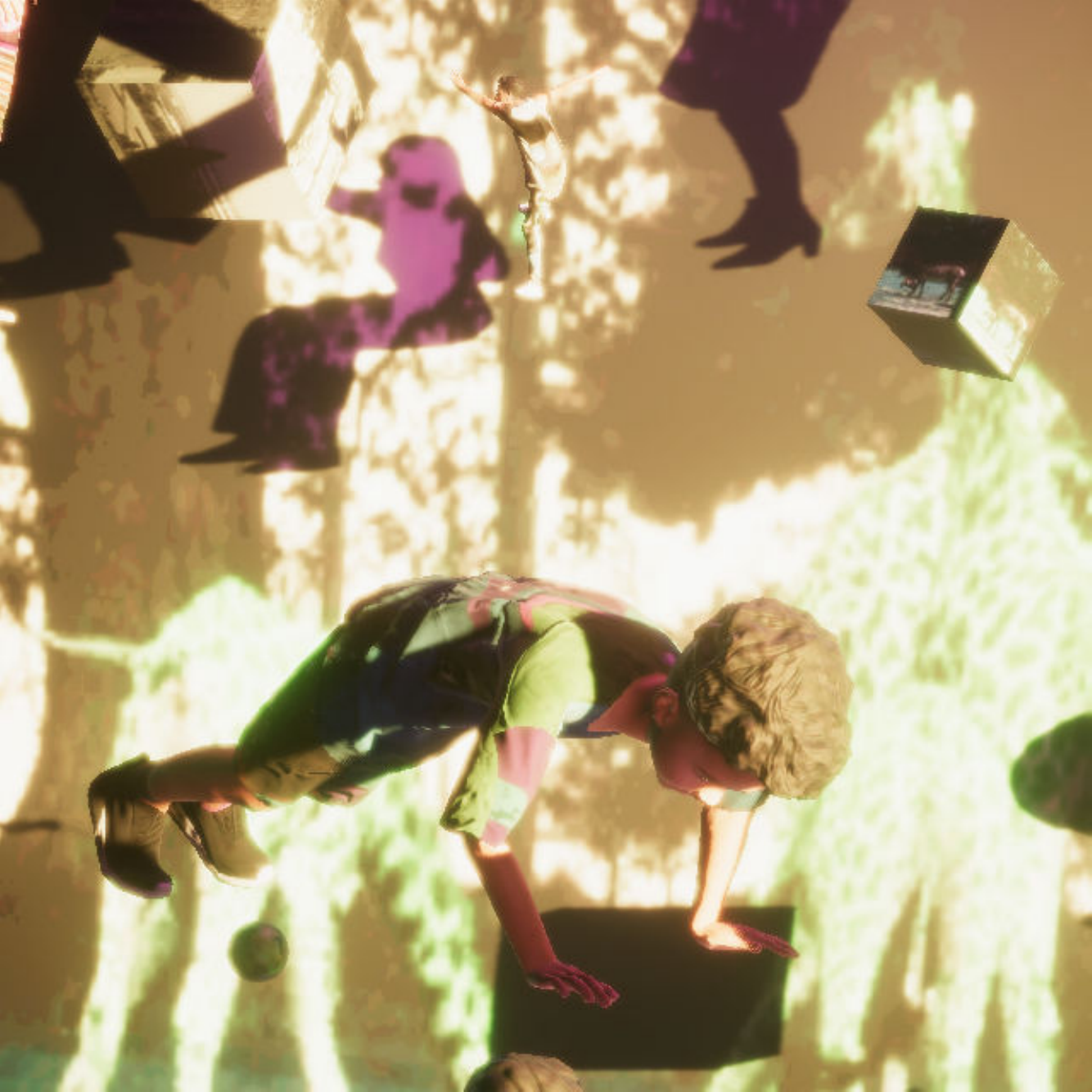}}
    \end{subfigure}
    \begin{subfigure}[t]{0.270\textwidth}
        \raisebox{-\height}{\includegraphics[width=\textwidth]{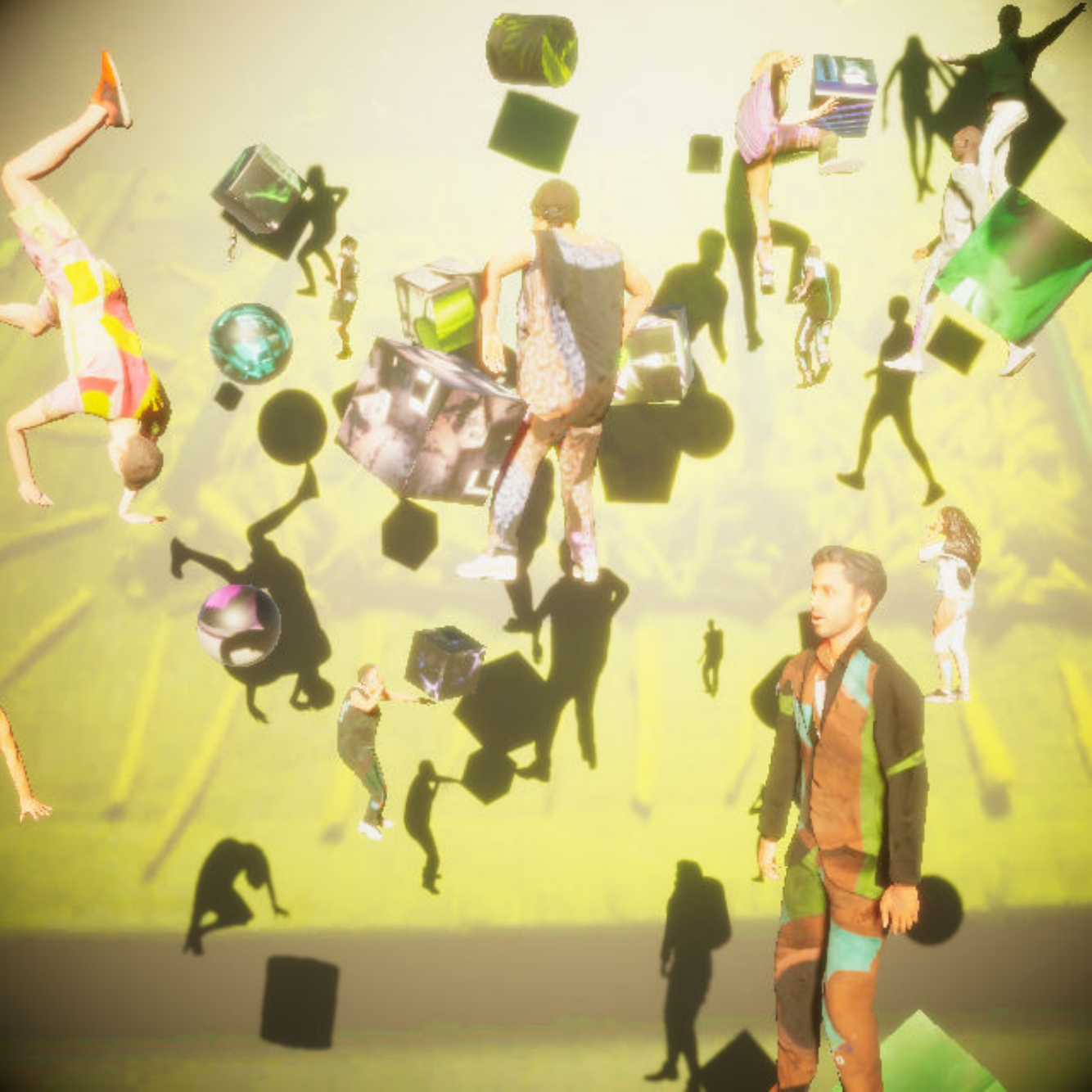}}
    \end{subfigure}
    \hfill
    \\
    \begin{subfigure}[t]{0.270\textwidth}
        \raisebox{-\height}{\includegraphics[width=\textwidth]{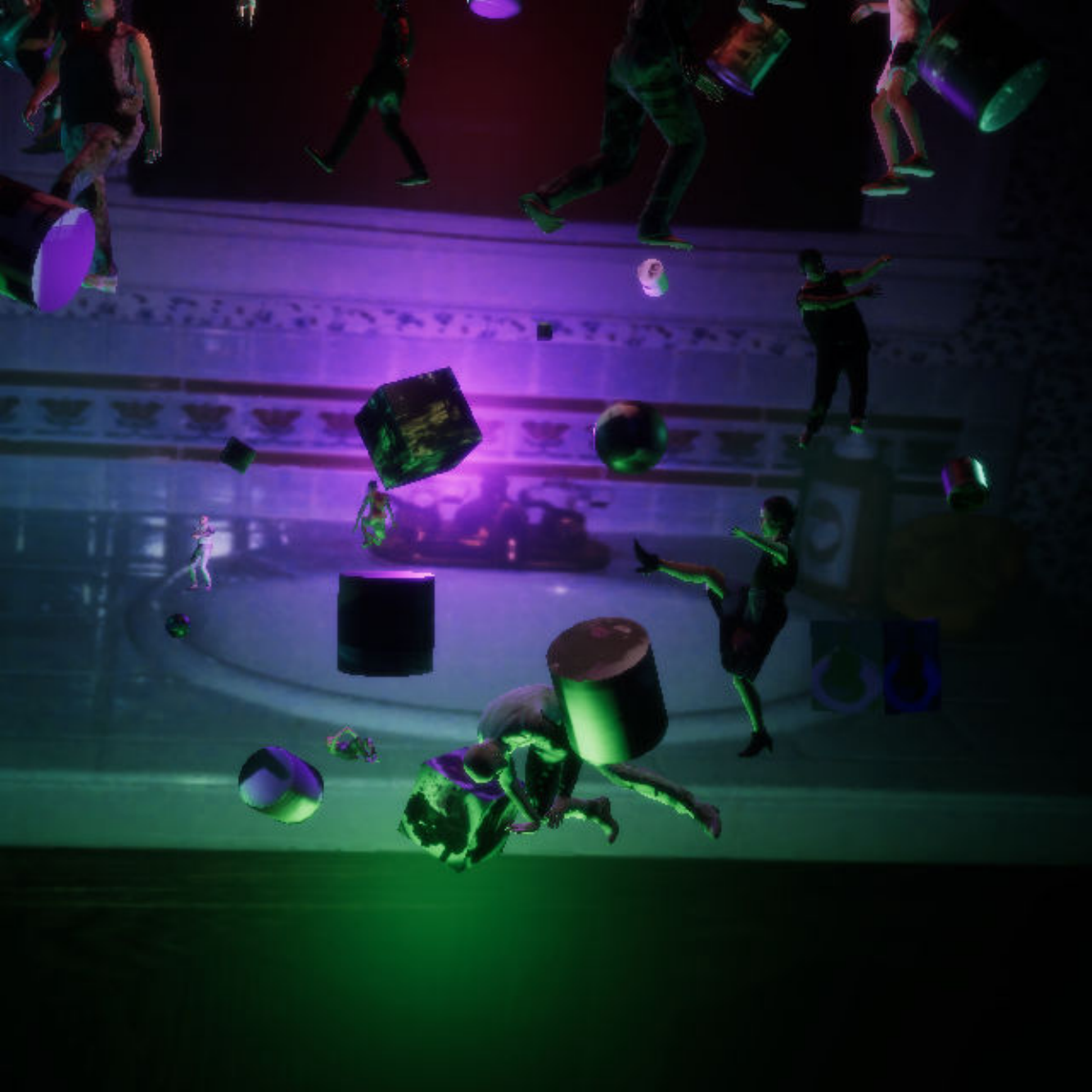}}
    \end{subfigure}
    \begin{subfigure}[t]{0.270\textwidth}
        \raisebox{-\height}{\includegraphics[width=\textwidth]{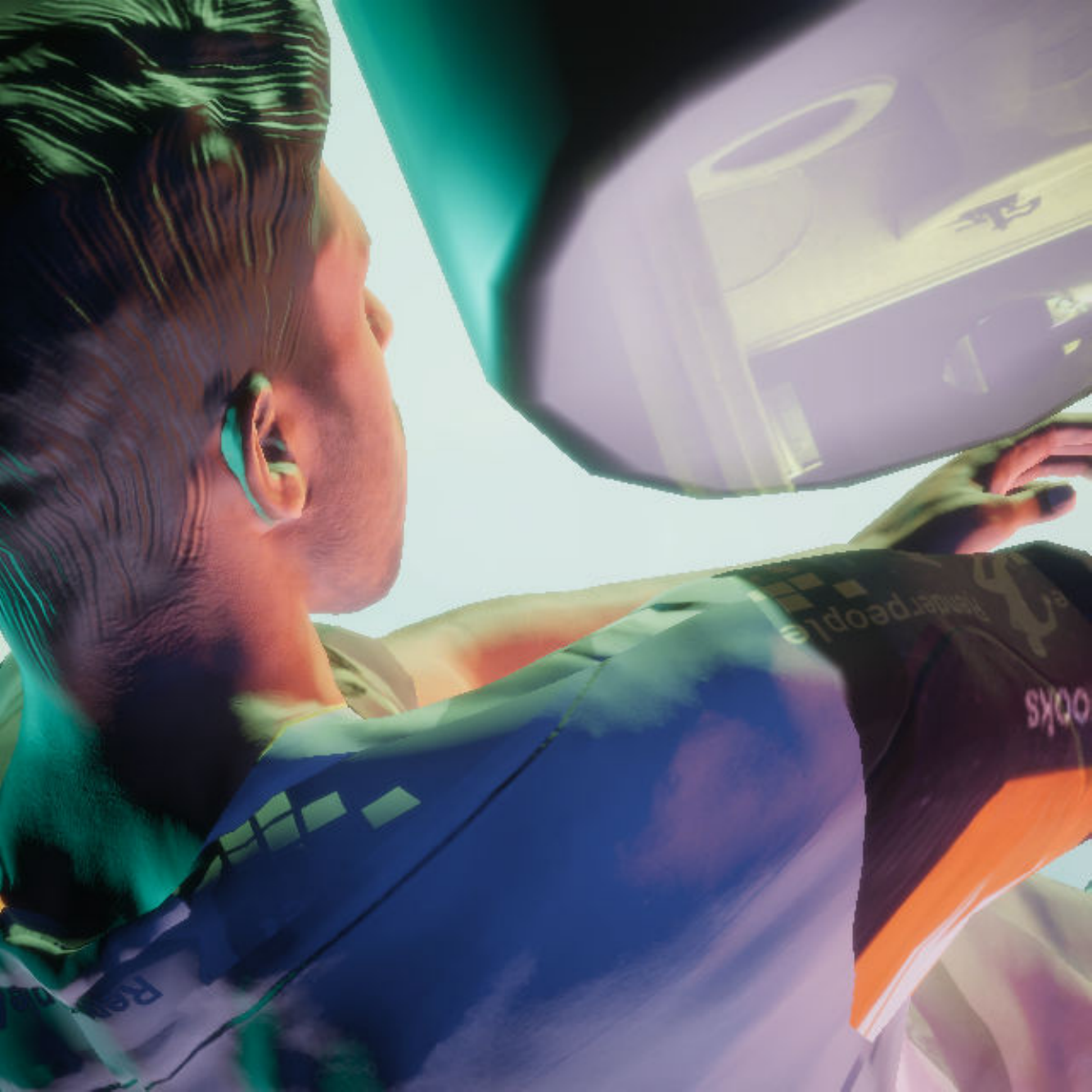}}
    \end{subfigure}
    \begin{subfigure}[t]{0.270\textwidth}
        \raisebox{-\height}{\includegraphics[width=\textwidth]{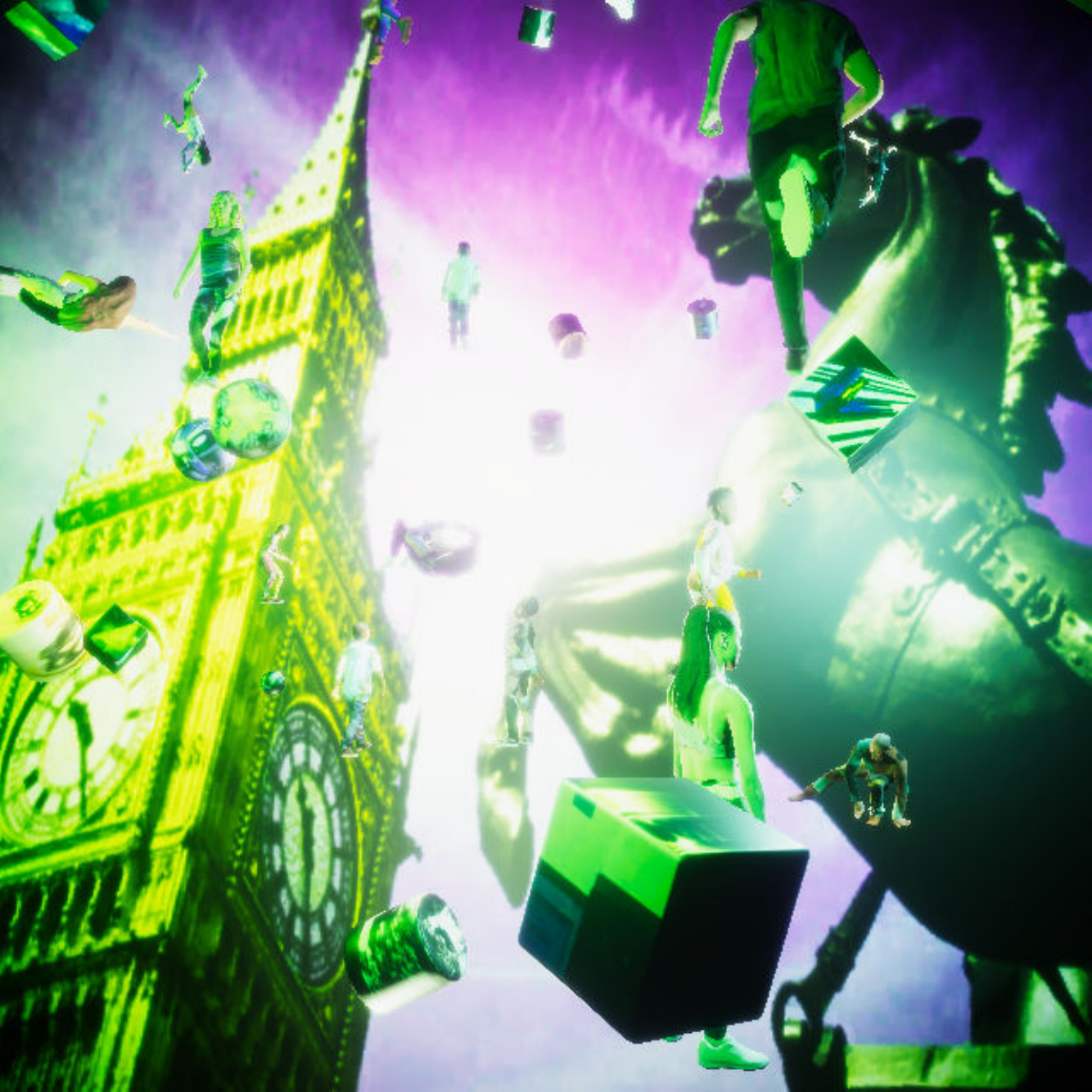}}
    \end{subfigure}
    \hfill
    \\
    \begin{subfigure}[t]{0.270\textwidth}
        \raisebox{-\height}{\includegraphics[width=\textwidth]{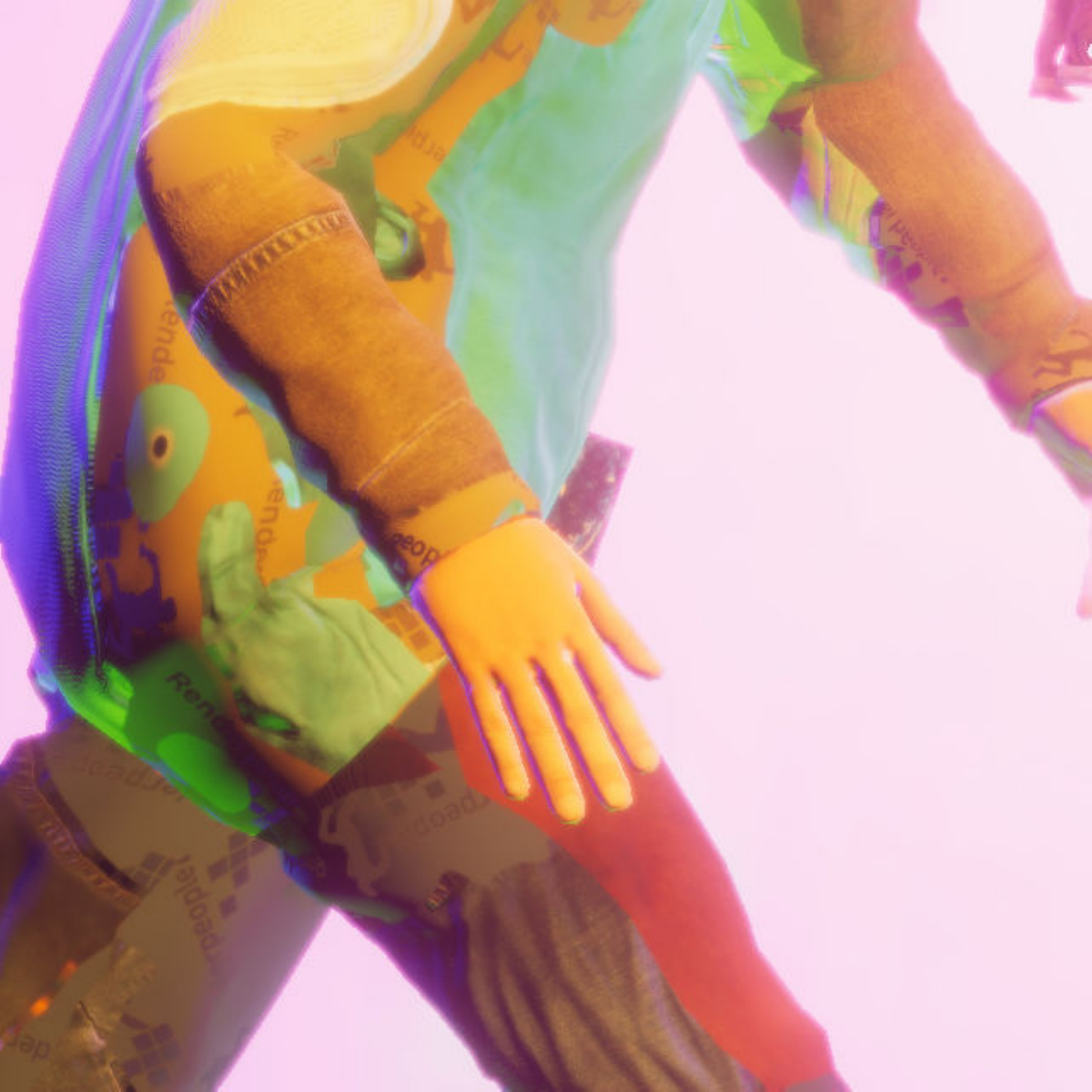}}
    \end{subfigure}
    \begin{subfigure}[t]{0.270\textwidth}
        \raisebox{-\height}{\includegraphics[width=\textwidth]{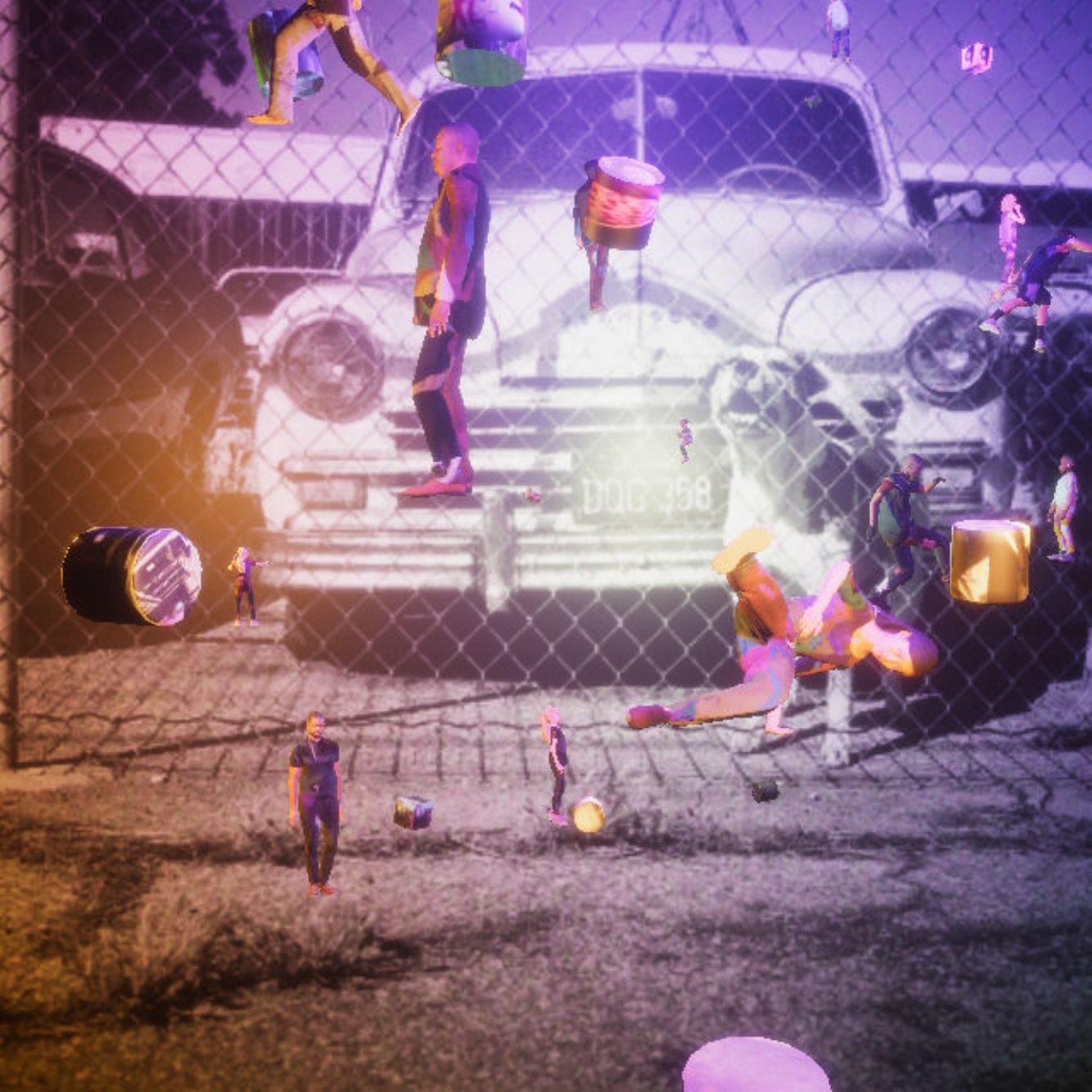}}
    \end{subfigure}
    \begin{subfigure}[t]{0.270\textwidth}
        \raisebox{-\height}{\includegraphics[width=\textwidth]{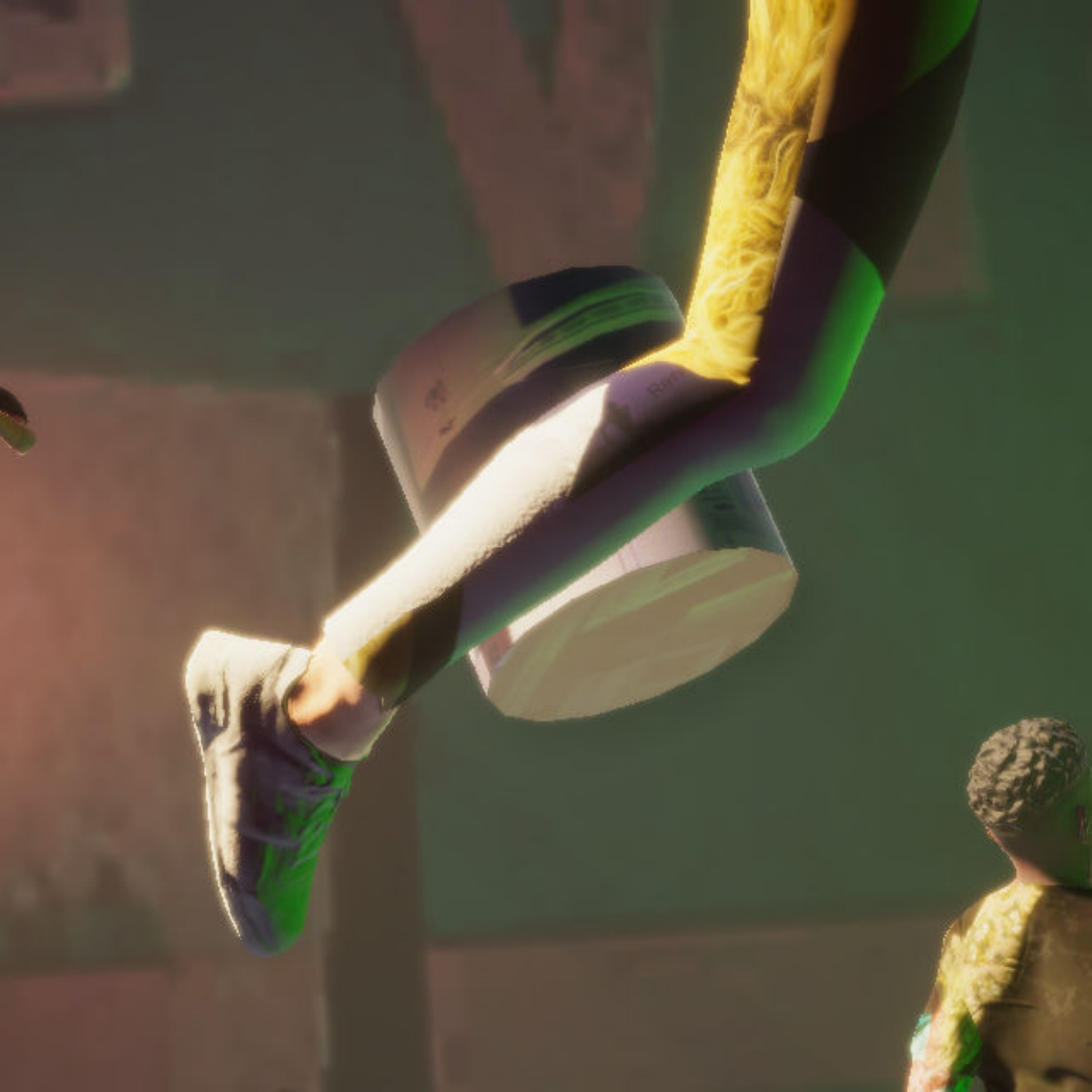}}
    \end{subfigure}
\caption{More examples of generated data 2/2.}
\label{fig:fig:moreteaser2}%
\end{figure}

In Fig.~\ref{fig:fig:moreteaser1} and \ref{fig:fig:moreteaser2} we show additional examples from our generated dataset. Note the variety of view perspectives enabled by the camera randomizer (translating and rotating to change perspective and zooming in and out, and adding blur and bloom with some objects out of focus). The lighting conditions are diversified thanks to the light randomizer, producing some over-exposed and under-exposed images and mimicking artificial and natural light settings and shadows. The lighting color changes and some post-processing effects also create unique looks for our scenes and augment the dataset by nature.

We use a set of primitive 3D game objects such as cubes, cylinders, and spheres provided by the Perception package in our scene to act as background or occluder/distractor objects. We spawn them at random positions with random scales, orientations, textures, and hue offsets in the scene. We use the same COCO unlabeled 2017 textures for these objects. In some generated frames the occluder objects likely obstruct much of the Perception camera's view. If this is not desired, it can be adjusted by modifying the parameters of Background/Occluder placement and intrinsic camera parameters described in the previous section.
The background texture changes in addition to the occluder/distractor objects random placement and texture changes increase the diversity of the scenes, producing some challenging examples for the model. 

The animation randomizer varies the pose of the characters, with some facing away from the camera. We encourage the readers to study the effect of the pose randomizer and the random rotation of people assets around the $Y$-axis. A model presented with these examples might perform better as this data exhibits a larger variety of poses with people partially visible in each scene.

The Shader Graph randomizer varies the texture of clothing, producing some camouflage-like textures, in total producing some \num[group-separator={,}]{21952} unique clothing textures, creases, and wrinkles -- from $28$ Albedos, $28$ Masks, and $28$ Normals -- that look different under lighting. We think of such variations in our data generator as a built-in data augmentation technique that could facilitate some research into the adversarial robustness of human-centric computer vision.

\subsection{Additional Examples from Lighting Diversity}
\begin{figure}[htb]
    \centering
    \begin{subfigure}[t]{0.230\textwidth}
        \raisebox{-\height}{\includegraphics[width=\textwidth]{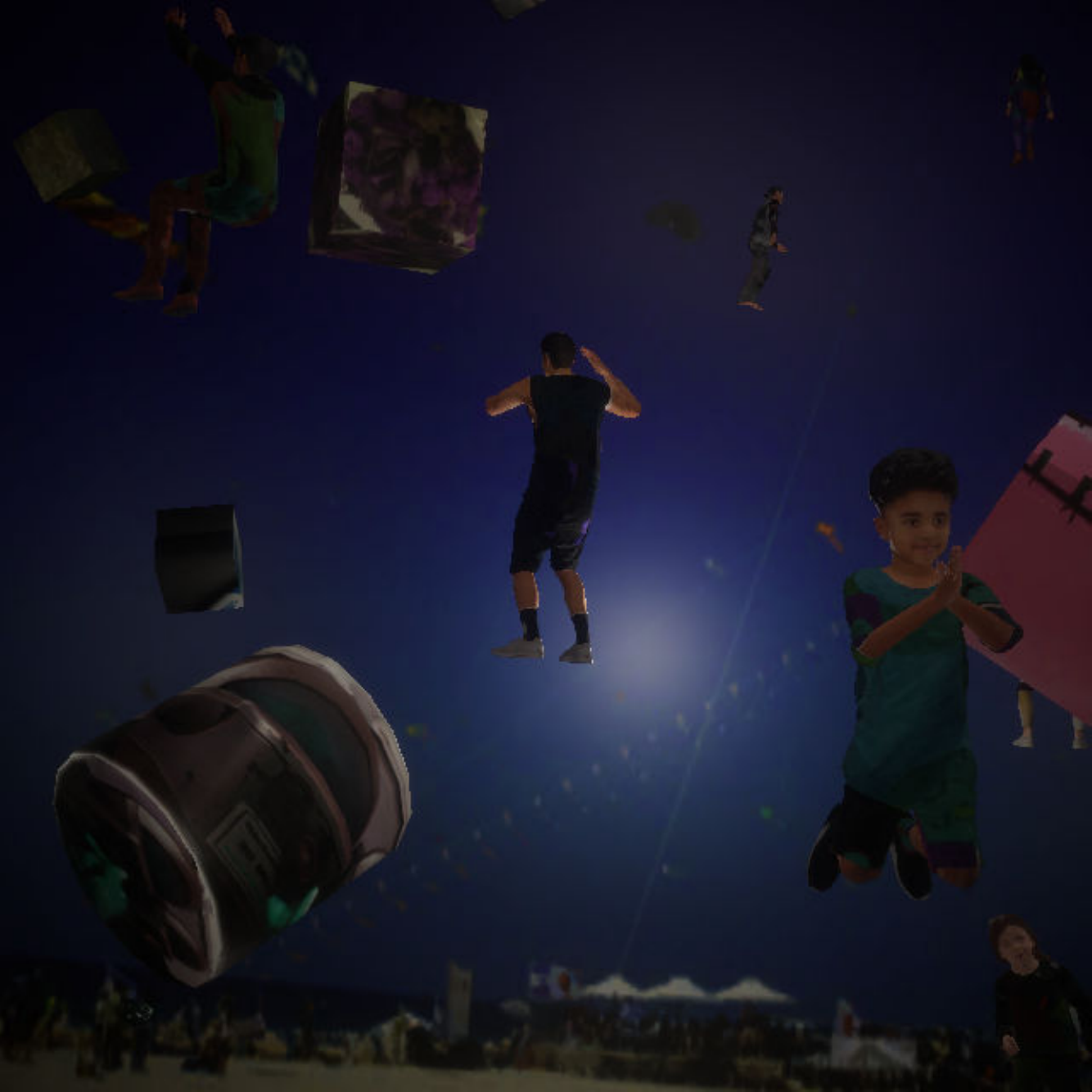}}
    \end{subfigure}
    \begin{subfigure}[t]{0.230\textwidth}
        \raisebox{-\height}{\includegraphics[width=\textwidth]{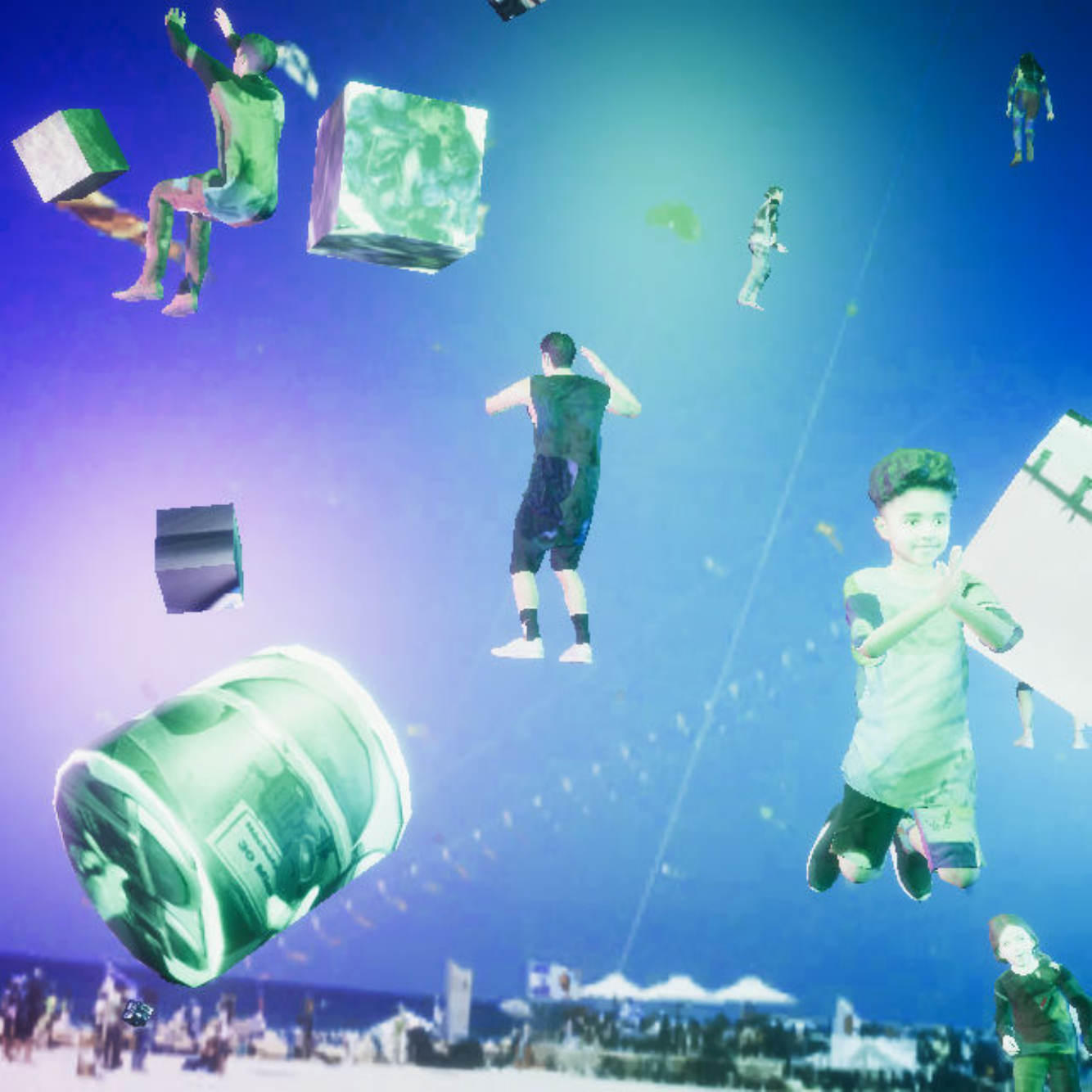}}
    \end{subfigure}
    \begin{subfigure}[t]{0.230\textwidth}
        \raisebox{-\height}{\includegraphics[width=\textwidth]{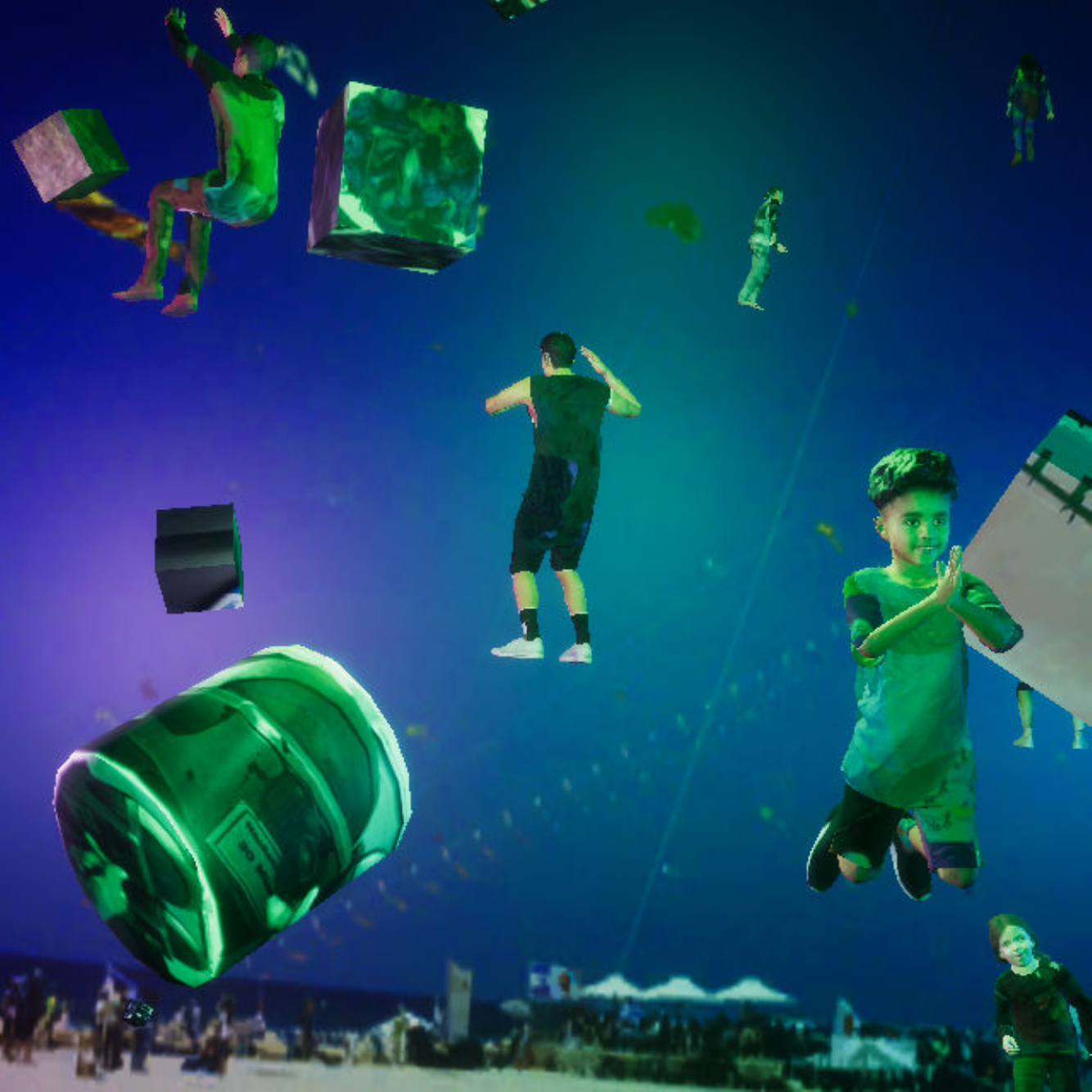}}
    \end{subfigure}
    \hfill
    \\
    \begin{subfigure}[t]{0.230\textwidth}
        \raisebox{-\height}{\includegraphics[width=\textwidth]{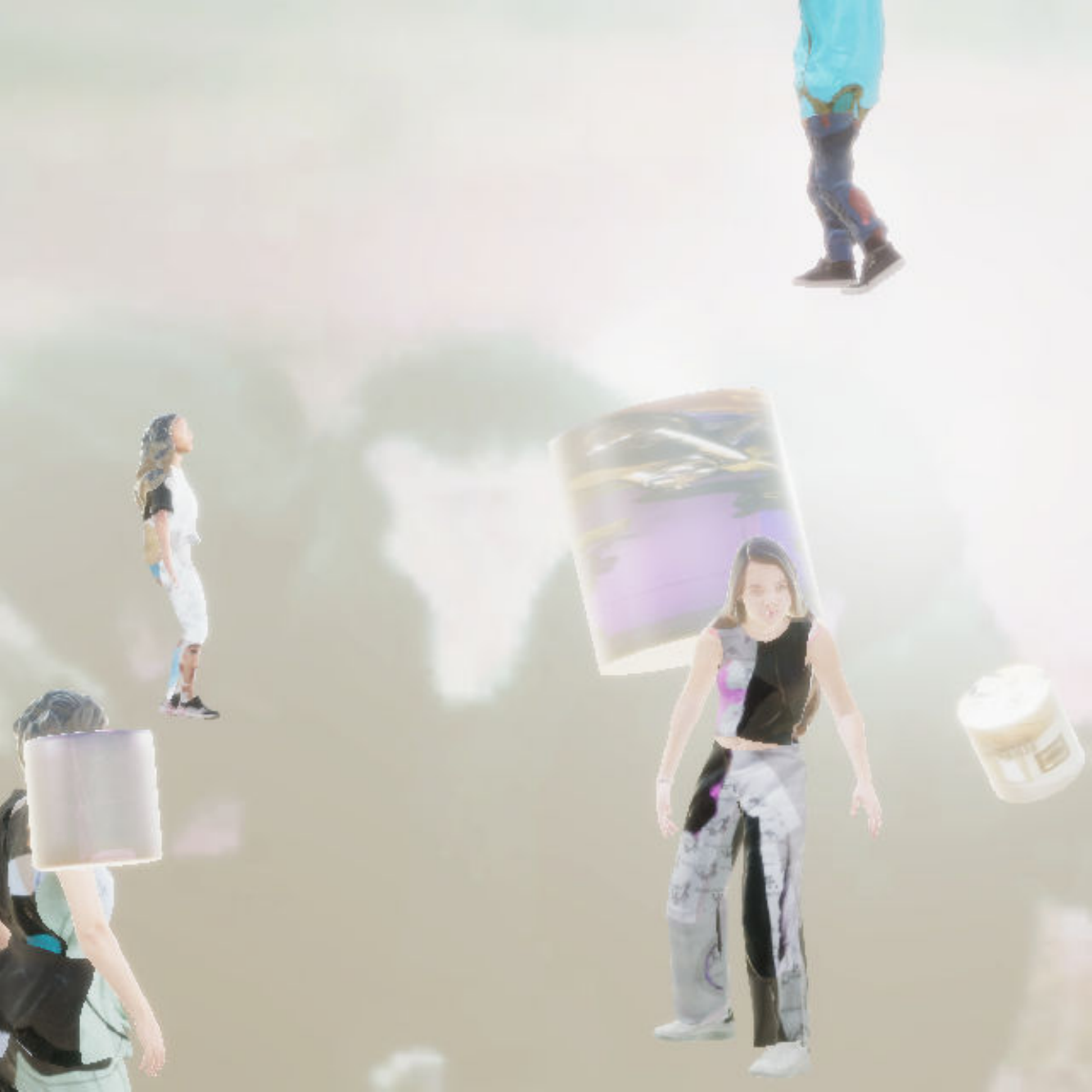}}
    \end{subfigure}
    \begin{subfigure}[t]{0.230\textwidth}
        \raisebox{-\height}{\includegraphics[width=\textwidth]{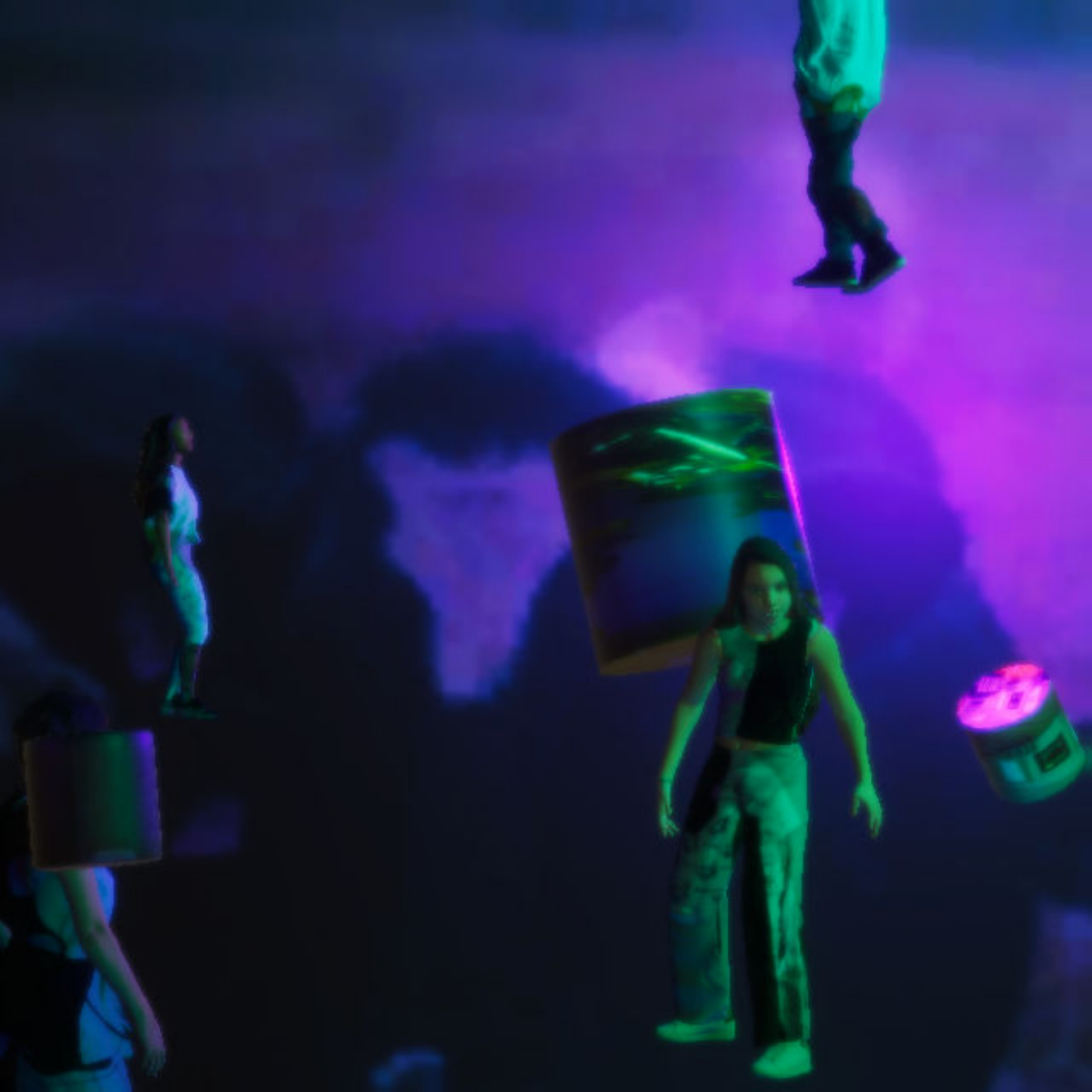}}
    \end{subfigure}
    \begin{subfigure}[t]{0.230\textwidth}
        \raisebox{-\height}{\includegraphics[width=\textwidth]{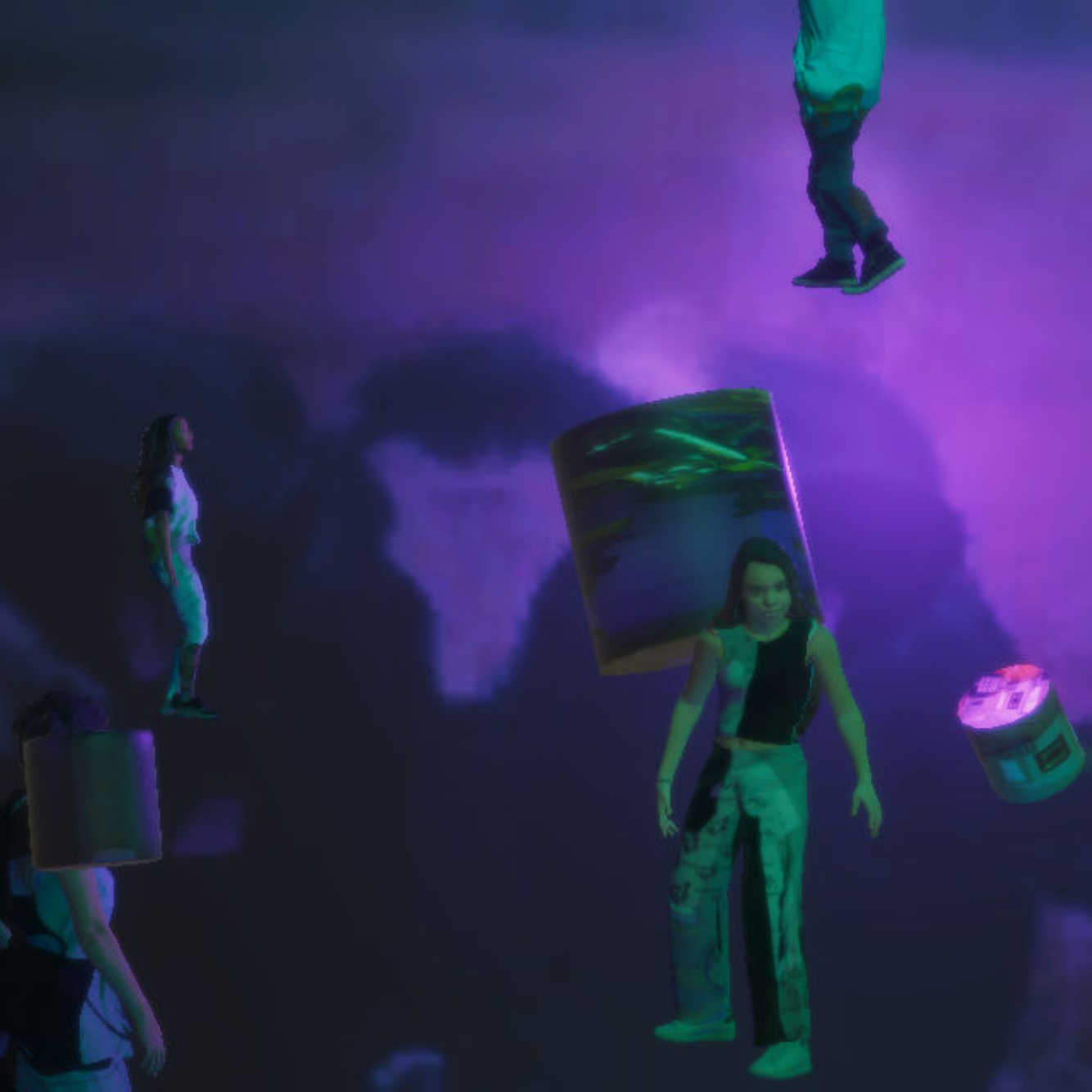}}
    \end{subfigure}
    \\
    \begin{subfigure}[t]{0.230\textwidth}
        \raisebox{-\height}{\includegraphics[width=\textwidth]{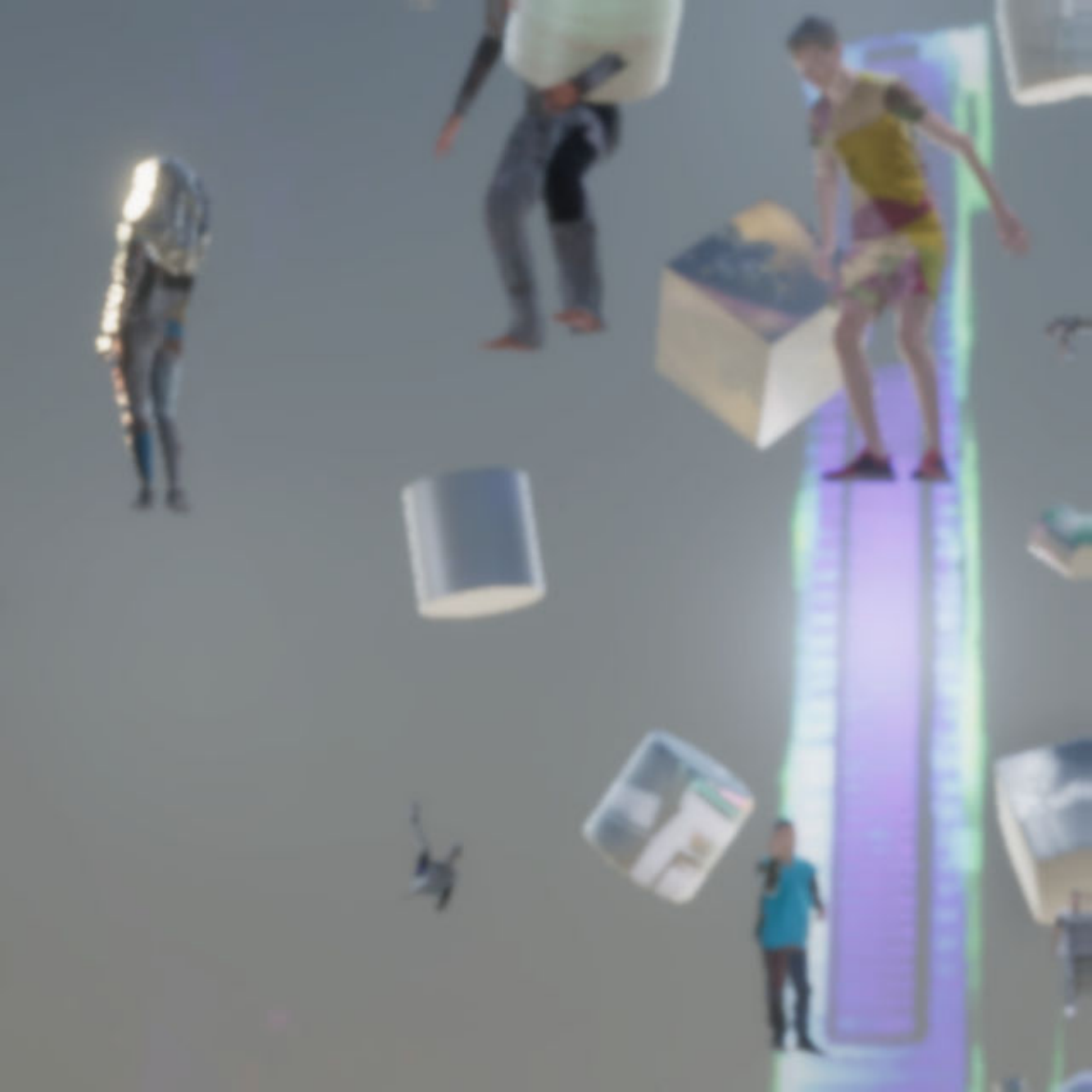}}
    \end{subfigure}
    \begin{subfigure}[t]{0.230\textwidth}
        \raisebox{-\height}{\includegraphics[width=\textwidth]{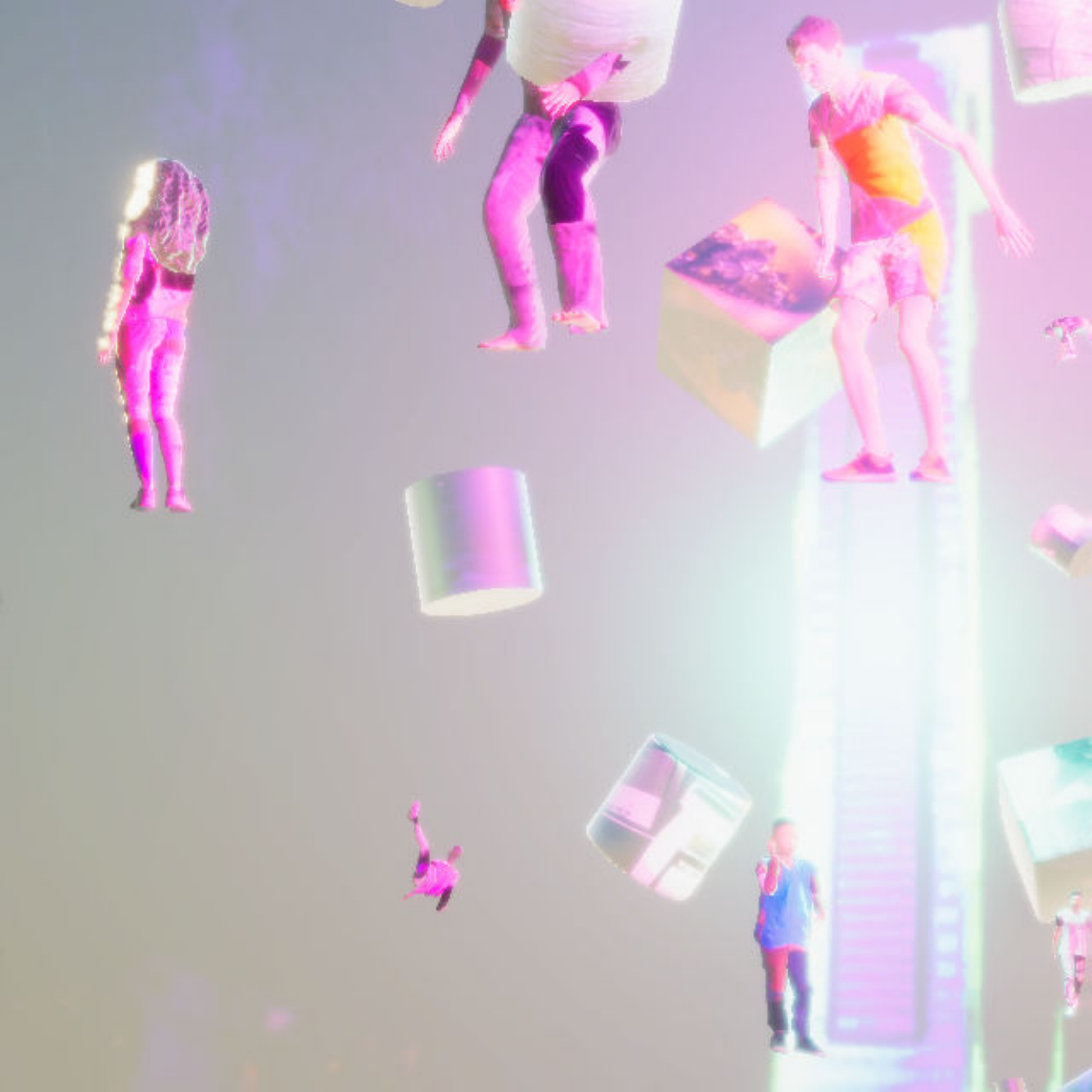}}
    \end{subfigure}
    \begin{subfigure}[t]{0.230\textwidth}
        \raisebox{-\height}{\includegraphics[width=\textwidth]{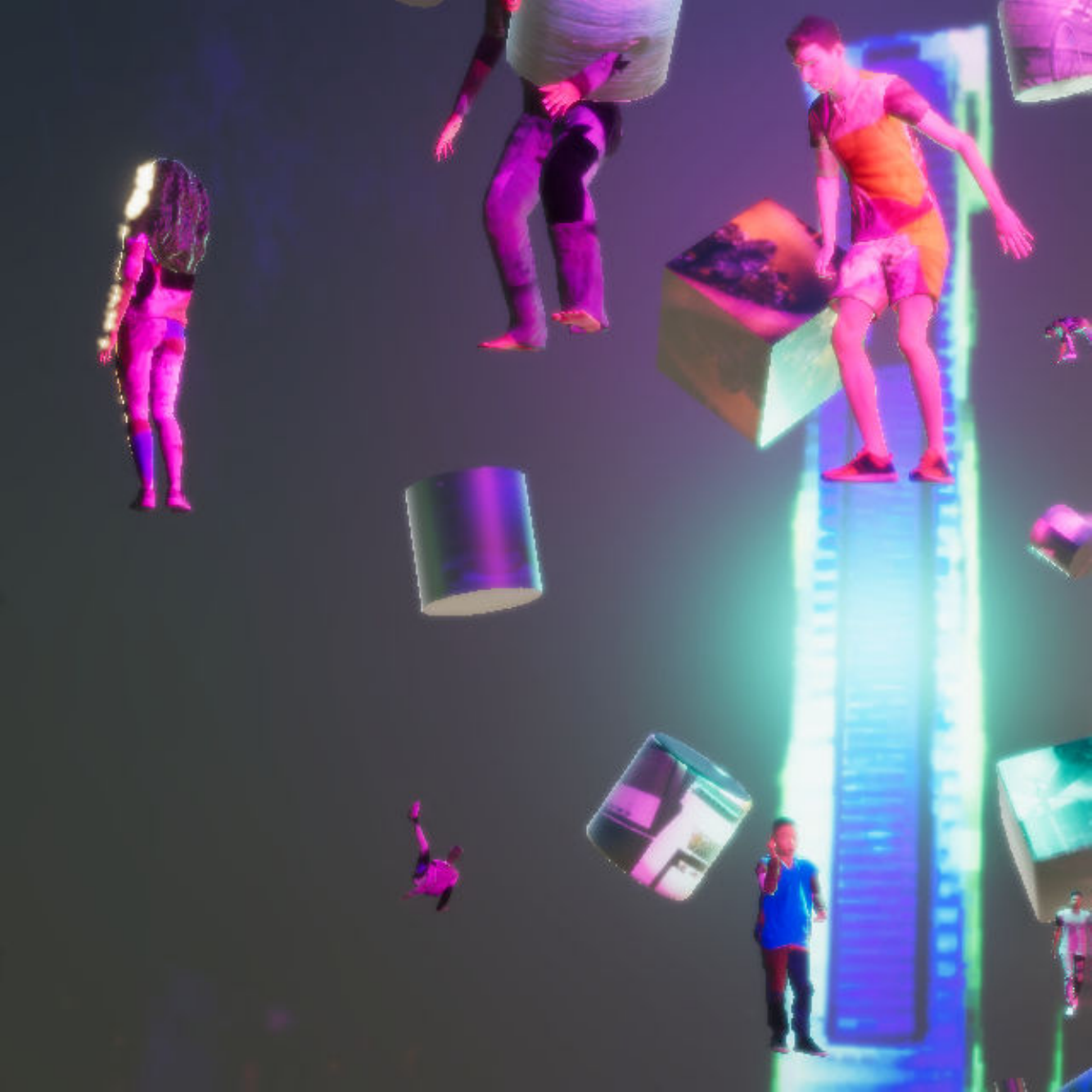}}
    \end{subfigure}
    \hfill \\
    \begin{subfigure}[t]{0.230\textwidth}
        \raisebox{-\height}{\includegraphics[width=\textwidth]{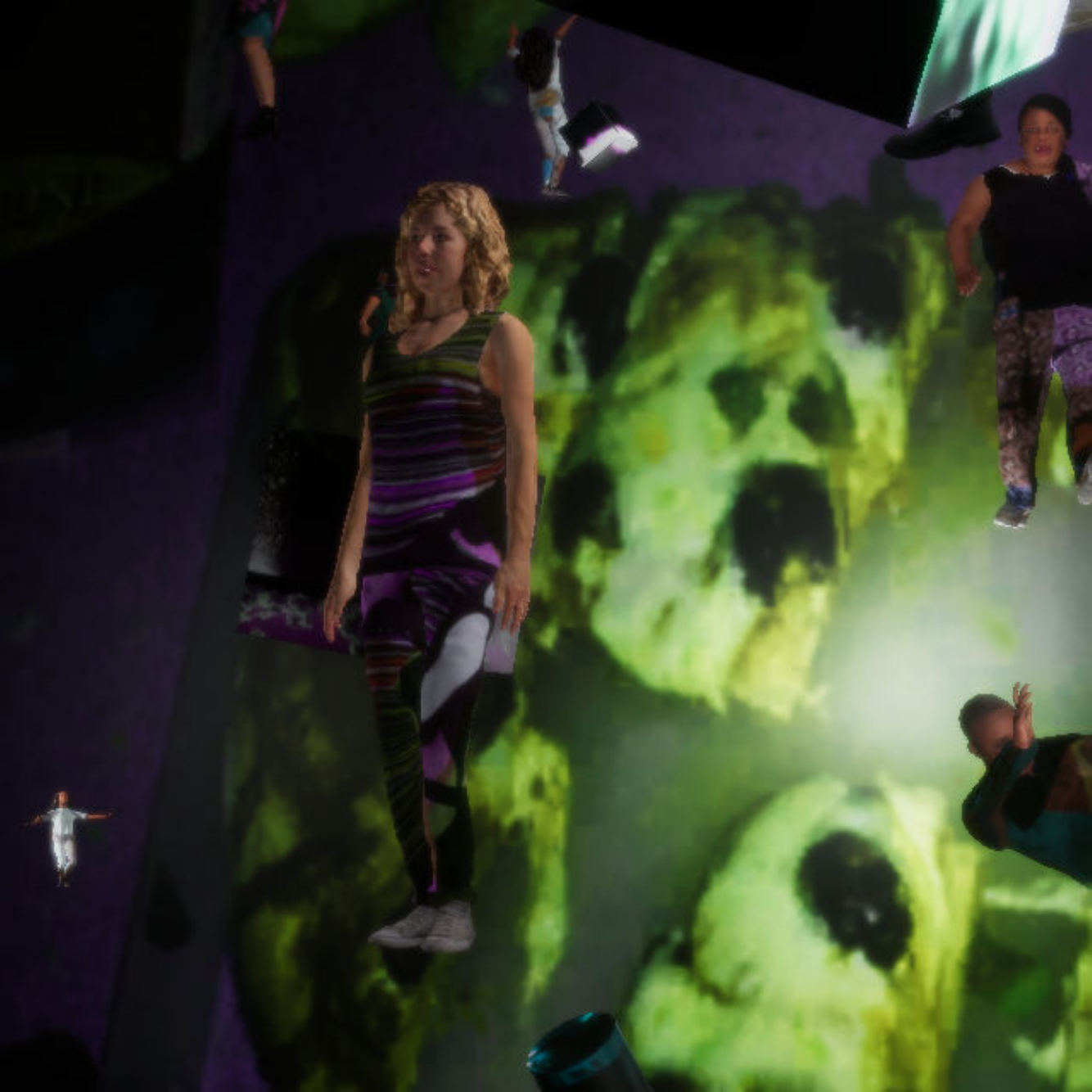}}
    \end{subfigure}
    \begin{subfigure}[t]{0.230\textwidth}
        \raisebox{-\height}{\includegraphics[width=\textwidth]{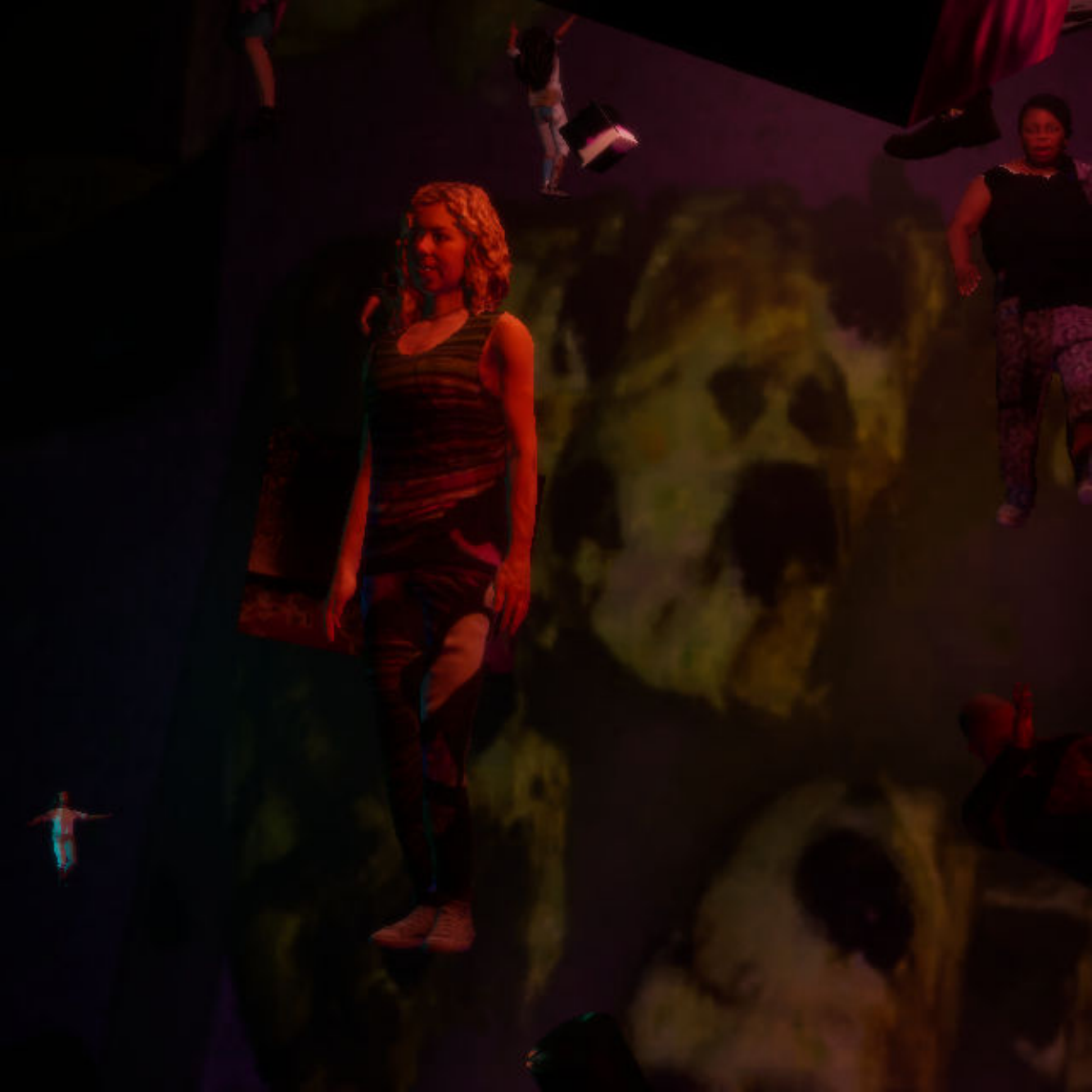}}
    \end{subfigure}
    \begin{subfigure}[t]{0.230\textwidth}
        \raisebox{-\height}{\includegraphics[width=\textwidth]{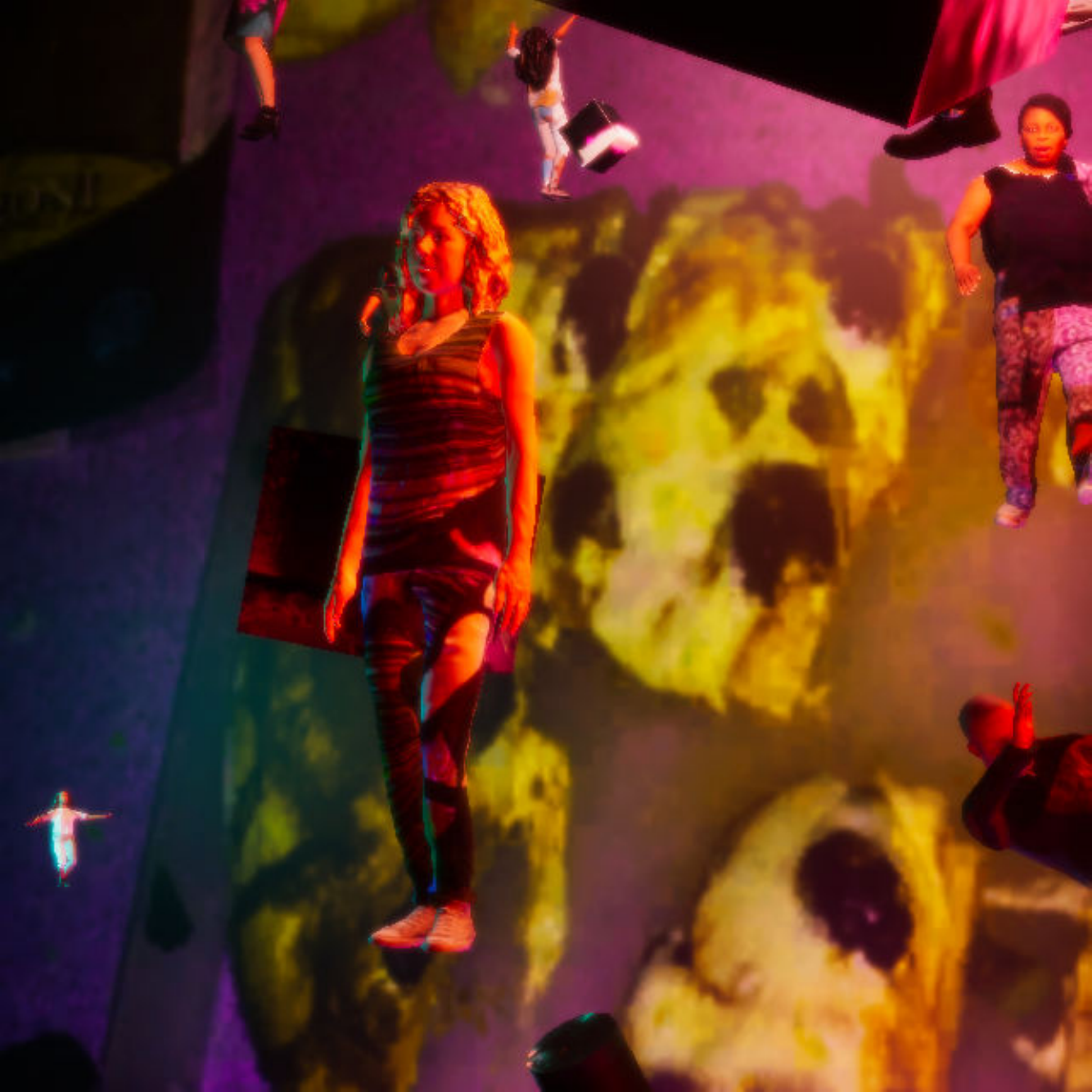}}
    \end{subfigure}
    \hfill \\
    \begin{subfigure}[t]{0.230\textwidth}
        \raisebox{-\height}{\includegraphics[width=\textwidth]{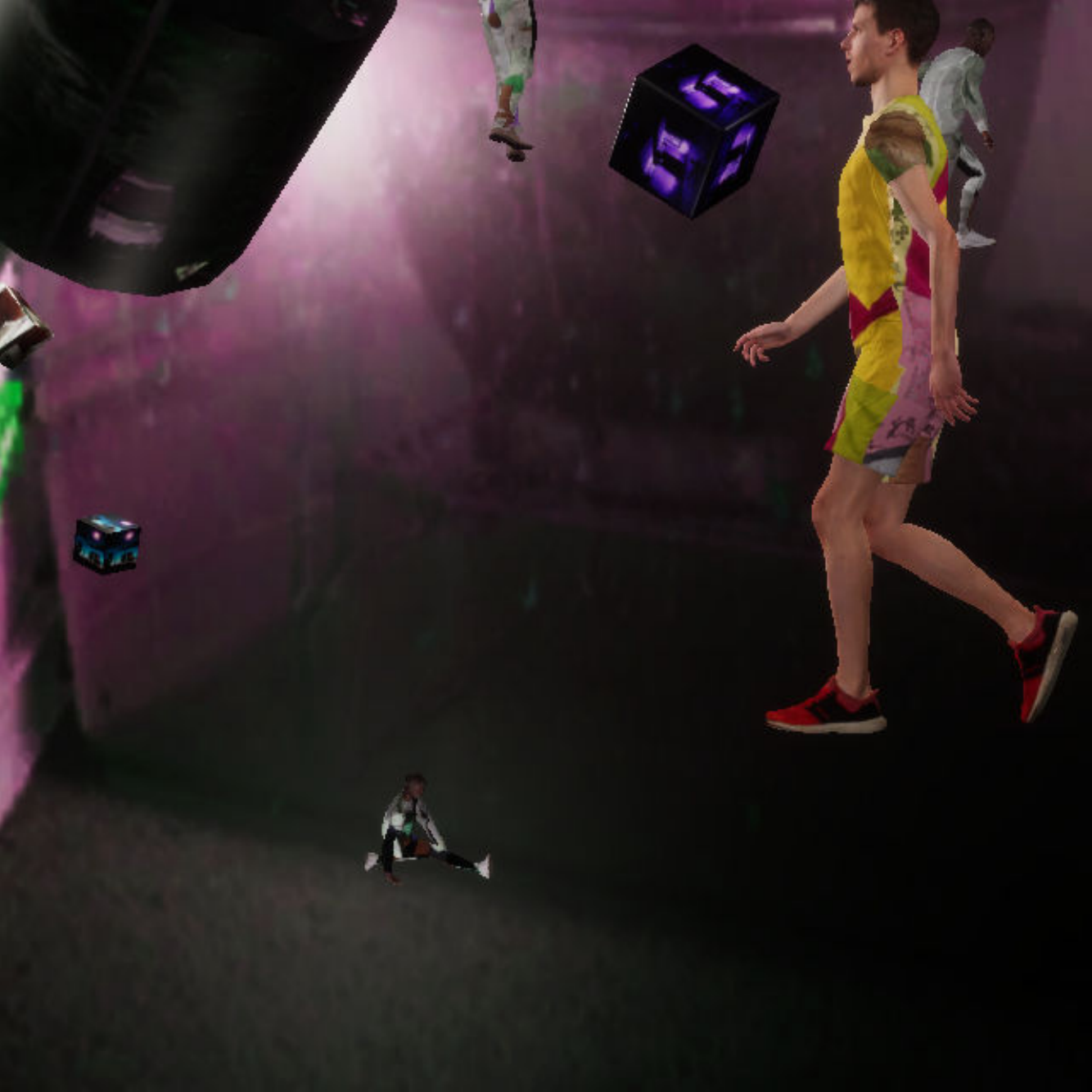}}
    \end{subfigure}
    \begin{subfigure}[t]{0.230\textwidth}
        \raisebox{-\height}{\includegraphics[width=\textwidth]{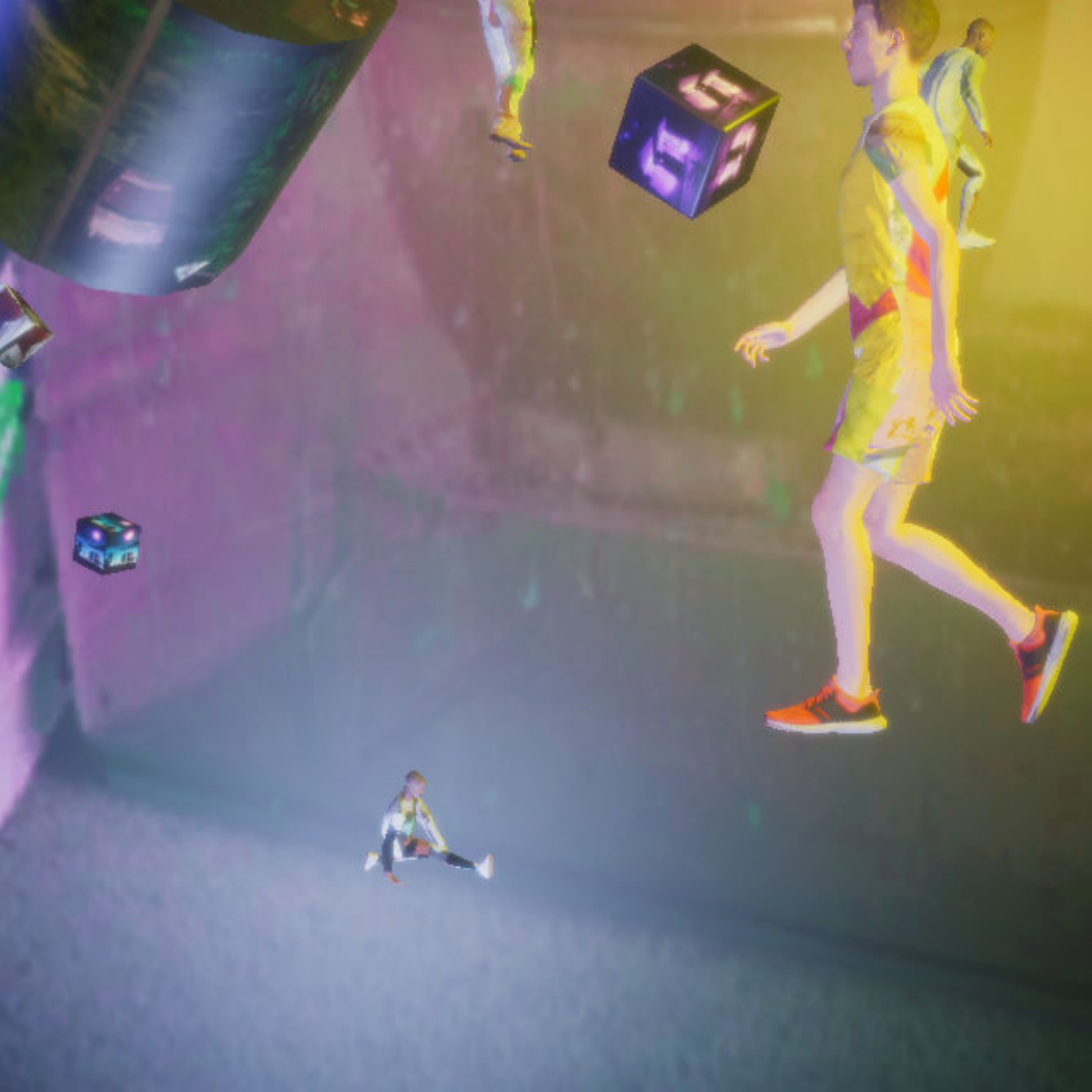}}
    \end{subfigure}
    \begin{subfigure}[t]{0.230\textwidth}
        \raisebox{-\height}{\includegraphics[width=\textwidth]{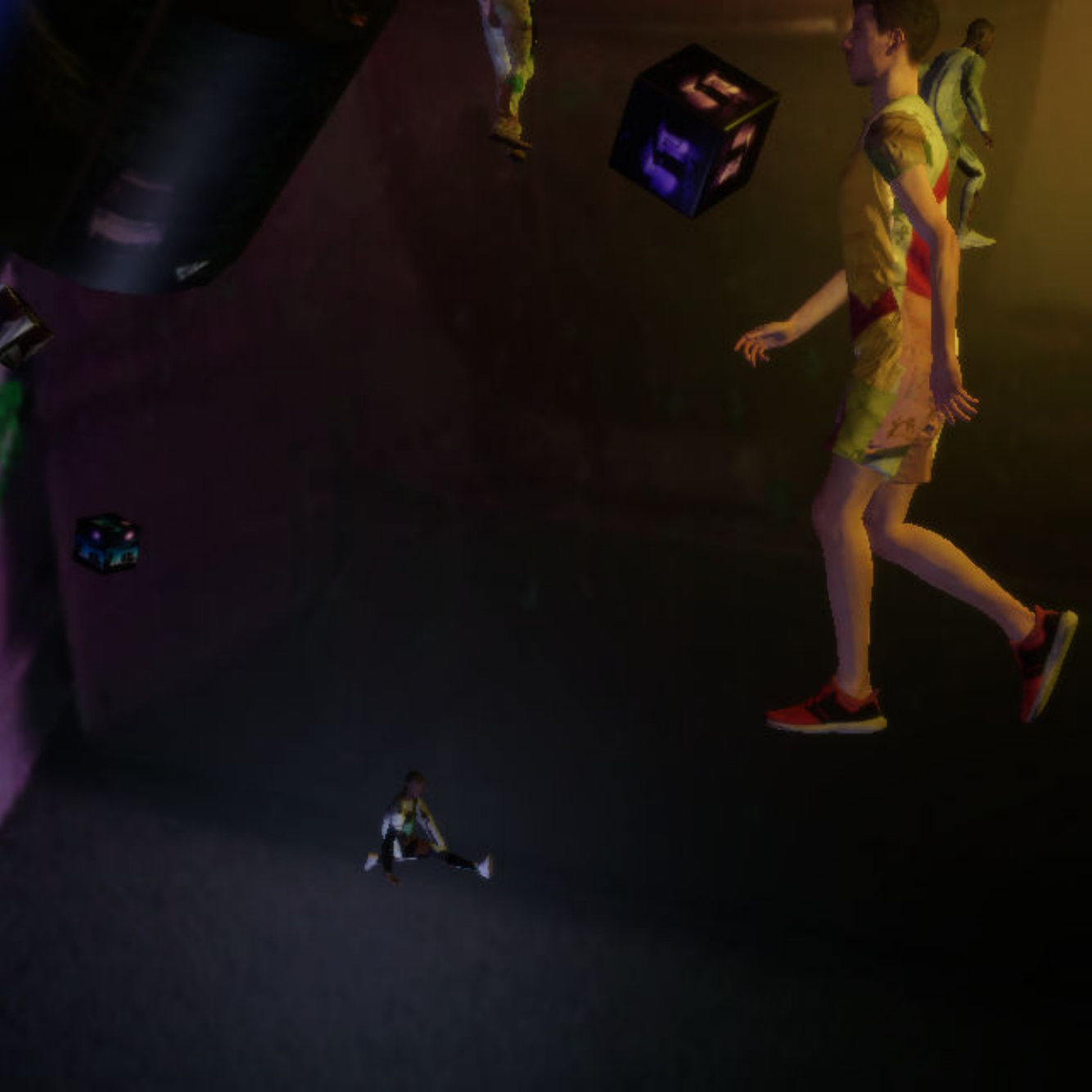}}
    \end{subfigure}
    \hfill \\
    \begin{subfigure}[t]{0.230\textwidth}
        \raisebox{-\height}{\includegraphics[width=\textwidth]{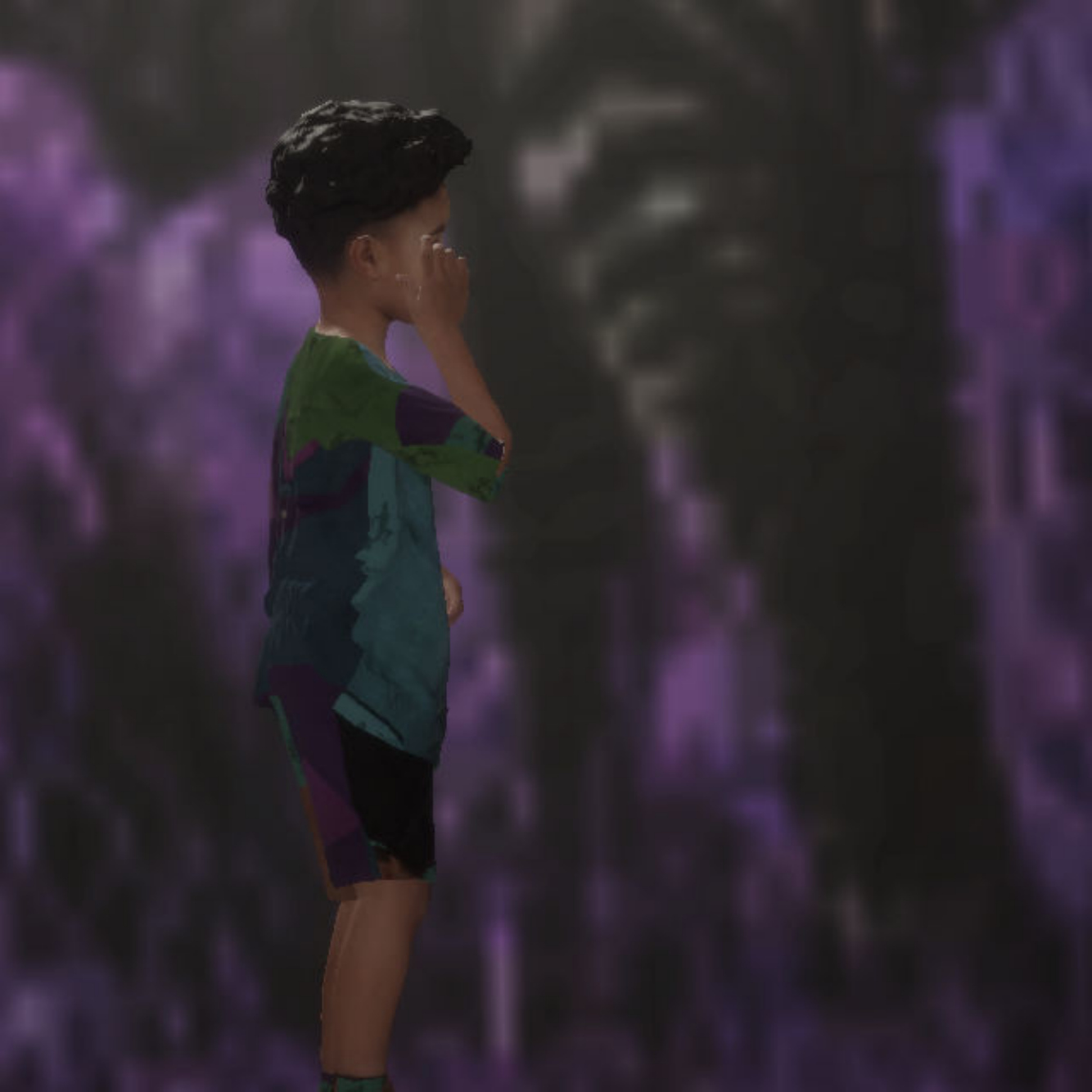}}
    \end{subfigure}
    \begin{subfigure}[t]{0.230\textwidth}
        \raisebox{-\height}{\includegraphics[width=\textwidth]{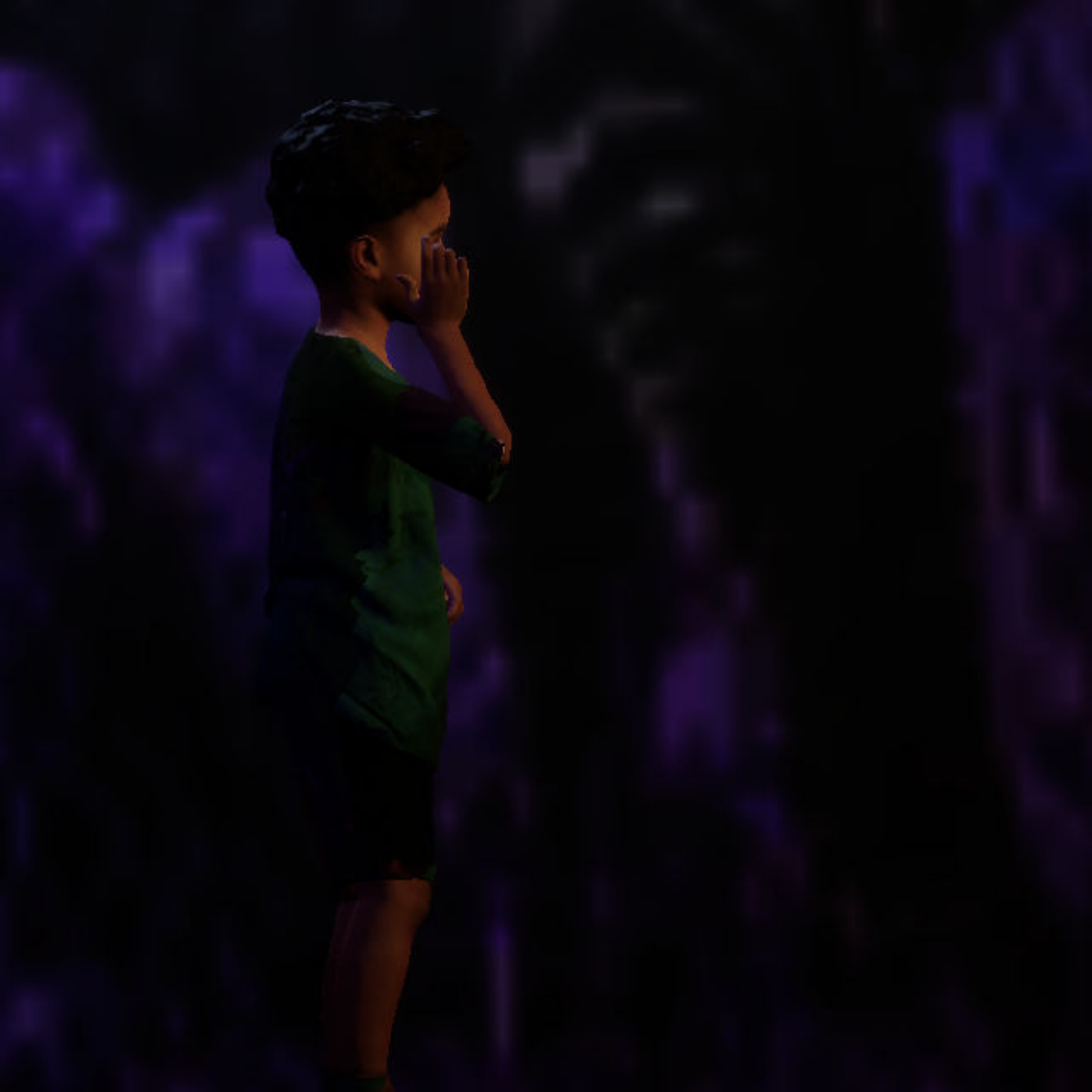}}
    \end{subfigure}
    \begin{subfigure}[t]{0.230\textwidth}
        \raisebox{-\height}{\includegraphics[width=\textwidth]{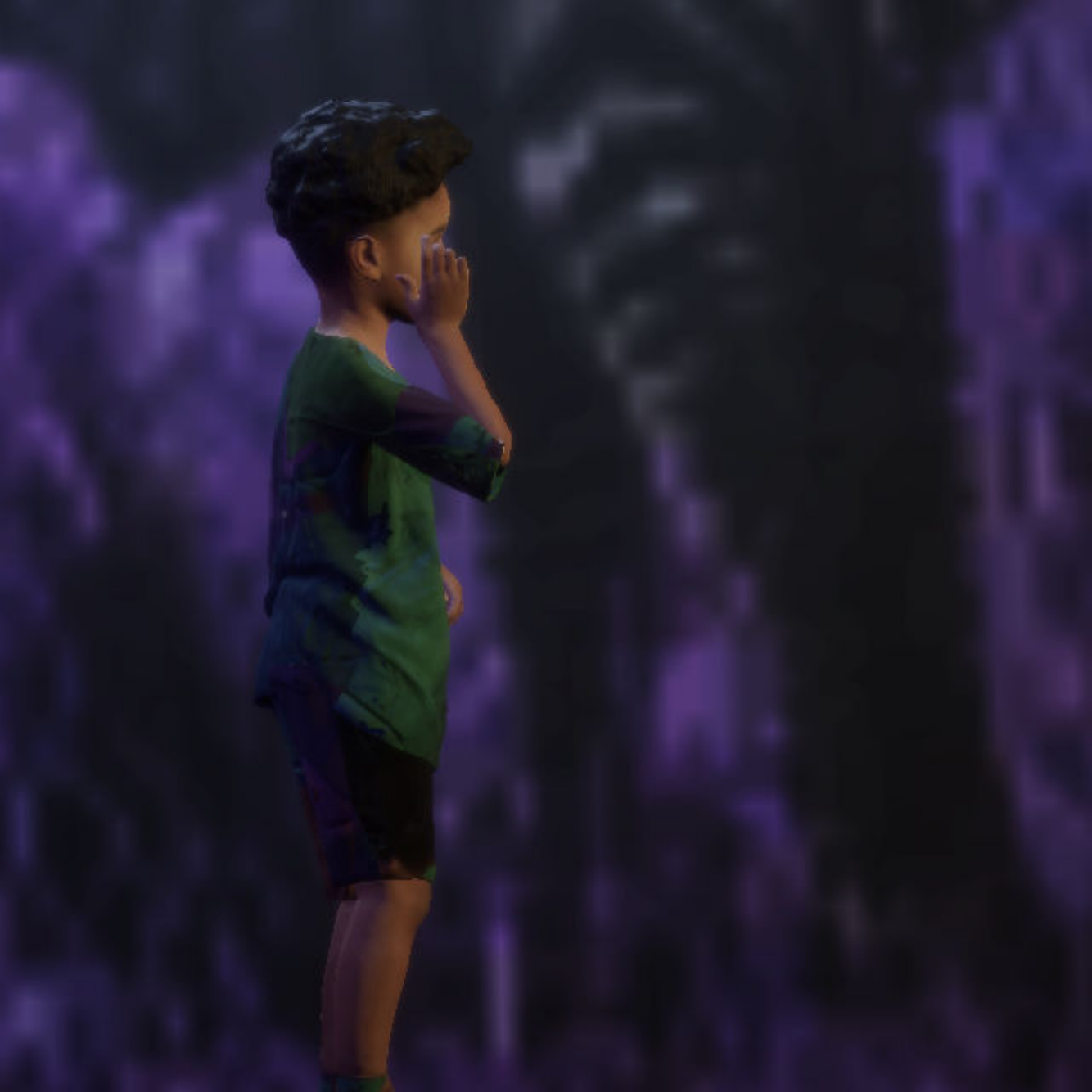}}
    \end{subfigure}
    \hfill \\   
    \begin{subfigure}[t]{0.230\textwidth}
        \raisebox{-\height}{\includegraphics[width=\textwidth]{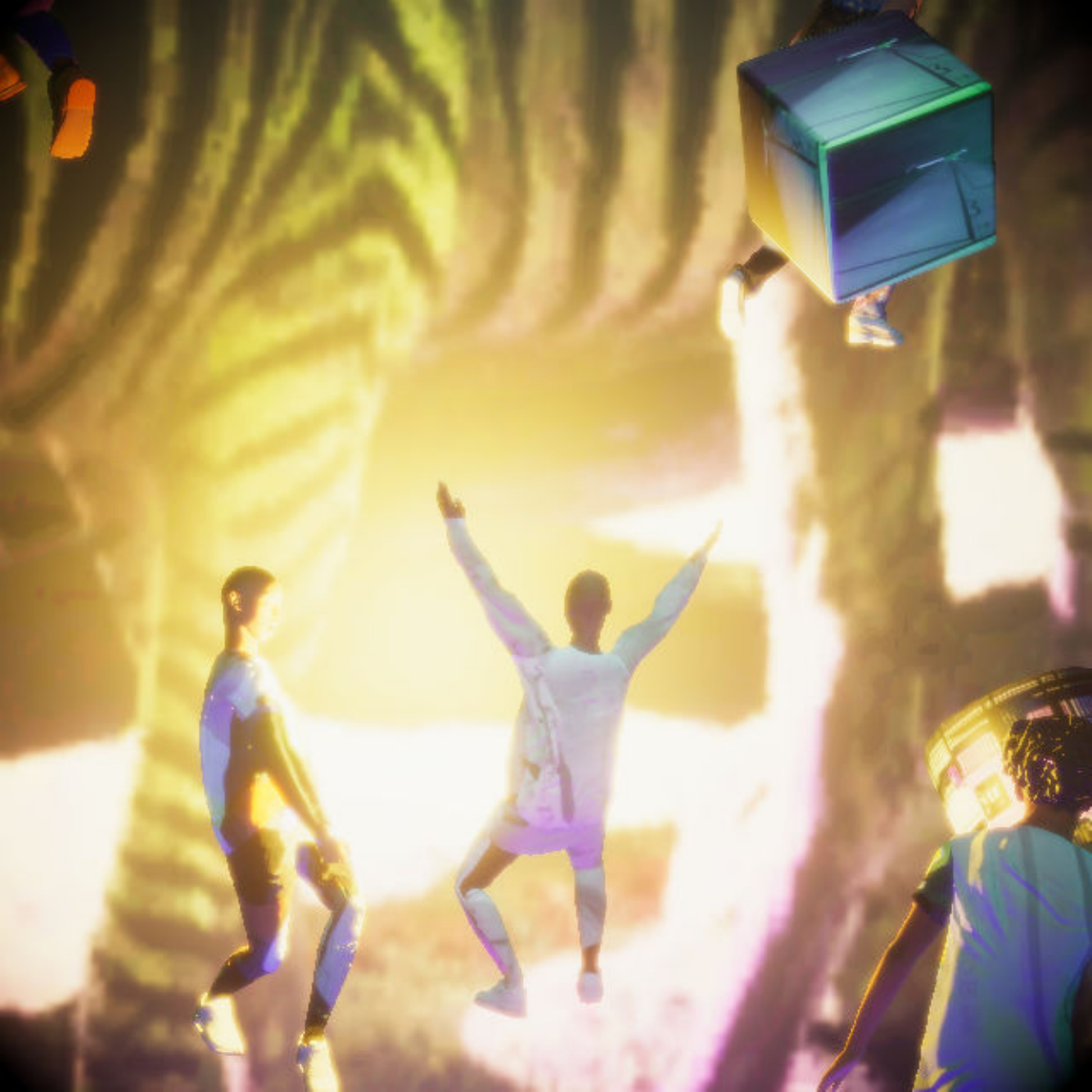}}
    \end{subfigure}
    \begin{subfigure}[t]{0.230\textwidth}
        \raisebox{-\height}{\includegraphics[width=\textwidth]{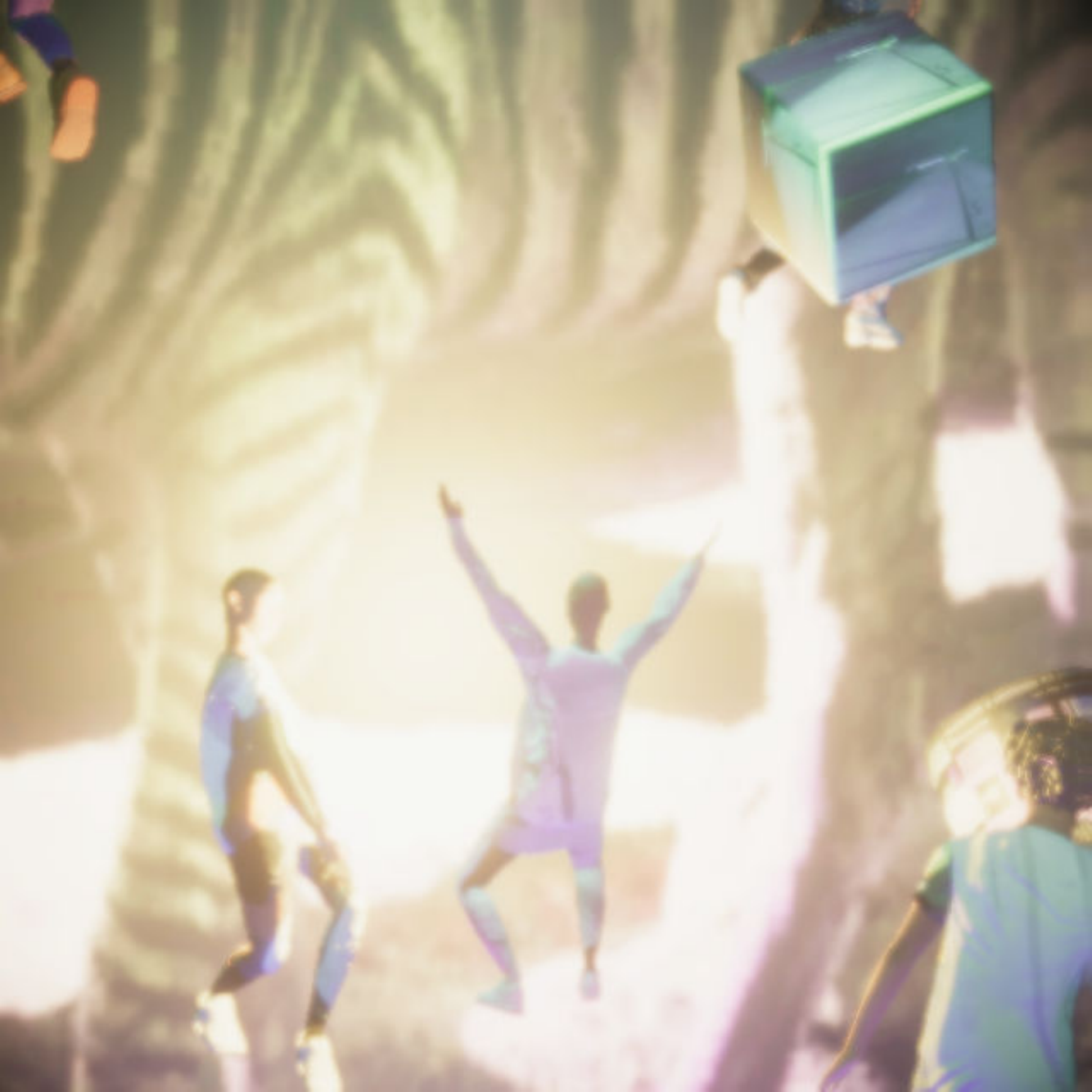}}
    \end{subfigure}
    \begin{subfigure}[t]{0.230\textwidth}
        \raisebox{-\height}{\includegraphics[width=\textwidth]{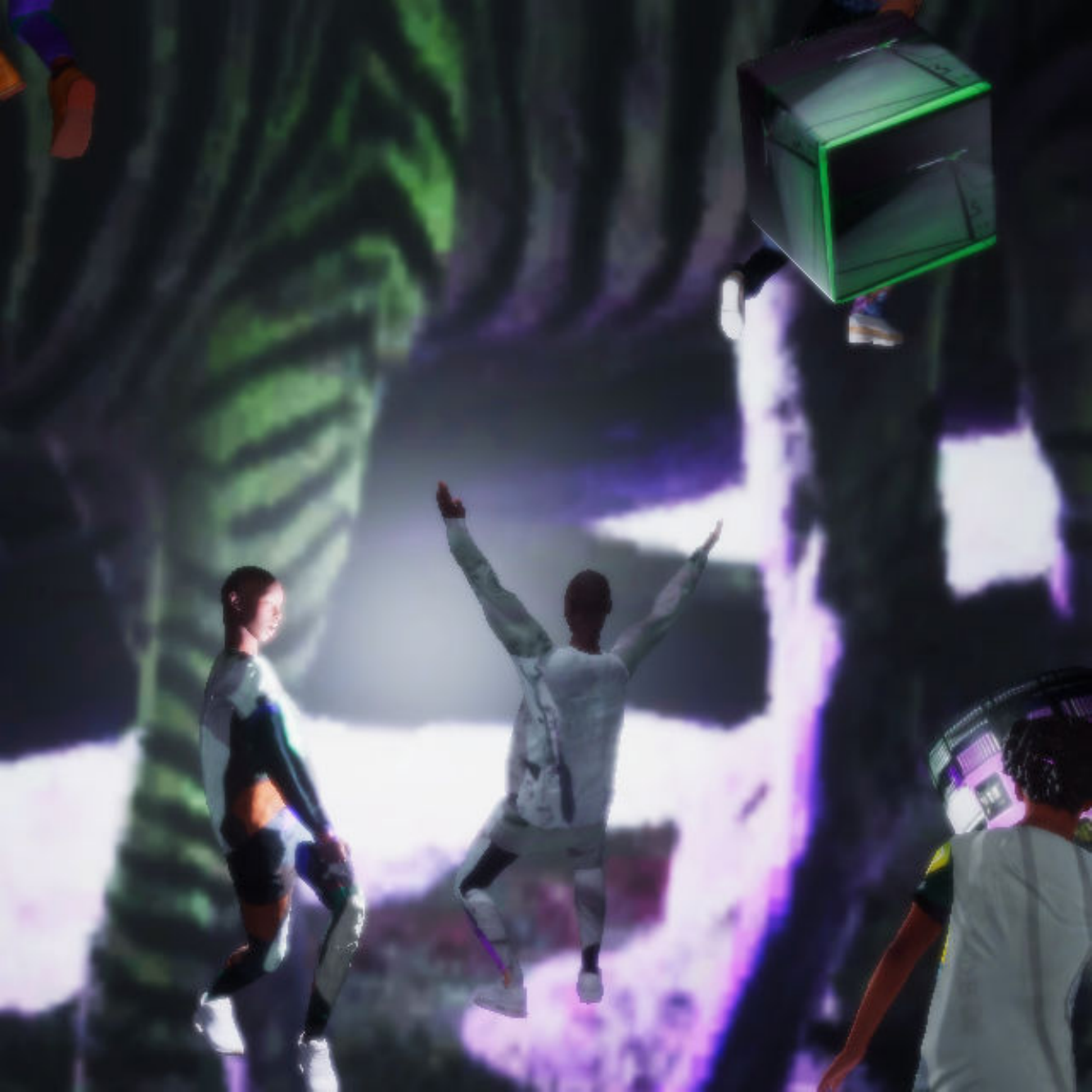}}
    \end{subfigure}
\caption{Examples of light randomization in the same scene 1/2. Each row shows three different lighting conditions while the rest of the scene is unchanged.}
\label{fig:fig:morelight1}%
\end{figure}

\begin{figure}[htb]
    \centering
    \begin{subfigure}[t]{0.230\textwidth}
        \raisebox{-\height}{\includegraphics[width=\textwidth]{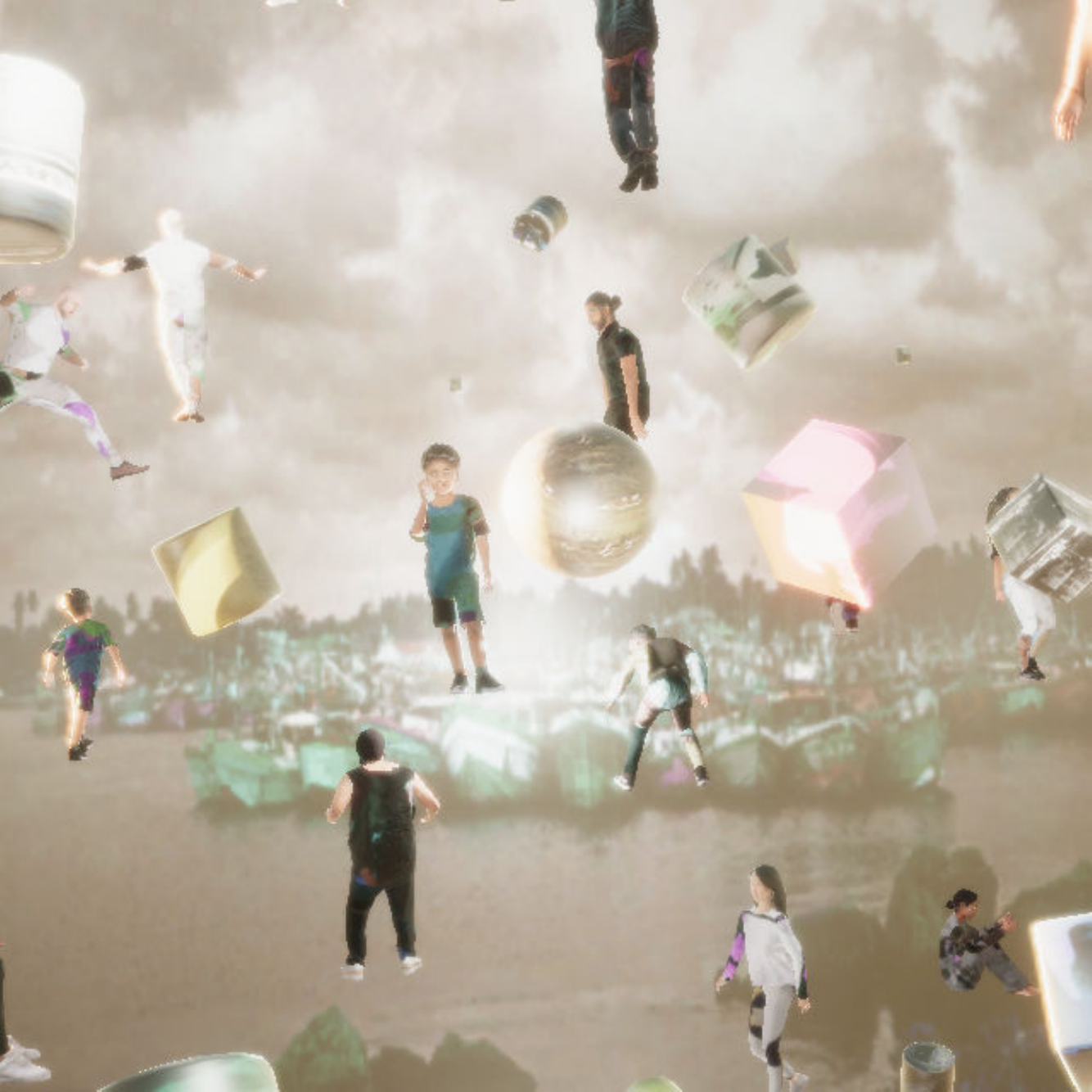}}
    \end{subfigure}
    \begin{subfigure}[t]{0.230\textwidth}
        \raisebox{-\height}{\includegraphics[width=\textwidth]{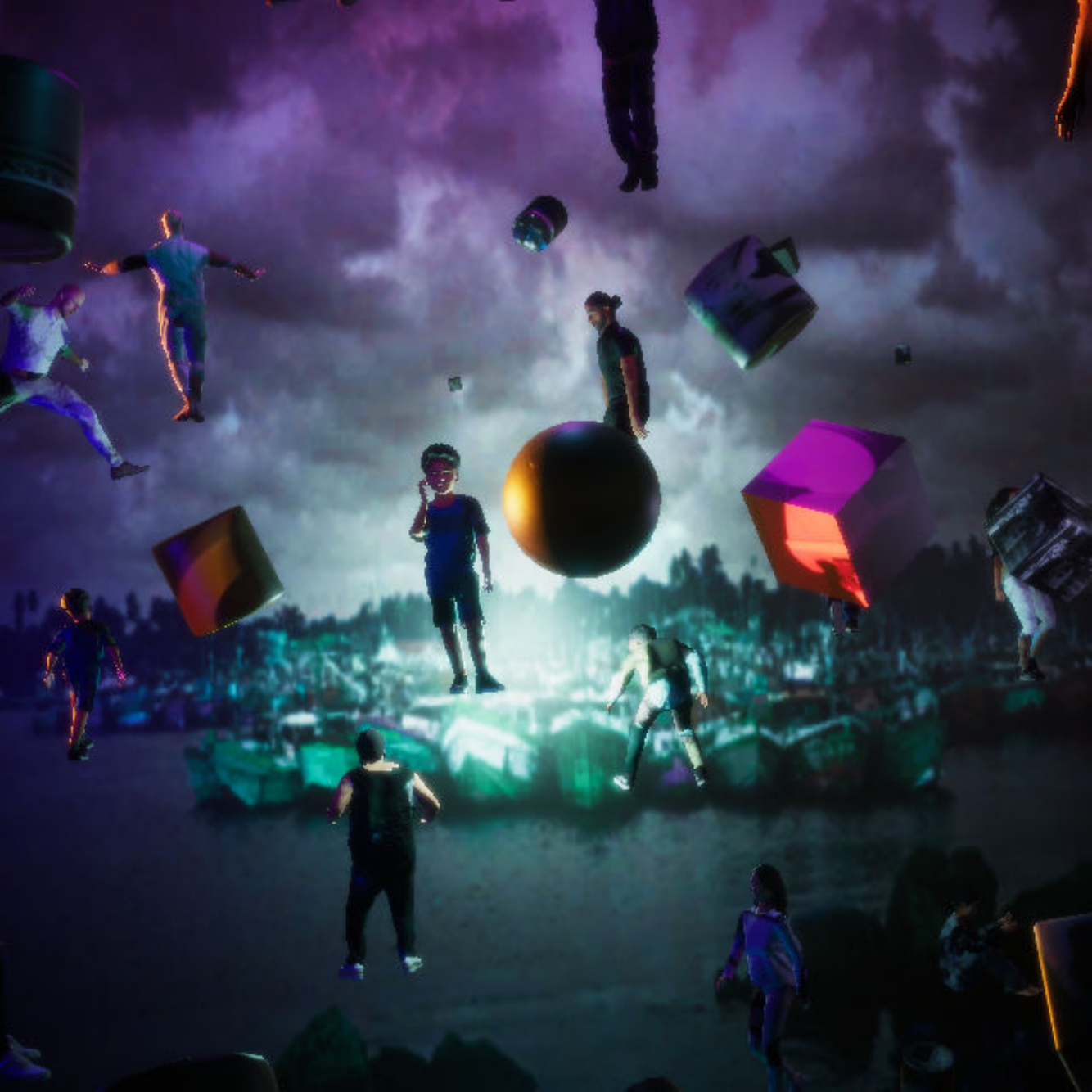}}
    \end{subfigure}
    \begin{subfigure}[t]{0.230\textwidth}
        \raisebox{-\height}{\includegraphics[width=\textwidth]{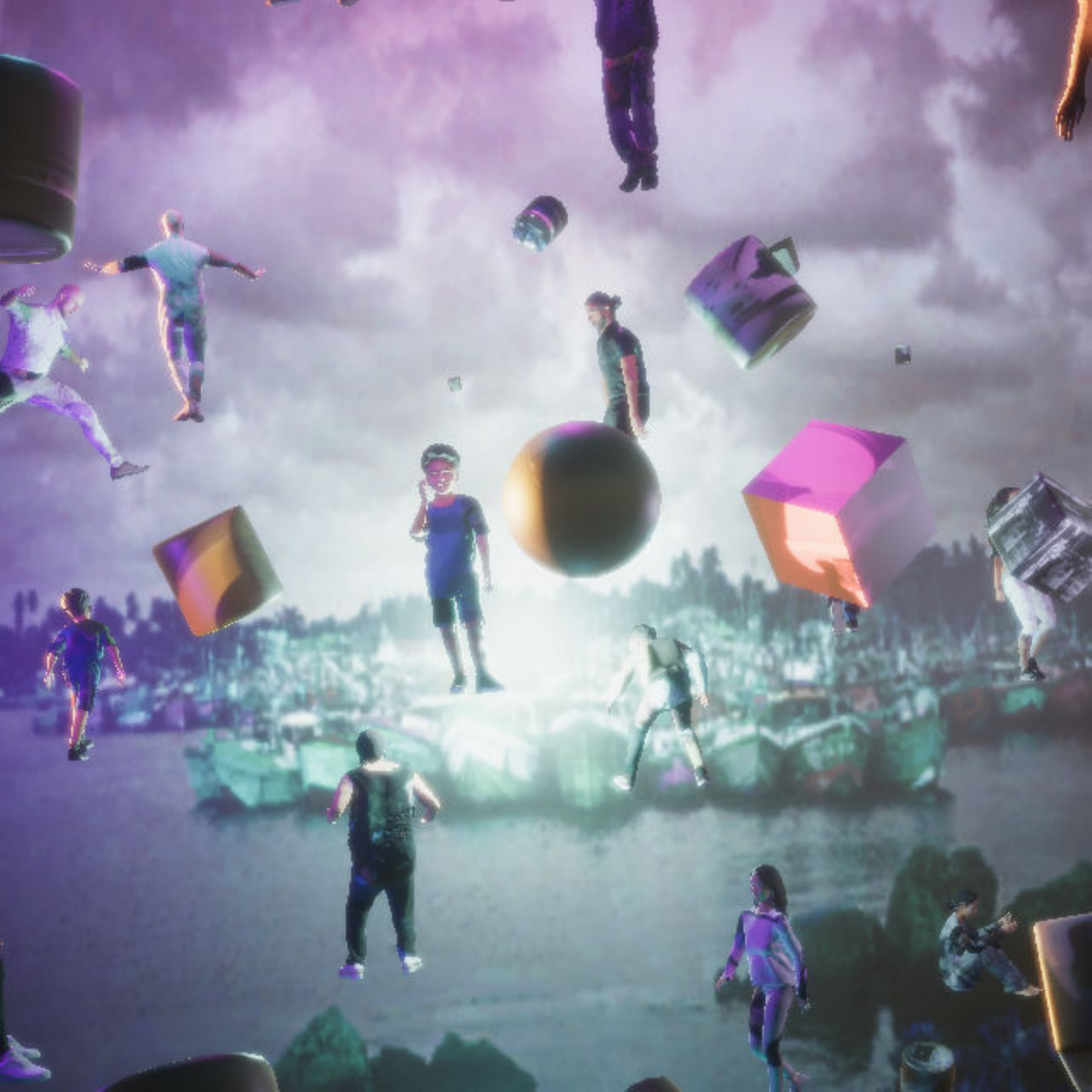}}
    \end{subfigure}
    \hfill \\
    \begin{subfigure}[t]{0.230\textwidth}
        \raisebox{-\height}{\includegraphics[width=\textwidth]{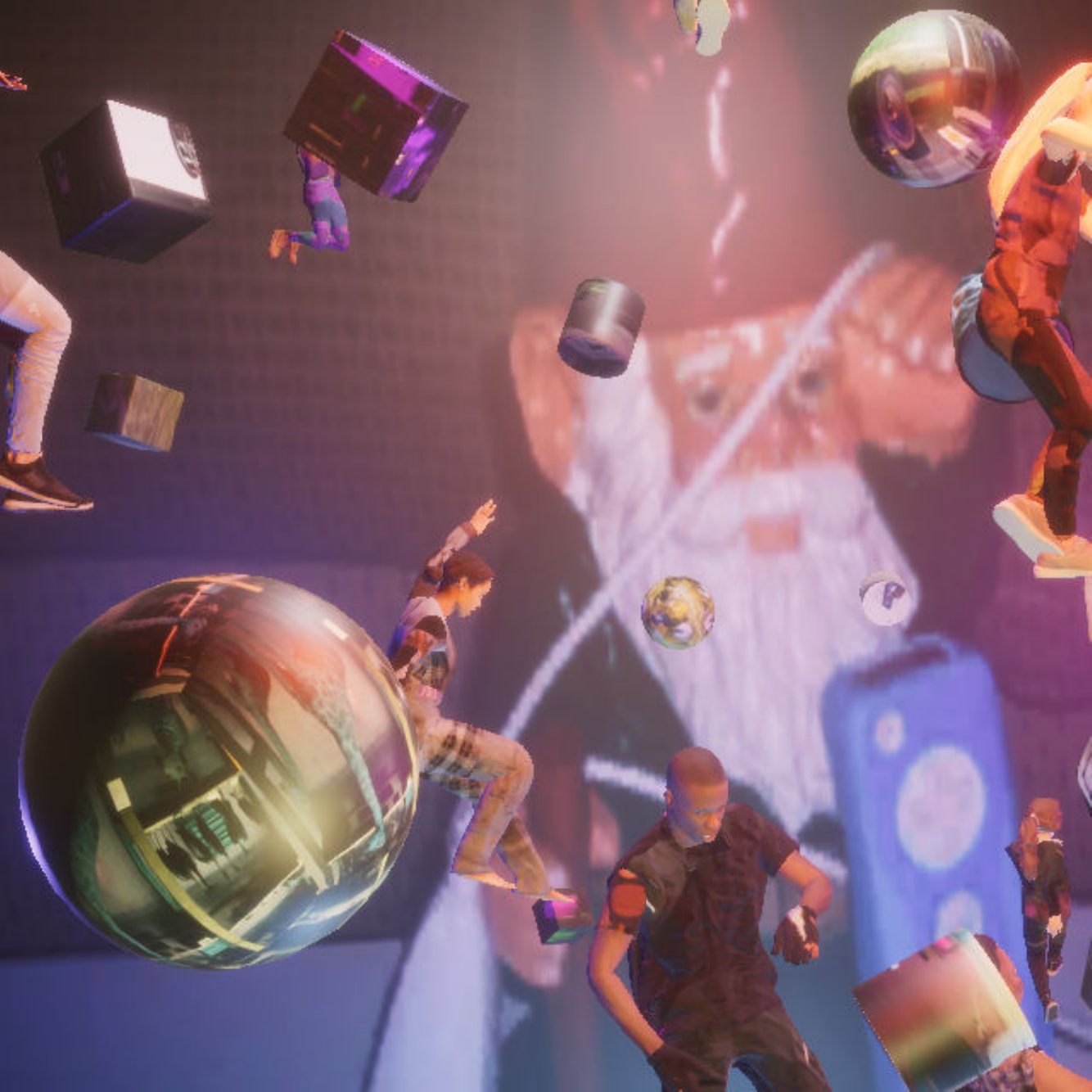}}
    \end{subfigure}
    \begin{subfigure}[t]{0.230\textwidth}
        \raisebox{-\height}{\includegraphics[width=\textwidth]{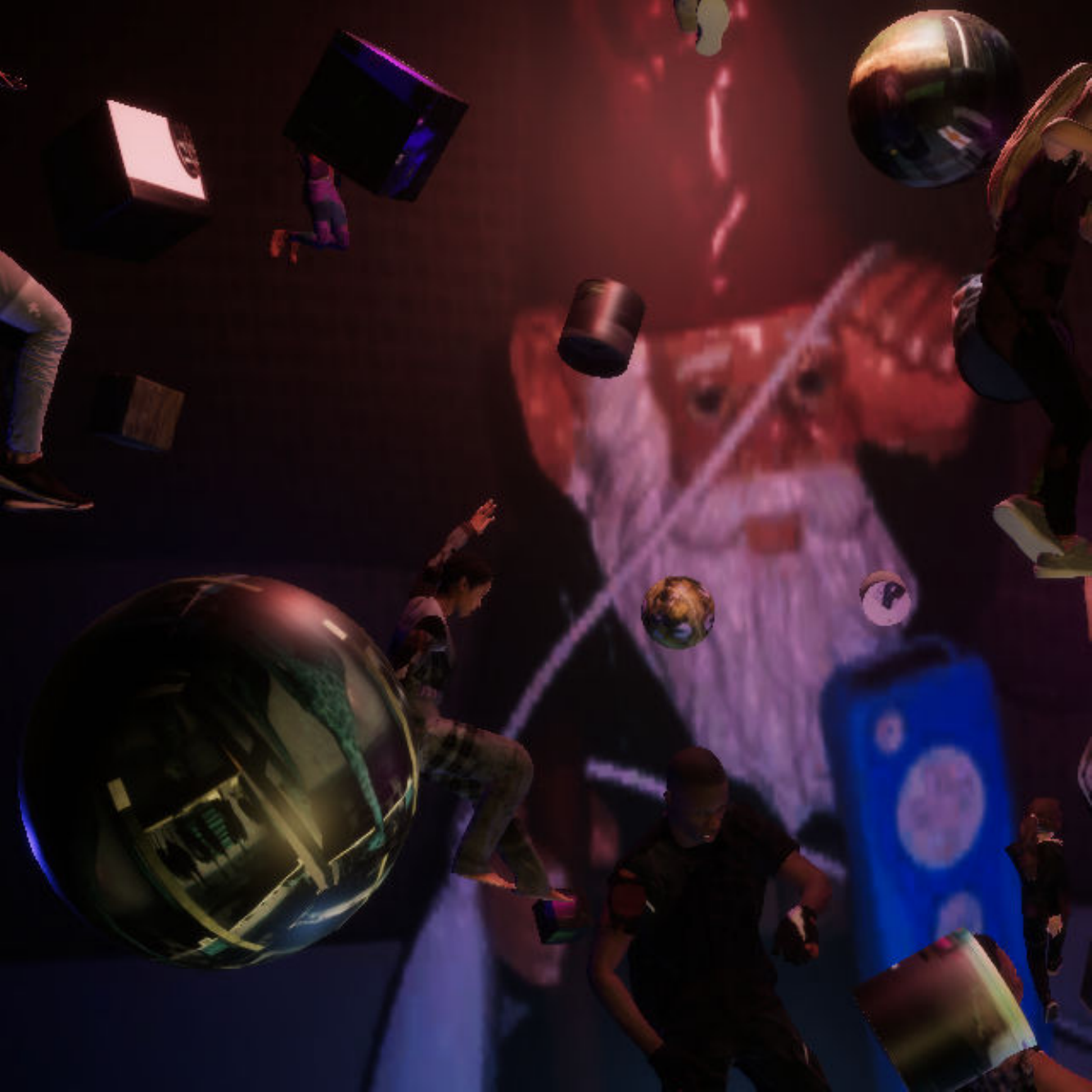}}
    \end{subfigure}
    \begin{subfigure}[t]{0.230\textwidth}
        \raisebox{-\height}{\includegraphics[width=\textwidth]{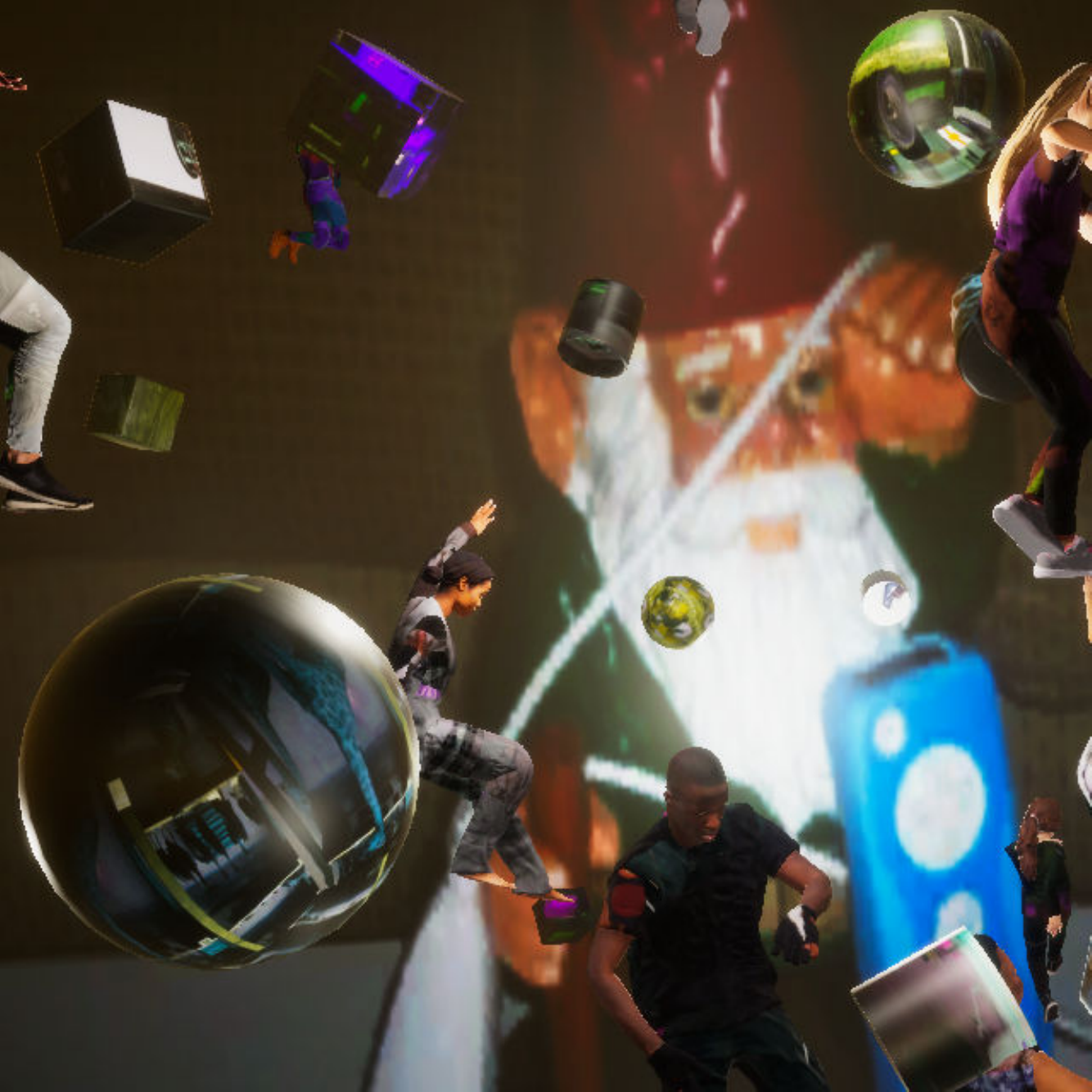}}
    \end{subfigure}
    \hfill \\
    \begin{subfigure}[t]{0.230\textwidth}
        \raisebox{-\height}{\includegraphics[width=\textwidth]{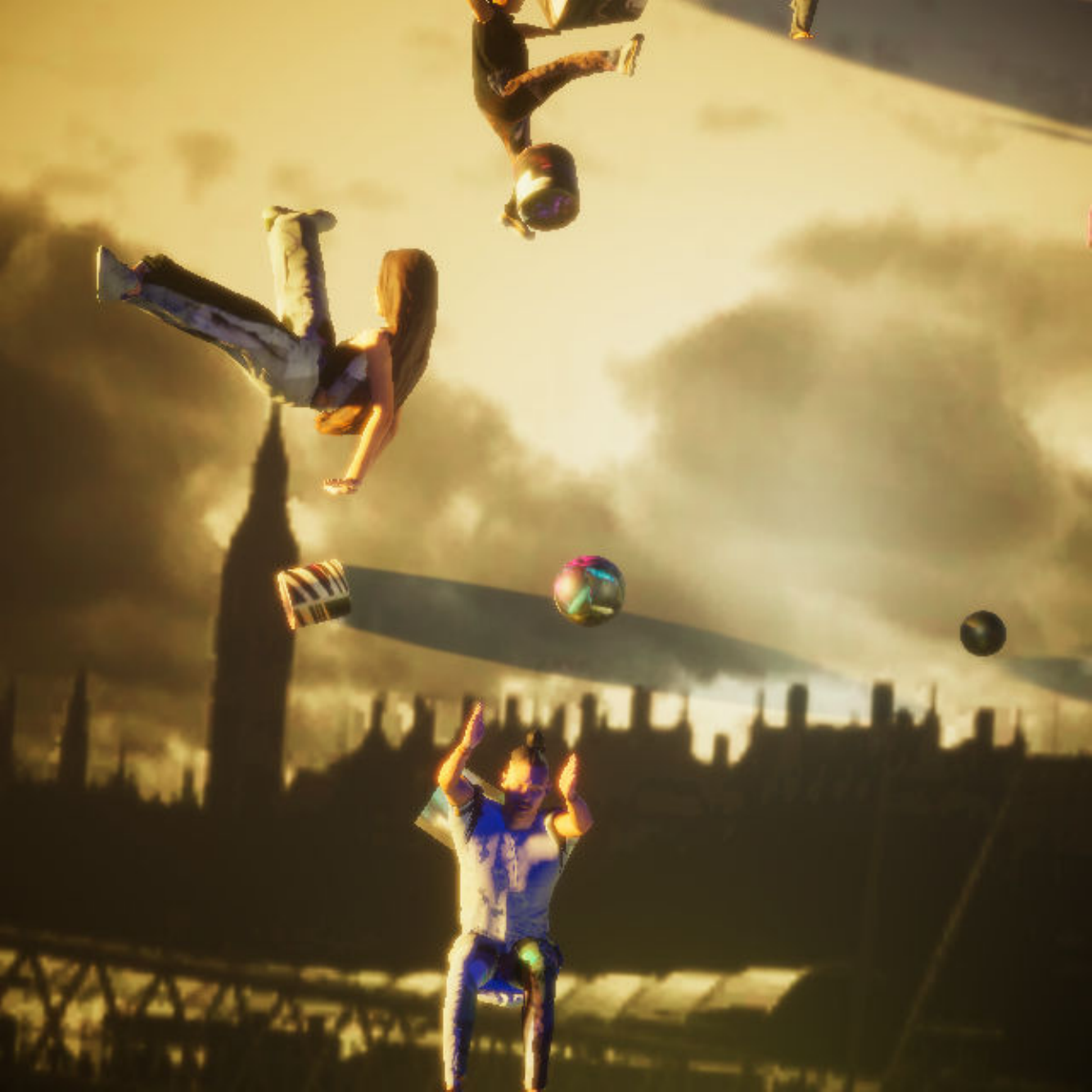}}
    \end{subfigure}
    \begin{subfigure}[t]{0.230\textwidth}
        \raisebox{-\height}{\includegraphics[width=\textwidth]{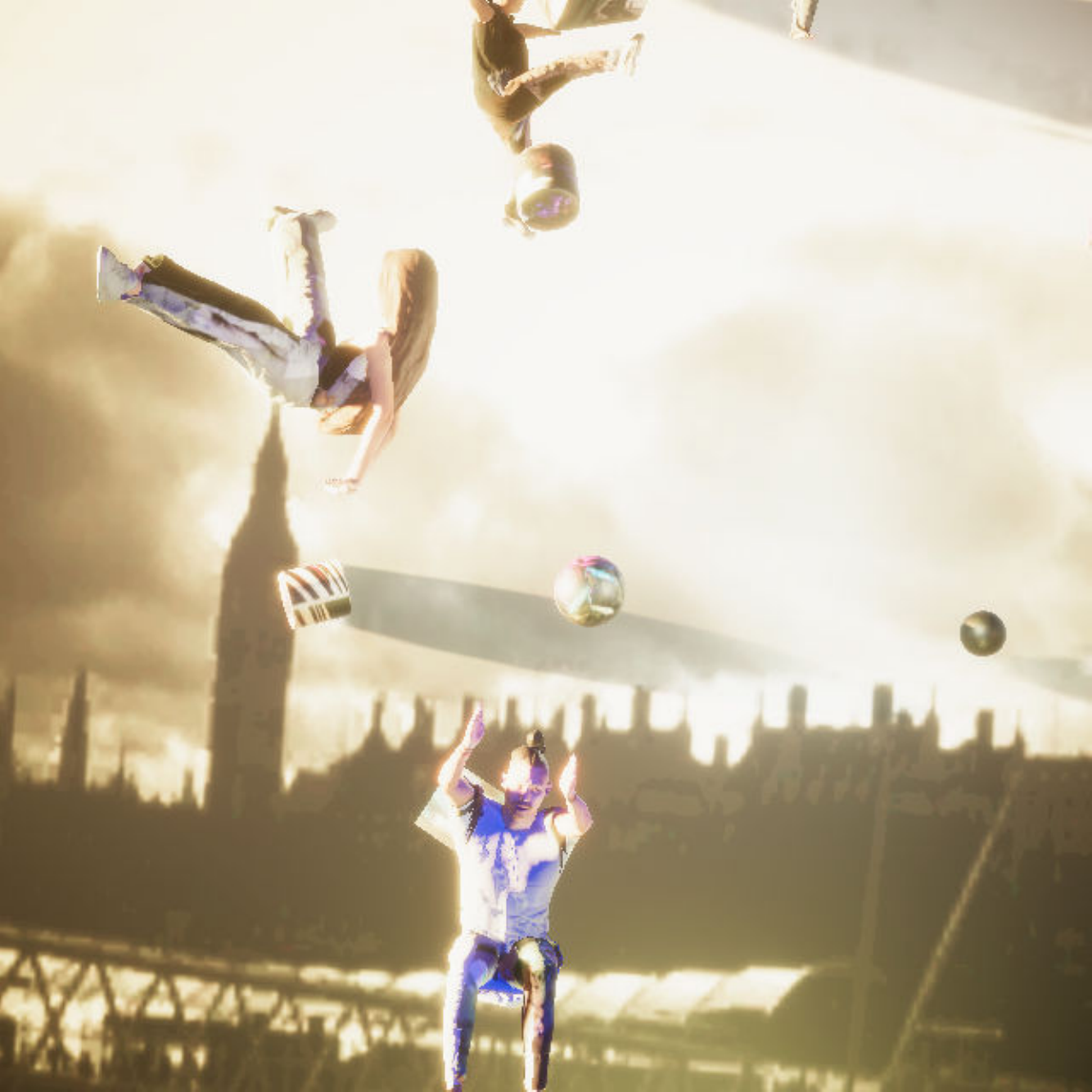}}
    \end{subfigure}
    \begin{subfigure}[t]{0.230\textwidth}
        \raisebox{-\height}{\includegraphics[width=\textwidth]{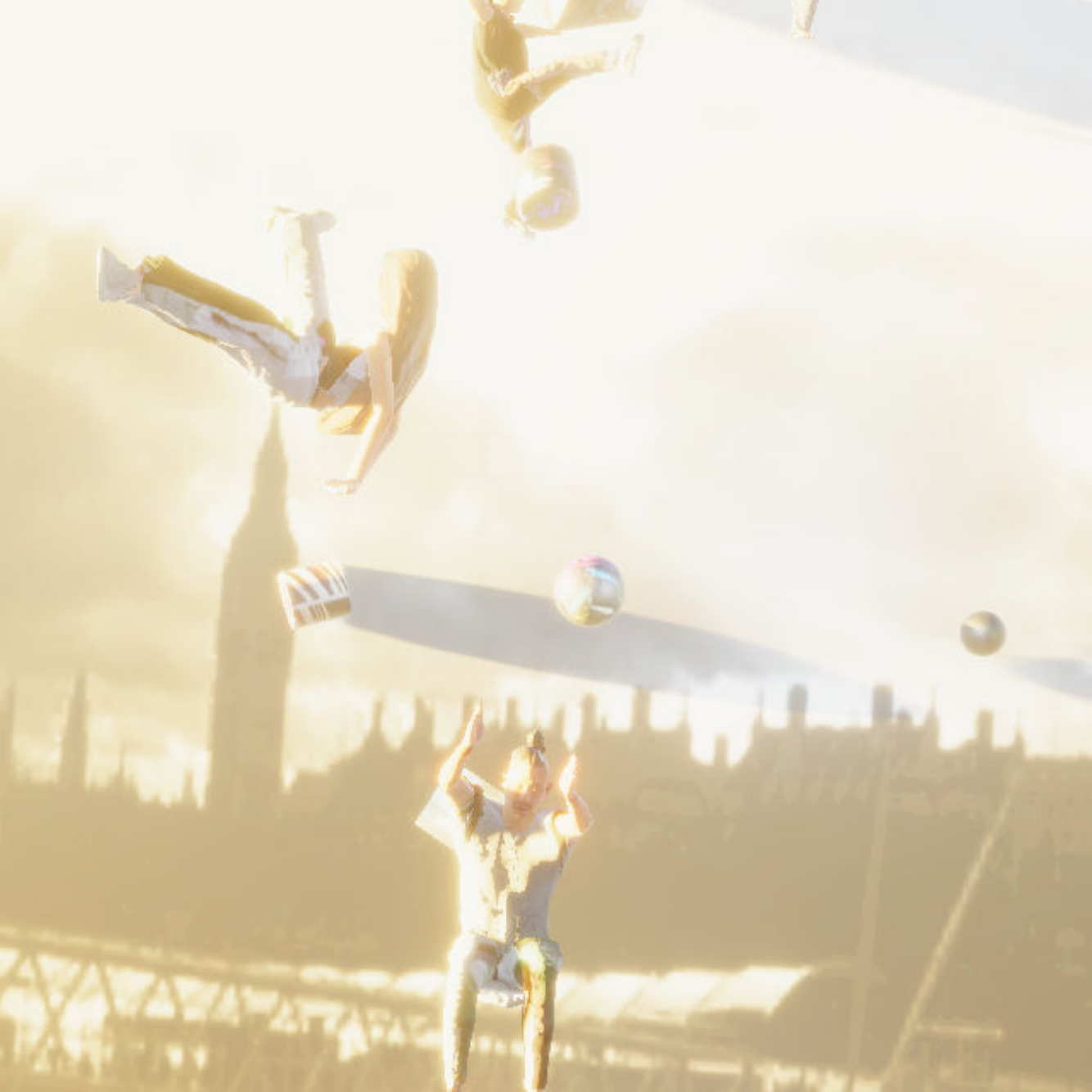}}
    \end{subfigure}
    \hfill \\
    \begin{subfigure}[t]{0.230\textwidth}
        \raisebox{-\height}{\includegraphics[width=\textwidth]{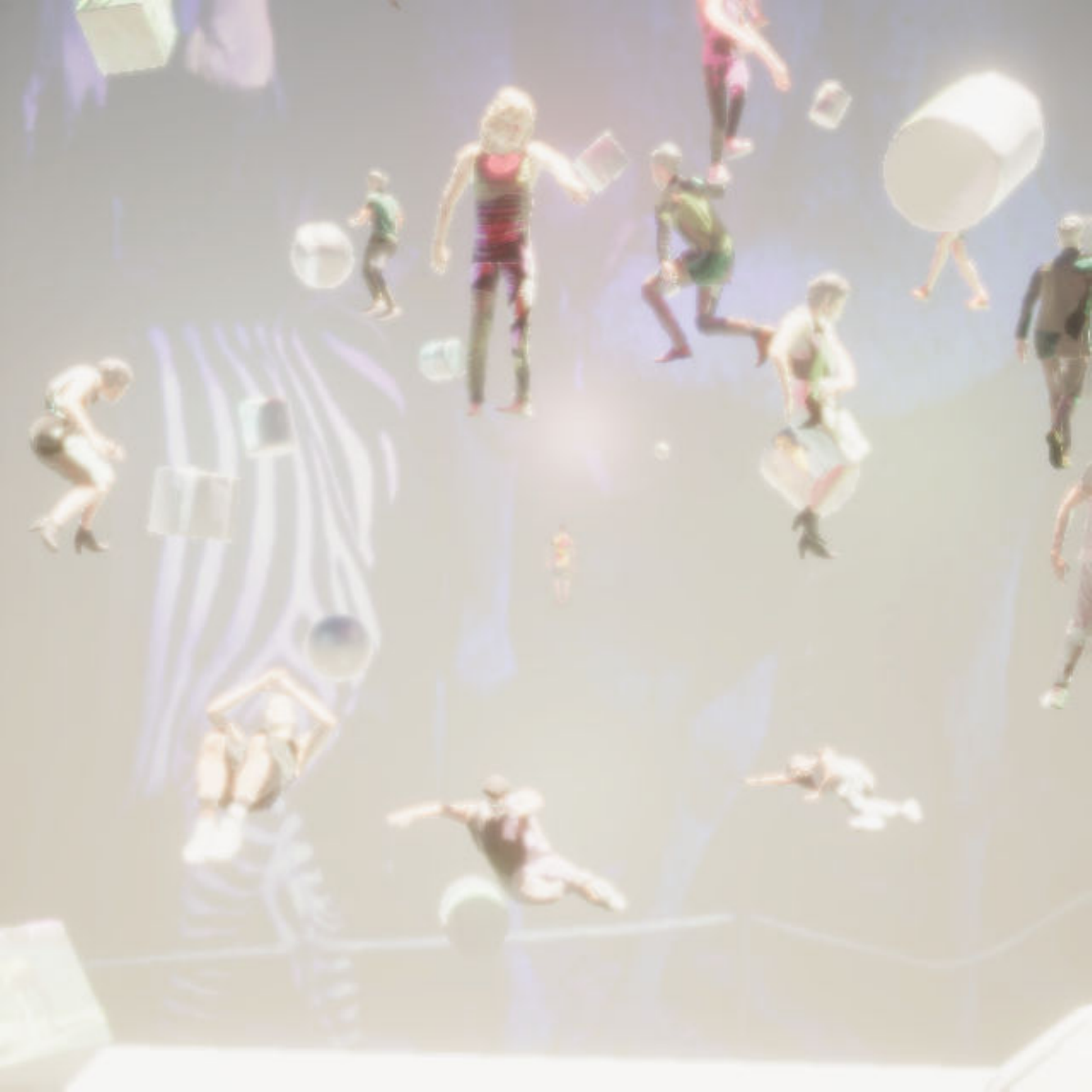}}
    \end{subfigure}
    \begin{subfigure}[t]{0.230\textwidth}
        \raisebox{-\height}{\includegraphics[width=\textwidth]{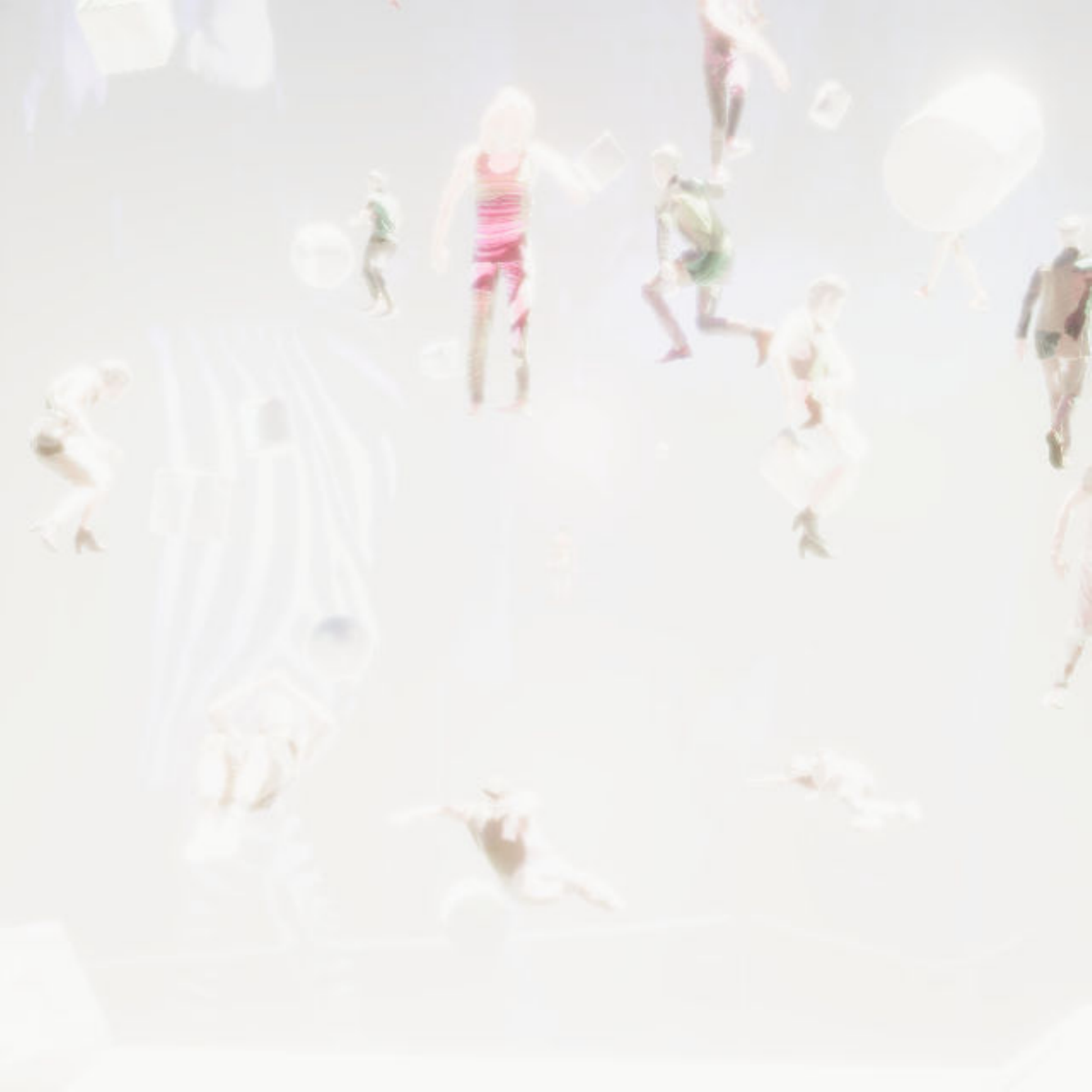}}
    \end{subfigure}
    \begin{subfigure}[t]{0.230\textwidth}
        \raisebox{-\height}{\includegraphics[width=\textwidth]{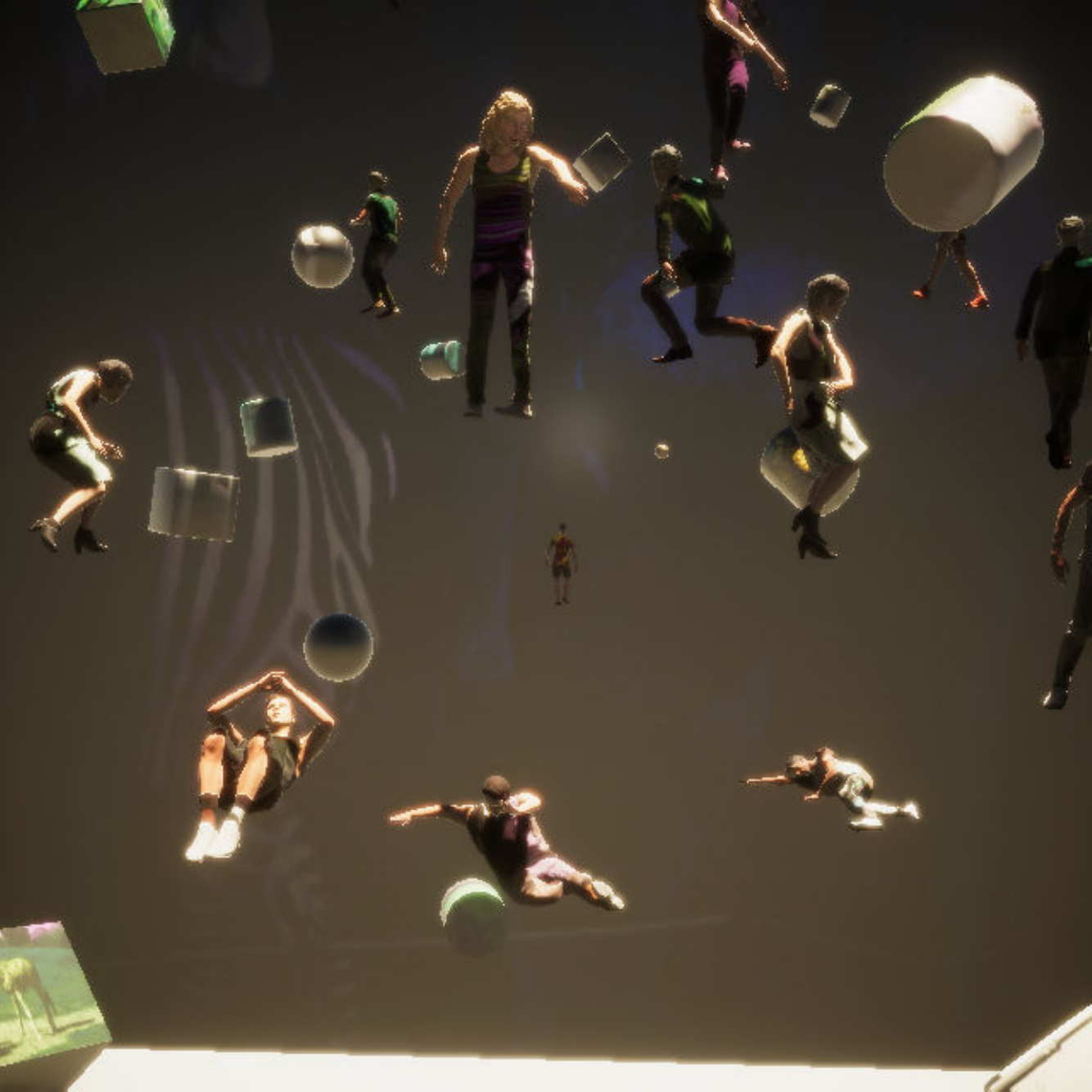}}
    \end{subfigure}
    \hfill \\
    \begin{subfigure}[t]{0.230\textwidth}
        \raisebox{-\height}{\includegraphics[width=\textwidth]{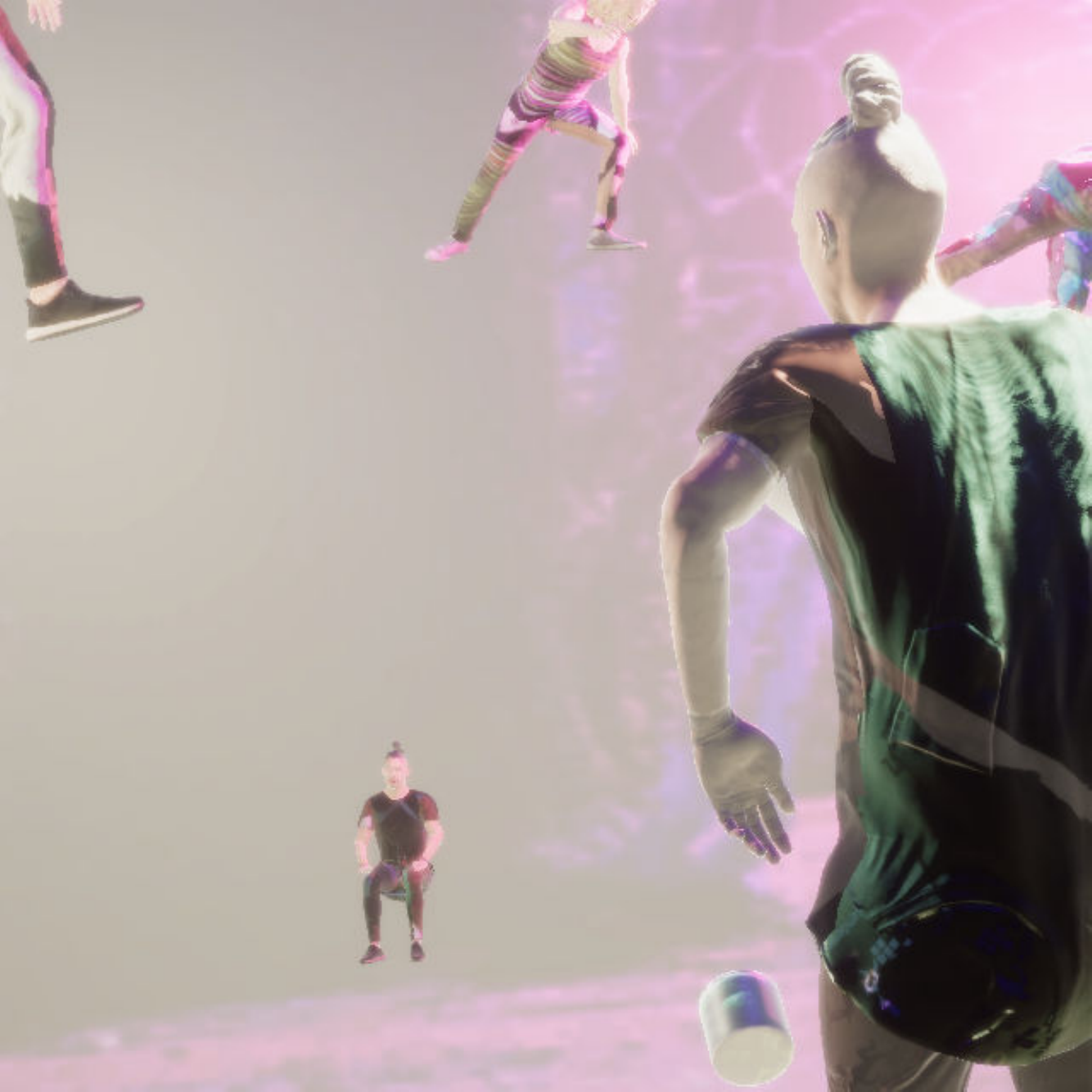}}
    \end{subfigure}
    \begin{subfigure}[t]{0.230\textwidth}
        \raisebox{-\height}{\includegraphics[width=\textwidth]{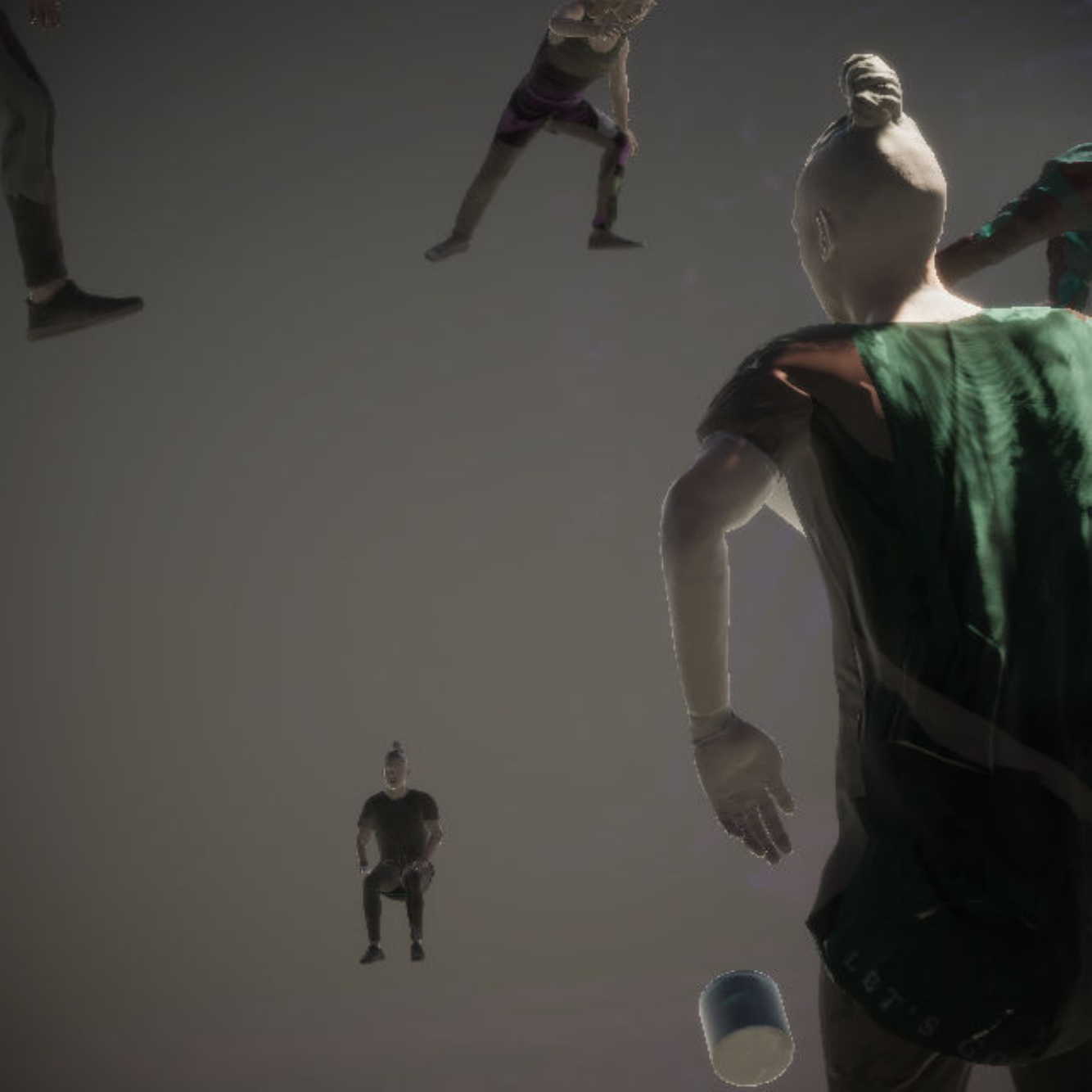}}
    \end{subfigure}
    \begin{subfigure}[t]{0.230\textwidth}
        \raisebox{-\height}{\includegraphics[width=\textwidth]{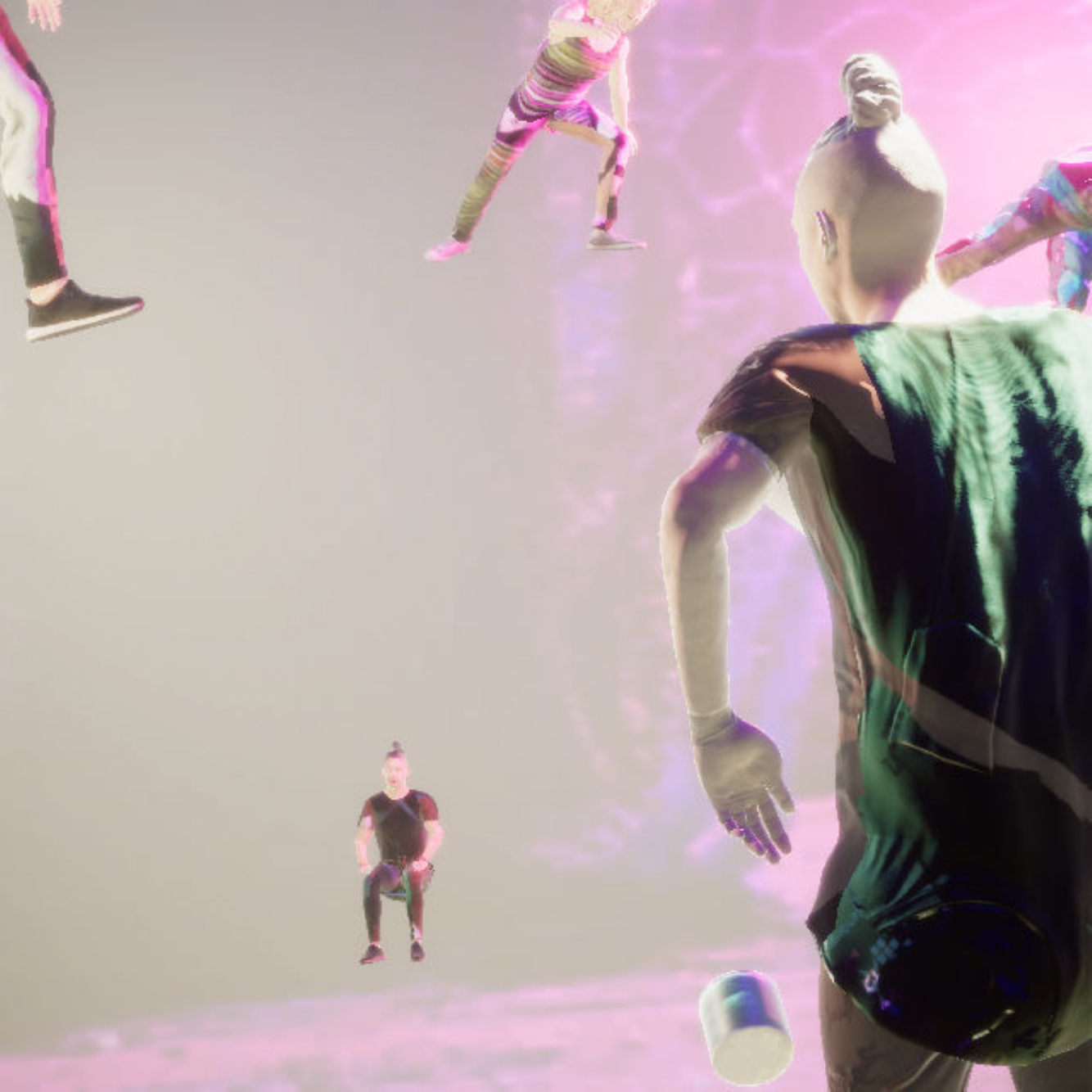}}
    \end{subfigure}
    \hfill \\
    \begin{subfigure}[t]{0.230\textwidth}
        \raisebox{-\height}{\includegraphics[width=\textwidth]{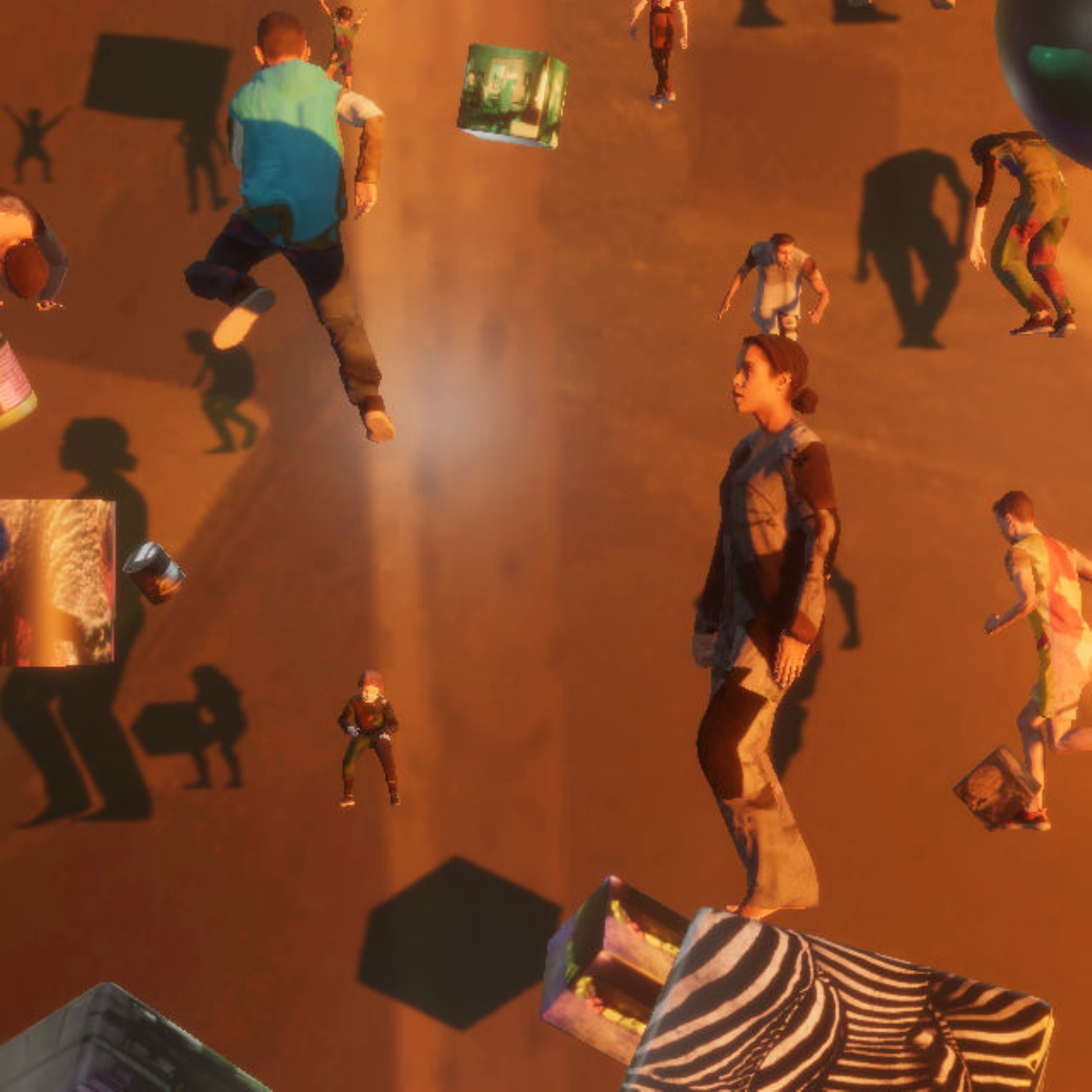}}
    \end{subfigure}
    \begin{subfigure}[t]{0.230\textwidth}
        \raisebox{-\height}{\includegraphics[width=\textwidth]{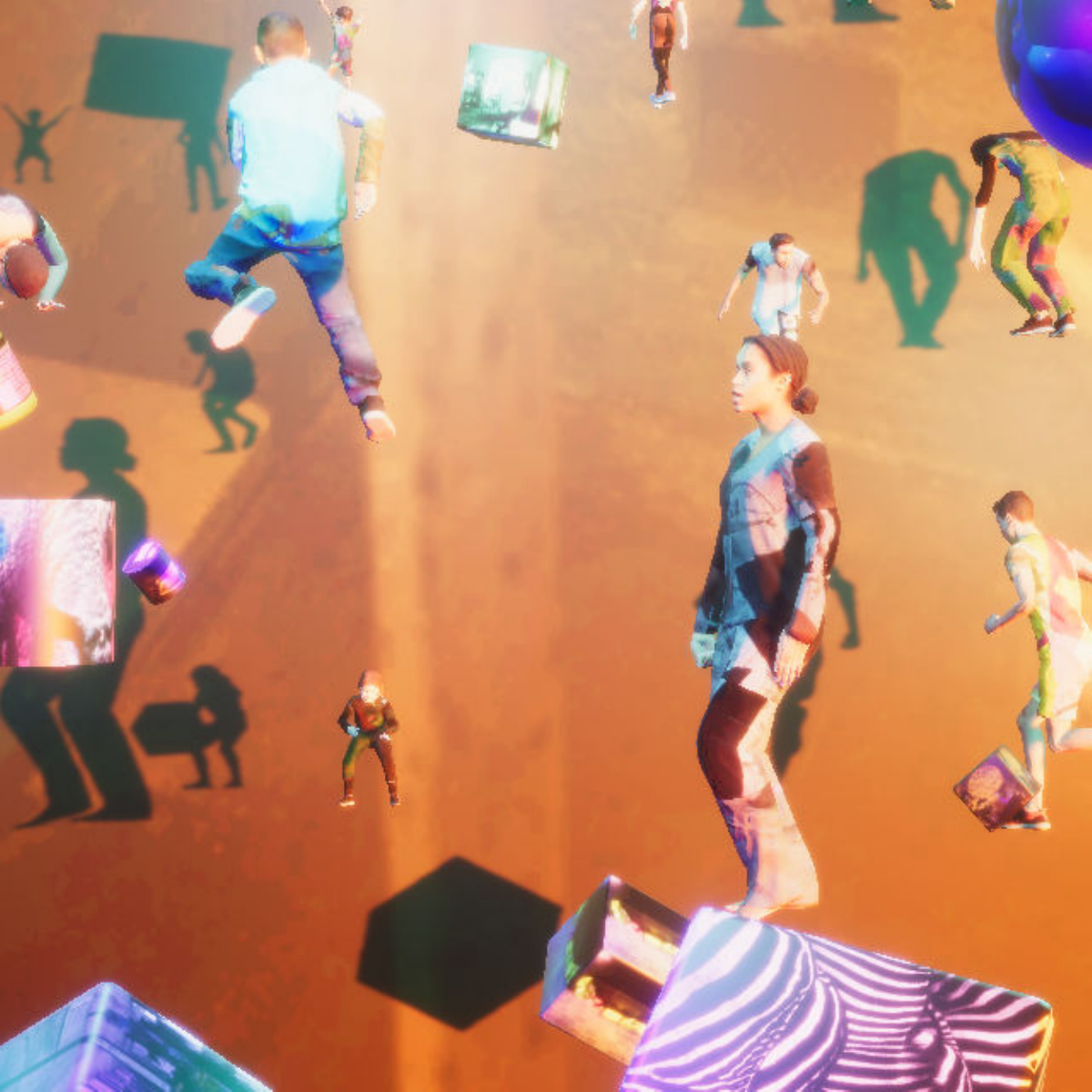}}
    \end{subfigure}
    \begin{subfigure}[t]{0.230\textwidth}
        \raisebox{-\height}{\includegraphics[width=\textwidth]{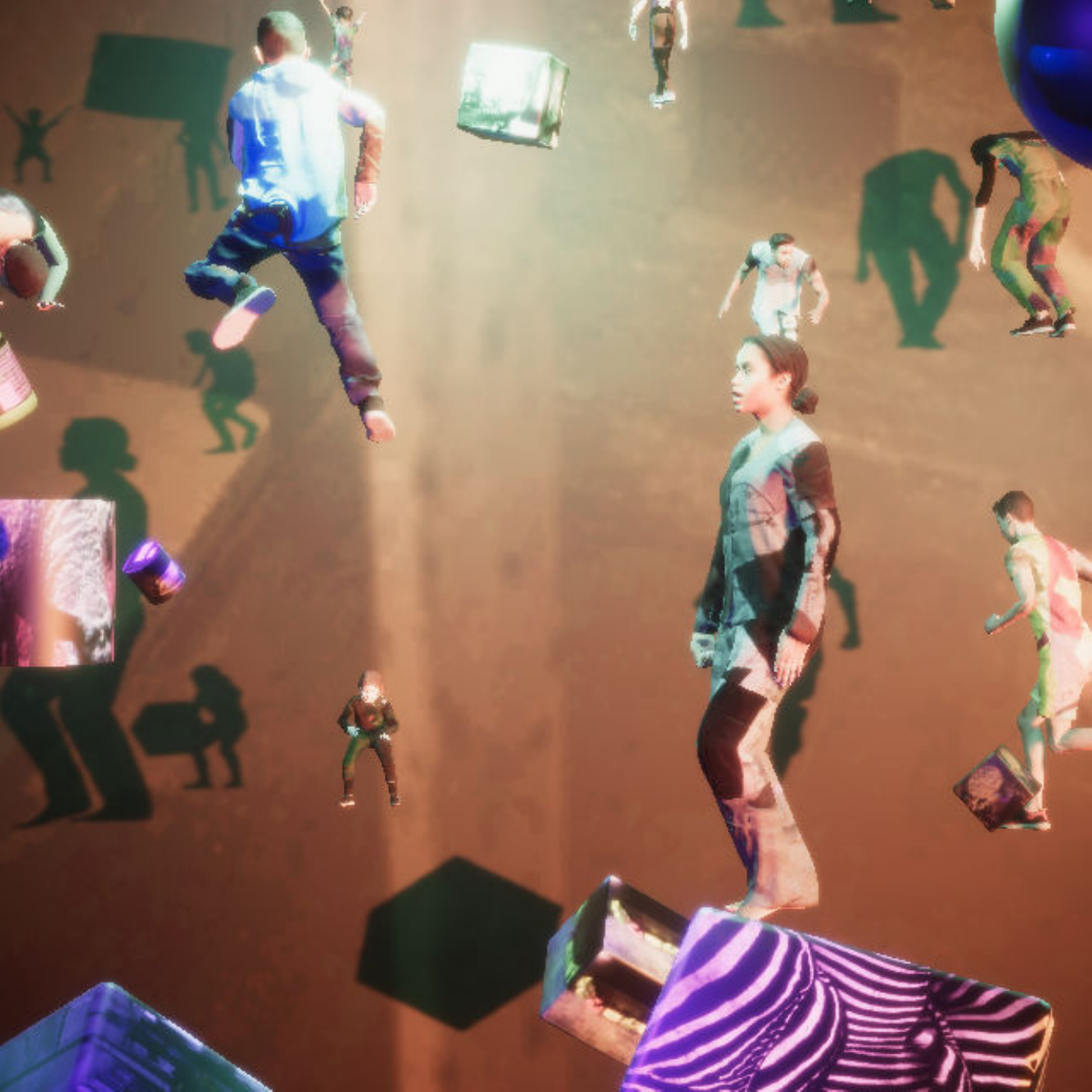}}
    \end{subfigure}
    \hfill \\
    \begin{subfigure}[t]{0.230\textwidth}
        \raisebox{-\height}{\includegraphics[width=\textwidth]{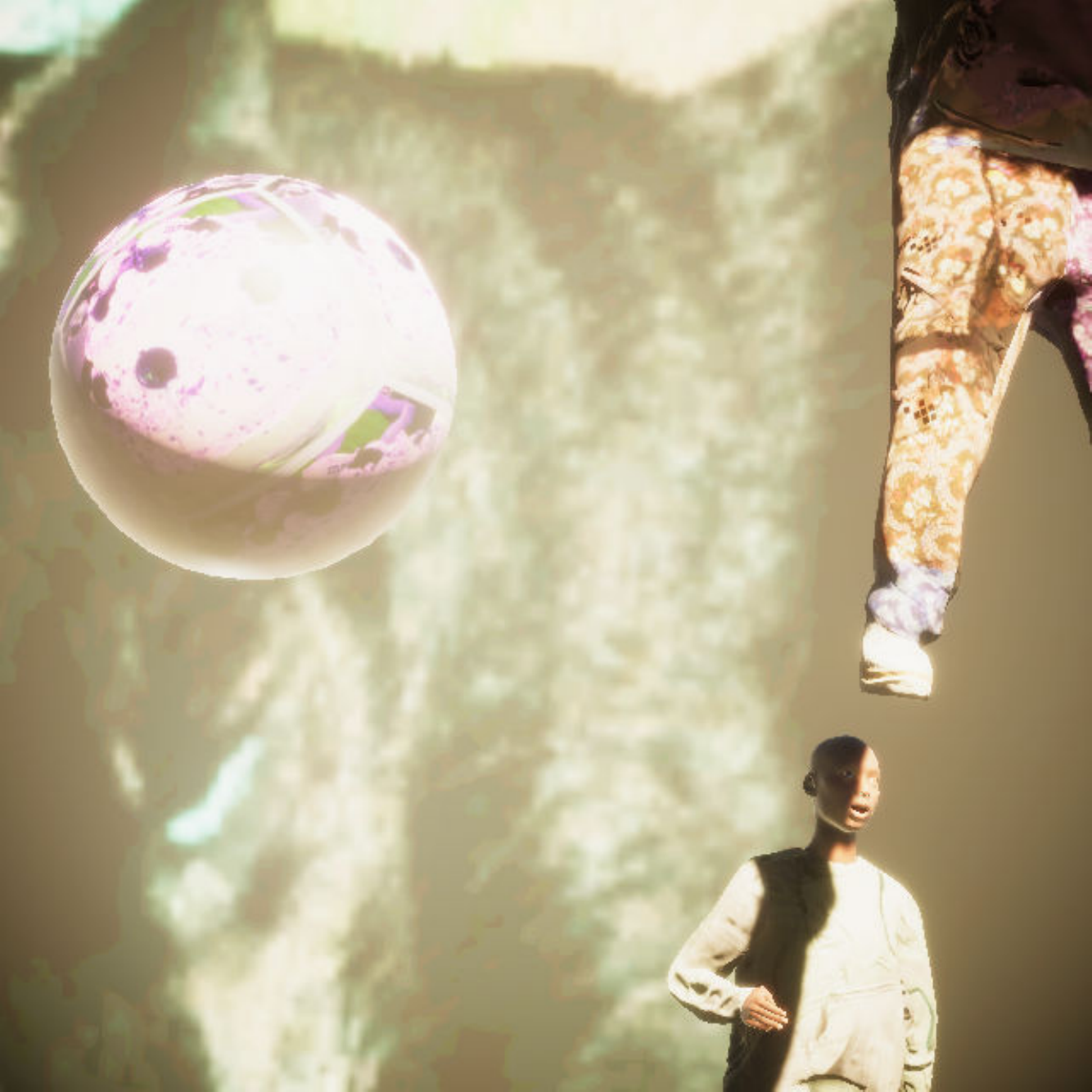}}
    \end{subfigure}
    \begin{subfigure}[t]{0.230\textwidth}
        \raisebox{-\height}{\includegraphics[width=\textwidth]{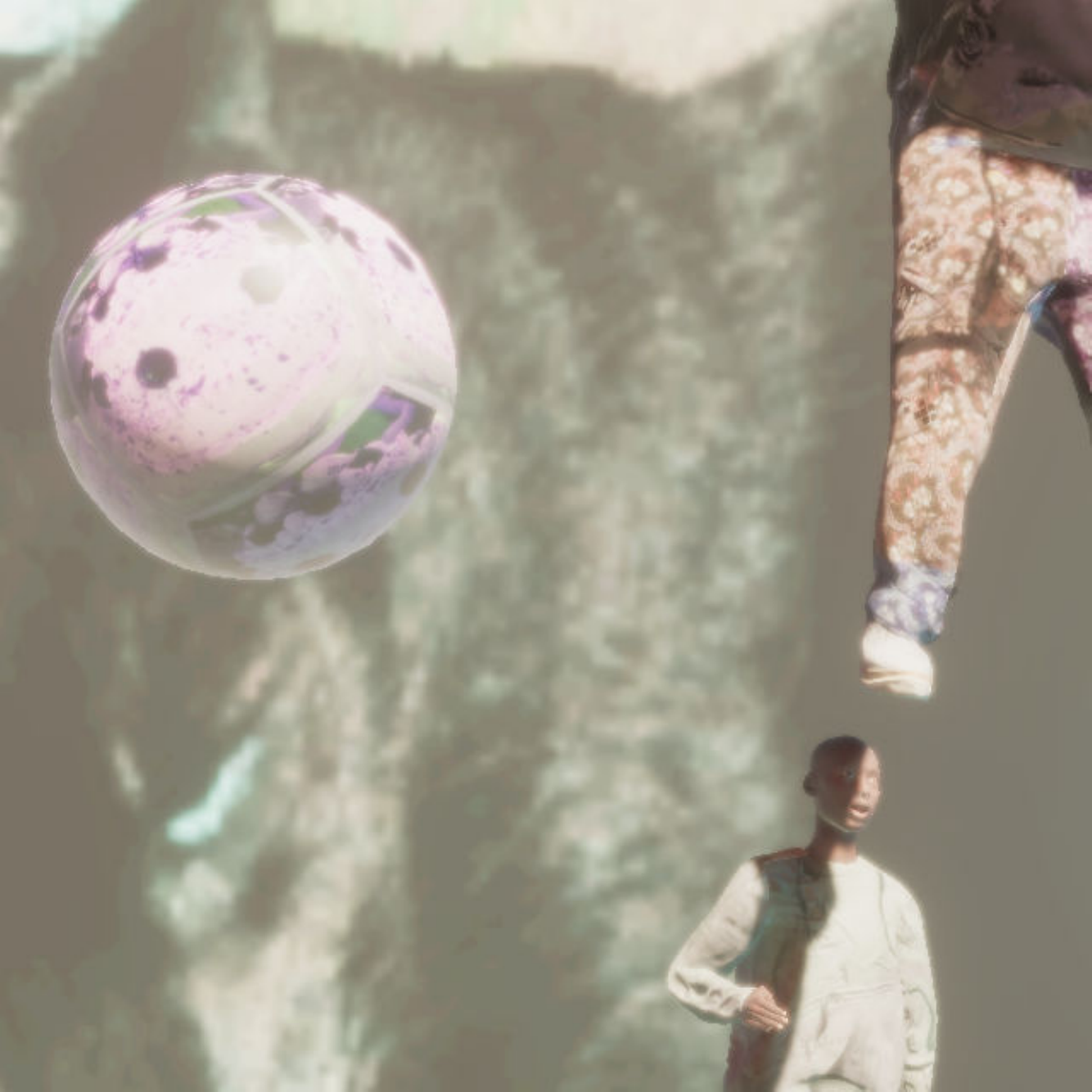}}
    \end{subfigure}
    \begin{subfigure}[t]{0.230\textwidth}
        \raisebox{-\height}{\includegraphics[width=\textwidth]{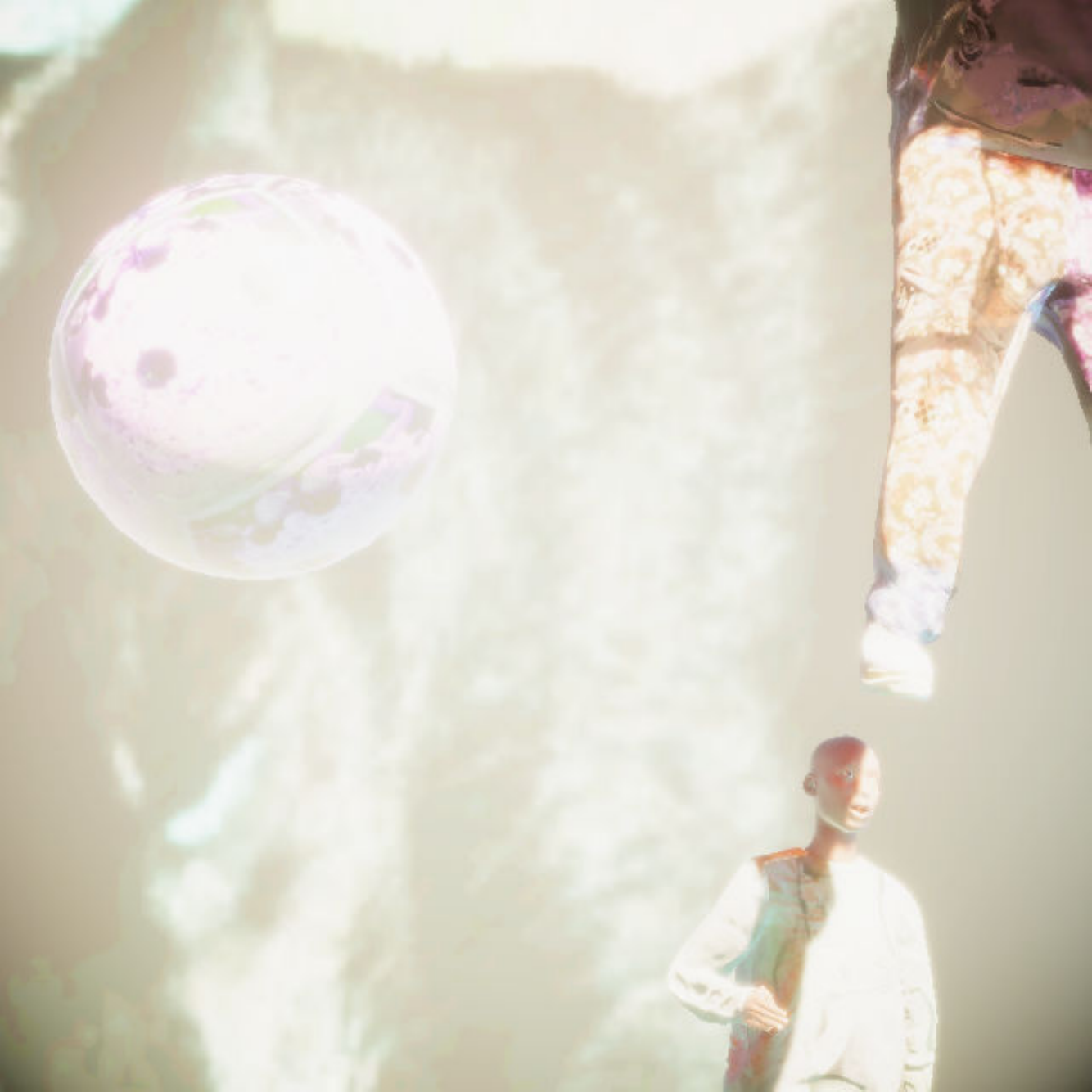}}
    \end{subfigure}
\caption{Examples of light randomization in the same scene 2/2. Each row shows three different lighting conditions while the rest of the scene is unchanged.}
\label{fig:fig:morelight2}%
\end{figure}

In Fig.~\ref{fig:fig:morelight1} and \ref{fig:fig:morelight2} we show additional examples for lighting diversity in our data. Each row shows the same scene under different types of lighting that are made available by our scene lighting design and light randomizers.

\subsection{Shader Graph Design}
\begin{figure}[htb]
    \centering
    \includegraphics[width=\textwidth]{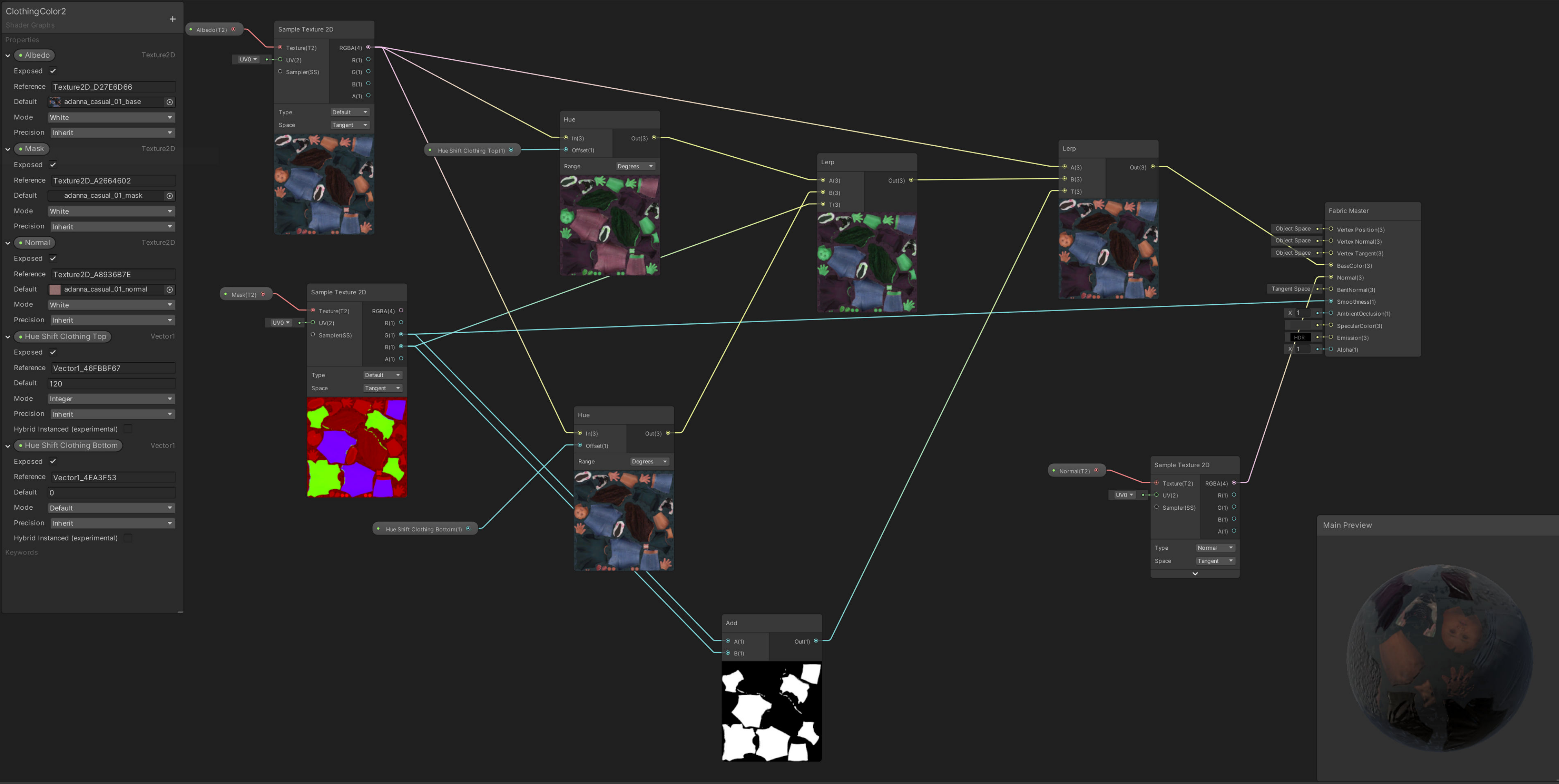}
    \caption{Shader Graph design in Unity.}
    \label{fig:shadergraph_design}
\end{figure}

We constructed a Shader Graph to define the shading methods for the materials used on the people. 
The Shader Graph has inputs for Albedo, Normal, and Mask textures plus exposures for hue offsets on top and bottom articles of clothing, such as shirts and pants. 
We use the mask texture per channel. 
Each hue offset is applied to an instance of the Albedo texture, and the resulting offset colors are then combined based on the mask channels to produce a single Albedo used in the material. 
This Shader Graph allows us programmatically alter the hue and texture of a masked region of a model. For the assets in \psp{} we have masked the clothing region and use this Shader Graph to change the clothing hue and texture. Fig.~\ref{fig:shadergraph_design} shows the design of the Shader Graph in Unity Editor. Examples of the Shader Graph effects on our human assets' clothing are shown in Fig.~\ref{fig:psp_synth_data_examples}, \ref{fig:renderpeople_assets}, \ref{fig:fig:moreteaser1}, \ref{fig:fig:moreteaser2}, \ref{fig:fig:morelight1}, and \ref{fig:fig:morelight2}.

\end{document}